\ifdefined\pdfobjcompresslevel
\fi
\documentclass[referee,pdflatex,sn-nature]{sn-jnl}

\usepackage{bibunits}
\defaultbibliographystyle{sn-nature}
\defaultbibliography{sn-bibliography}
\usepackage{graphicx}%
\usepackage{multirow}%
\usepackage{amsmath,amssymb,amsfonts}%
\usepackage{amsthm}%
\usepackage{mathrsfs}%
\usepackage[title]{appendix}%
\usepackage{xcolor}%
\usepackage{textcomp}%
\usepackage{manyfoot}%
\usepackage{booktabs}
\usepackage{graphicx}
\usepackage{algorithm}%
\usepackage{algorithmicx}%
\usepackage{algpseudocode}%
\usepackage{listings}%
\usepackage{adjustbox}
\usepackage{booktabs}
\usepackage{makecell}
\usepackage[capitalize]{cleveref}
\usepackage{longtable}
\usepackage{array}
\usepackage{needspace}
\usepackage{pdfpages}
\usepackage{titletoc}
\usepackage[all]{hypcap}
\usepackage{changepage}

\makeatletter
\newcommand{\DeclareName}[3]{%
  \expandafter\def\csname nm@#1@sg\endcsname{#2}%
  \expandafter\def\csname nm@#1@pl\endcsname{#3}%
}
\newcommand{\Name}[1]{\csname nm@#1@sg\endcsname}
\newcommand{\Names}[1]{\csname nm@#1@pl\endcsname}
\makeatother
\DeclareName{cnn}{nnFoundation\textsubscript{CNN}}{nnFoundation\textsubscript{CNN}}
\DeclareName{vit}{nnFoundation\textsubscript{ViT}}{nnFoundation\textsubscript{ViT}}

\begin{document}

\title[]{\makebox[\linewidth][c]{\mbox{nnFoundation: 3D Foundation Models for Radiology}}}



\author[1,*]{\fnm{Constantin} \sur{Ulrich Harsy}}
\author[1,3,*]{\fnm{Tassilo} \sur{Wald}}
\author[1,3,*]{\fnm{Karol} \sur{Gotkowski}}
\author[1,2,5,*]{\fnm{Yannick} \sur{Kirchhoff}}
\author[3,4,5,*]{\fnm{Marcel} \sur{Knopp}}
\author[1,3,*]{\fnm{Maximilian} \sur{Rokuss}}

\author[1]{\fnm{Elisa} \sur{Stegmeier}}
\author[1,5]{\fnm{Philipp} \sur{Schader}}
\author[6]{\fnm{Dasha} \sur{Trofimova}}
\author[1,3,5,7]{\fnm{Raphael} \sur{Stock}}
\author[6]{\fnm{Kim-Celine} \sur{Kahl}}
\author[6]{\fnm{Stephen} \sur{Schaumann}}
\author[6,8]{\fnm{Selen} \sur{Erkan}}
\author[1,3]{\fnm{David} \sur{Zimmerer}}
\author[6]{\fnm{Stefan} \sur{Denner}}
\author[1,5]{\fnm{Moritz} \sur{Langenberg}}
\author[1,3]{\fnm{Sebastian} \sur{Ziegler}}
\author[1,9,10]{\fnm{Katharina} \sur{Eckstein}}
\author[1]{\fnm{Maximilian} \sur{Fischer}}
\author[1,3]{\fnm{Jonathan} \sur{Suprijadi}}

\author[1,2,7,10]{\fnm{Bálint} \sur{Kovács}}
\author[1]{\fnm{Benjamin} \sur{Hamm}}
\author[1,11]{\fnm{Anand} \sur{Deshpande}}
\author[1,7,10]{\fnm{Dimitrios} \sur{Bounias}}
\author[1,2]{\fnm{Nico} \sur{Disch}}
\author[6]{\fnm{Shuhan} \sur{Xiao}}
\author[1,9,10]{\fnm{Jessica} \sur{Kächele}}
\author[8]{\fnm{Jan} \sur{Sellner}}
\author[6]{\fnm{Rajesh} \sur{Baidya}}
\author[1,3,4]{\fnm{Jeremias} \sur{Traub}}
\author[1,3]{\fnm{Lars} \sur{Krämer}}
\author[6,12]{\fnm{Maximilian} \sur{Zenk}}
\author[8]{\fnm{Tim} \sur{Rädsch}}
\author[6]{\fnm{Stefan} \sur{Dvoretskii}}
\author[1,7]{\fnm{Robin} \sur{Peretzke}}
\author[1]{\fnm{Jonathan} \sur{Deissler}}
\author[1,10]{\fnm{Alexandra} \sur{Ertl}}
\author[6]{\fnm{Partha} \sur{Ghosh}}
\author[1,3]{\fnm{Kris} \sur{Dreher}}
\author[1]{\fnm{Stefan} \sur{Dinkelacker}}
\author[3,4]{\fnm{Annika} \sur{Reinke}}
\author[4]{\fnm{Evangelia} \sur{Christodoulou}}

\author[13]{\fnm{Numan} \sur{Saeed}}
\author[14]{\fnm{Yoland} \sur{Savriama}}
\author[15]{\fnm{Santiago} \sur{Estrada}}
\author[15]{\fnm{David} \sur{Kügler}}
\author[16,17]{\fnm{Laura Alexandra} \sur{Daza Barragan}}
\author[16,17]{\fnm{Cristina Isabel Gonzalez} \sur{Osorio}}
\author[18]{\fnm{Jan} \sur{Peeken}}
\author[6,19]{\fnm{Michael} \sur{Baumgartner}}
\author[19]{\fnm{Marvin} \sur{Teichmann}}
\author[19]{\fnm{Guillaume} \sur{Chabin}}
\author[12]{\fnm{Matthias} \sur{Kirchler}}
\author[12]{\fnm{Valentin} \sur{Koch}}

\author[]{\fnm{} \sur{for the ALFA study}}

\author[20]{\fnm{Markus} \sur{Hohenhaus}}
\author[20]{\fnm{Dimitri} \sur{Koslov}}
\author[6,7]{\fnm{Nina} \sur{Decker}}

\author[21]{\fnm{Mohammad} \sur{Yaqub}}
\author[14]{\fnm{Arnd} \sur{Heuser}}
\author[15,22,23]{\fnm{Martin} \sur{Reuter}}
\author[16,17,24,25]{\fnm{Julia A.} \sur{Schnabel}}
\author[19]{\fnm{Tobias} \sur{Heimann}}
\author[19]{\fnm{Florin} \sur{Ghesu}}
\author[12]{\fnm{Paul} \sur{Brachmann}}

\author[26]{\fnm{Claus P.} \sur{Heußel}}
\author[27]{\fnm{Alexander} \sur{Radbruch}}
\author[1,27]{\fnm{Gianluca} \sur{Brugnara}}
\author[27]{\fnm{Aditya} \sur{Rastogi}}
\author[27]{\fnm{Martha} \sur{Foltyn-Dumitru}}
\author[28]{\fnm{Heinz-Peter} \sur{Schlemmer}}
\author[28]{\fnm{Ignaz} \sur{Reicht}}
\author[28]{\fnm{Julius C.} \sur{Holzschuh}}
\author[29]{\fnm{Michael} \sur{Bach}}
\author[30]{\fnm{Bram} \sur{Stieltjes}}
\author[26,27]{\fnm{Kai} \sur{Schlamp}}

\author[3,4,5,7,10,21]{\fnm{Lena} \sur{Maier-Hein}}
\author[1,11]{\fnm{Marco} \sur{Nolden}}
\author[1]{\fnm{Ralf} \sur{Floca}}
\author[6]{\fnm{Paul F.} \sur{Jäger}}

\author[1,27,†]{\fnm{Philipp} \sur{Vollmuth}}
\author[1,3,7,†]{\fnm{Fabian} \sur{Isensee}}
\author[1,2,3,5,7,10,11,21,31,†]{\fnm{Klaus H.} \sur{Maier-Hein}}




\affil[1]{\orgdiv{Division of Medical Image Computing}, \orgname{German Cancer Research Center (DKFZ)}, \orgaddress{{Heidelberg}, \country{Germany}}}
\affil[2]{\orgdiv{HIDSS4Health - Helmholtz Information and Data Science School for Health}, \orgname{Helmholtz Association}, \orgaddress{{Karlsruhe/Heidelberg}, \country{Germany}}}
\affil[3]{\orgdiv{Helmholtz Imaging}, \orgname{German Cancer Research Center (DKFZ)}, \orgaddress{{Heidelberg}, \country{Germany}}}
\affil[4]{\orgdiv{Division of Intelligent Medical Systems}, \orgname{German Cancer Research Center (DKFZ)}, \orgaddress{{Heidelberg}, \country{Germany}}}
\affil[5]{\orgdiv{Faculty of Mathematics and Computer Science}, \orgname{Heidelberg University}, \orgaddress{{Heidelberg}, \country{Germany}}}
\affil[6]{\orgdiv{Formerly: Division of Medical Image Computing}, \orgname{German Cancer Research Center (DKFZ)}, \orgaddress{{Heidelberg}, \country{Germany}}}
\affil[7]{\orgname{National Center for Tumor Diseases (NCT) Heidelberg, a partnership between the German Cancer Research Center (DKFZ) and Heidelberg University Hospital (UKHD)}, \orgaddress{{Heidelberg}, \country{Germany}}}
\affil[8]{\orgdiv{Formerly: Division of Intelligent Medical Systems}, \orgname{German Cancer Research Center (DKFZ)}, \orgaddress{{Heidelberg}, \country{Germany}}}
\affil[9]{\orgname{German Cancer Consortium (DKTK), DKFZ, core center Heidelberg}, \orgaddress{{Heidelberg}, \country{Germany}}}
\affil[10]{\orgdiv{Medical Faculty Heidelberg}, \orgname{Heidelberg University}, \orgaddress{{Heidelberg}, \country{Germany}}}
\affil[11]{\orgdiv{Helmholtz Metadata Collaboration (HMC) Hub Health}, \orgname{German Cancer Research Center (DKFZ)}, \orgaddress{{Heidelberg}, \country{Germany}}}
\affil[12]{\orgname{Floy GmbH}, \orgaddress{{Munich}, \country{Germany}}}
\affil[13]{\orgdiv{Department of Computer Vision}, \orgname{Mohamed bin Zayed University of Artificial Intelligence (MBZUAI)}, \orgaddress{{Abu Dhabi}, \country{United Arab Emirates}}}
\affil[14]{\orgdiv{Animal Phenotyping Platform}, \orgname{Max-Delbrück-Centrum für Molekulare Medizin in der Helmholtz-Gemeinschaft}, \orgaddress{{Berlin}, \country{Germany}}}
\affil[15]{\orgdiv{AI in Medical Imaging}, \orgname{German Center for Neurodegenerative Diseases (DZNE)}, \orgaddress{{Bonn}, \country{Germany}}}
\affil[16]{\orgdiv{Institute of Machine Learning for Biomedical Imaging}, \orgname{Helmholtz Munich}, \orgaddress{{Neuherberg}, \country{Germany}}}
\affil[17]{\orgdiv{School of Computation, Information and Technology}, \orgname{Technical University of Munich}, \orgaddress{{Munich}, \country{Germany}}}
\affil[18]{\orgdiv{Department of Radiation Oncology}, \orgname{Technical University of Munich (TUM), School of Medicine and Health, Klinikum rechts der Isar}, \orgaddress{{Munich}, \country{Germany}}}
\affil[19]{\orgdiv{Digital Technology and Innovation}, \orgname{Siemens Healthineers}, \orgaddress{{Erlangen}, \country{Germany}}}
\affil[20]{\orgdiv{IT Core Facility}, \orgname{German Cancer Research Center (DKFZ)}, \orgaddress{{Heidelberg}, \country{Germany}}}
\affil[21]{\orgdiv{Division of Computing and Mathematical Sciences}, \orgname{Mohamed bin Zayed University of Artificial Intelligence (MBZUAI)}, \orgaddress{{Abu Dhabi}, \country{United Arab Emirates}}}
\affil[22]{\orgname{A.A. Martinos Center for Biomedical Imaging, Massachusetts General Hospital}, \orgaddress{{Boston, MA}, \country{United States}}}
\affil[23]{\orgdiv{Department of Radiology}, \orgname{Harvard Medical School}, \orgaddress{{Boston, MA}, \country{United States}}}
\affil[24]{\orgname{Munich Center for Machine Learning}, \orgaddress{{Munich}, \country{Germany}}}
\affil[25]{\orgdiv{School of Biomedical Engineering and Imaging Sciences}, \orgname{King's College London}, \orgaddress{{London}, \country{United Kingdom}}}
\affil[26]{\orgdiv{Diagnostic and Interventional Radiology with Nuclear Medicine}, \orgname{Heidelberg Thoracic Clinic, Heidelberg University}, \orgaddress{{Heidelberg}, \country{Germany}}}
\affil[27]{\orgdiv{Division for Computational Radiology \& Clinical AI (CCIBonn.ai), Department of Neuroradiology}, \orgname{University Hospital Bonn}, \orgaddress{{Bonn}, \country{Germany}}}
\affil[28]{\orgdiv{Department of Radiology}, \orgname{German Cancer Research Center (DKFZ)}, \orgaddress{{Heidelberg}, \country{Germany}}}
\affil[29]{\orgdiv{Department of Radiology and Nuclear Medicine}, \orgname{University Hospital Basel}, \orgaddress{{Basel}, \country{Switzerland}}}
\affil[30]{\orgdiv{D\&ICT}, \orgname{University Hospital Basel}, \orgaddress{{Basel}, \country{Switzerland}}}
\affil[31]{\orgdiv{Pattern Analysis and Learning Group, Department of Radiation Oncology}, \orgname{Heidelberg University Hospital}, \orgaddress{{Heidelberg}, \country{Germany}}}


\abstract{Radiological artificial intelligence has advanced rapidly, yet most systems remain narrowly task-specific, data-intensive, and fragile under domain shift. Foundation models offer a path toward more transferable and data-efficient radiological models, but existing approaches are limited in scale, evaluated narrowly, and often assume that a single pretrained model can support diverse downstream tasks.
\\
Here we present nnFoundation, a pair of complementary 3D radiological foundation models designed for transferable representation learning across heterogeneous tasks and datasets. Developed within the Human Radiome Project (THRP), nnFoundation is trained on 2.1 million CT, MRI, and PET image volumes from 125 institutional and public datasets, representing the largest radiological pretraining resource to date. We systematically evaluate performance across a comprehensive suite of 108 tasks spanning segmentation, detection, classification, report generation, and image retrieval, including evaluations under domain shift, by external partners and in low-data and low-compute regimes.
\\
Across all task types, our convolution- and transformer-based nnFoundation models consistently outperform both prior 3D foundation models and training from scratch, establishing state-of-the-art performance for radiological imaging. 
However, performance follows a consistent task-dependent structure: the convolutional nnFoundation model dominates spatially localized tasks, whereas the transformer-based nnFoundation model excels in tasks requiring global semantic reasoning and in frozen-feature settings. Dynamically aligning the foundation model topology with the dataset characteristics post-hoc further improves transfer across heterogeneous 3D settings.
\\
These results show that transferable 3D radiological performance is governed not by a single universal model, but by the interplay of scalable pretraining, complementary architectures, and dataset-aware adaptation. We release nnFoundation models integrated into nnU-Net and nnDetection, enabling immediate application across established radiology workflows.
}

\makeatletter
\let\originalartauthors\artauthors

\renewcommand{\artauthors}{%
  \begin{adjustwidth}{-1.7cm}{-1.7cm}%
    \originalartauthors\par
    \vspace{4pt}
    {\normalfont\small
      \noindent\textsuperscript{*}Equal contribution, Author order among the co-first authors may be adjusted for individual use.\par
      \noindent\textsuperscript{$\dagger$} Shared last. Equal supervision.\par
      Affiliations and individual author contributions are provided in \cref{affil}.\par\par
    }
  \end{adjustwidth}%
}
\makeatother

\makeatletter
\let\savedaffiliations\auaddress
\gdef\auaddress{}
\makeatother

\makeatletter
\let\originalprintabstract\printabstract

\renewcommand{\printabstract}{%
  \begin{adjustwidth}{-1.7cm}{-1.7cm}
    \originalprintabstract
  \end{adjustwidth}%
}
\makeatother

\makeatletter
\let\originalprintkeywords\printkeywords

\renewcommand{\printkeywords}{%
  \begin{adjustwidth}{-1.7cm}{-1.7cm}%
    \originalprintkeywords
  \end{adjustwidth}%
}
\makeatother
\maketitle
\section{Introduction}\label{intro}
Radiological imaging informs clinical decision-making across medical specialties and is leading the adoption of artificial intelligence (AI), with more than three-quarters of over 1,400 regulatory-approved AI algorithms in medicine targeting radiological tasks \citep{FDAAIEnabledMedicalDevices2025}. Yet, these algorithms face key limitations: they are typically trained from scratch using fully supervised learning on expert-annotated data for a single task \cite{WANG2024103201}. This narrow training regime limits both performance and generalizability across scanners, institutions, and patient populations \citep{Generalization_survey,Yu2022-sj,Suleman2025-kj}. Enabling these models to generalize robustly to new clinical settings, with minimal additional labeled data remains a key translational challenge for radiology AI, and continues to hinder broader clinical adoption \cite{doi:10.1148/radiol.242961}. \\
\begingroup
\renewcommand{\thefootnote}{}
\footnotetext{The nnSSL pretraining framework and downstream repositories are available on \href{https://github.com/MIC-DKFZ/nnssl}{GitHub}.}
\endgroup
\noindent Foundation models offer a potential path to address this challenge, characterized by self-supervised training on unlabeled data at scale, yielding general-purpose feature representations that can be adapted to a wide array of downstream tasks with minimal additional labeled data \citep{assran2023self,oquab2023dinov2,dinov3,he2022mae,grill2020byol,chen2020simclr}. In radiology, foundation-model development has primarily concentrated on 2D chest data \citep{perez2025exploring, tiu2022expert, zhou2023advancing, boecking2022making, bannur2023learning, moon2022multi, wang-etal-2022-medclip, zhang2023biomedclip, lozano2025biomedica}, where pretraining is computationally tractable and public data are abundant. Three-dimensional (3D) cross-sectional radiology (CT, MRI, PET), which produces volumetric data and requires 3D architectures to exploit spatial structure, has received comparatively limited attention. Initial 3D radiology foundation models have established the promise of self-supervised pretraining for volumetric imaging through focused demonstrations on specific anatomies and modalities \citep{ munk2024amaes, wang2023mis, pai2025vision,Blankemeier2026, wald2024openmind,Tak2026BrainIAC}. However, extending these models across the full anatomical and diagnostic spectrum of cross-sectional radiology, systematically evaluating them within this spectrum, and rigorously benchmarking them against established fully supervised methods \cite{wald2025primus,isensee2021nnu} remains an unresolved challenge. \\

\noindent Beyond scale, the central question is how foundation models should be designed to accommodate heterogeneous task requirements and dataset characteristics. Clinical CT, MRI, and PET data vary widely in resolution, contrast mechanisms, anatomical coverage, and acquisition protocols, while downstream tasks impose heterogeneous requirements on learned representations. Localization task types such as segmentation and detection demand high spatial precision, whereas classification, retrieval, and vision-language modeling rely on global semantic understanding. Addressing this heterogeneity requires dataset-adaptive design and controlled comparison across architectures, pretraining strategies, and transfer mechanisms. Neither has been systematically developed for 3D cross-sectional radiology, leaving open how to resolve the tension between model generality and task-specific demands. \\

\begin{figure}[h]
\centering
\includegraphics[width=\textwidth]{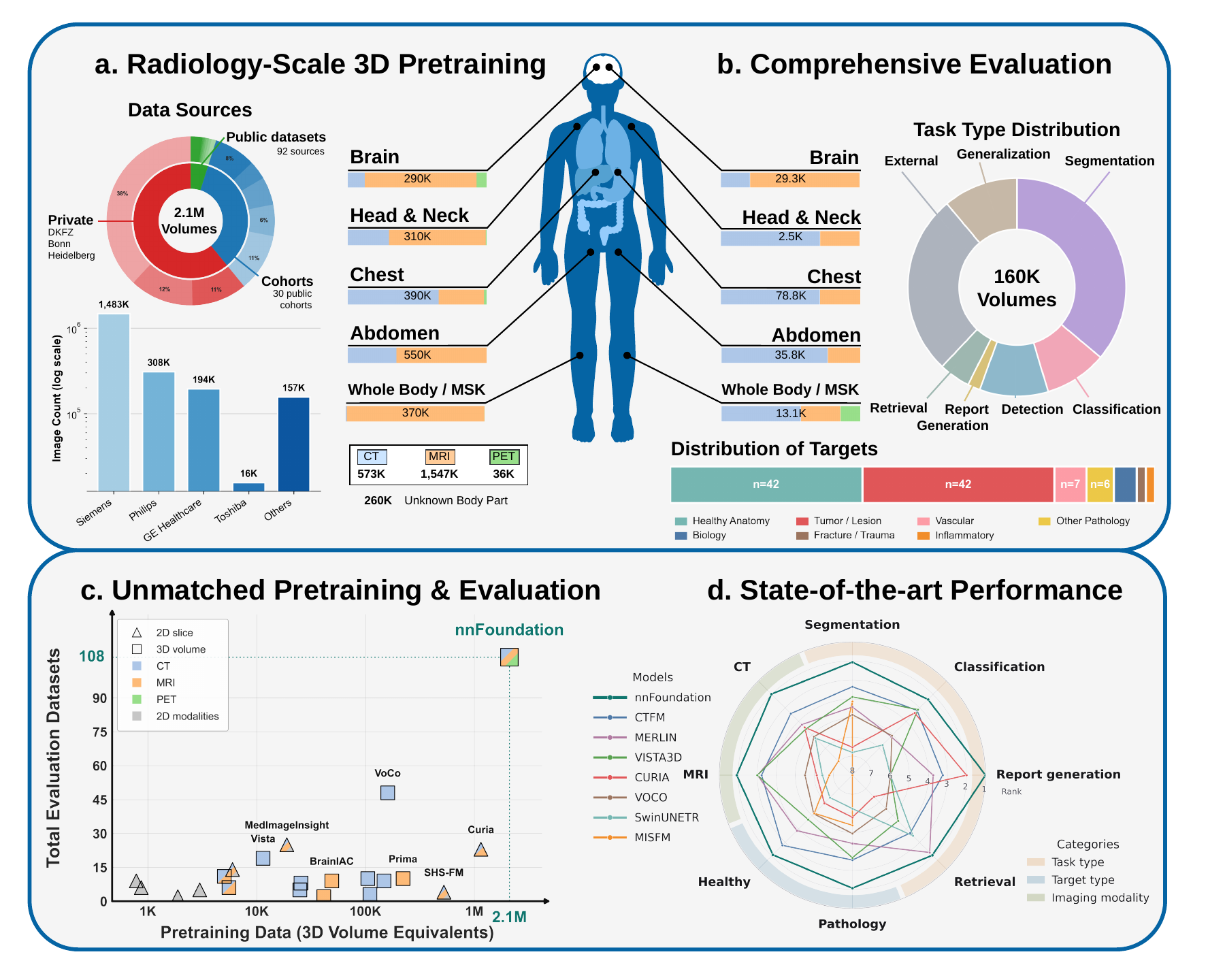}
\caption{\textbf{Radiology-scale pretraining, comprehensive evaluation, and SoTa performance.}
a) Pretraining dataset composition and diversity. The dataset comprises 2.1 million 3D imaging series from public sources, clinical cohorts, and research datasets, spanning CT, MRI, and PET, multiple scanner vendors, and broad anatomical coverage. b) Downstream evaluation protocol. A large-scale benchmark covering 160,000 volumes across diverse task types, including segmentation, detection, classification, retrieval, report generation, generalization under domain shift and to biological imaging domains, and external validation with representation of multiple anatomical regions and both healthy and pathological structures.
c) Pretraining and evaluation scale relative to prior work. nnFoundation substantially expands both pretraining data scale and evaluation breadth compared to existing radiological foundation models. d) Performance across tasks and settings. Pretrained models achieve state-of-the-art performance across task types, imaging modalities, anatomical regions, and target types.}
\label{fig1}
\end{figure}

\noindent Here, we present nnFoundation, \Name{vit} and \Name{cnn}, a pair of complementary 3D radiological foundation models based on Vision Transformer (ViT) and Convolutional Neural Network (CNN) architectures, designed to enable transferable representation learning across heterogeneous tasks and datasets. nnFoundation is developed within the Human Radiome Project (THRP), a large-scale effort providing the data, infrastructure, and evaluation protocols required to study radiological foundation models at scale. We assemble large-scale pretraining data spanning 2.1 million image volumes including CT, MRI, and PET data from 125 institutional and public sources, exceeding prior 3D radiological foundation models in pretraining scale by a wide margin (\cref{fig1}c). On this dataset, we systematically improve foundation model design by investigating various design choices, including architectures, pretraining objectives, and hyperparameters, and evaluating their effects rigorously under controlled conditions and matched compute budgets. \\ 

\noindent To assess general-purpose capability, we perform a comprehensive evaluation of the nnFoundation models, \Name{cnn} and \Name{vit}, across 108 downstream tasks and 160,000 evaluation volumes, exceeding prior work not only in pretraining scale but also in downstream validation breadth (\cref{fig1}c). The evaluation  spans all body regions, cross-sectional modalities (CT, MRI, PET), and major radiological task types: segmentation, detection, classification, image retrieval, and report generation. Evaluations further include robustness under domain shift, validation by external partners and performance in low-data and low-compute regimes. We show that nnFoundation models consistently outperform prior 3D foundation models and training from scratch, with masked autoencoding (MAE) emerging as the most effective pretraining objective. However, peak performance follows a task-dependent pattern with complementary strengths across model architectures. \\

\noindent Finally, we show that aligning model topology with dataset characteristics improves transfer across heterogeneous 3D data and introduce dynamic dataset-specific adaptation as a mechanism to achieve this alignment without additional training. By integrating all pretrained models into widely used frameworks, including nnU-Net \citep{isensee2021nnu,isensee2024nnunet} and nnDetection \citep{nndet}, we enable immediate application to downstream tasks. Together, our results establish a design principle for radiological foundation models: general-purpose performance is governed not by a single universal model, but by the combination of scalable pretraining, task-aligned architectures, and dataset-aware adaptation.

\section{Results}
\label{results}

Large-scale self-supervised pretraining on 2.1 million CT, MRI, and PET volumes yields strong and transferable 3D radiological representations. Across all modalities, tasks, and target types, our foundation models exceed training from scratch and prior foundation models, setting a new state of the art (\cref{fig1}d). We find that performance is mainly governed by three factors: pretraining scale and strategy, task-specific architecture selection, and alignment between model topology and dataset characteristics. In the following, we analyze transfer performance across tasks and architectures, show how dataset-aware adaptation enhances performance across heterogeneous 3D settings, and assess data efficiency, convergence, and robustness.

\subsection*{Large-scale multimodal 3D radiological data enables general-purpose pretraining}
\addcontentsline{toc}{subsection}{Large-scale multimodal 3D radiological data enables general-purpose pretraining}
\label{results_1}

To enable large-scale pretraining of radiological foundation models, we assembled the largest collection of 3D radiological imaging data reported to date (\cref{fig1}a). The dataset comprises 2.1 million CT, MRI, and PET image volumes across 125 institutional and public sources, spanning a wide range of scanner types, acquisition protocols, anatomical regions, and patient populations. The final dataset was obtained through extensive curation and harmonization of heterogeneous sources (see Online Methods \ref{methods_1}), ensuring consistent data quality across modalities and institutions. While metadata availability varies across sources, this heterogeneity reflects real-world clinical data conditions. This scale and diversity provide a foundation for learning transferable representations across body regions, modalities, and task types. \\

\noindent The dataset integrates three complementary data streams (\cref{fig1}a, Supplementary \ref{appendix_pretraining_datasets}): clinical partnerships with academic medical centers (60.9\%; 1,315,227 volumes), specialized research cohorts (34.2\%; 737,714 volumes), and a collection of small public datasets (4.9\%; 105,629 volumes). The dataset spans all major 3D imaging modalities, with MRI (71.7\%), CT (26.6\%), and PET (1.7\%), and provides full-body anatomical coverage, including head, chest, abdomen, pelvis, and extremities. It captures substantial variability in scanner manufacturers, acquisition protocols, and reconstruction settings, reflecting the heterogeneity encountered in routine clinical practice. \\

\noindent This combination of scale and heterogeneity directly overcomes a key limitation of prior radiological foundation models \cite{chen2023masked, zhuang2023advancing,tang2024hyspark, zhang2024mapsegunifiedunsuperviseddomain, tang2022self, zhou2021models,wu2024voco,he2023geometric}, which were trained on comparatively narrow datasets, and provides the basis for the pretraining strategy that underlies all subsequent results.

\begin{figure}[h]
\centering
\includegraphics[width=\textwidth]{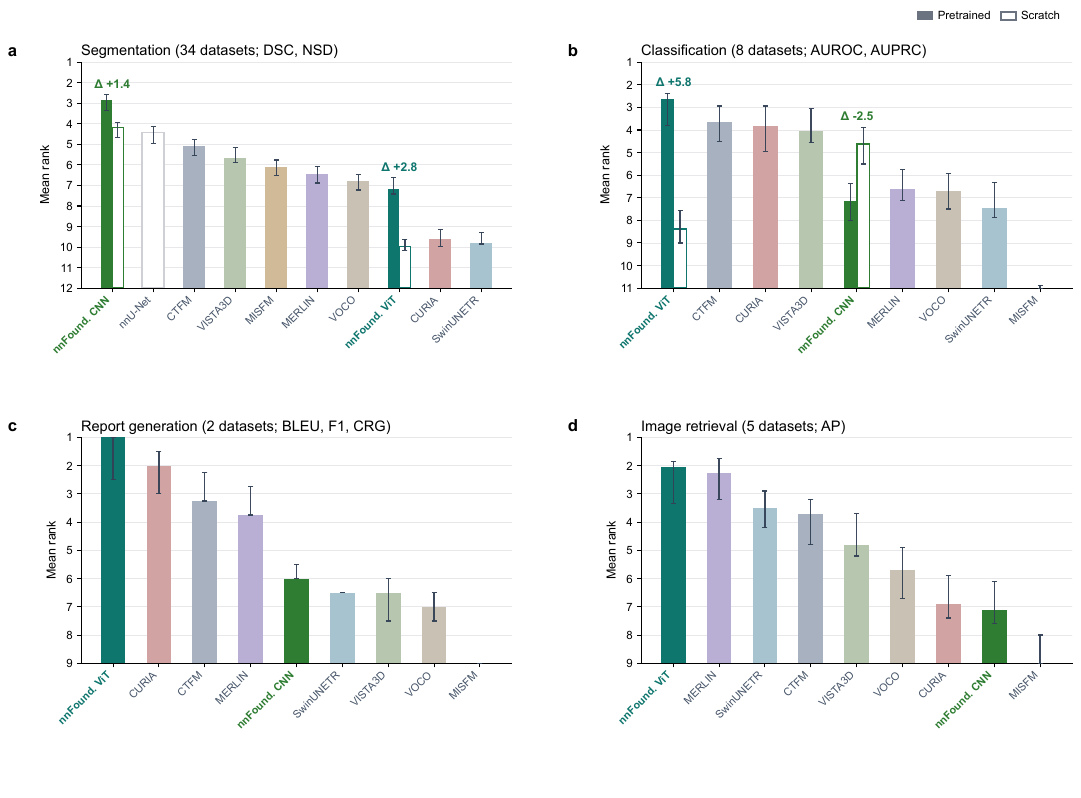}
\caption{\textbf{State-of-the-art performance and task-dependent pattern.} 
Our pretrained models establish state-of-the-art performance across \textbf{a}: segmentation, \textbf{b}: classification, \textbf{c} report generation, and \textbf{d}: image retrieval. A clear task-dependent pattern emerges: The \Name{cnn} excels on spatially localized tasks, while the \Name{vit} dominates tasks requiring global semantic understanding.}
\label{fig2}
\end{figure}

\subsection*{A complementary pair of pretrained 3D radiological foundation models}
The nnFoundation models were developed on an internal benchmark spanning 5 segmentation, 2 detection, and 2 classification datasets, with no overlap with the main results in \cref{fig1,fig2,fig3} (Online Methods \ref{methods_2} and \ref{methods_3}). On this benchmark, masked autoencoding emerged as the most performant pretraining strategy, while convolutional and transformer backbones revealed complementary strengths across task types. Motivated by these findings, we trained two complementary foundation models on the full 2.1M-volume corpus: \Name{cnn} and \Name{vit}.

\subsection*{Large-scale pretraining establishes strong transferable representations}
\addcontentsline{toc}{subsection}{Large-scale pretraining establishes strong transferable representations}
\label{results_2}

\noindent To assess transfer performance across clinically relevant settings, we constructed a large-scale downstream benchmark comprising 108 total datasets spanning five task types: segmentation, classification, report generation, image-to-image retrieval, and detection across multiple data regimes and imaging modalities (\cref{fig1}b). Because dedicated detection fine-tuning integrations are not available for any baseline models, detection experiments are reported for nnFoundation within nnDetection (\cref{fig3}) but excluded from cross-model comparisons. This is the largest benchmark to date for evaluating medical 3D foundation models, with a downstream evaluation suite spanning 34 segmentation datasets~\cite{panther2025task1,panther2025task2,betancourt_tarifa_2025_panther,antonelli2022medical,hernandezpetsche2022isles,mahmoud2025mugliomapost,kuijf2019standardized,lesjak2018novel,cipriano2022mandibular,bolelli2025toothfairy,bolelli2025cvpr_toothfairy2,podobnik2023hanseg,wahid2024hntsmrg,bernard2018deep,zhao2008rider,garrucho2025large,roth2021covid1920,campello2021multi,moawad2023hcctace,wawtace2024,luo2022word,desai2022skm,sang2025pengwin,kustner2025longitudinalct,kazerooni2024bratspeds,verdier2024bratsgli,deeppsma2025}, 8 classification datasets~\cite{hamamci2026generalist,rsna2025intracranialaneurysm,rsna2019intracranialhemorrhage,revel2021stoic,muellerfranzes2025odelia,lldmmri2023,heller2023kits21}, 2 report generation datasets~\cite{hamamci2026generalist,Blankemeier2026}, 5 image-to-image retrieval datasets~\cite{antonelli2022medical,garrucho2025large}, 9 detection  datasets~\cite{saha2024picai,mei2022sanet,lausanne2021tofmraaneurysm,ivantsits2022cada,sudre2021valdo,moawad2023bratsmets,he2021meta,he2020dense,shao2011laparoscopic,shao2012precise,yang2024ribfrac,armato2011lidcidri,saha2018radiogenomics,mela2022}, 7 datasets introducing a domain shift~\cite{campello2021multi,bernard2018deep,care2025whs,bakas2024bratsafrica,zhang2025pansegnet,haouchine2024spinemets,cao2024bladdermri}, 5 biological imaging datasets~\cite{yang2024lungvis10,cremi,urocell2,urocell1,Chen2025SELMA3D,lapdmouse1,lapdmouse2} and 29 datasets for external validation in addition to the 9 development datasets~\cite{gatidis2022fdgpetctlesions,heller2023kits21,ji2022amos,yang2023topcow,stanfordaimi2023brainmetshare,grovik2020deep,saha2024picai,mei2022sanet,lin2023rsnacervicalspine,bien2018skmtea}, covering diverse anatomical regions, acquisition protocols, and both healthy and pathological structures (Supplementary \cref{appendix_downstream_datasets}). Together with the 2.1M-volume multimodal pretraining corpus, this represents, to our knowledge, the most comprehensive combination of 3D radiological pretraining scale and downstream evaluation breadth reported to date (\cref{fig1}c). Because datasets and task metrics differ in scale, we summarized cross-dataset performance using dataset-level ranks: models were ranked within each dataset using task-specific metrics, metric ranks were fused where multiple metrics were available, and ranks were then averaged across datasets (also see Online Methods~\ref{methods_10}). Pairwise outperformance was considered statistically supported only when the paired-bootstrap 95\% confidence interval for the corresponding nnFoundation-versus-comparator difference was entirely above zero (Online Methods~\ref{methods_10}). Baseline models and downstream fine-tuning protocols are described in Online Methods~\cref{methods_4,methods_5,methods_6}.\\

\noindent Across this comprehensive evaluation, pretrained nnFoundation models consistently outperform both models trained from scratch and existing foundation models across task types (\cref{fig1}d, \cref{fig2}, Supplementary \cref{tab:fig2-panel-rank-overview}). For segmentation, \Name{cnn} obtained the best average performance across all datasets, outperforming the next-best baseline model, CTFM, by 3.0 Dice points and the nnU-Net baseline by 1.9 Dice points. Notably, under our broad and standardized evaluation protocol, none of the competing baselines outperformed nnU-Net \cite{isensee2021nnu}, a strong dataset-adaptive segmentation baseline trained from scratch for each individual dataset. This reveals a key limitation of prior evaluations: performance gains observed in narrow benchmarks are not sufficient evidence for broadly transferable representations. In contrast, \Name{cnn} is the only foundation model in our comparison to exceed nnU-Net. For classification, \Name{vit} achieved the best mean rank across datasets and improved over training from scratch by 8.3 area under the receiver operating characteristic curve (AUROC) points. While several foundation models reached similar mean AUROC values across all classification datasets, including CTFM (0.749), CURIA (0.739), VISTA3D (0.744), and \Name{vit} (0.748), their performance varied substantially across datasets with a standard deviation of more than 0.06 for all models (Supplementary \cref{tab:fig2-summary-classification}). Averaged over all classification datasets (\cref{fig2}b), \Name{vit} achieved the strongest mean rank (2.63), followed by CTFM (3.63), CURIA (3.81), and VISTA3D (4.06), although the differences between these top-performing models were within the confidence intervals of the pairwise rank deltas.
In image-to-image retrieval (\cref{fig2}d), \Name{vit} achieved the best mean ranking, placing first or second on every dataset, with Merlin showing comparable performance. For report generation (\cref{fig2}c), \Name{vit} achieved the best overall performance across both datasets. On CT-RATE, it achieved a F1 score of 43.7 compared with 42.8 for the next-best method, CURIA. Notably, \Name{vit} is also competitive to Merlin on the MERLIN dataset itself, despite Merlin having been trained on this data, achieving a bilingual evaluation understudy (BLEU) score of 0.172 compared with 0.170.\\
\noindent nnFoundation's performance advantage was consistent across modalities and target types (\cref{fig1}d and Supplementary \cref{tab:supp_task_modality_rankings}). After stratification by imaging modality, nnFoundation models maintained the strongest performance across both CT and MRI. This included CT-only evaluation, which represents the most direct comparison to the pretraining domain of most competing baselines. A similar pattern was observed when stratifying targets into healthy anatomy and pathology-focused structures (\cref{fig1}d).
Dataset-wise results for all experiments are provided in the Supplementary (starting with \cref{tab:fig2-panel-rank-overview}).\\
\noindent Together, these results establish nnFoundation's large-scale pretraining as a general strategy for transferable 3D representations, forming the first of three factors governing performance.

\subsection*{Task-dependent performance reveals complementary model strengths}
\addcontentsline{toc}{subsection}{Task-dependent performance reveals complementary model strengths}
\label{results_3}

\noindent The results suggest that architecture remains an important determinant of downstream performance. A clear task-dependent pattern emerges, reflecting complementary strengths of convolutional and transformer-based models (\cref{fig2}). \\

\noindent For spatially localized tasks such as semantic segmentation, convolutional architectures achieve the strongest performance. Across 34 segmentation datasets, the \Name{cnn} model consistently outperforms both the \Name{vit} model and all baselines (\cref{fig2}a). The detection results (\cref{fig3}d) further confirm this observation. This reflects the strong inductive bias of convolutional networks for precise spatial modeling in 3D volumes and is consistent with prior comparisons of CNNs and transformers trained without pretraining \cite{isensee2024nnunet}. \\

\noindent In contrast, tasks requiring global semantic understanding, such as classification, favor transformer-based architectures. The \Name{vit} achieves the strongest overall global task performance, with substantial gains from pretraining (\cref{fig2}b). While convolutional models remain competitive, particularly when paired with contrastive pretraining objectives such as in CT-FM, \Name{cnn} can underperform in this regime, suggesting a mismatch between reconstruction-based pretraining and global semantic tasks in CNNs. \\

\noindent The dominance of the \Name{vit} becomes even more pronounced in settings that do not allow task-specific fine-tuning. In image-to-image retrieval and radiological report generation, where encoder features are typically not adapted to the downstream task, the \Name{vit} outperforms all baselines. These tasks require capturing global semantic relationships across the entire volume, a capability more naturally supported by transformer architectures (\cref{fig2}c-d). \\

\noindent Taken together, these results demonstrate that the observed task-dependent pattern reflects fundamental differences in how CNN and transformer architectures encode 3D medical images. Large-scale pretraining does not eliminate architectural trade-offs but instead accentuates complementary strengths: convolutional models excel at spatially precise prediction, whereas transformers provide superior global semantic representations. These findings identify architecture selection as the second key factor governing performance.

\subsection*{Dataset-aware adaptation enhances transfer across heterogeneous 3D imaging settings}
\addcontentsline{toc}{subsection}{Dataset-aware adaptation enhances transfer across heterogeneous 3D imaging settings}
\label{results_4}

\begin{figure}[h]
\centering
\includegraphics[width=\textwidth]{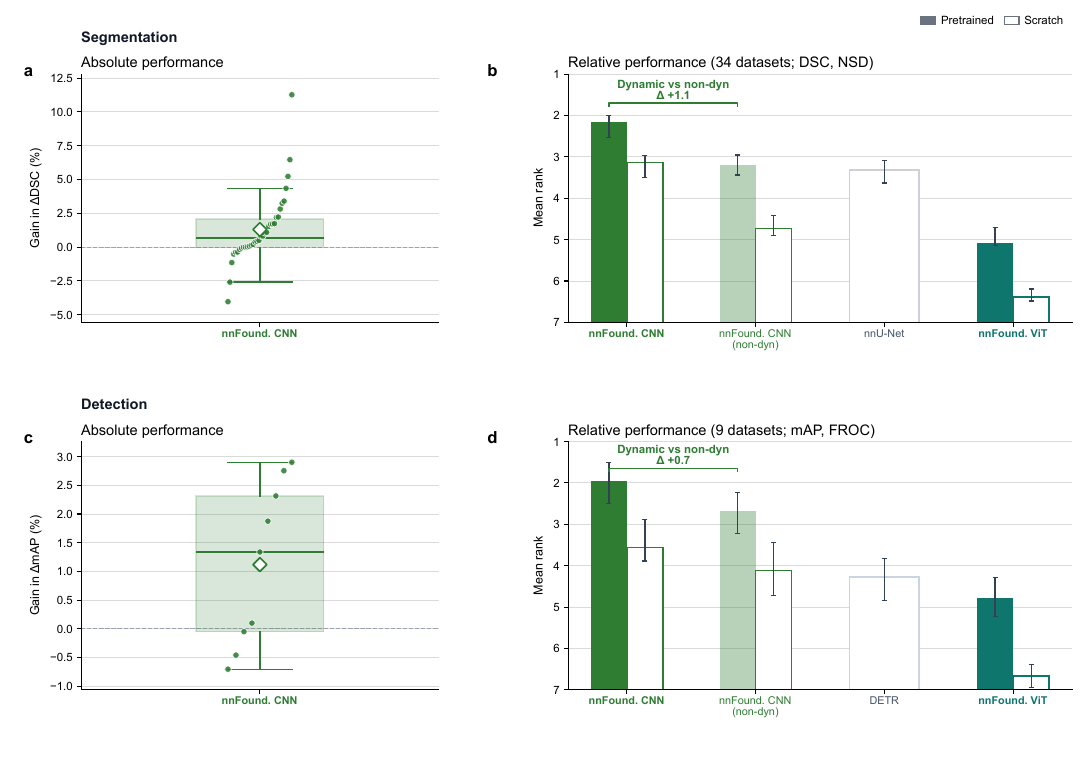}
\caption{\textbf{Dynamic adaptation enhances transfer performance.}
Dataset-specific adaptation of pretrained models improves performance compared to fixed-topology transfer across both segmentation (\textbf{a, b}) and detection (\textbf{c, d}) tasks. Building on dataset-adaptive paradigms introduced by nnU-Net and nnDetection, we extend this concept to pretrained models by introducing a heuristic-based adaptation of network topology and kernel parameters during weight transfer. This approach departs from conventional fixed-architecture foundation models and enables alignment to downstream dataset characteristics without additional training, resulting in improved transfer performance across heterogeneous 3D settings.
}\label{fig3}
\end{figure}

\noindent Beyond pretraining and architecture, performance is further shaped by alignment between model structure and dataset characteristics. In 3D radiology, downstream datasets differ substantially in spatial resolution, anisotropy, and anatomical extent, requiring dataset-specific configurations for optimal performance. Prior work, particularly nnU-Net \citep{isensee2021nnu} and nnDetection \citep{nndet}, has shown that adapting network topology to these properties is critical. Fixed-topology transfer imposes a structural mismatch that limits performance \cite{stunet}. Therefore, a key limitation of all current foundation models remains in their fixed architecture. 
 \\

\noindent We illustrate this effect with an example: the Medical Segmentation Decathlon (MSD) Pancreas dataset~\cite{antonelli2022medical}, which exhibits strong anisotropy. A static model trained from scratch performs poorly under this mismatch, achieving 56.5 Dice points. Pretraining substantially improves performance in this setting (61.8), but the fixed-topology pretrained model still underperforms a dataset-adapted nnU-Net ResEnc-L configuration trained from scratch (63.7), indicating that pretraining alone cannot compensate for architectural misalignment. Once the pretrained model is adapted to the dataset-specific topology, however, it surpasses the corresponding dynamically configured nnU-Net baseline (67.0), demonstrating that unlocking dataset-aligned architectures is necessary to fully realize the benefits of pretraining. \\

\noindent To achieve this, we introduce dynamic weight adaptation, which aligns pretrained encoders with dataset-specific target topologies at initialization. During weight transfer, pretrained kernels and encoder stages are mapped to the architecture planned for each downstream dataset using heuristic-based adaptation rules (Online Methods \ref{methods_9}). This allows a single pretrained checkpoint to be reused across heterogeneous 3D datasets while preserving the benefits of dataset-adaptive planning, rather than constraining transfer to a fixed foundation-model topology. \\

\noindent Dynamic weight adaptation increases transfer performance by an average of 1.2 Dice points across 34 segmentation datasets and improves mean average precision (mAP) by 1.2 points across 9 detection datasets~\cite{lausanne2021tofmraaneurysm,ivantsits2022cada,sudre2021valdo,moawad2023bratsmets,he2021meta,he2020dense,shao2011laparoscopic,shao2012precise,yang2024ribfrac,armato2011lidcidri,saha2018radiogenomics,mela2022} when combined with the dataset-specific topology planning in nnU-Net and nnDetection (\cref{fig3}). The strongest gains are observed for datasets with pronounced anisotropy or extreme aspect ratios as in the MSD Pancreas dataset. \\

\noindent These findings demonstrate that pretraining and architecture selection alone are insufficient for robust transfer in 3D radiology. Instead, they establish dataset-specific adaptation as the third factor governing performance. Effective deployment requires explicit alignment between model architecture and dataset characteristics, which can be achieved through zero-shot adaptation during weight transfer.

\subsection*{Pretraining improves data efficiency, convergence, and robustness}
\addcontentsline{toc}{subsection}{Pretraining improves data efficiency, convergence, and robustness}
\label{results_5}

\noindent Having established the pretraining, task-specific architecture selection, and dataset-aware adaptation as the three main factors governing performance, we next assess how they translate into performance under practical constraints. Specifically, performance was evaluated under limited annotations, in low-compute-resource settings, and under distribution shifts (also see Online Methods \ref{methods_7} and \ref{methods_8}).\\

\begin{figure}[h!]
\centering
\includegraphics[width=\textwidth]{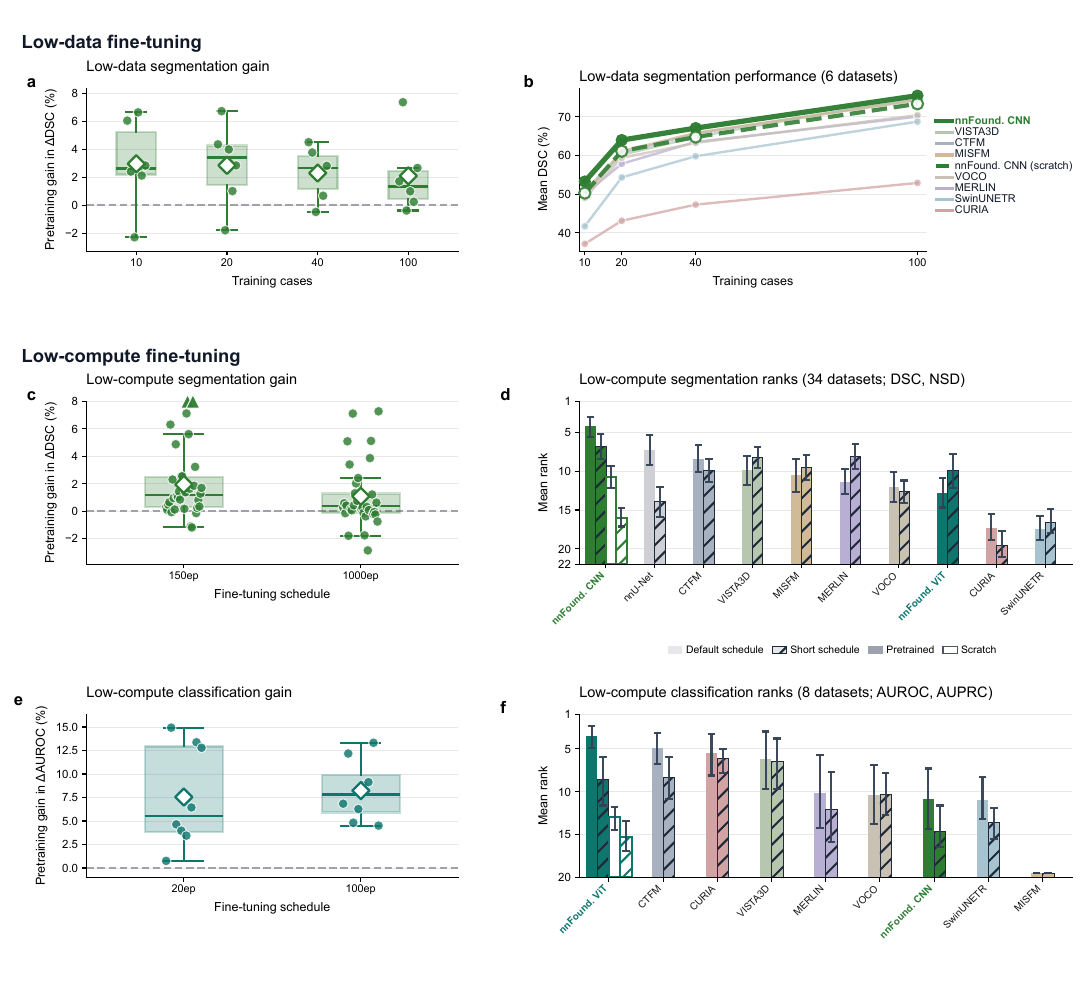}
\caption{\textbf{Pretraining improves data and compute efficiency.}
Large-scale pretraining enhances transfer not only in terms of peak accuracy, but also under the practical constraints of scarce annotations and limited fine-tuning compute.
\textbf{a, b} In low-data segmentation experiments across six datasets, pretrained \Name{cnn} models show consistent gains over training from scratch, improving performance by an average of 3 Dice points with only 10 training cases and maintaining the strongest absolute performance across label budgets.
\textbf{c, d} In low-compute segmentation experiments across 34 datasets, \Name{cnn} preserves 99.16\% of its full-schedule performance when fine-tuning is reduced from 1000 to 150 epochs and maintains the best overall rank relative to competing methods under both default and short schedules.
\textbf{e, f} In low-compute classification experiments across eight datasets, \Name{vit} improves over training from scratch under both 20- and 100-epoch schedules and achieves the strongest ranking among pretrained models, although several baselines remain competitive under short fine-tuning.
}
\label{fig4}
\end{figure}

\noindent In low-data regimes, pretrained models substantially outperform training from scratch. Across six segmentation datasets, performance gains are largest at very small training set sizes, demonstrating a strong dependence on self-supervised initialization (\cref{fig4}a). The \Name{cnn} outperforms all baselines (\cref{fig4}b) in this low-data scenario. This underscores the potential for applications where annotations are scarce or costly. \\

\noindent Pretraining also accelerates convergence and reduces training cost. When fine-tuned with only a fraction of the standard training iterations, pretrained models preserve a large proportion of their full-performance levels. For segmentation, the \Name{cnn} retains on average 99.16\% of its full fine-tuning performance using only 15\% of the training compute (\cref{fig4}c), while still outperforming all baselines, including models trained from scratch with full training schedule (\cref{fig4}d). For classification, the \Name{vit} consistently outperforms its from-scratch counterpart across both short and extended training schedules (\cref{fig4}e), although convergence is slower than for some baselines, reflecting a slower adaptation explained by the higher model capacity and reconstruction-based pretraining objective (\cref{fig4}f). This observation is consistent with prior work on masked autoencoding, which found that reconstruction-based pretraining can produce representations that are less linearly separable than contrastive representations, while becoming highly competitive once encoder layers are adapted during fine-tuning \cite{he2022mae}.

\begin{figure}[t!]
\centering
\includegraphics[width=\textwidth]{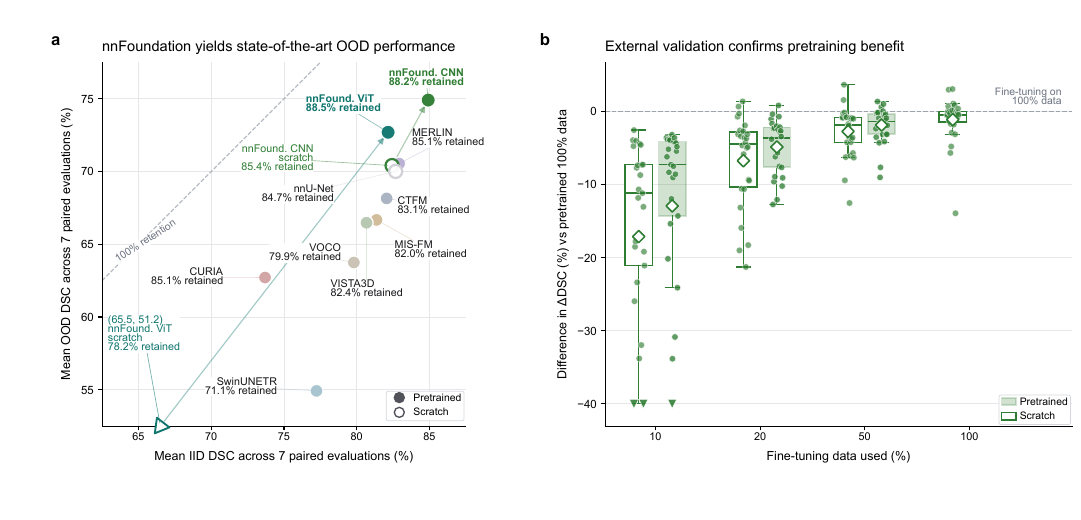}
\caption{\textbf{nnFoundation improves OOD transfer and shows consistent benefits in external validation.}
\textbf{a} Large-scale pretraining enhances transfer not only in-distribution (ID), but also in clinically realistic out-of-distribution (OOD) settings. Across paired ID and OOD evaluations, pretrained nnFoundation models show clear gains compared to trained from scratch under distribution shift, with the \Name{cnn} achieving the strongest absolute OOD performance and the \Name{vit} showing a smaller relative drop from ID to OOD performance, resulting in comparable OOD accuracy while outperforming all other baselines. 
\textbf{b} Independent external validation across heterogeneous partner pipelines further confirms consistent benefits of pretraining, with the largest gains observed in low-data regimes.
}\label{fig5}
\end{figure}

\noindent Pretraining further improves robustness under clinically realistic domain shift. To assess out-of-distribution (OOD) generalization, models were fine-tuned on one dataset and evaluated on a separate dataset, introducing shifts in site, scanner, acquisition protocol, and patient population. Across seven transfer settings (\cref{fig5}a, Supplementary \ref{app:figure5-ood-summary} and \ref{app:figure5-ood-dataset-level-results}), nnFoundation pretraining combined high in-distribution (ID) accuracy with strong retention under distribution shift. \Name{cnn} achieved the strongest overall robustness profile, reaching 74.9 OOD Dice points while retaining 88.8\% of its ID performance, corresponding to an absolute shift of only 9.5 Dice points from 84.4 ID Dice points. This made \Name{cnn} the best model in absolute OOD performance while achieving the highest mean retention. The next-best baseline model by retention, CURIA, retained 87.5\% but reached only 62.7 OOD Dice points from 71.7 ID Dice points. Conversely, the next-best foundation model by absolute OOD performance, Merlin, reached 70.5 OOD Dice points from 81.4 ID Dice points, but retained only 86.7\%. Compared with training from scratch, which retained 87.4\% and reached 70.4 OOD Dice points, nnFoundation pretraining improved both robustness ($+$1.4\%) and absolute accuracy ($+$4.5 OOD Dice). These results indicate that large-scale heterogeneous pretraining does not merely reduce sensitivity to distribution shift, but yields the strongest overall domain-shift performance among all evaluated models. Additional exploratory experiments on 3D microscopy tasks further suggest that nnFoundation representations transfer beyond radiology, with fine-tuning improving performance despite substantial shifts in modality and acquisition scale, and biological target structures (Supplementary \cref{app:extra-bio-segmentation}).\\

\noindent These findings are further supported by independent external validation across multiple academic and industrial partner institutions. \cref{fig5}b reports the segmentation results, while a lung nodule detection experiment is summarized in Supplementary \cref{tab:fig5-siemens-detection-validation}. Pretrained nnFoundation models were distributed to external partners and evaluated independently within local pipelines. Where applicable, partners received fine-tuning code, but all model training and evaluation were conducted locally at each site, without the core authors accessing the downstream data. Across settings, pretraining improved performance over training from scratch, with the largest gains observed in low-data regimes. While effect sizes varied between sites, likely due to differences in fine-tuning protocols, task-specific optimization, and integration into local pipelines, the overall benefit remained consistent. This confirms that nnFoundation pretraining generalizes beyond the controlled experimental setup and remains useful when evaluated independently at partner institutions on their own datasets, tasks, and training pipelines. \\

\subsection*{Framework integration and usability }
\addcontentsline{toc}{subsection}{Framework integration and usability }
\label{results_6}

\noindent A key barrier to the adoption of foundation models in radiological workflows is the need for substantial engineering effort to integrate pretrained models into existing pipelines. To address this, all pretrained models are directly integrated into the official nnU-Net and nnDetection repositories, two widely used frameworks for 3D medical image segmentation and detection.
This integration enables immediate use of the pretrained models within established automated planning, training, and inference workflows, without requiring architectural redesign or additional implementation effort. In particular, the models are fully compatible with dataset-specific planning and dynamic adaptation, allowing seamless alignment with downstream data characteristics.
For classification, retrieval, and vision-language applications, we provide dedicated fine-tuning pipelines within complementary task type-specific frameworks. This ensures that pretrained representations can be consistently applied across the full range of radiological tasks evaluated in this study.
By embedding foundation models directly into widely adopted tools, we enable direct translation of performance improvements into practical use, facilitating scalable development and deployment of radiological AI across diverse research and clinical settings.

\section{Discussion}
\label{discussion}

\noindent In this study, we present nnFoundation, two complementary 3D radiological foundation models pretrained on 2.1 million CT, MRI, and PET image volumes within the Human Radiome Project. Across 108 downstream tasks, nnFoundation established state-of-the-art performance across major radiological task types and improved transfer under low-data, low-compute, and domain-shift settings. These results suggest that transferable 3D radiological performance is governed by the interplay of scalable pretraining, task-aligned architectures, and dataset-aware adaptation. \\ 

\noindent Our evaluation benchmark highlights why broad, standardized validation is essential for radiological foundation models. Under this protocol, no competing foundation model surpassed nnU-Net for segmentation, despite several having previously reported competitive performance on narrower benchmarks; only \Name{cnn} exceeded this strong dataset-adaptive baseline. More broadly, the benchmark revealed a consistent task-dependent structure: convolutional architectures performed best on spatially localized tasks such as segmentation and detection, whereas transformer-based architectures excelled on tasks requiring global semantic reasoning, including classification, retrieval, and report generation from frozen features. This suggests that no single foundation-model architecture is uniformly optimal across the full spectrum of radiological tasks. More broadly, these findings challenge the assumption that increasingly capable foundation models will converge toward a single dominant architecture. Instead, different representation requirements appear to favor different architectural inductive biases even at scale.\\

\noindent Pretraining, however, was not uniformly beneficial: its value depended on the interaction between pretraining objective, architecture, and task. Although masked autoencoding emerged as the most effective pretraining strategy overall, on classification it improved the \Name{vit} substantially yet reduced the performance of the \Name{cnn} relative to its from-scratch counterpart (Fig. 2b). We attribute this negative-transfer effect to a mismatch between reconstruction-based pretraining and the global semantic representations required for classification in convolutional backbones. \\

\noindent Our results also speak to how foundation models relate to dedicated, task-specific approaches. nnFoundation provides a strong generalist, but a single generalist applied uniformly across more than one hundred downstream tasks cannot incorporate the domain-specific optimizations that individual applications admit, and such optimization will remain valuable. We therefore view foundation models not as a replacement for task-specific modeling, but as a new starting point from which specialized solutions can be developed. We expect the next generation of domain-specific models to be constructed around pretrained generalists, adding the task-specific components required for peak performance. Our results indicate that a generalist validated across breadth provides a stronger substrate for this process than a narrowly trained model.
A central implication of our results is that fixed-topology foundation models may be fundamentally mismatched to heterogeneous volumetric imaging data. Unlike natural-image benchmarks, radiological datasets vary substantially in anisotropy, spatial resolution, field of view, and anatomical coverage, creating structural mismatches that scaling alone cannot resolve.
More broadly, our results suggest that foundation models for volumetric imaging should be viewed not as fixed pretrained architectures, but as adaptable representation systems whose optimal realization depends on the downstream data. Our dynamic dataset-specific adaptation provides one practical realization of this principle, specializing a pretrained generalist to downstream datasets without additional training and improving or matching fixed-topology transfer on 27 of 34 segmentation datasets, particularly under strong anisotropy or extreme aspect ratios. \\

\noindent Realizing the impact of radiology foundation models ultimately depends on their adoption in existing workflows. To facilitate this, we release nnFoundation both as a standalone pretrained encoder and directly integrated into nnU-Net and nnDetection, the de facto standard frameworks for 3D medical image segmentation and detection. This places nnFoundation within the nn-* ecosystem of self-configuring medical imaging methods and lowers the barrier to practical deployment and evaluation.

\noindent Our study has limitations, which fall into two groups: constraints of the present evaluation, and questions deferred to future work. As constraints of the evaluation, although our benchmark is substantially broader than prior evaluations of radiological foundation models, evidence remains uneven across task families: public 3D radiological benchmarks are concentrated in segmentation and, to a lesser extent, classification, while high-quality detection, retrieval, and report-generation datasets remain scarce. Detection presents a further constraint on cross-model comparison: baseline models have generally not been evaluated in a dedicated detection framework and do not provide detection-specific integration code, protocols, or recommended fine-tuning settings, so we report detection results for nnFoundation within nnDetection but exclude detection from the main cross-model rankings. In addition, while external validation across independent partner pipelines supports the transferability of our representations, the present study is based on retrospective benchmarks, and prospective validation in real-world clinical workflows remains a crucial next step. \\
As an initial study, our scope was deliberately empirical and resource-oriented: we aimed to introduce and release the nnFoundation models, provide practical integration into established training frameworks, and evaluate their transfer behavior across a broad set of downstream benchmarks. By contrast, we did not aim to fully disentangle the causal contribution of each pretraining design choice. Several controlled ablations that would be required for such mechanistic understanding therefore lay beyond the scope of this work. In particular, we did not isolate the contributions of pretraining scale and encoder capacity, which would require retraining at multiple data and model sizes; the effect of modality composition, including modality-specific pretraining such as CT-only or MRI-only versus the full multimodal corpus and its influence on per-modality transfer; the relative roles of scale, diversity, and data curation, including model sensitivity to biased or imbalanced data; or pretraining-objective comparisons at full scale. Each of these analyses would require retraining foundation models at scale. The evaluation suite alone, covering all baselines as well as low-data, low-compute, and out-of-distribution regimes, already consumed more than 50,000 A100 GPU-hours. A full ablation grid would multiply this cost substantially, and the required computational resources were not available within the scope of the present study. The heterogeneity and variable metadata of the corpus further limit how cleanly these factors can be separated post hoc. Consequently, we cannot determine whether the observed gains arise primarily from scale, diversity, multimodality, or data curation, nor whether a modality-specific pretraining strategy would outperform the multimodal corpus for individual target modalities. Relatedly, the corpus is imbalanced across modalities, with MRI heavily represented and PET strongly underrepresented, and across sources. In addition, we did not perform formal subgroup or fairness audits across demographic or protocol-defined cohorts: characterizing representativeness and ensuring equitable performance across populations, modalities, and acquisition protocols will be essential prior to clinical translation. Finally, although we assess robustness under domain shift and out-of-distribution transfer, systematic robustness to acquisition artifacts and to rare, long-tail pathologies remains to be established through expanded and standardized benchmarks.\\

\noindent In conclusion, our results indicate that transferable 3D radiological performance emerges from the combined effect of scalable pretraining, complementary task-aligned architectures, and dataset-aware adaptation rather than from a single universal model. 
While developed for radiology, these findings may have broader implications for foundation models in volumetric biomedical imaging, where heterogeneous data characteristics similarly challenge the assumption of fixed architectures and universal transfer strategies. More broadly, our results suggest that foundation models for volumetric imaging should be viewed not as static pretrained networks, but as adaptable representation systems whose success depends jointly on scale, architecture, and alignment to downstream data.

\section{Methods}
\label{methods}

\subsection{Curating real-world clinical and internet-scale data}
\addcontentsline{toc}{subsection}{Curating real-world clinical and internet-scale data}
\label{methods_1}

\noindent To construct the pretraining dataset, we aggregated 3,271,454 volumetric imaging series from three primary sources: publicly available small datasets, large public cohorts, and clinical partner institutions. After systematic filtering and quality control, this resulted in a final dataset of 2,158,570 3D imaging volumes. \\

\noindent Given the scale and heterogeneity of the data, we applied a sequence of filtering steps to remove irrelevant, corrupted, or non-representative samples. While defining universally harmful data remains an open question, we applied a series of pragmatic filtering steps to improve data consistency and relevance for 3D radiological representation learning. First, we performed metadata-based modality filtering to retain CT, MRI, and PET, while excluding modalities that may appear volumetric but are not suitable for 3D radiological representation learning, including ultrasound, microscopy, and mammography, as well as non-image data such as segmentation masks and radiotherapy dose maps. This reduced the dataset to 3,091,403 volumes. We then removed volumes containing non-finite values (NaN or Inf; 67 volumes) and additional segmentation masks identified via low unique-value statistics (4,192 volumes). Next, we excluded volumes derived from four-dimensional image series such as diffusion MRI and time-resolved acquisitions. (458,754 volumes). \\

\noindent We subsequently applied an out-of-distribution detection pipeline to identify samples that deviate from the bulk of the dataset. Manual inspection of high-scoring outliers resulted in the exclusion of 25,456 volumes, including 9,110 non-human scans. To prevent data leakage into downstream evaluation, we performed hash-based duplicate detection and removed all volumes overlapping with downstream datasets or known benchmark collections (88,477 volumes). We then applied a body-part regression model\footnote{\url{https://github.com/MIC-DKFZ/BodyPartRegression}} to identify reconstruction failures or anatomically implausible samples, leading to the exclusion of 99,716 volumes. Finally, we applied geometric filtering by removing volumes with fewer than 10 slices along any spatial axis (229,693 volumes) and volumes with voxel spacing greater than 8 mm along any axis (26,478 volumes).

\subsection{Controlled benchmarking motivates a complementary model set}
\addcontentsline{toc}{subsection}{Controlled benchmarking motivates a complementary model set}
\label{methods_2}

\begin{figure}[t!]
\centering
\includegraphics[width=\textwidth]{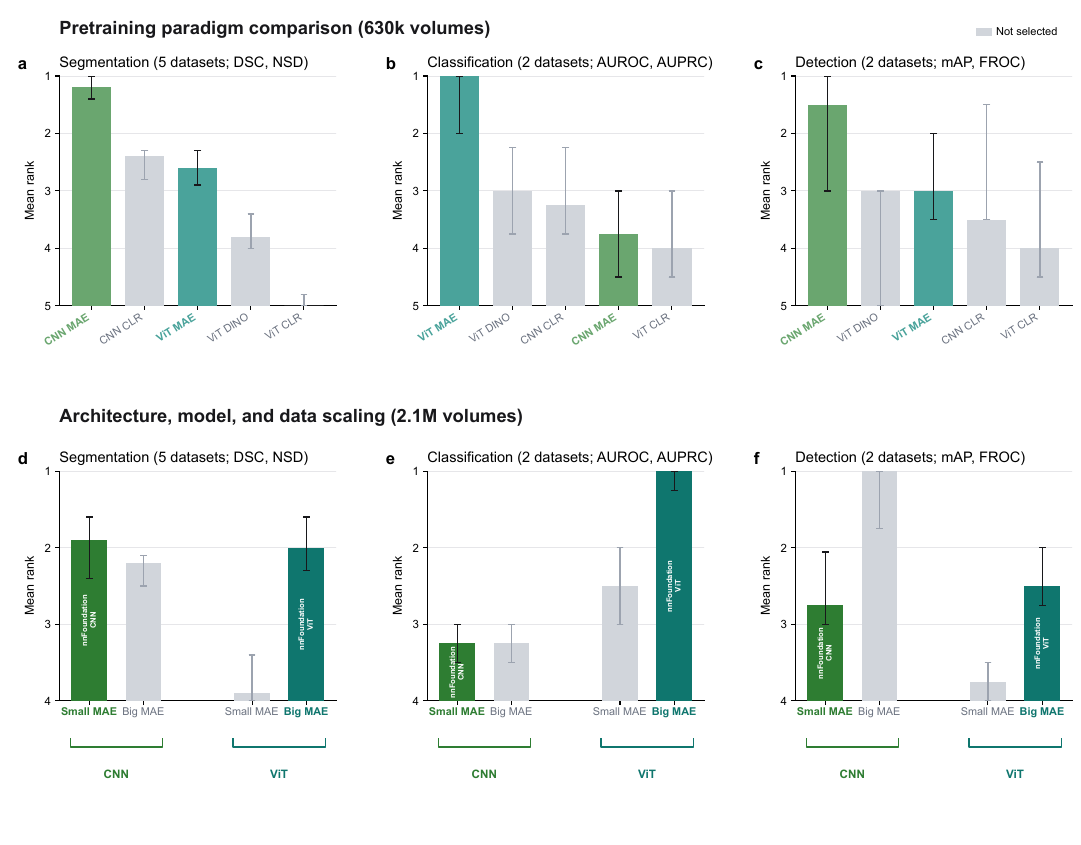}
\caption{\textbf{Pretraining paradigm comparison and scaling behavior.}
At an initial scale of 630k public volumes, MAE pretraining outperforms DINO and CLR on a development benchmark (five segmentation, two classification, two detection tasks), motivating its selection for further scaling. CNNs and transformers show complementary strengths for local and global tasks, supporting the scaling of both architectures. Scaling model size and compute yields strong gains for transformers but limited improvements for CNNs, leading to the selection of a smaller CNN and a larger transformer as final models.
}\label{fig6}
\end{figure}

\noindent Before scaling to the full dataset, we conducted a controlled proxy benchmark to evaluate architectural choices and self-supervised learning paradigms (\cref{fig6}a-c). The benchmark compared two architecture families: a CNN based on the nnU-Net ResEnc-L preset~\cite{isensee2024nnunet} and a transformer based on Primus~\cite{wald2025primus}. For both architectures, independent teams optimized MAE, DINOv2, and contrastive learning (CLR) on a public subset of approximately 630,000 3D volumes, with each team allocated approximately 25,000 A100 GPU hours for method development. \\

\noindent To ensure fairness, models were constrained to fit on a single A100 40GB GPU with batch size two during downstream evaluation, and pretraining was limited to a budget of 400 A100 GPU hours per method. As different paradigms require method-specific optimization, including masking strategies for MAE, multi-crop augmentations for DINOv2, and batch size scaling for contrastive learning, further details are provided in the Supplementary Information. \\

\noindent Downstream evaluation in this development stage covered nine dedicated validation tasks, distinct from the test datasets used for the main results, spanning segmentation~\cite{gatidis2022fdgpetctlesions,heller2023kits21,ji2022amos,yang2023topcow,stanfordaimi2023brainmetshare,grovik2020deep}, detection~\cite{saha2024picai,mei2022sanet}, and classification~\cite{lin2023rsnacervicalspine,bien2018skmtea}, providing a compact but diverse benchmark. The results revealed clear task-dependent trends: for spatially localized tasks, convolutional MAE models performed best, whereas classification tasks favored transformer-based models, again with MAE pretraining achieving the strongest performance (\cref{fig6}a,b,c). These findings motivated the selection of complementary architectures and pretraining strategies for large-scale training.

\subsection{Model training and scaling}
\addcontentsline{toc}{subsection}{Model Training and Scaling}
\label{methods_3}

\noindent Based on the controlled benchmark, we selected masked autoencoding for both architecture families and scaled the complementary CNN and transformer backbones to the full dataset. All models were trained with a global batch size of 96 on cropped input patches of size $192\times192\times192$ for 625,000 training steps. For each architecture, two model sizes were evaluated on the initial benchmark (CNN small: 102M parameters, CNN large: 696M parameters; ViT small: 205M parameters, ViT large: 674M parameters). Model size scaling substantially improved performance for the transformer models, whereas no consistent benefit was observed for the convolutional models (\cref{fig6}b). The scaled CNN outperformed the smaller variant only on the two detection tasks (\cref{fig6}f). Given that this advantage was limited to two detection datasets (vs 5 segmentation datasets), which also showed relatively high variance (Supplementary \cref{app:figure6-metric-tables}), and came with a substantial increase in fine-tuning compute requirements, we selected the smaller model as the final \Name{cnn}.

Consequently, the final \Name{cnn} corresponds to the smaller configuration with 102M parameters, while the larger transformer model was selected as the final \Name{vit}. \\

\noindent The final transformer follows the Primus design with a layer depth of 40, a token patch size of $8\times8\times8$, and an embedding dimension of 1056. Pretraining was performed with a masking ratio of 0.8 using random patch masking and a decoder of depth 8 used only during pretraining. Optimization used AdamW with weight decay $5\times10^{-2}$, a peak learning rate of $2\times10^{-3}$, 50,000 linear warm-up steps followed by polynomial decay, mixed-precision training, and gradient clipping of 1. The final training required approximately 12,500 A100 GPU hours corresponding to 130 hours on 96 GPUs. \\

\noindent The \Name{cnn} was trained using masked autoencoding with a masking ratio of 0.75 (masked patch size: 16x16x16). Optimization used stochastic gradient descent (SGD) with weight decay $3\times10^{-5}$, an initial learning rate of $1\times10^{-2}$ followed by polynomial decay, mixed-precision training, and gradient clipping of 12. In contrast to the transformer, increasing batch size did not improve performance for the CNN; the final model was therefore trained with a reduced batch size of 8, resulting in approximately 500 A100 GPU hours. For both nnFoundation models, the pretraining data was not resampled to a fixed target spacing, as the heterogeneous datasets and downstream tasks require substantially different spatial resolutions. Z-score normalization was applied to all images. \\

\subsection{Baselines}
\addcontentsline{toc}{subsection}{Baselines}
\label{methods_4}

We compared nnFoundation against recent publicly available 3D radiological foundation models spanning convolutional, Swin-transformer, ViT, and hybrid architectures. The selected baselines cover diverse pretraining paradigms, including contrastive learning, DINOv2-style self-distillation, reconstruction-based objectives, volume-fusion pretraining, vision--language pretraining, and supervised segmentation pretraining. All baselines were integrated into the same downstream evaluation pipelines, with task-specific learning rates selected from small validation sweeps, as summarized in Table~\ref{tab:baselines}. Because CURIA is a 2D image encoder, classification was performed on frozen slice-level embeddings aggregated by mean pooling, whereas segmentation used fine-tuning with slice-wise supervision.

\begin{table}[t]
\caption{Overview of baseline foundation models and final nnFoundation models (a), together with downstream settings (b). Parameter counts refer to the evaluated checkpoint and are rounded to the nearest million. Learning rates correspond to the peak learning rates selected by validation sweeps. VoCo-B was used for the reported experiments because it was the largest VoCo variant that trained reliably under our hardware constraints; VoCo-L did not converge in our setup, while VoCo-H exceeded the available GPU memory. Abbreviations: Arch., architecture; Pretr., pretraining; Vol., volumes; Seg., segmentation; Cls., classification; Ret., retrieval; Rep., report generation; FOV, field of view; lr, learning rate; avg, average pooling; max, maximum pooling; std, standard-deviation pooling.}
\label{tab:baselines}
\centering

{\fontsize{5.8}{6.6}\selectfont
\setlength{\tabcolsep}{1.4pt}
\renewcommand{\arraystretch}{0.95}

\begin{tabular*}{\linewidth}{@{\extracolsep{\fill}}lllllll@{}}
\multicolumn{7}{@{}l}{\textbf{a. Baseline characteristics}}\\
\toprule
Name & Params & Arch. & Pretr. Objective & Modality & Pretrain Vol. & Pretr. patch \\
\midrule
CTFM~\cite{pai2025vision} & 87M & 3D CNN & Contrastive & CT & 148k & [24, 128, 128] \\
CURIA~\cite{dancette2025curia} & 86M & 2D ViT & DINOv2 & CT, MRI & 1M & [1, 512, 512] \\
MERLIN~\cite{Blankemeier2026} & 271M & 3D CNN & Vision--language & CT & 15k & [160, 224, 224] \\
MIS-FM~\cite{wang2023mis} & 142M & 3D Hybrid & Volume fusion & CT & 110k & [64, 128, 128] \\
SwinUNETR~\cite{tang2022self} & 102M & 3D Swin-T & Inpaint.+contr.+rot. & CT & 5k & [96, 96, 96] \\
VISTA3D~\cite{he2025vista3d} & 218M & 3D CNN & Sup. segmentation & CT & 11k & [128, 128, 128] \\
VoCo-B~\cite{wu2024voco,voco2} & 54M & 3D Swin-T & Vol. contrastive & CT & 160k & [96, 96, 96] \\
\midrule
\Name{cnn} & 102M & 3D CNN & MAE & CT, MRI, PET & 2.1M & [192, 192, 192] \\
\Name{vit} & 674M & 3D ViT & MAE & CT, MRI, PET & 2.1M & [192, 192, 192] \\
\bottomrule
\end{tabular*}

\vspace{0.8em}

\begin{tabular*}{\linewidth}{@{\extracolsep{\fill}}lllllll@{}}
\multicolumn{7}{@{}l}{\textbf{b. Downstream settings}}\\
\toprule
Name & Seg. lr & Cls. lr & Ret. FOV & Ret. pool & Rep. lr & Rep. FOV \\
\midrule
CTFM & 2e-4 & 1e-4 & 192-crop & avg & 4e-4 & 192-crop \\
CURIA & 3e-4 & 1e-1 & resize & std & 1e-3 & resized \\
MERLIN & 5e-5 & 3e-4 & native crop & max & 4e-4 & native crop \\
MIS-FM & 1e-3 & 1e-3 & 192-crop & avg & 4e-4 & 192-crop \\
SwinUNETR & 3e-4 & 3e-4 & 192-crop & max & 4e-4 & 192-crop \\
VISTA3D & 5e-5 & 3e-4 & native crop & std & 4e-4 & 192-crop \\
VoCo-B & 3e-4 & 1e-3 & native crop & max & 4e-4 & 192-crop \\
\midrule
\Name{cnn} & 1e-3 & 3e-4 & resize & avg & 4e-4 & resized \\
\Name{vit} & 1e-3 & 1e-4 & 192-crop & avg & 1e-3 & native crop \\
\bottomrule
\end{tabular*}
}
\end{table}

\subsection{Downstream Experiments}
\addcontentsline{toc}{subsection}{Downstream Experiments}
\label{methods_5}

\subsubsection{Segmentation}
\addcontentsline{toc}{subsubsection}{Segmentation}
\label{methods_5_1}

\noindent For the large-scale downstream benchmark, we evaluated segmentation on 34 datasets, excluding the five datasets used in the initial development-stage benchmark. All experiments were conducted within the nnU-Net framework using a 50/50 train-test split. Unless noted otherwise, baselines and fixed-topology \Name{cnn} models were fine-tuned for 1,000 epochs using 250 steps per epoch, a batch size of 2, and input patches of size $192 \times 192 \times 192$. In contrast, dynamically adapted \Name{cnn} and nnU-Net models followed the dataset-specific nnU-Net-configured patch size. All data were resampled to the dataset-specific target spacing determined by nnU-Net and intensity-normalized using z-score normalization. For the reduced-compute experiments, the same protocol was used, but training was limited to 150 epochs. \\

\noindent For all methods, only encoder weights were transferred from the pretrained foundation model, while decoder weights were initialized randomly. The only exceptions were SwinUNETR and VOCO, which use a SwinUNETR(V2) backbone and therefore relied on their original weight-loading procedure. Training employed a linear warm-up over the first 15\% of epochs, followed by polynomial learning-rate decay. Peak learning rates were chosen method-specifically, either following the original work or determined by a small learning-rate sweep on the development datasets, and are summarized in \cref{tab:baselines}. For the pretrained models, we followed the optimization settings of the respective original works and used AdamW throughout. For nnU-Net models trained from scratch, we used the recommended SGD optimizer. For the \Name{cnn}, we used the default nnU-Net SGD setup for the full 1,000-epoch fine-tuning experiments but switched to AdamW for the 150-epoch reduced-compute experiments to enable a fairer comparison with the pretrained baselines. If a baseline showed unstable training or clear failure on a given dataset, the peak learning rate was reduced by a factor of 10.

\subsubsection{Classification}
\addcontentsline{toc}{subsubsection}{Classification}
\label{methods_5_2}

 \noindent For the large-scale downstream benchmark, we evaluated classification on eight datasets, excluding the two datasets used during the validation phase of the initial development-stage benchmark. All models were evaluated under full fine-tuning using a 40/10/50 train/validation/test split. For CT-RATE we used the official test split. Input volumes were resampled to 1 mm isotropic resolution and cropped to $192 \times 192 \times 192$ around the target region identified with tools like TotalSegmentator for organ structures and HD-BET for brain extraction. The encoder output was processed with adaptive average pooling followed by a multilayer perceptron (MLP) classification head. Training used AdamW, mixed precision, default nnU-Net augmentations, and a linear warm-up over the first 10\% of epochs followed by cosine learning-rate decay. Models were trained for 100 epochs with 50 steps per epoch and an effective batch size of 32 implemented via gradient accumulation. For reduced-compute experiments, the same setup was used with 20 epochs. Peak learning rates were optimized on an internal CT-RATE validation split for all methods and are summarized in \cref{tab:baselines}.

\subsubsection{Detection}
\addcontentsline{toc}{subsubsection}{Detection}
\label{methods_5_3}

\noindent For the large-scale downstream benchmark, we evaluated detection on nine datasets, excluding the two datasets used during the validation phase of the initial development-stage benchmark. All experiments were performed within the nnDetection framework using the default Deformable DEtection TRansformer-based (DETR) detector. Because related 3D foundation models had not previously been assessed in a dedicated detection framework, we restricted comparisons to this common detector setup to isolate the effect of encoder pretraining. For datasets with an official test split, we retained the official test set and split the remaining training data into an 80/20 train/validation split. For the development dataset PN9, we used the official train/validation/test split. For datasets without an official test split, we used a 40/10/50 train/validation/test split. 
For all methods, only encoder weights were transferred from the pretrained models, while the remaining detector components were initialized randomly. We used the target spacing and normalization scheme recommended by nnDetection. Static models were trained with a fixed patch size of $128 \times 128 \times 128$, whereas dynamically adapted models followed the dataset-specific nnDetection planning configuration. All models were trained for 100 epochs with 2500 steps per epoch and a batch size of 4. We used AdamW, mixed-precision training, and the default nnDetection augmentation pipeline. Primus models and all pretrained ResEnc models were trained with a linear warm-up over 10,000 steps followed by polynomial learning-rate decay. ResEnc models trained from scratch used polynomial learning-rate decay without warm-up, except on VALDO and PN9, where the nnDetection default configuration includes a 10,000-step warm-up. Peak learning rates were set to $1\times10^{-4}$ for Primus and $3\times10^{-4}$ for ResEnc-L, except for the VALDO and PN9 ResEnc from-scratch experiments, where the default nnDetection configuration uses a learning rate of $1\times10^{-4}$ to ensure stable convergence. Because some models showed signs of overfitting, we selected the best checkpoint based on validation performance.

\subsection{Field of View for report generation and image-to-image retrieval}
\addcontentsline{toc}{subsection}{Field of View for report generation and image-to-image retrieval}
\label{methods_6}
\noindent To accommodate differences in input resolution and field of view across encoders, we standardized preprocessing and evaluated multiple field of view configurations per encoder. Volumes were resampled to each encoder's native target spacing, or to 1 mm isotropic when no fixed native spacing was defined during pretraining (as is the case for the nnFoundation models) and normalized using its original intensity scheme (z-score for \Name{vit}, \Name{cnn}, SwinUNETR, VoCo, CTFM, MisFM, and Curia; fixed-range scaling to $[0, 1]$ for VISTA3D and Merlin). Each volume was then cropped to the bounding box of an anatomical region-of-interest (ROI), expanded by 4\% to include surrounding tissue. The ROI was defined as the union of TotalSegmentator \cite{wasserthal2023totalsegmentator} lung labels on CT-RATE and as the union of all TotalSegmentator foreground labels on MERLIN or the corresponding target organ for the retrieval datasets with connected components smaller than 400 voxels removed as outliers prior to bounding-box extraction.

\noindent We evaluated three field of view modes on the ROI crop at each encoder's native spacing: (i) a resize mode, which trilinearly interpolated the ROI crop to the encoder's native patch size, (ii) a center crop to the encoder's native patch size, and (iii) a center crop to a fixed $192 \times 192 \times 192$ patch size. Because the resulting MERLIN ROI crops span the full abdomen and substantially exceed all encoders' native patch sizes, only the resize mode was evaluated on MERLIN.

\noindent Two encoder-specific adaptations were required. Because Curia is a 2D encoder, its per-slice spatial tokens were aggregated by mean pooling into 8, 12, or 16 axial bins to obtain a volumetric token grid, with the number of bins swept alongside the field of view modes. For the Merlin encoder evaluated on the MERLIN dataset, on which it was originally pretrained, we used the preprocessing of its original pretraining pipeline~\cite{Blankemeier2026}, which brings volumes to a fixed $224 \times 224 \times 160$ shape via per-axis padding and center-cropping without any ROI crop, to avoid a distribution shift relative to its pretraining conditions.

\subsubsection{Report Generation}
\addcontentsline{toc}{subsubsection}{Report Generation}
\label{methods_6_1}

\noindent Report generation experiments were conducted on the CT-RATE and MERLIN datasets. For all methods, the image encoder was kept fully frozen and used to replace the original vision encoder in Qwen2.5-VL-3B. Vision-language alignment is facilitated by a 3D-optimized adapter (following Qwen2.5-VL). It utilizes a spatial-to-channel pooling mechanism where $2^3$ neighboring representations are concatenated and projected into the large language model's (LLM) hidden space. This configuration ensures that local volumetric context is preserved and properly scaled for the language model, while the underlying image encoder remains fixed. This protocol isolates the semantic quality of the pretrained visual representations by preventing task-specific adaptation of the image encoder. To account for differences in representation structure across encoders, we performed a sweep over learning rates and input fields of view for each model on internal validation splits. All reported results correspond to the best-performing learning rate and field-of-view configuration for each encoder, as summarized in \cref{tab:baselines}. \\
Training was performed with supervised next-token prediction in two stages. In the first stage, the adapter was trained for 1.2 dataset epochs to align the frozen visual features with the language model input space, serving as a preemptive measure to avoid feature collapse of the LLM. In the second stage, Low-Rank Adaptation (LoRA) fine-tuning of the language model was performed for 10.8 additional epochs while keeping the image encoder frozen. The adapter remained trainable throughout both stages.

\subsubsection{Image-to-image retrieval}
\addcontentsline{toc}{subsubsection}{Image-to-image retrieval}
\label{methods_6_2}

\noindent For image-to-image retrieval, we followed the MedImageInsight benchmark and evaluated retrieval on its default datasets, including Medical Segmentation Decathlon Liver, Colon, Pancreas, and Lung, and additionally extended the benchmark with the MAMA-MIA dataset. For the MAMA-MIA dataset, we derived clinically meaningful retrieval targets by assigning tumor stage labels based on tumor volume measured from the segmentation masks and grouping cases into three stages. All images were cropped to a single breast, irrespective of whether a tumor mask was present, and crops without tumor involvement were used as tumor-free samples. In contrast to the original MedImageInsight benchmark, where negative samples originate from a different dataset, this design avoids introducing an additional dataset-shift confounder into the tumor retrieval task. \\
No task-specific training was performed for retrieval, and all encoders were evaluated in a fully frozen setting. Retrieval was based on pooled image embeddings derived from encoder features. We evaluated multiple aggregation strategies, including average pooling, max pooling, median pooling, and standard-deviation pooling, and additionally varied the input field of view as described in \cref{methods_6}. For each encoder, we report the best-performing combination of field of view and aggregation strategy, as summarized in \cref{tab:baselines}.

\subsection{Label and compute efficiency experiments}
\addcontentsline{toc}{subsection}{Label and compute efficiency experiments}
\label{methods_7}
\noindent To evaluate label efficiency, we selected six segmentation datasets and randomly sampled 10, 20, 40, and 100 training cases from a fixed 50/50 train/test split. For each subset size, three random seeds were used, and the same sampled subsets were shared across all methods within each seed. All low-data experiments otherwise followed the standard segmentation protocol, except that training was limited to 300 nnU-Net epochs with 250 steps per epoch. To evaluate compute efficiency, we repeated the downstream fine-tuning experiments under reduced training budgets for both segmentation and classification. Segmentation training was reduced from 1,000 to 150 epochs, and classification training from 100 to 20 epochs, while all other settings remained unchanged.

\subsection{Robustness under domain shift}
\addcontentsline{toc}{subsection}{Robustness under domain shift}
\label{methods_8}

\noindent To evaluate robustness beyond in-distribution performance, we constructed a set of out-of-distribution segmentation transfer experiments using the default segmentation fine-tuning setup described above. To reduce the influence of source-domain overfitting, all OOD experiments used the reduced-compute setting with 150 fine-tuning epochs instead of 1,000.

\noindent We included seven transfer settings spanning multiple clinically relevant types of distribution shift. These included held-out-center transfer for heart structures segmentation in cardiac MRI, whole-heart segmentation in CT and MRI, bladder cancer segmentation in MRI, and pancreas segmentation in MRI; cross-dataset transfer for vertebral segmentation in CT; heart structures segmentation in cardiac MRI and a combined dataset; and cross-geographic population, country and site shift for brain glioma segmentation in MRI. Across all settings, source and target tasks were chosen such that label definitions were aligned, enabling direct transfer evaluation without modifying the prediction target.

\subsection{Dynamic adaptation for \Name{cnn}}
\addcontentsline{toc}{subsection}{Dynamic adaptation for \Name{cnn}}
\label{methods_9}
\noindent All segmentation and detection experiments were conducted within the standardized nnU-Net and nnDetection evaluation pipelines. These frameworks support dataset-adaptive planning, in which target spacing and downstream topology, including patch size, kernel sizes, pooling scheme, and network depth, are derived from dataset properties such as spacing, image shapes, and anisotropy. As the base architecture of the \Name{cnn} follows the ResEnc-L configuration of nnU-Net, we enabled transfer to these planned target architectures through dynamic weight adaptation at initialization. When planned convolutional kernels were smaller than in the pretrained model, weights were adapted by averaging along reduced spatial axes, for example, when reducing a 3×3×3 kernel to 1×3×3. When encoder depth differed, only overlapping stages were transferred, while additional downstream stages remained randomly initialized and excess pretrained stages were discarded. For input-channel mismatches, input projection weights were repeated and rescaled, and learnable positional embeddings were resized by trilinear interpolation when patch size changed.

\subsection{Evaluation}
\addcontentsline{toc}{subsection}{Evaluation}
\label{methods_10}
\noindent We internally evaluated nnFoundation across 58 downstream tasks spanning segmentation (34), classification (8), report generation (2), image-to-image retrieval (5), and detection (9), as well as experiments assessing dynamic adaptation, low-data and low-compute transfer, and robustness under domain shift.

\noindent Complementary task-native metrics were used throughout. Segmentation was evaluated using Dice similarity coefficient (DSC) and normalized surface Dice (NSD), macro-averaged over foreground classes within each dataset. Classification was evaluated using AUROC and area under the precision-recall curve (AUPRC), computed one-vs-rest for multi-class tasks and label-wise for multi-label tasks, with macro-averaging across classes or labels. Detection was evaluated using mean average precision over intersection-over-union (IoU) thresholds from 0.10 to 0.50 in steps of 0.05 and free-response receiver operating characteristic (FROC) performance, defined as the mean sensitivity at 1/8, 1/4, 1/2, 1, 2, 4, and 8 false positives per image at IoU 0.10. Report generation was evaluated using corpus-level BLEU. Where structured annotations were available, we additionally computed abnormality-label F1 and the clinical report generation (CRG) score. Retrieval was evaluated using average precision for each retrieval task.

\noindent Model rankings were computed by averaging dataset-level ranks with equal weight. Within each dataset, models were ranked separately for each metric, with ties assigned average ranks. Metric ranks were averaged and re-ranked to obtain a single rank per dataset, ensuring that each dataset contributed one rank on the same ordinal scale. This avoided giving disproportionate influence to datasets with larger sample sizes, broader metric ranges, or larger absolute performance differences. Final task-family rankings were obtained by averaging these dataset-level ranks across datasets.

\noindent 95\% confidence intervals were calculated using bootstrap resampling with 1,000 replicates. Datasets were held fixed, and resampling was performed within each dataset at the case level with known hierarchical structures preserved, such as patient-level clustering. Metrics were recomputed and passed through the full metric-to-rank aggregation procedure within each bootstrap replicate. Confidence intervals for the summary rankings were calculated from the resulting bootstrap distribution of mean ranks.

\noindent The performance of nnFoundation was further assessed using paired bootstrap intervals for differences between the nnFoundation reference and each comparator in the Supplementary Detailed Results. Outperformance was claimed only when the 95\% confidence interval from the paired bootstrap for the corresponding nnFoundation-versus-comparator difference was entirely above zero.

\backmatter

\section*{Affiliations}
\label{affil}
\begingroup
\parindent=0pt
\savedaffiliations
\endgroup

\newcounter{savedpage}
\setcounter{savedpage}{\value{page}}



# Authorship and Credit Attribution

Contributions are organised by work package and category. Core contributors made sustained, central contributions. Contributors provided meaningful support. Core contributors and contributors are sorted alphabetically.

## Project Management

*Leads:* Klaus H. Maier-Hein, Fabian Isensee

**Project supervision**
*Core contributors:* Fabian Isensee, Klaus H. Maier-Hein
*Contributors:* Ralf Floca, Paul F. Jäger, Lena Maier-Hein, Marco Nolden, Philipp Vollmuth

**Project conceptualisation**
*Core contributors:* Fabian Isensee, Paul F. Jäger, Klaus H. Maier-Hein, Constantin Ulrich, Tassilo Wald

**Project coordination**
*Core contributors:* Nina Decker, Fabian Isensee

**Paper writing**
*Core contributors:* Fabian Isensee, Marcel Knopp, Klaus H. Maier-Hein, Constantin Ulrich, Philipp Vollmuth
*Contributors:* Kris Dreher, Katharina Eckstein, Karol Gotkowski, Yannick Kirchhoff, Maximilian Rokuss, Numan Saeed, Elisa Stegmeier, Tassilo Wald, Mohammad Yaqub

## Data Acquisition & Governance

*Lead:* Maximilian Rokuss[1], *Initial Lead:* Tassilo Wald[2]

**Clinical data sourcing**
*Core contributors:* Bálint Kovacs, Ignaz Reicht, Maximilian Rokuss, Raphael Stock
*Contributors:* Klaus H. Maier-Hein

**Open-source data sourcing**
*Core contributors:* Kim-Celine Kahl, Raphael Stock, Tassilo Wald, Moritz Langenberg

**Big-cohort data sourcing**
*Core contributors:* Stefan Dvoretskii, Benjamin Hamm, Yannick Kirchhoff, Tassilo Wald
*Contributors:* Marco Nolden

**Downstream dataset preparation**
*Contributors:* Alexandra Ertl, Maximilian Fischer, Benjamin Hamm, Jessica Kächele, Yannick Kirchhoff, Bálint Kovacs, Moritz Langenberg, Robin Peretzke, Maximilian Rokuss

[1] Lead from 07/2024
[2] Lead from 02/2024 to 07/2024, switch to lead Pre-training

**Data governance & compliance**
*Core contributors:* Ralf Floca, Julius C. Holzschuh, Maximilian Rokuss, Dasha Trofimova, Tassilo Wald

## Data Infrastructure

*Lead:* Karol Gotkowski

**Dataset conversion**
*Core contributors:* Stephen Schaumann, Elisa Stegmeier, Sebastian Ziegler
*Contributors:* Dimitrios Bounias, Nico Disch, Katharina Eckstein, Alexandra Ertl, Benjamin Hamm, Jessica Kächele, Bálint Kovacs, Moritz Langenberg, Raphael Stock, Shuhan Xiao, Maximilian Zenk

**Database systems**
*Core contributors:* Karol Gotkowski, Philipp Schader
*Contributors:* Michael Baumgartner, Anand Deshpande, Elisa Stegmeier

**Query systems & metadata analysis**
*Core contributors:* Elisa Stegmeier
*Contributors:* Anand Deshpande, Karol Gotkowski, Marco Nolden, Philipp Schader

**Storage systems**
*Core contributors:* Ralf Floca, Karol Gotkowski
*Contributors:* Michael Baumgartner

**Infrastructure tooling**
*Contributors:* Stefan Dinkelacker, Jan Sellner

**IT coordination**
*Core contributors:* Ralf Floca, Markus Hohenhaus, Fabian Isensee, Dimitri Koslov

## Data Cleaning & Quality Control

*Lead:* Constantin Ulrich

**Out-of-distribution detection**
*Core contributors:* Selen Erkan, Elisa Stegmeier, Constantin Ulrich, David Zimmerer

**Visual inspection**
*Core contributors:* Stefan Dinkelacker, Elisa Stegmeier, Constantin Ulrich, David Zimmerer

**LLM-assisted metadata curation**
*Core contributors:* Selen Erkan, Elisa Stegmeier
*Contributors:* Rajesh Baidya, Stefan Dvoretskii, Philipp Schader

**Deduplication**

*Core contributors:* Nico Disch, Yannick Kirchhoff, Elisa Stegmeier, Constantin Ulrich

**Other curation steps**

*Core contributors:* Moritz Langenberg, Elisa Stegmeier, Constantin Ulrich

## Model Pre-training

*Leads:* Constantin Ulrich, Tassilo Wald

**Self-supervised learning methods**

*Core contributors:* Karol Gotkowski, Constantin Ulrich, Yannick Kirchhoff, Raphael Stock, Dasha Trofimova, Tassilo Wald

*Contributors:* David Zimmerer

**Training infrastructure & HPC compute acquisition**

*Core contributors:* Karol Gotkowski, Fabian Isensee, Yannick Kirchhoff, Raphael Stock, Dasha Trofimova, Constantin Ulrich, Tassilo Wald

*Contributors:* Jeremias Traub

**Data preparation for pre-training**

*Core contributors:* Karol Gotkowski, Raphael Stock, Constantin Ulrich, Tassilo Wald

## Model Fine-tuning

*Leads:* Constantin Ulrich, Yannick Kirchhoff, Karol Gotkowski, Tassilo Wald

**Segmentation fine-tuning**

*Core contributors:* Jonathan Deissler, Maximilian Fischer, Karol Gotkowski, Constantin Ulrich

*Contributors:* Lars Krämer, Tassilo Wald

**Classification fine-tuning**

*Core contributors:* Karol Gotkowski, Yannick Kirchhoff, Dasha Trofimova, Constantin Ulrich, Tassilo Wald, Sebastian Ziegler

*Contributors:* Stefan Denner

**Detection fine-tuning**

*Core contributors:* Katharina Eckstein, Constantin Ulrich

**Downstream partner coordination**

*Core contributors:* Yannick Kirchhoff, Constantin Ulrich

**Foundation model baseline implementation**

*Contributors:* Jonathan Deissler, Partha Ghosh, Karol Gotkowski, Moritz Langenberg, Maximilian Rokuss, Jeremias Traub, Constantin Ulrich

## Vision-Language Models & Retrieval

**VLM development & training**

*Core contributors:* Stefan Denner, Selen Erkan, Kim-Celine Kahl, Moritz Langenberg, Raphael Stock, Jonathan Suprijadi, David Zimmerer

*Contributors:* Julius C. Holzschuh, Constantin Ulrich

**Embedding generation & retrieval**

*Core contributors:* Stefan Denner, Moritz Langenberg

*Contributors:* Raphael Stock, Constantin Ulrich

## Validation & Benchmarking

*Leads:* Yannick Kirchhoff, Marcel Knopp

**Dataset selection**

*Core contributors:* Fabian Isensee, Yannick Kirchhoff, Constantin Ulrich

*Contributors:* Marcel Knopp, Tassilo Wald

**Validation strategy**

*Core contributors:* Yannick Kirchhoff, Marcel Knopp

*Contributors:* Evangelia Christodoulou, Lena Maier-Hein, Tim Rädsch, Annika Reinke, Constantin Ulrich

**Results analysis & reporting**

*Core contributors:* Marcel Knopp

*Contributors:* Kris Dreher, Numan Saeed, Mohammad Yaqub

## Project Partners

**External validation**

*PIs:* Arnd Heuser, Martin Reuter, Julia A. Schnabel, Tobias Heimann, Florin Ghesu, Paul Brachmann

*Core contributors:* Laura Alexandra Daza Barragan, Michael Baumgartner, Guillaume Chabin, Santiago Estrada, Cristina Isabel Gonzalez Osorio, Matthias Kirchler, Valentin Koch, David Kügler, Jan Peeken, Yoland Savriama, Marvin Teichmann

**Data providers**

*PIs:* Heinz-Peter Schlemmer, Kai Schlamp, Michael Bach, Bram Stieltjes, Philipp Vollmuth, Claus P. Heußel, Alexander Radbruch

*Core contributors:* Gianluca Brugnara, Martha Foltyn-Dumitru, Julius C. Holzschuh, Aditya Rastogi, Ignaz Reicht



\setcounter{page}{\value{savedpage} + 2}

\bmhead*{Acknowledgments}

\bmhead*{Funding}
\noindent C.U., Y.K., K.G. and J.D. disclose support for the research of this work from the Helmholtz Association under the Helmholtz Foundation Model Initiative (HFMI), project ``The Human Radiome Project'' (THRP). Y.K., B.K. and N. Disch disclose support for the research of this work from the Helmholtz Association under the joint research program ``HIDSS4Health -- Helmholtz Information and Data Science School for Health''. T.W., K.G., M. Rokuss, D.T., R.S., K.-C.K., S.Z., S. Erkan, D.Z., J. Suprijadi, J.T., L.K., K.D., A. Reinke and F.I. disclose support for the research of this work from HELMHOLTZ IMAGING, a platform of the Helmholtz Information \& Data Science Incubator. A.D., S.Dvoretskii and M.N. disclose support for the research of this work from HELMHOLTZ METADATA COLLABORATION (HMC), a platform of the Helmholtz Information \& Data Science Incubator.

\noindent Y.K. discloses support for the research of this work from the Deutsche Forschungsgemeinschaft (DFG, German Research Foundation; 402688427). M.F. and M. Baumgartner disclose support for the research of this work from the Deutsche Forschungsgemeinschaft (DFG, German Research Foundation; 410981386). A.E. discloses support for the research of this work from the Deutsche Forschungsgemeinschaft (DFG, German Research Foundation; 428224476).

\noindent E.S. discloses support for the research of this work from the Medical Informatics Initiative project ``PrivateAIM'' (01ZZ2316M). P.S., S. Denner and B.H. disclose support for the research of this work from NUM 3.0 (01KX2524). S. Denner, B.H. and R.B. disclose support for the research of this work from NUM 2.0 (01KX2121). R.S., B.K., D.B., R.P., F.I. and N. Decker disclose support for the research of this work from the NCT Heidelberg Image Data Analysis Infrastructure (IDAI). K.E. and J.K. disclose support for the research of this work from the German Cancer Consortium (DKTK, Strategic Initiative Joint Imaging Platform). K.E. discloses support for the research of this work from the European Union project EUCAIM (101100633). M.F. discloses support for the research of this work from Research Campus M2OLIE, funded by the German Federal Ministry of Research, Technology and Space (BMFTR) within the framework ``Research Campus -- Public-Private Partnership for Innovation'' (13GW0388A).

\noindent M. Rokuss discloses support for the research of this work from the Google PhD Fellowship Program. T.R. discloses support for the research of this work from a scholarship from the Hanns Seidel Foundation, funded by the German Federal Ministry of Research, Technology and Space (BMFTR). J.C.H. discloses support for the research of this work from a fellowship of the DKFZ Clinician Scientist Program, supported by the Dieter Morszeck Foundation. P.V. discloses support for the research of this work from an Else Kr\"oner Clinician Scientist Endowed Professorship from the Else Kr\"oner-Fresenius Foundation (2022 EKCS.17). A. Rastogi discloses support for the research of this work from the Bonfor Startup Postdoc Fellowship (2024-1B-10). L.A.D.B. discloses support for the research of this work from the German Federal Ministry of Research, Technology and Space (DECIPHER-M, 01KD2420G).

\noindent M. Knopp, S.S., M.L., S.X., J. Sellner, M.Z., P.G., S. Dinkelacker, E.C., L.M.-H., R.F., P.J., K.M.-H., H.-P.S., I.R., M.H., D. Koslov, K.S., M. Bach, B.S., M.F.-D., G.B., C.P.H., A. Radbruch, A.H., Y.S., M. Reuter, S. Estrada, D. K\"ugler, J.A.S., C.I.G.O., J.P., T.H., F.G., M.T., G.C., P.B., M. Kirchler, V.K., M.Y. and N.S. declare no relevant funding.

\bmhead*{Dataset acknowledgments}
\noindent Data used in this work were obtained from the A4 Study and companion Longitudinal Evaluation of Amyloid Risk and Neurodegeneration (LEARN) Study. The A4 Study was funded by a public-private-philanthropic partnership, including the National Institutes of Health--National Institute on Aging, Eli Lilly and Company, the Alzheimer's Association, the Accelerating Medicines Partnership, the GHR Foundation, an anonymous foundation and additional private donors, with in-kind support from Avid Radiopharmaceuticals, Cogstate, Albert Einstein College of Medicine and the Foundation for Neurologic Diseases. The LEARN Study was funded by the Alzheimer's Association and the GHR Foundation. The A4 and LEARN Studies were led by Reisa Sperling at Brigham and Women's Hospital, Harvard Medical School, and Paul Aisen at the Alzheimer's Therapeutic Research Institute (ATRI), University of Southern California, coordinated by ATRI, and made available under the auspices of the Alzheimer's Clinical Trial Consortium through the Global Research \& Imaging Platform (GRIP). The authors acknowledge the dedication of the A4 and LEARN study participants and study partners.

\noindent Data used in this work were obtained from the Adolescent Brain Cognitive Development\textsuperscript{TM} (ABCD) Study, held in the NIH Brain Development Cohorts Data Sharing Platform. The ABCD Study\textsuperscript{\textregistered} is supported by the National Institutes of Health and additional federal partners under award numbers U01DA041048, U01DA050989, U01DA051016, U01DA041022, U01DA051018, U01DA051037, U01DA050987, U01DA041174, U01DA041106, U01DA041117, U01DA041028, U01DA041134, U01DA050988, U01DA051039, U01DA041156, U01DA041025, U01DA041120, U01DA051038, U01DA041148, U01DA041093, U01DA041089, U24DA041123 and U24DA041147. ABCD Consortium investigators designed and implemented the study and/or provided data but did not necessarily participate in the analysis or writing of this report. This manuscript reflects the views of the authors and may not reflect the opinions or views of the NIH or ABCD Consortium investigators.

\noindent Data used in this work were obtained from the NYU fastMRI Initiative database. NYU fastMRI investigators provided data but did not participate in the analysis or writing of this report. The primary goal of fastMRI is to test whether machine learning can aid in the reconstruction of medical images.

\noindent Data used in the preparation of this work were obtained from the Human Connectome Project (HCP) database (https://ida.loni.usc.edu/login.jsp). Data collection and sharing for this project were provided by the Human Connectome Project (HCP; Principal Investigators: Bruce Rosen, M.D., Ph.D., Arthur W. Toga, Ph.D., and Van J. Wedeen, M.D.). HCP funding was provided by the National Institute of Dental and Craniofacial Research (NIDCR), the National Institute of Mental Health (NIMH), and the National Institute of Neurological Disorders and Stroke (NINDS). HCP is the result of efforts of co-investigators from the University of Southern California, the Martinos Center for Biomedical Imaging at Massachusetts General Hospital (MGH), Washington University, and the University of Minnesota. HCP data are disseminated by the Laboratory of Neuro Imaging at the University of Southern California.

\noindent Data used in the preparation of this work were obtained from the International Consortium for Brain Mapping (ICBM) database (www.loni.usc.edu/ICBM). Data collection and sharing for this project were provided by the International Consortium for Brain Mapping (ICBM; Principal Investigator: John Mazziotta, M.D., Ph.D.). ICBM funding was provided by the National Institute of Biomedical Imaging and Bioengineering. ICBM is the result of efforts of co-investigators from UCLA, Montreal Neurologic Institute, University of Texas at San Antonio, and the Institute of Medicine, Juelich/Heinrich Heine University, Germany. ICBM data are disseminated by the Laboratory of Neuro Imaging at the University of Southern California.

\noindent This manuscript was prepared using an Osteoarthritis Initiative (OAI) public-use data set and does not necessarily reflect the opinions or views of the OAI investigators, the NIH or the private funding partners. The authors thank the participants, principal investigators, co-investigators and staff of all hospitals that contributed data to the OAI. The OAI is a public-private partnership comprising five contracts (N01-AR-2-2258, N01-AR-2-2259, N01-AR-2-2260, N01-AR-2-2261 and N01-AR-2-2262), funded by the National Institutes of Health and conducted by the OAI Study Investigators. Private funding partners include Merck Research Laboratories, Novartis Pharmaceuticals Corporation, GlaxoSmithKline and Pfizer. Private-sector funding for the OAI is managed by the Foundation for the National Institutes of Health.

\noindent Data used in this work were generated by the National Cancer Institute's Cancer Moonshot Biobank. Data used in this work were generated by the National Cancer Institute Clinical Proteomic Tumor Analysis Consortium (CPTAC). Data used in this work were collected and shared with support from award U01 CA151261, PI Fiona Fennessy. Data used in this work were collected and shared with support from award U01 CA154602, PI Wei Huang. The results shown here are in whole or in part based upon data generated by the TCGA Research Network. Data used in this work were obtained from The Cancer Imaging Archive (TCIA) \citep{clark2013tcia}, supported by the National Cancer Institute.

\noindent This project used data funded in whole or in part with federal funds from the National Center for Advancing Translational Sciences (UL1 TR003107) and the National Cancer Institute, Contract No. 75N91019D00024, Subcontract 20X023F.

\noindent Data used in the preparation of this manuscript were obtained from the 4-Repeat Neuroimaging Initiative (4RTNI) database. The primary goal of 4RTNI is to identify neuroimaging and biomarker indicators of disease progression in 4-repeat tauopathy neurodegenerative diseases, including progressive supranuclear palsy and corticobasal degeneration. Data collection and sharing for this project were funded by the 4-Repeat Tauopathy Neuroimaging Initiative (4RTNI; National Institutes of Health grant R01 AG038791) and through generous contributions from the Tau Research Consortium. The study is coordinated through the University of California, San Francisco, Memory and Aging Center. 4RTNI data are disseminated by the Laboratory of Neuro Imaging at the University of Southern California.

\noindent We also thank Istanbul Medipol University for their provision of the CT-RATE data and the MELA project teams for providing their dataset.

\bmhead*{Compute acknowledgments}
\noindent The authors gratefully acknowledge the computing time provided on the high-performance computer HoreKa by the National High-Performance Computing Center at KIT (NHR@KIT). This center is jointly supported by the Federal Ministry of Research, Technology and Space and the Ministry of Science, Research and the Arts of Baden-Württemberg, as part of the National High-Performance Computing (NHR) joint funding program. HoreKa is partly funded by the German Research Foundation (DFG).

\noindent The authors gratefully acknowledge the Gauss Centre for Supercomputing e.V. (www.gauss-centre.eu) for funding this project by providing computing time on the GCS Supercomputer JUWELS \citep{JUWELS} at Jülich Supercomputing Centre (JSC).

\bmhead*{Data availability}
All publicly available datasets used in this study are listed in Supplementary (\cref{appendix_downstream_datasets} and \cref{tab:predatasets}). The private pretraining data and data provided by external partners cannot be shared due to data privacy regulations and data usage agreements.

\bmhead*{Code availability}
Code for this work will be made publicly available, in particular through the public repositories \hyperlink{https://github.com/MIC-DKFZ/nnssl}{nnssl}, 
\hyperlink{https://github.com/mic-dkfz/nnunet}{nnU-Net} and \hyperlink{https://github.com/MIC-DKFZ/nnDetection}{nnDetection}







\begin{appendices}
\clearpage
\pagenumbering{arabic}
\setcounter{page}{1}
\renewcommand{\thepage}{S\arabic{page}}
\section*{Supplementary information}
\label{app:cover}
\addcontentsline{toc}{section}{Appendix}

\begingroup
\singlespacing

\startcontents[appendix]
\setcounter{tocdepth}{2}
\printcontents[appendix]{}{1}{}

\endgroup
\clearpage


\section{Initial development stage}
\label{sup_1}
\noindent This section summarizes the exploratory optimization experiments that guided the design of the final pretraining strategies. All experiments were conducted on the internal proxy benchmark described in the main text, comprising five segmentation, two detection, and two classification datasets. The purpose of this benchmark was to support controlled model and training design before scaling to the full dataset. \\

\noindent Because this benchmark was used for iterative method development, not all experimental conditions remained fixed throughout. In particular, some downstream settings, preprocessing choices, and optimization details were revised over the course of experimentation. We therefore do not report individual benchmark results in this section. Instead, we summarize which configurations were tested, the main empirical findings, and the observations that informed the final large-scale model design. \\

\noindent Initial developing benchmark experiments were performed on data resampled to a fixed isotropic target spacing of isotropic 1mm. In later experiments, we tested pretraining without enforcing a fixed target spacing and did not observe a substantial reduction in downstream performance. Because this setting was expected to produce more robust representations across heterogeneous acquisition spacings, we did not apply fixed-spacing resampling in the final large-scale training.

\subsection{MAE Optimization} 
\noindent Optimization of the transformer-based MAE models started from a base Primus-M configuration with encoder depth 16, 12 attention heads, embedding dimension 864, and a decoder with depth 8 and 12 attention heads.\\ 
Among the tested masking ratios, 0.8 yielded the strongest overall downstream behavior and was therefore selected for subsequent experiments. For this base configuration, the most effective learning rate was $3 \times 10^{-4}$. \\
Increasing encoder depth showed a clear positive scaling trend, with improvements observed up to depth 28 in this benchmark setting. In contrast, varying the number of attention heads between 12 and 16 did not produce meaningful differences. Similarly, moderate changes in embedding dimension had little measurable impact on downstream behavior. \\ 
The decoder configuration had a stronger effect on downstream task behavior. For shallower encoders, particularly at depth 16, a smaller decoder with depth 2 performed better for segmentation than deeper decoders with depth 4 or 8, whereas classification benefited from a deeper decoder. This pattern suggests that decoder capacity influences the balance between preserving fine-grained localized information and encouraging more abstract semantic representations. At an encoder depth of 28, a decoder depth of 4 provided the best trade-off for both segmentation and classification. Based on this trend, we increased decoder depth during the later scaling phase to encoder depth 40 for the final model and used a decoder depth of 8. \\
Longer pretraining schedules also improved downstream behavior up to the GPU-hour budget allocated for this scaling stage.\\

\begin{figure}[t!]
\centering
\includegraphics[width=\textwidth]{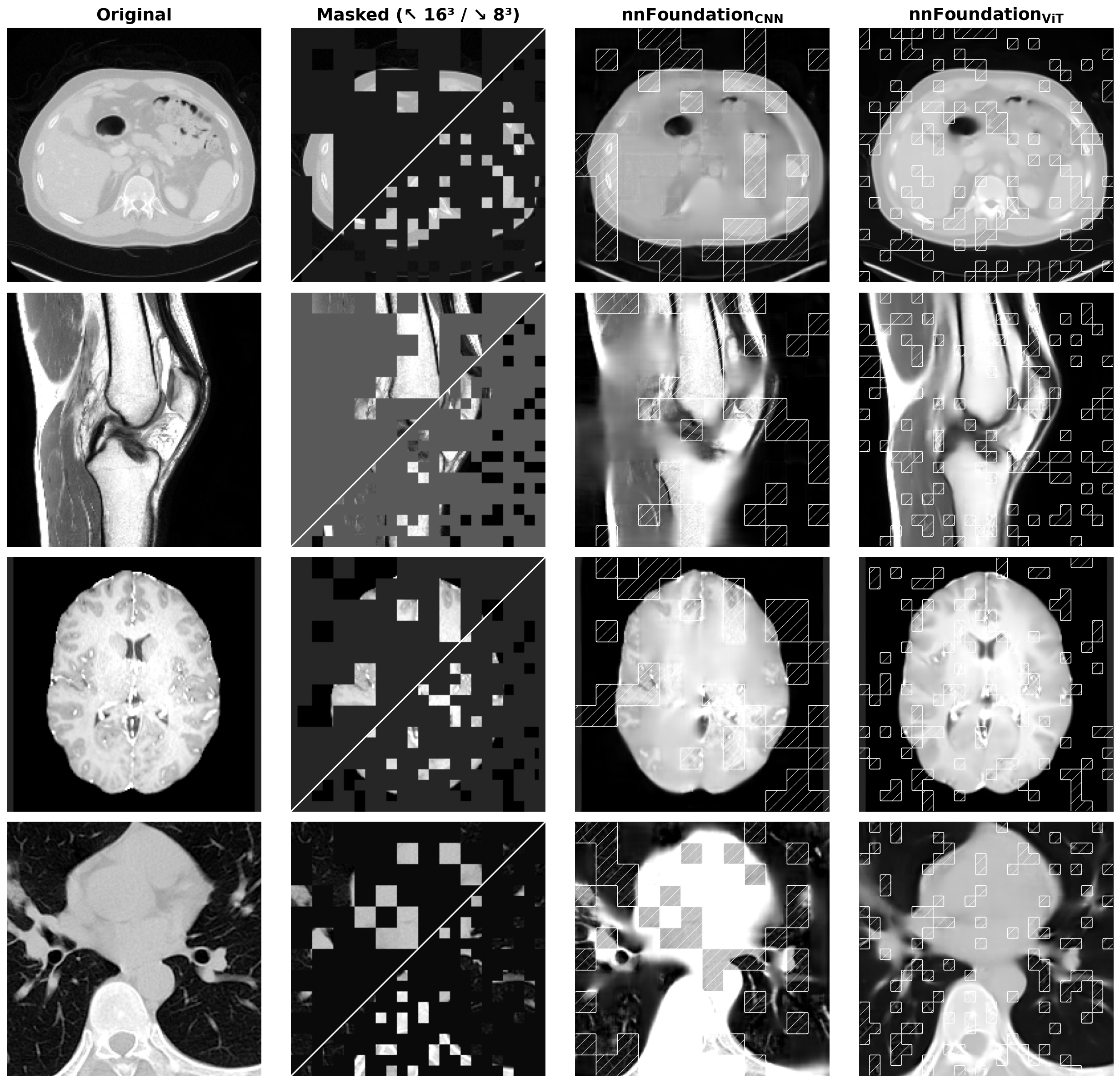}
\caption{\textbf{MAE mask reconstruction}
MAE reconstruction examples for four imaging modalities (abdominal CT, knee MRI, brain MRI, chest CT). The masked input (column 2) uses a diagonal split to show masking behavior of \Name{cnn} (top) and \Name{vit} (bottom) simultaneously. Columns 3 and 4 show reconstructions by \Name{cnn} and \Name{vit}, respectively, with hatched rectangles indicating non masked regions. \Name{cnn} produces smoother reconstructions, while \Name{vit} better preserves fine structural detail, confirming that both architectures learn meaningful anatomical representations through our self-supervised MAE pretraining.
}\label{appendix_fig7}
\end{figure}

\noindent Optimization of the CNN MAE began from an established configuration based on the ResEnc-L CNN architecture with a batch size of 8 and training for 1,000 epochs with 250 steps per epoch \cite{wald2024openmind}. Bold values indicate the best-performing configuration. We evaluated combinations of learning rates in $\{7 \times 10^{-2}, 3 \times 10^{-2}, \mathbf{1 \times 10^{-2}}, 7 \times 10^{-3}, 3 \times 10^{-3}, 1 \times 10^{-3}\}$ and weight decay values in $\{1 \times 10^{-3}, 3 \times 10^{-4}, 7 \times 10^{-5}, \mathbf{3 \times 10^{-5}}\}$. In addition, we performed a grid search over mask ratios of $\{60\%, \mathbf{75\%}, 90\%\}$ and masking strategies including grid masking, same-axis slice masking, and all-axis slice masking. For grid masking, we additionally evaluated cubic mask sizes of $\{2, 4, 8,  \mathbf{16}, 32, 64\}$.

\noindent We further evaluated the CNN sparsity adaptation proposed in Revisiting MAE \cite{wald2025revisiting}, input patch sizes of $\{128, 160, \mathbf{192}, 224\}$, and different reconstruction losses. The loss functions included L1, \textbf{L2}, SSIM, L1 NoMask, and L2 NoMask, where NoMask denotes losses computed only over masked regions, while the remaining variants were computed over the full input. We also explored augmentation strength through a grid search over intensity augmentation levels \{\textbf{off}, low, medium, high\} and spatial augmentation levels \{\textbf{low}, medium, high\}. Finally, we assessed architecture scaling by varying the depth, defined as the number of layers per stage, and the width, defined as the number of channels per stage, using \{\textbf{base}, medium, large\} settings for each while keeping the total number of stages fixed at six.

\noindent Despite this extensive set of ablations, none of the modifications produced a consistent improvement in downstream performance. We therefore retained the original configuration, but used an input patch size of $192 \times 192 \times 192$.

\noindent The qualitative reconstruction examples in Figure \ref{appendix_fig7} demonstrate that both CNN- and transformer-based MAE successfully recover anatomical structures across diverse imaging modalities. The CNN-based model tends to produce smoother reconstructions, while the ViT-based model preserves finer structural detail, particularly in high-frequency regions such as bone edges in the knee MRI and cortical structures in the brain MRI. Together, these examples confirm that both architectures have learned meaningful representations of medical image anatomy through self-supervised pretraining.

\subsection{Contrastive learning Optimization}
\noindent The contrastive learning was inspired by SimCLR and the model optimization started from a base Primus-M configuration with encoder depth 16, 12 attention heads, embedding dimension 864, and an MLP projection head with hidden and output dimension of 2048. For each image, we extracted two crops with a crop size of 128 cubic and a batch size of 4. \\
Increasing the number of heads to 24 while lowering the learning rate to $3 \times 10^{-5}$ yielded most stable training and best performance. Changing the minimum amount of overlap between crops from 0.25 to 0.75 did not have a measurable influence, therefore we chose 0.75 for further experiments to guarantee maximum foreground for each crop even for larger crop sizes. \\
Lowering the hidden and output dimension of the MLP projection head yielded slightly better performance while significantly reducing memory footprint and epoch times.\\
Within a fixed compute budget, increasing batch size up to 256 while reducing crop size to 32 cubic yielded the best performance. Especially the localized tasks (segmentation and detection) benefitted from the smaller crop size. \\
Extending pretraining time within the allocated compute budget yielded minimal performance improvements, while model scaling up to a depth of 28 had the strongest impact, aligning with the findings of the ViT MAE development.\\
We transferred the configuration used for the ViT SimCLR to the ResEncL architecture to also provide a CNN CLR baseline.

\subsection{DINOv2 optimization}
\noindent Optimization of the modified DINOv2 models started from a default Primus-M-style configuration with 24 transformer blocks, 12 attention heads, an embedding dimension of 864, a patch size of 8, global and local crop sizes of 96 and 48 voxels cubic, respectively, and crop scale ranges of $[0.7, 1.3]$ for both global and local crops. Pretraining was performed for 100 epochs with a per-GPU batch size of 4, a weight decay of 0.04, a learning rate of $5 \times 10^{-4}$, 10 warm-up epochs, and the nnU-Net-based medical augmentation pipeline. DINOv2 pretraining already provided a clear benefit over training from-scratch and was therefore used as the basis for subsequent experiments. \\
The most effective improvements came from optimization and model scaling rather than from changing the crop setup. Increasing the warm-up period to 50 epochs, reducing the weight decay to 0.005, and increasing the per-GPU batch size to 6 each improved downstream behavior relative to the default configuration. Extending pretraining from 100 to 400 epochs yielded an additional strong improvement, indicating that DINOv2 benefited substantially from longer training in this setting. These changes were combined in the final selected configuration, which used 400 pretraining epochs, a per-GPU batch size of 6, a weight decay of 0.005, and 50 warm-up epochs. \\
In addition, increasing the number of attention heads from 12 to 16 while keeping the depth fixed at 24 was part of the best-performing configuration and was therefore adopted in the final model. In contrast, the crop design was retained from the default configuration, with global and local crop sizes of 96 and 48 voxels cubic and crop scale ranges of $[0.7, 1.3]$, respectively. Although alternative crop settings were explored, they did not provide sufficiently robust evidence to replace the default crop setup in the final model. \\
Overall, DINOv2 performance was improved mainly through a longer and more stable optimization schedule together with a moderate increase in model capacity, while the default crop and augmentation design remained unchanged in the final selected configuration.

\clearpage

\section{Scale comparison with prior foundation models in radiology}
\label{app:model_scale_comparison}

\begin{figure}[!htbp]
\centering
\makebox[\textwidth][c]{\includegraphics[width=1.15\textwidth]{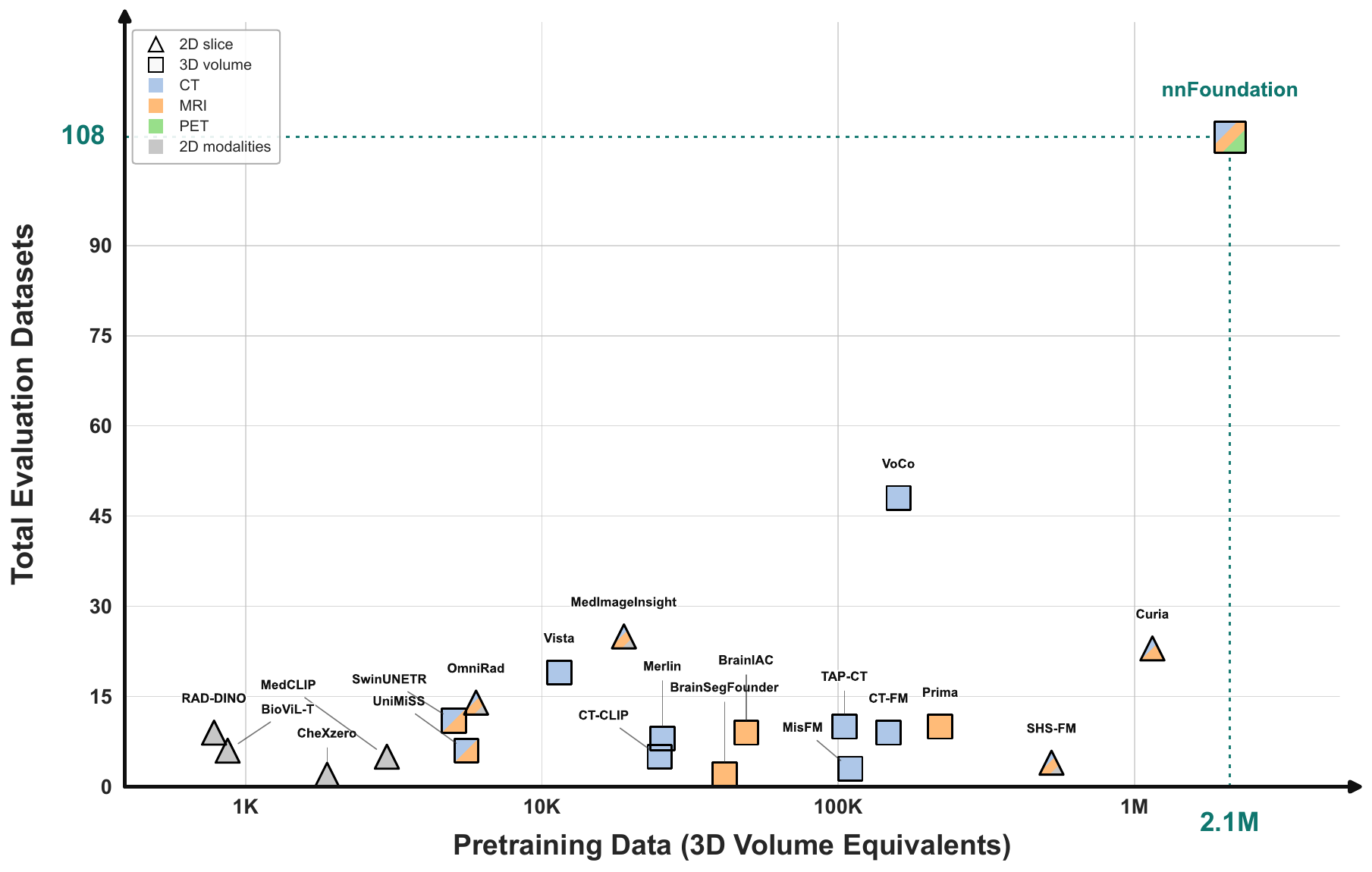}}
\caption{\textbf{Pretraining and evaluation scale across prior foundation models in radiology.}
Enlarged and fully labeled variant of Fig.~\ref{fig1}c. To the best of our knowledge, nnFoundation represents the largest radiology foundation model by reported 3D pretraining scale and was evaluated on the largest number of downstream datasets. Pretraining scale is plotted in 3D-volume equivalents, computed as
\(N_{\mathrm{3D\ equiv.}} = N_{\mathrm{3D\ volumes}} + N_{\mathrm{2D\ slices}}/200\).
The divisor of 200 is the visualization convention used throughout this comparison to normalize reported 2D slice counts to an approximate 3D-volume-equivalent scale; raw pretraining counts are provided in Table~\ref{tab:model_scale_plot_points}. The y-axis reports the number of downstream evaluation datasets counted for each model.}
\label{appendix_fig_model_scale_all_labels}
\end{figure}

\begin{table}[p]
\caption{Source values for prior foundation models in radiology included in the model-scale comparison in Fig.~\ref{appendix_fig_model_scale_all_labels}. For each model, the table reports input dimensionality, imaging modality groups, raw 3D and 2D pretraining counts, the corresponding 3D-volume-equivalent count, and the number of downstream evaluation datasets. The 3D-equivalent value is computed as \(N_{\mathrm{3D\ equiv.}} = N_{\mathrm{3D\ volumes}} + N_{\mathrm{2D\ slices}}/200\).}
\label{tab:model_scale_plot_points}
\centering
\small
\setlength{\tabcolsep}{2pt}

\begin{tabular}{@{}>{\raggedright\arraybackslash}p{2.35cm}>{\raggedright\arraybackslash}p{1.55cm}>{\raggedright\arraybackslash}p{2.15cm}rrrr@{}}
\toprule
Model & Dimension & Modality & \makecell{\(N_{\mathrm{3D\ volumes}}\)} & \makecell{\(N_{\mathrm{2D\ slices}}\)} & \makecell{\(N_{\mathrm{3D\ equiv.}}\)} & \makecell{Evaluation\\datasets} \\
\midrule
nnFoundation & 3D & CT; MRI; PET & 2,100,000 & -- & 2,100,000 & 108 \\
Curia~\cite{dancette2025curia} & 2D & CT; MRI & 150,000 & 200,000,000 & 1,150,000 & 23 \\
SHS-FM~\cite{ghesu2022contrastive} & 2D & CT; MRI; 2D modalities & -- & 105,006,320 & 525,032 & 4 \\
Prima \cite{lyu2025learningneuroimagingmodelshealth}& 3D & MRI & 221,147 & -- & 221,147 & 10 \\
VoCo~\cite{wu2024voco,voco2} & 3D & CT & 160,000 & -- & 160,000 & 48 \\
CT-FM~\cite{pai2025vision} & 3D & CT & 148,000 & -- & 148,000 & 9 \\
MisFM~\cite{wang2023mis} & 3D & CT & 110,000 & -- & 110,000 & 3 \\
TAP-CT \cite{veenboer2026tapct} & 3D & CT & 105,000 & -- & 105,000 & 10 \\
BrainIAC~\cite{Tak2026BrainIAC} & 3D & MRI & 48,965 & -- & 48,965 & 9 \\
BrainSeg\newline Founder \cite{COX2024103301} & 3D & MRI & 41,400 & -- & 41,400 & 2 \\
Merlin~\cite{Blankemeier2026} & 3D & CT & 25,528 & -- & 25,528 & 8 \\
CT-CLIP~\cite{hamamci2026generalist} & 3D & CT & 25,000 & -- & 25,000 & 5 \\
MedImage\newline Insight \cite{codella2024medimageinsightopensourceembeddingmodel}& 2D & CT; MRI; 2D modalities & -- & 3,786,210 & 18,931.0 & 25 \\
Vista~\cite{he2025vista3d} & 3D & CT & 11,454 & -- & 11,454 & 19 \\
OmniRad \cite{zedda2026omniradradiologicalfoundationmodel} & 2D & CT; MRI; 2D modalities & -- & 1,200,000 & 6,000 & 14 \\
UniMiSS \cite{unimiss}& 3D & CT; MRI & 5,022 & 108,948 & 5,566.7 & 6 \\
SwinUNETR~\cite{tang2022self} & 3D & CT; MRI & 5,050 & -- & 5,050 & 11 \\
MedCLIP~\cite{wang-etal-2022-medclip} & 2D & 2D modalities & -- & 600,000 & 3,000 & 5 \\
CheXzero~\cite{tiu2022expert} & 2D & 2D modalities & -- & 377,110 & 1,885.5 & 2 \\
BioViL-T~\cite{bannur2023learning} & 2D & 2D modalities & -- & 174,100 & 870.5 & 6 \\
RAD-DINO~\cite{perez2025exploring} & 2D & 2D modalities & -- & 156,620 & 783.1 & 9 \\
\bottomrule
\end{tabular}
\end{table}
\clearpage

\section{Additional Experiment Details}

\subsection{Generalization}

We evaluate across six ID/OOD axes derived from five datasets, listed in
Table~\ref{tab:datasets}, covering diverse anatomies, modalities, and shift types.
Models are fine-tuned exclusively on the ID training split and evaluated on
held-out ID and OOD test partitions.

\begin{table}[ht]
\caption{Datasets and ID/OOD evaluation axes. The number of cases are reported as:
OOD evaluation cases/ID evaluation cases/finetuning cases, for each of the evaluated. Classes are multi-class (M)
or binary (B).}
\label{tab:datasets}
\centering
\small
\setlength{\tabcolsep}{2pt}

\begin{tabular}{llllcl}
\toprule
Dataset & Mod. & Task & \#OOD/ID/FT & Cls. & Shift \\
\midrule
MMs~\cite{campello2021multi}                            & MRI    & Cardiac     & 150/75/75   & M & Held-out center  \\
MMs $\to$ ACDC~\cite{bernard2018deep}                   & MRI    & Cardiac     & 200/75/75   & M & Cross-dataset    \\
WHS~\cite{care2025whs}                                  & CT+MRI & Whole-heart & 46/20/20    & M & Held-out center  \\
BraTS-GLI $\to$ Africa~\cite{bakas2024bratsafrica}      & MRI    & Glioma      & 60/675/675  & M & Geographic       \\
Pancreas-T1~\cite{zhang2025pansegnet}                   & MRI    & Pancreas    & 72/156/156  & B & Held-out centers \\
CTSpine1K $\to$ Spine-Mets~\cite{haouchine2024spinemets} & CT     & Vertebrae   & 55/503/502  & M & Cross-dataset    \\
BladderCancerMRI~\cite{cao2024bladdermri} & MRI     &  Bladder Cancer    & 55/83/83  & B & Held-out center     \\
\bottomrule
\end{tabular}
\end{table}

\subsection{Exploratory Out-of-Domain Biology Segmentation}
\label{app:extra-bio-segmentation}
\begin{figure}[!htbp]
\centering
\includegraphics[width=\linewidth,height=0.70\textheight,keepaspectratio]{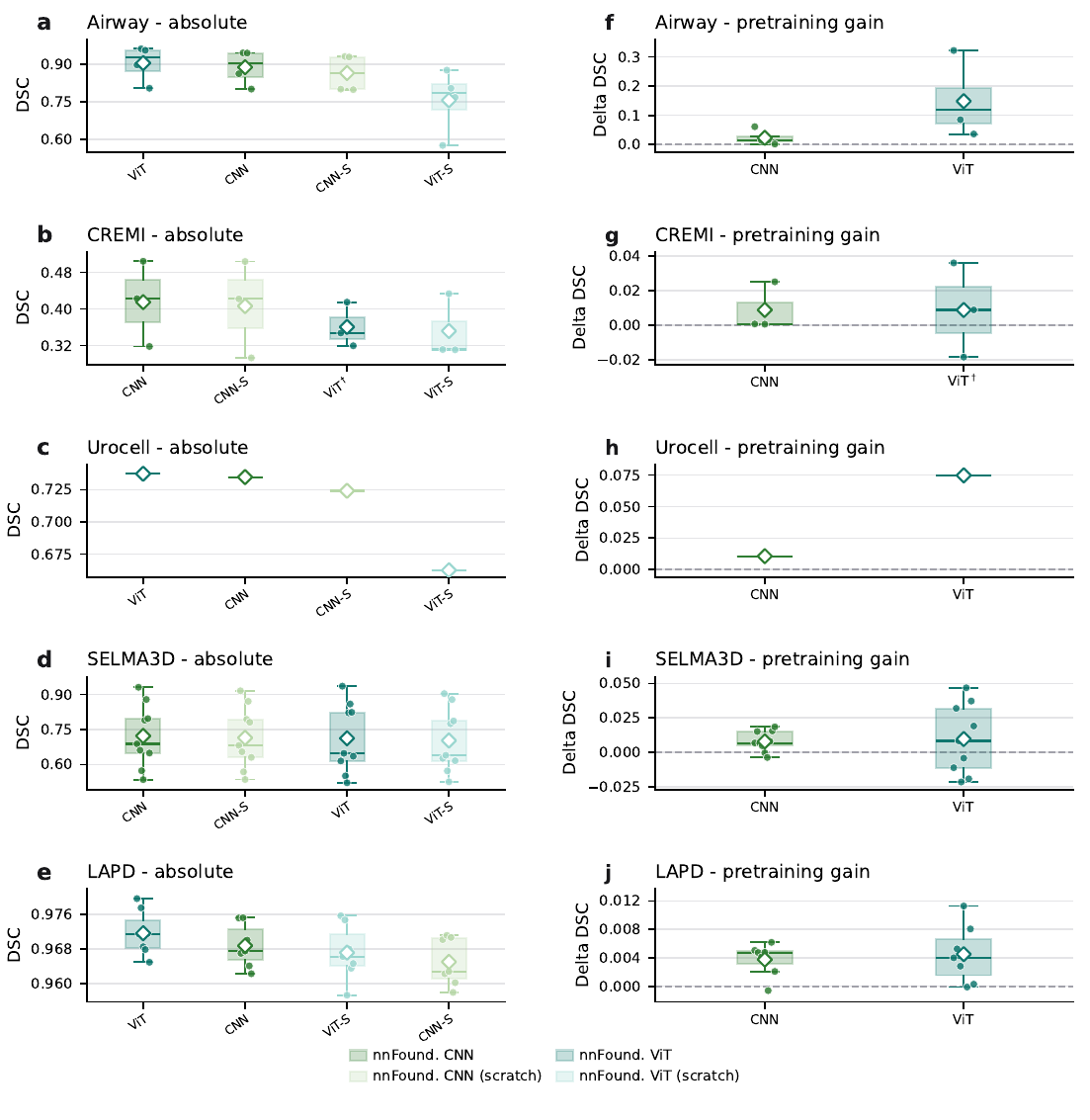}
\caption{\textbf{DSC comparisons on exploratory out-of-domain biology segmentation tasks suggest that pretraining benefits generalize beyond radiology.} Rows show datasets; left panels show absolute case-level DSC and right panels show paired pretrained-minus-scratch DSC differences. Positive deltas favor the pretrained model. Points denote validation cases, boxes show case-level distributions, and diamonds denote means. Models in the absolute DSC panels are ordered by descending mean DSC within each dataset. The CREMI nnFoundationViT$^\dagger$ mark denotes a 30-epoch fine-tuning run; the scratch comparator and all other downstream runs used 150 epochs.}
\label{fig:extra-bio-segmentation}
\end{figure}

To further assess transferability beyond radiology, we extend the evaluation to out-of-domain 3D bioimaging tasks. These experiments cover bioimaging domains with fundamentally different acquisition characteristics from CT, MRI, and PET, including microscopy-based volumetric data with different spatial scales, contrast mechanisms, and biological target structures (\cref{tab:extra-bio-model-summary}). Despite this substantial domain and modality gap, fine-tuning nnFoundation on these tasks still yields improvements, suggesting that radiology-scale 3D pretraining learns volumetric representations that transfer beyond clinical imaging (\cref{fig:extra-bio-segmentation}).
\clearpage
\begingroup
\singlespacing
\tablebodyfont
\setlength{\tabcolsep}{2pt}
\renewcommand{\arraystretch}{0.90}
\setlength{\LTcapwidth}{\linewidth}
\begin{longtable}{@{}>{\raggedright\arraybackslash}p{0.18\linewidth}>{\raggedright\arraybackslash}p{0.27\linewidth}>{\raggedright\arraybackslash}p{0.15\linewidth}c>{\raggedright\arraybackslash}p{0.27\linewidth}@{}}
\caption{Exploratory out-of-domain biology segmentation dataset profile. All results are validation summaries from small microscopy or electron microscopy datasets. Validation design reports validation cases used for the summary tables.}\\
\toprule
\textbf{Dataset} & \textbf{Input domain} & \textbf{Target} & \textbf{Train $n$} & \textbf{Validation design} \\
\midrule
\endfirsthead
\caption[]{Exploratory out-of-domain biology segmentation dataset profile. All results are validation summaries from small microscopy or electron microscopy datasets. Validation design reports validation cases used for the summary tables. (continued)}\\
\toprule
\textbf{Dataset} & \textbf{Input domain} & \textbf{Target} & \textbf{Train $n$} & \textbf{Validation design} \\
\midrule
\endhead
\midrule
\multicolumn{5}{r}{\footnotesize Continued on next page} \\
\endfoot
\bottomrule
\endlastfoot
Airway \cite{yang2024lungvis10} & Light sheet fluorescence microscopy & Airway & 22 & Single split, 4 val. cases \\
CREMI synapses  \cite{cremi} & Electron microscopy & Synapses & 3 & 3-fold CV, 3 val. cases \\
Urocell mitochondria \cite{urocell2,urocell1} & Electron microscopy & Mitochondria & 6 & Single split, 1 val. case \\
SELMA3D \cite{Chen2025SELMA3D} & Light sheet microscopy & Structure & 44 & Single split, 9 val. cases \\
LAPD mouse \cite{lapdmouse1,lapdmouse2} & Fluorescence microscopy & Airway & 34 & Single split, 7 val. cases \\
\end{longtable}
\endgroup

\begingroup
\singlespacing
\tablebodyfont
\setlength{\tabcolsep}{2.2pt}
\renewcommand{\arraystretch}{0.90}
\setlength{\LTcapwidth}{\linewidth}
\begin{longtable}{@{}>{\raggedright\arraybackslash}p{0.42\linewidth}ccc@{}}
\caption{Exploratory out-of-domain biology segmentation validation performance. Values are case-level means with SDs in parentheses across validation cases. All downstream runs used 150 epochs except the pretrained \Name{vit} run on CREMI, marked with $^\dagger$, which used 30 epochs.}\label{tab:extra-bio-model-summary}\\
\toprule
\textbf{Model} & \textbf{$n$ cases} & \textbf{DSC} & \textbf{NSD} \\
\midrule
\endfirsthead
\caption[]{Exploratory out-of-domain biology segmentation validation performance. Values are case-level means with SDs in parentheses across validation cases. All downstream runs used 150 epochs except the pretrained \Name{vit} run on CREMI, marked with $^\dagger$, which used 30 epochs. (continued)}\\
\toprule
\textbf{Model} & \textbf{$n$ cases} & \textbf{DSC} & \textbf{NSD} \\
\midrule
\endhead
\midrule
\multicolumn{4}{r}{\footnotesize Continued on next page} \\
\endfoot
\bottomrule
\endlastfoot
\multicolumn{4}{@{}l}{\textbf{Airway}} \\
\addlinespace[0.10em]
\Name{cnn} & 4 & 0.887 (0.070) & 0.412 (0.153) \\
\Name{cnn} (scratch) & 4 & 0.864 (0.075) & 0.378 (0.133) \\
\Name{vit} & 4 & 0.904 (0.073) & 0.445 (0.169) \\
\Name{vit} (scratch) & 4 & 0.755 (0.128) & 0.334 (0.125) \\
\addlinespace[0.25em]
\multicolumn{4}{@{}l}{\textbf{CREMI synapses}} \\
\addlinespace[0.10em]
\Name{cnn} & 3 & 0.415 (0.093) & 0.462 (0.122) \\
\Name{cnn} (scratch) & 3 & 0.406 (0.106) & 0.453 (0.138) \\
\Name{vit}$^\dagger$ & 3 & 0.361 (0.049) & 0.377 (0.061) \\
\Name{vit} (scratch) & 3 & 0.352 (0.071) & 0.363 (0.074) \\
\addlinespace[0.25em]
\multicolumn{4}{@{}l}{\textbf{Urocell mitochondria}} \\
\addlinespace[0.10em]
\Name{cnn} & 1 & 0.735 (--) & 0.526 (--) \\
\Name{cnn} (scratch) & 1 & 0.724 (--) & 0.526 (--) \\
\Name{vit} & 1 & 0.737 (--) & 0.552 (--) \\
\Name{vit} (scratch) & 1 & 0.663 (--) & 0.497 (--) \\
\addlinespace[0.25em]
\multicolumn{4}{@{}l}{\textbf{SELMA3D}} \\
\addlinespace[0.10em]
\Name{cnn} & 9 & 0.723 (0.135) & 0.774 (0.164) \\
\Name{cnn} (scratch) & 9 & 0.715 (0.133) & 0.761 (0.174) \\
\Name{vit} & 9 & 0.713 (0.149) & 0.762 (0.182) \\
\Name{vit} (scratch) & 9 & 0.703 (0.136) & 0.747 (0.181) \\
\addlinespace[0.25em]
\multicolumn{4}{@{}l}{\textbf{LAPD mouse}} \\
\addlinespace[0.10em]
\Name{cnn} & 7 & 0.969 (0.005) & 1.000 (0.000) \\
\Name{cnn} (scratch) & 7 & 0.965 (0.006) & 1.000 (0.000) \\
\Name{vit} & 7 & 0.972 (0.005) & 1.000 (0.000) \\
\Name{vit} (scratch) & 7 & 0.967 (0.006) & 1.000 (0.000) \\
\end{longtable}
\endgroup

\begingroup
\singlespacing
\tablebodyfont
\setlength{\tabcolsep}{1.8pt}
\renewcommand{\arraystretch}{0.90}
\setlength{\LTcapwidth}{\linewidth}
\begin{longtable}{@{}>{\raggedright\arraybackslash}p{0.42\linewidth}ccc@{}}
\caption{Pretrained-minus-scratch deltas for exploratory out-of-domain biology segmentation. Positive values favor the pretrained model. Cells report mean delta with SD in parentheses. The CREMI \Name{vit}$^\dagger$ row used 30 fine-tuning epochs; the scratch comparator and all other downstream runs used 150 epochs.}\label{tab:extra-bio-pretraining-deltas}\\
\toprule
\textbf{Pair} & \textbf{$n$ paired} & \textbf{$\Delta$ DSC} & \textbf{$\Delta$ NSD} \\
\midrule
\endfirsthead
\caption[]{Pretrained-minus-scratch deltas for exploratory out-of-domain biology segmentation. Positive values favor the pretrained model. Cells report mean delta with SD in parentheses. The CREMI \Name{vit}$^\dagger$ row used 30 fine-tuning epochs; the scratch comparator and all other downstream runs used 150 epochs. (continued)}\\
\toprule
\textbf{Pair} & \textbf{$n$ paired} & \textbf{$\Delta$ DSC} & \textbf{$\Delta$ NSD} \\
\midrule
\endhead
\midrule
\multicolumn{4}{r}{\footnotesize Continued on next page} \\
\endfoot
\bottomrule
\endlastfoot
\multicolumn{4}{@{}l}{\textbf{Airway}} \\
\addlinespace[0.10em]
\Name{cnn} & 4 & 0.023 (0.026) & 0.034 (0.027) \\
\Name{vit} & 4 & 0.149 (0.125) & 0.111 (0.078) \\
\addlinespace[0.25em]
\multicolumn{4}{@{}l}{\textbf{CREMI synapses}} \\
\addlinespace[0.10em]
\Name{cnn} & 3 & 0.009 (0.014) & 0.009 (0.016) \\
\Name{vit}$^\dagger$ & 3 & 0.009 (0.027) & 0.014 (0.036) \\
\addlinespace[0.25em]
\multicolumn{4}{@{}l}{\textbf{Urocell mitochondria}} \\
\addlinespace[0.10em]
\Name{cnn} & 1 & 0.011 (--) & 0.000 (--) \\
\Name{vit} & 1 & 0.075 (--) & 0.055 (--) \\
\addlinespace[0.25em]
\multicolumn{4}{@{}l}{\textbf{SELMA3D}} \\
\addlinespace[0.10em]
\Name{cnn} & 9 & 0.008 (0.007) & 0.013 (0.017) \\
\Name{vit} & 9 & 0.010 (0.025) & 0.015 (0.044) \\
\addlinespace[0.25em]
\multicolumn{4}{@{}l}{\textbf{LAPD mouse}} \\
\addlinespace[0.10em]
\Name{cnn} & 7 & 0.004 (0.002) & 0.000 (0.000) \\
\Name{vit} & 7 & 0.005 (0.004) & 0.000 (0.000) \\
\end{longtable}
\endgroup

\noindent Abbreviations: CI, confidence interval; DSC, Dice similarity coefficient; NSD, normalized surface Dice; AUROC, area under the receiver operating characteristic curve; AUPRC, area under the precision-recall curve; mAP, mean average precision; FROC, free-response receiver operating characteristic; IoU, intersection over union; AP, average precision.

\subsection{External validation}
\noindent To assess whether our pretrained models can be successfully applied by independent users, both within our recommended fine-tuning pipeline and as drop-in components of pre-existing custom pipelines, we collaborated with multiple partner institutions: Siemens Healthineers AG, Floy GmbH, Helmholtz Munich (Machine Learning in Biomedical Imaging group), the German Center for Neurodegenerative Diseases (DZNE, Artificial Intelligence in Medical Imaging group), and the Max Delbrück Center (MDC, Animal Phenotyping group). An overview of all external validation tasks is provided in \cref{tab:downstreampartner}. Partners were provided with the \Name{cnn} and the small version of the \Name{vit}, as the larger transformer model was not compatible with the compute budget available in all external settings. In most cases, partners used the pretrained checkpoints in combination with the nnU-Net fine-tuning code from our internal experiments, while Siemens Healthineers AG integrated the pretrained model into their own existing downstream pipeline. This reflects the heterogeneous ways in which released models are typically adopted in practice. \\
In all cases, partners compared the pretrained model against the same architecture trained from scratch, isolating the effect of pretraining without introducing confounders from differing backbone designs. All external experiments were conducted without influence from the core authors or developers of the pretrained models. In particular, data preprocessing, model integration, training, and evaluation were carried out independently by the respective partners. Compared with the internal benchmark, these results showed greater fluctuation across datasets and institutions. We attribute this to the broader diversity of downstream data, differences in experimental pipelines, saturation of performance, and the fact that no centralized troubleshooting or error correction was performed when individual models, whether pretrained or trained from scratch, did not behave as expected. As such, these experiments provide a realistic assessment of transferability under heterogeneous downstream conditions, complementing the more controlled internal benchmark.
In addition to the segmentation experiments, Siemens Healthineers performed an external lung nodule detection experiment on chest CT using an existing RetinaNet-based detection pipeline with the \Name{cnn} model used as backbone. The data (5480/641/260 train/val/test volumes), was sourced from NLST \cite{team2013data} and evaluated on an independently annotated test set of 260 images, for which the reference standard was defined as the union of annotations from three readers. Models were trained with different data regimes, corresponding to 100\%, 10\%, and 1\% of the available training data, and compared against the same architectures trained from scratch. Consistent with the segmentation experiments, pretraining was evaluated under realistic external conditions and showed benefits in low-data settings (\cref{tab:fig5-siemens-detection-validation}). This experiment extends the external validation beyond dense segmentation and indicates that the pretrained encoders can also be integrated into object-detection workflows.

\begin{table}[!htbp]
\centering
\caption{Overview of external partner validation datasets. For each task, the imaging modality, number of segmentation labels, label names, and train/test set sizes are reported. Datasets span a broad range of anatomical structures and pathologies across CT and MRI, and were evaluated independently at external partner institutions without involvement from the core authors.}
\label{tab:downstreampartner}

\scriptsize
\setlength{\tabcolsep}{2pt}
\renewcommand{\arraystretch}{0.90}

\begin{tabular*}{\textwidth}{@{\extracolsep{\fill}} >{\raggedright\arraybackslash}p{3cm} c c >{\raggedright\arraybackslash}p{3.8cm} c c @{}}
\toprule
Name & Modality & Labels & Label Names & Train & Test \\
\midrule
Siemens Pulmonary & CT & 1 & Pulmonary Artery & 150 & 33 \\
Siemens BP & CT & 2 & Brachial Plexus L; Brachial Plexus R & 317 & 44 \\
Siemens Bowel Bag & CT & 1 & Abdominal Cavity & 544 & 25 \\
Siemens Brain & CT & 1 & Brain & 200 & 9 \\
Siemens Chest Wall LR & CT & 2 & Chestwall Left; Chestwall Right & 48 & 11 \\
Siemens Constrictor Musc. & CT & 3 & Musc Cnstr I; Musc Cnstr M; Musc Cnstr S & 355 & 89 \\
Siemens Esophagus & CT & 1 & Esophagus & 1160 & 22 \\
Siemens Jaws & CT & 9 & Mandible; Parotid Left; Parotid Right; Supraglottic Larynx; Glottis; Submandibular Left; Submandibular Right; Oral Cavity; Lips & 399 & 50 \\
Siemens Ribs & CT & 24 & 24 ribs & 553 & 18 \\
Siemens Pancreas & CT & 1 & Pancreas & 369 & 93 \\
Siemens Pelvic & CT & 6 & Bladder; Prostate; Rectum; Femur Head Left; Femur Head Right; Seminal Vesicles & 692 & 14 \\
Siemens Penile Bulb & CT & 1 & Penile Bulb & 857 & 208 \\
Siemens Spinal Cord & CT & 1 & Spinal Cord & 1508 & 51 \\
Siemens Sternum & CT & 1 & Sternum & 1184 & 22 \\
Siemens Stomach & CT & 1 & Stomach & 371 & 92 \\
Siemens Thyroid & CT & 1 & Thyroid & 293 & 56 \\
Siemens Trachea Bronchus & CT & 1 & Trachea & 150 & 34 \\
Siemens Uterus & CT & 1 & Uterus & 381 & 94 \\
Siemens Inf. Venacava & CT & 1 & Inferior Vena Cava & 149 & 34 \\
Siemens Sup. Venacava & CT & 1 & Superior Vena Cava & 151 & 33 \\
Siemens Whole Bowel & CT & 5 & Duodenum; Small Bowel; Large Bowel; Sigmoid; Rectum & 332 & 84 \\
Siemens Lung Nodule Detection & CT & 1 & lung nodules & 5480/641 & 260 \\
Floy Abdomen & CT & 3 & Muscle, Subcutaneous fat, Visceral fat  & 92 & 29 \\
Floy Spine lesion & CT & 2 & Benign spine lesions; Indeterminate or malignant spine lesions & 910 & 431 \\
Floy Spine lesion & MR & 2 & Benign spine lesions; Indeterminate or malignant spine lesions & 1160 & 245 \\
MDC Mouse & microCT & 2 & Skull; Brain Endocast & 50 & 50 \\
DZNE Cerebellum~\cite{FABER2022119703} & MR & 18 & 18 cerebellum substructures & 18 & 8 \\
DZNE Hypothalamus~\cite{hyper} & MR & 17 & 17 hypothalamus substructures & 44 & 6 \\
Munich Sarcoma & MR & 1 & Sarcoma & 503 & 503 \\
\bottomrule
\end{tabular*}

\end{table}

\clearpage
\section{Pretraining Datasets}
\label{appendix_pretraining_datasets}

\begingroup
\singlespacing
\tablebodyfont
\setlength{\tabcolsep}{2pt}
\renewcommand{\arraystretch}{0.90}
\setlength{\LTcapwidth}{\linewidth}
\begin{longtable}{@{}l@{\hspace{1.20em}}r@{\hspace{1.20em}}l@{\hspace{1.20em}}l@{}}
\caption{Overview of datasets used during pretraining. Shown are the dataset name, number of images used after all curation steps and available modalities.}\label{tab:predatasets}\\
\toprule
\textbf{Dataset} & \multicolumn{1}{c}{\textbf{\# images}} & \textbf{Modalities} & Data source \\
\midrule
\endfirsthead
\caption[]{Overview of datasets used during pretraining. Shown are the dataset name, number of images used after all curation steps and available modalities. (continued)}\\
\toprule
\textbf{Dataset} & \multicolumn{1}{c}{\textbf{\# images}} & \textbf{Modalities} & Data source \\
\midrule
\endhead
\midrule
\multicolumn{3}{r}{\footnotesize Continued on next page} \\
\endfoot
\bottomrule
\endlastfoot
A4~\cite{sperling2014a4} & 28,153 & MR, PET & Public cohort \\
ABCD\_NIH~\cite{casey2018abcd} & 41,581 & MR & Public cohort \\
BIMCV-R~\cite{chen2024bimcvr} & 6,603 & CT & Public dataset \\
FastMRI\_Brain~\cite{knoll2020fastmri,zbontar2018fastmri} & 10,052 & MR & Public dataset \\
FastMRI\_Prostate~\cite{knoll2020fastmri,zbontar2018fastmri,tibrewala2024fastmriProstate} & 310 & MR & Public dataset \\
HCP~\cite{vanessen2013wuminn} & 67 & MR & Public dataset \\
ICBM~\cite{mazziotta2001icbm} & 899 & MR & Public dataset \\
nlm\_visible\_human\_project~\cite{clunie2024dicom} & 13 & CT, MR & Public dataset \\
openneuro~\cite{markiewicz2021openneuro} & 115,639 & CT, MR, PET & Public cohort \\
OsteoArthritisInitiative~\cite{oai2006} & 240,867 & MR & Public cohort \\
acrin\_6698~\cite{newitt2021acrin} & 41,532 & MR & Public cohort \\
acrin\_contralateral\_breast\_mr~\cite{kinahan2021acrincontralateralbreastmr} & 10,422 & MR & Public cohort \\
acrin\_flt\_breast~\cite{kinahan2017data} & 8,051 & CT, PET & Public cohort \\
acrin\_nsclc\_fdg\_pet~\cite{kinahan2019data} & 2,712 & CT, MR, PET & Public cohort \\
adrenal\_acc\_ki67\_seg~\cite{moawad2023voxellevel} & 49 & CT & Public dataset \\
advanced\_mri\_breast\_lesions~\cite{daniels2024standard} & 11,137 & MR & Public dataset \\
anti\_pd\_1\_lung~\cite{madhavi2019data} & 528 & CT, PET & Public dataset \\
breast\_diagnosis~\cite{bloch2015breastdiagnosis} & 1,185 & CT, MR, PET & Public dataset \\
breast\_mri\_nact\_pilot~\cite{newitt2016single} & 1,427 & MR & Public dataset \\
c4kc\_kits~\cite{heller2019data} & 177 & CT & Public dataset \\
cc\_tumor\_heterogeneity~\cite{mayr2023cervical} & 4,473 & CT, MR, PET & Public dataset \\
cmb\_aml~\cite{biobank2023cancer} & 3 & CT & Public dataset \\
cmb\_crc~\cite{biobank2022cancer} & 593 & CT, MR, PET & Public dataset \\
cmb\_gec~\cite{biobank2022cancer_2} & 8 & CT & Public dataset \\
cmb\_lca~\cite{biobank2022cancer_3} & 209 & CT, MR & Public dataset \\
cmb\_mel~\cite{biobank2022cancer_4} & 197 & CT, PET & Public dataset \\
cmb\_mml~\cite{biobank2022cancer_5} & 91 & CT, MR & Public dataset \\
cmb\_pca~\cite{biobank2022cancer_6} & 64 & CT, MR, PET & Public dataset \\
colorectal\_liver\_metastases~\cite{simpson2023preoperative} & 1 & CT & Public dataset \\
covid\_19\_ar~\cite{desai2020data} & 128 & CT & Public dataset \\
covid\_19\_ny\_sbu~\cite{saltz2021stony} & 2,871 & CT, MR, PET & Public dataset \\
cptac\_ccrcc~\cite{cptac2018the} & 558 & CT, MR & Public dataset \\
cptac\_cm~\cite{cptac2018the_2} & 164 & CT, MR, PET & Public dataset \\
cptac\_lscc~\cite{cptac2018the_3} & 189 & CT, PET & Public dataset \\
cptac\_luad~\cite{cptac2018the_4} & 182 & CT, MR, PET & Public dataset \\
cptac\_pda~\cite{cptac2018the_5} & 910 & CT, MR, PET & Public dataset \\
cptac\_sar~\cite{cptac2019the} & 156 & CT, MR, PET & Public dataset \\
cptac\_ucec~\cite{cptac2019the_2} & 1,684 & CT, MR, PET & Public dataset \\
ct\_colonography~\cite{k2015data} & 582 & CT & Public dataset \\
ct\_lymph\_nodes~\cite{roth2015a} & 1 & CT & Public dataset \\
ct\_vs\_pet\_ventilation\_imaging~\cite{eslick2022ct} & 230 & CT, PET & Public dataset \\
ctpred\_sunitinib\_pannet~\cite{chen2022prediction} & 73 & CT & Public dataset \\
ea1141~\cite{comstock2023abbreviated} & 6,987 & MR & Public cohort \\
hcc\_tace\_seg~\cite{moawad2021multimodality} & 396 & CT & Public dataset \\
ispy1~\cite{newitt2016multicenter} & 5,815 & MR & Public dataset \\
ispy2~\cite{li2022ispy2,newitt2021acrin} & 77,433 & MR & Public cohort \\
lctsc~\cite{yang2017data} & 54 & CT & Public dataset \\
lung\_fused\_ct\_pathology~\cite{madabhushi2018fused} & 16 & CT & Public dataset \\
lung\_pet\_ct\_dx~\cite{li2020a} & 1,062 & CT, PET & Public dataset \\
lungct\_diagnosis~\cite{o2015data} & 57 & CT & Public dataset \\
midrc\_ricord\_1a~\cite{tsai2020data} & 202 & CT & Public dataset \\
midrc\_ricord\_1b~\cite{tsai2021medical} & 116 & CT & Public dataset \\
mouse\_astrocytoma~\cite{jansen2015tcia} & 3 & MR & Public dataset \\
naf\_prostate~\cite{kurdziel2015data} & 198 & CT, PET & Public dataset \\
nlst~\cite{team2013data} & 129,592 & CT & Public cohort \\
nsclc\_radiogenomics~\cite{bakr2017data} & 864 & CT, PET & Public dataset \\
nsclc\_radiomics~\cite{aerts2014data} & 5 & CT & Public dataset \\
nsclc\_radiomics\_genomics~\cite{hjwl2015data} & 86 & CT & Public dataset \\
nsclc\_radiomics\_interobserver1~\cite{wee2019data} & 21 & CT & Public dataset \\
pancreatic\_ct\_cbct\_seg~\cite{hong2021breathhold} & 117 & CT & Public dataset \\
pediatric\_ct\_seg~\cite{jordan2021pediatric} & 9 & CT & Public dataset \\
pelvic\_reference\_data~\cite{yorke2019pelvic} & 71 & CT & Public dataset \\
prostate\_3t~\cite{litjens2015data} & 24 & MR & Public dataset \\
prostate\_diagnosis~\cite{bloch2015data} & 941 & MR & Public dataset \\
prostate\_fused\_mri\_pathology~\cite{madabhushi2016fused} & 1,126 & MR & Public dataset \\
prostate\_mri~\cite{p2016data} & 564 & MR & Public dataset \\
prostate\_mri\_us\_biopsy~\cite{natarajan2020prostate} & 3,037 & MR & Public dataset \\
prostatex~\cite{litjens2017spieaapm} & 18,312 & MR & Public dataset \\
pseudo\_phi\_dicom\_data~\cite{rutherford2021a} & 12 & CT, MR, PET & Public dataset \\
qin\_breast~\cite{li2016data} & 4,727 & CT, MR, PET & Public dataset \\
qin\_breast\_dce\_mri~\cite{huang2014variations} & 632 & MR & Public dataset \\
qin\_lung\_ct~\cite{goldgof2015data,kalpathycramer2015qin} & 44 & CT & Public dataset \\
qin\_prostate\_repeatability~\cite{fedorov2018data} & 1,539 & MR & Public dataset \\
remind~\cite{juvekar2023the} & 476 & MR & Public dataset \\
rider\_breast\_mri~\cite{meyer2015rider} & 10 & MR & Public dataset \\
rider\_lung\_ct~\cite{zhao2015coffeebreak} & 4 & CT & Public dataset \\
rider\_lung\_pet\_ct~\cite{p2015data} & 1,777 & CT, PET & Public dataset \\
rider\_pilot~\cite{lidc2023rider} & 114 & CT & Public dataset \\
soft\_tissue\_sarcoma~\cite{vallières2015a} & 122 & CT, MR, PET & Public dataset \\
spie\_aapm\_lung\_ct\_challenge~\cite{iii2015spieaapmnci} & 70 & CT & Public dataset \\
stageii\_colorectal\_ct~\cite{t2022abdominal} & 227 & CT & Public dataset \\
tcga\_blca~\cite{kirk2016the} & 724 & CT, MR, PET & Public cohort \\
tcga\_brca~\cite{lingle2016the} & 1,912 & MR & Public cohort \\
tcga\_cesc~\cite{lucchesi2016the} & 416 & MR & Public cohort \\
tcga\_coad~\cite{kirk2016the_2} & 55 & CT & Public cohort \\
tcga\_esca~\cite{lucchesi2016the_2} & 92 & CT & Public cohort \\
tcga\_kich~\cite{tcga_kich_data} & 93 & CT, MR & Public cohort \\
tcga\_kirc~\cite{akin2016the} & 1,744 & CT, MR & Public cohort \\
tcga\_kirp~\cite{linehan2016the} & 262 & CT, MR, PET & Public cohort \\
tcga\_lihc~\cite{erickson2016the} & 1,150 & CT, MR, PET & Public cohort \\
tcga\_luad~\cite{albertina2016the} & 283 & CT, PET & Public cohort \\
tcga\_lusc~\cite{kirk2016the_3} & 180 & CT, PET & Public cohort \\
tcga\_ov~\cite{holback2016the} & 563 & CT, MR & Public cohort \\
tcga\_prad~\cite{zuley2016the} & 412 & CT, MR, PET & Public cohort \\
tcga\_read~\cite{kirk2016the_4} & 19 & CT, MR & Public cohort \\
tcga\_sarc~\cite{roche2016the} & 12 & CT, MR & Public cohort \\
tcga\_stad~\cite{lucchesi2016the_3} & 231 & CT & Public cohort \\
tcga\_thca~\cite{kirk2016the_5} & 15 & CT, PET & Public cohort \\
tcga\_ucec~\cite{erickson2016the_2} & 581 & CT, MR, PET & Public cohort \\
upenn\_gbm~\cite{bakas2021multiparametric} & 26,001 & MR & Public cohort \\
vestibular\_schwannoma\_mc\_rc~\cite{kujawa2023segmentation} & 340 & MR & Public dataset \\
vestibular\_schwannoma\_seg~\cite{shapey2021segmentation} & 379 & MR & Public dataset \\
aapm-rt-mac~\cite{cardenas2019data} & 55 & MR & Public dataset \\
anti-pd-1\_melanoma~\cite{patnana2019antipd1} & 1,116 & CT, MR, PET & Public dataset \\
Brain-TR-GammaKnife~\cite{wang2023brain} & 76 & MR & Public dataset \\
GLIS-RT~\cite{shusharina2021glioma} & 667 & CT, MR & Public dataset \\
Head-Neck-Cetuximab~\cite{bosch2015headneck} & 979 & CT, PET & Public dataset \\
Healthy-Total-Body-CTs~\cite{selfridge2023lowdose} & 130 & CT & Public dataset \\
HNC-IMRT-70-33~\cite{buatti2024ctrtstructrtdosertplan} & 208 & CT & Public dataset \\
HNSCC~\cite{a2020hnscc} & 2,744 & CT, MR, PET & Public dataset \\
HNSCC-3DCT-RT~\cite{t2018headandneck} & 60 & CT & Public dataset \\
lgg-1p19qdeletion~\cite{erickson2017data} & 304 & MR & Public dataset \\
Meningioma-SEG-CLASS~\cite{vassantachart2023segmentation} & 479 & MR & Public dataset \\
mri-dir~\cite{ger2018data} & 40 & MR & Public dataset \\
OPC-Radiomics~\cite{jyy2019data} & 591 & CT & Public dataset \\
qin-brain-dsc-mri~\cite{schmainda2016glioma} & 48 & MR & Public dataset \\
QIN-GBM-Treatment-Response~\cite{ab2016data} & 1,613 & MR & Public dataset \\
qin-sarcoma~\cite{huang2016qinsarcoma} & 2,167 & MR & Public dataset \\
RADCURE~\cite{welch2023computed} & 3,268 & CT & Public dataset \\
RPA-Head-and-Neck-Lymph-Nodes~\cite{maroongroge2024the} & 45 & CT & Public dataset \\
4RTNI~\cite{4rtni} & 1,652 & MR & Public dataset \\
DKFZ Radiological Research PACS & 821,692 & CT, MR, PET & Private \\
Bonn & 264,803 & CT, MR & Private \\
ThoraxKlinik Heidelberg& 228,732 & CT, MR & Private \\
\end{longtable}
\endgroup

\clearpage

\section{Downstream Datasets}
\label{appendix_downstream_datasets}

\begingroup
\singlespacing
\tablebodyfont
\setlength{\tabcolsep}{2pt}
\renewcommand{\arraystretch}{0.90}
\setlength{\LTcapwidth}{\linewidth}
\begin{longtable}{@{}>{\raggedright\arraybackslash}p{3.0cm}@{\hspace{0.55em}}r@{\hspace{0.55em}}>{\raggedright\arraybackslash}p{1.45cm}@{\hspace{0.55em}}>{\raggedright\arraybackslash}p{1.65cm}@{\hspace{0.55em}}>{\raggedright\arraybackslash}p{1.35cm}@{\hspace{0.55em}}>{\raggedright\arraybackslash}p{2.05cm}@{}}
\caption{Overview of all downstream datasets. Shown are the dataset name, number of images, modalities, target and task types.}\\
\toprule
\textbf{Dataset} & \multicolumn{1}{c}{\textbf{\# images}} & \textbf{Modality} & \makecell[l]{\textbf{Anatomical}\\\textbf{region}} & \textbf{Target} & \makecell[l]{\textbf{Task}\\\textbf{type}} \\
\midrule
\endfirsthead
\caption[]{Overview of all downstream datasets. Shown are the dataset name, number of images, modalities, target and task types. (continued)}\\
\toprule
\textbf{Dataset} & \multicolumn{1}{c}{\textbf{\# images}} & \textbf{Modality} & \makecell[l]{\textbf{Anatomical}\\\textbf{region}} & \textbf{Target} & \makecell[l]{\textbf{Task}\\\textbf{type}} \\
\midrule
\endhead
\midrule
\multicolumn{6}{r}{\footnotesize Continued on next page} \\
\endfoot
\bottomrule
\endlastfoot
\addlinespace[0.18em]
\multicolumn{6}{@{}l}{\textbf{Segmentation}}\\
\addlinespace[0.02em]
AutoPET FDG~\cite{gatidis2022fdgpetctlesions} & 1014 & CT + PET & whole body & pathology & segmentation \\
KiTS~\cite{heller2023kits21} & 489 & CT & abdomen & pathology & segmentation \\
AMOS~\cite{ji2022amos} & 360 & CT + MR & abdomen & anatomy & segmentation \\
TopCoW~\cite{yang2023topcow} & 250 & CT + MR & brain & anatomy & segmentation \\
BrainMetShare~\cite{stanfordaimi2023brainmetshare,grovik2020deep} & 105 & MR & brain & pathology & segmentation \\
Panther Task 1~\cite{panther2025task1,betancourt_tarifa_2025_panther} & 92 & MR & abdomen & pathology & segmentation \\
Panther Task 2~\cite{panther2025task2,betancourt_tarifa_2025_panther} & 50 & MR & abdomen & pathology & segmentation \\
MSD Brain~\cite{antonelli2022medical} & 484 & MR & brain & pathology & segmentation \\
MSD Hippocampus~\cite{antonelli2022medical} & 260 & MR & brain & anatomy & segmentation \\
ISLES2022~\cite{hernandezpetsche2022isles} & 250 & MR & brain & pathology & segmentation \\
MU-Glioma-Post~\cite{mahmoud2025mugliomapost} & 593 & MR & brain & pathology & segmentation \\
WMHSegChallenge~\cite{kuijf2019standardized} & 60 & MR & brain & pathology & segmentation \\
MS Ljubljana~\cite{lesjak2018novel} & 331 & MR & brain & pathology & segmentation \\
ToothFairy~\cite{cipriano2022mandibular,bolelli2025toothfairy,bolelli2025cvpr_toothfairy2} & 532 & CBCT & head & anatomy & segmentation \\
HaNSeg~\cite{podobnik2023hanseg} & 42 & MR & head & anatomy & segmentation \\
HNTSMRG24~\cite{wahid2024hntsmrg} & 300 & MR & head \& neck & pathology & segmentation \\
MSD Heart~\cite{antonelli2022medical} & 20 & MR & chest & anatomy & segmentation \\
MSD Lung~\cite{antonelli2022medical} & 63 & CT & chest & pathology & segmentation \\
ACDC~\cite{bernard2018deep} & 200 & MR & chest & anatomy & segmentation \\
Rider Lung~\cite{zhao2008rider} & 65 & CT & chest & pathology & segmentation \\
Mama Mia~\cite{garrucho2025large} & 1506 & MR & chest & pathology & segmentation \\
Covid-19-20~\cite{roth2021covid1920} & 199 & CT & chest & pathology & segmentation \\
M\&Ms challenge~\cite{campello2021multi} & 300 & MR & chest & anatomy & segmentation \\
MSD Liver~\cite{antonelli2022medical} & 131 & CT & abdomen & pathology & segmentation \\
HCC-TACE-Seg~\cite{moawad2023hcctace} & 65 & CT & abdomen & pathology & segmentation \\
HCC-TACE-MRI~\cite{wawtace2024} & 34 & MR & abdomen & pathology & segmentation \\
MSD Hepatic Vessel~\cite{antonelli2022medical} & 303 & CT & abdomen & pathology & segmentation \\
MSD Pancreas~\cite{antonelli2022medical} & 281 & CT & abdomen & pathology & segmentation \\
MSD Spleen~\cite{antonelli2022medical} & 41 & CT & abdomen & anatomy & segmentation \\
MSD Colon~\cite{antonelli2022medical} & 126 & CT & abdomen & pathology & segmentation \\
WORD~\cite{luo2022word} & 120 & CT & abdomen & anatomy & segmentation \\
Stanford Knee~\cite{desai2022skm} & 155 & MR & knee & anatomy & segmentation \\
MSD Prostate~\cite{antonelli2022medical} & 32 & MR & pelvis & pathology & segmentation \\
Pengwin~\cite{sang2025pengwin} & 100 & CT & pelvis & pathology & segmentation \\
AutoPETIV Task 2~\cite{kustner2025longitudinalct} & 670 & CT & whole body & pathology & segmentation \\
BraTS2024 Pediatric~\cite{kazerooni2024bratspeds} & 261 & MR & brain & pathology & segmentation \\
BraTS2024 Glioma~\cite{verdier2024bratsgli} & 1350 & MR & brain & pathology & segmentation \\
DeepPSMA FDG~\cite{deeppsma2025} & 100 & CT + PET & whole body & pathology & segmentation \\
DeepPSMA PSMA~\cite{deeppsma2025} & 100 & CT + PET & whole body & pathology & segmentation \\
\addlinespace[0.18em]
\multicolumn{6}{@{}l}{\textbf{Scaling}}\\
\addlinespace[0.02em]
KiTS~\cite{heller2023kits21} & 489 & CT & abdomen & pathology & scaling \\
AMOS~\cite{ji2022amos} & 360 & CT + MR & abdomen & anatomy & scaling \\
BraTS2024 Glioma~\cite{verdier2024bratsgli} & 1350 & MR & brain & pathology & scaling \\
CTSpine1K~\cite{deng2025ctspine1k} & 1005 & CT & spine & anatomy & scaling \\
HNTSMRG24~\cite{wahid2024hntsmrg} & 300 & MR & head \& neck & pathology & scaling \\
Mama Mia~\cite{garrucho2025large} & 1506 & MRI & chest & pathology & scaling \\
\addlinespace[0.18em]
\multicolumn{6}{@{}l}{\textbf{Generalization}}\\
\addlinespace[0.02em]
M\&Ms challenge~\cite{campello2021multi} & 150 & MR & chest & anatomy & generalization \\
BraTS2024 Africa~\cite{bakas2024bratsafrica} & 60 & MR & brain & pathology & generalization \\
Spine-Mets-CT~\cite{haouchine2024spinemets} & 55 & CT & spine & anatomy & generalization \\
ACDC~\cite{bernard2018deep} & 200 & MR & Chest & anatomy & generalization \\
Care2025 WHS~\cite{care2025whs} & 46 & CT + MR & Chest & anatomy & generalization \\
BladderCancerMRI~\cite{cao2024bladdermri} & 55 & MR & lower abdomen / pelvis & pathology & generalization \\
PancreasSegMRI~\cite{zhang2025pansegnet} & 72 & MR & abdomen & anatomy & generalization \\
\addlinespace[0.18em]
\multicolumn{6}{@{}l}{\textbf{Classification}}\\
\addlinespace[0.02em]
RSNA Spine~\cite{lin2023rsnacervicalspine} & 2019 & CT & spine & pathology & classification \\
MRNet~\cite{bien2018skmtea} & 1370 & MR & knee & pathology & classification \\
CT-RATE~\cite{hamamci2026generalist} & 25000 & CT & chest & pathology & classification \\
RSNA Aneurysm~\cite{rsna2025intracranialaneurysm} & 4344 & CTA & brain & pathology & classification \\
RSNA Hemorrhage~\cite{rsna2019intracranialhemorrhage} & 18938 & CT & brain & pathology & classification \\
STOIC 2021~\cite{revel2021stoic} & 2000 & CT & chest & pathology & classification \\
ODELIA~\cite{muellerfranzes2025odelia} & 1022 & MR & chest & pathology & classification \\
LLD-MMRI2023~\cite{lldmmri2023} & 498 & MR & abdomen & pathology & classification \\
KiTS Liver~\cite{heller2023kits21} & 489 & CT & abdomen & pathology & classification \\
KiTS pancreas~\cite{heller2023kits21} & 489 & CT & abdomen & pathology & classification \\
\addlinespace[0.18em]
\multicolumn{6}{@{}l}{\textbf{Detection}}\\
\addlinespace[0.02em]
PI-CAI~\cite{saha2024picai} & 1295 & MR & pelvis & pathology & detection \\
PN9~\cite{mei2022sanet} & 8798 & CT & chest & pathology & detection \\
MRAAneurysms~\cite{lausanne2021tofmraaneurysm} & 296 & MRA & brain & pathology & detection \\
CADA~\cite{ivantsits2022cada} & 109 & CT & brain & pathology & detection \\
VALDO~\cite{sudre2021valdo} & 72 & MR & brain & pathology & detection \\
BraTSMets~\cite{moawad2023bratsmets} & 561 & MR & brain & pathology & detection \\
KIPA~\cite{he2021meta,he2020dense,shao2011laparoscopic,shao2012precise} & 70 & CT & abdomen & pathology & detection \\
Ribfrac~\cite{yang2024ribfrac} & 500 & CT & chest & pathology & detection \\
LIDC~\cite{armato2011lidcidri} & 1018 & CT & chest & pathology & detection \\
DUKE~\cite{saha2018radiogenomics} & 911 & MR & chest & pathology & detection \\
MELA~\cite{mela2022} & 880 & CT & chest & pathology & detection \\
\addlinespace[0.18em]
\multicolumn{6}{@{}l}{\textbf{Retrieval}}\\
\addlinespace[0.02em]
MSD Lung~\cite{antonelli2022medical} & 63 & CT & chest & pathology & retrieval \\
MSD Liver~\cite{antonelli2022medical} & 131 & CT & abdomen & pathology & retrieval \\
MSD Pancreas~\cite{antonelli2022medical} & 281 & CT & abdomen & pathology & retrieval \\
MSD Colon~\cite{antonelli2022medical} & 126 & CT & abdomen & pathology & retrieval \\
Mama Mia~\cite{garrucho2025large} & 1506 & MR & Breasts & pathology & retrieval \\
\addlinespace[0.18em]
\multicolumn{6}{@{}l}{\textbf{Report generation}}\\
\addlinespace[0.02em]
CT-Rate~\cite{hamamci2026generalist} & 25692 & CT & Chest & pathology & report generation \\
Merlin~\cite{Blankemeier2026} & 25528 & CT & Abdomen & pathology & report generation \\
\end{longtable}
\endgroup

\section{Detailed Results}
\label{app:detailed-results}
\setcounter{table}{0}
\renewcommand{\theHtable}{suppF.\arabic{table}}
\subsection{Rank Summary by Task Type and Modality}
\label{app:rank-summary-task-modality}

\begin{table}[!htbp]
\caption{Supplementary ranking overview for each evaluated model. Ranks are reported globally and separately for segmentation, classification, retrieval, and report generation, each split into overall, CT, and MRI performance where applicable. Cells report mean rank, bootstrap 95\% CI in parentheses, and SD across contributing datasets; for global ranks, SD is computed across task-type components. For nnFoundation we report results achieved by the recommended model, i.e. \Name{cnn} for segmentation and \Name{vit} for classification, retrieval and report generation. Global ranks apply equal task-type weighting.}
\label{tab:supp_task_modality_rankings}
\centering
{\fontsize{5.0}{5.6}\selectfont
\setlength{\tabcolsep}{0.55pt}
\renewcommand{\arraystretch}{0.90}
\begin{tabular*}{\textwidth}{@{\extracolsep{\fill}}l ccc ccc ccc c ccc@{}}
\toprule
\textbf{Model}
& \multicolumn{3}{c}{\makecell[c]{\textbf{Segmentation}\\\textbf{Rank}}}
& \multicolumn{3}{c}{\makecell[c]{\textbf{Classification}\\\textbf{Rank}}}
& \multicolumn{3}{c}{\makecell[c]{\textbf{Retrieval}\\\textbf{Rank}}}
& \multicolumn{1}{c}{\makecell[c]{\textbf{Report Gen}\\\textbf{Rank}}}
& \multicolumn{3}{c}{\makecell[c]{\textbf{Global}\\\textbf{Rank}}} \\
\cmidrule(lr){2-4}
\cmidrule(lr){5-7}
\cmidrule(lr){8-10}
\cmidrule(lr){11-11}
\cmidrule(lr){12-14}
& \textbf{Overall} & \textbf{CT} & \textbf{MRI}
& \textbf{Overall} & \textbf{CT} & \textbf{MRI}
& \textbf{Overall} & \textbf{CT} & \textbf{MRI}
& \makecell[c]{\textbf{Overall}\\\textbf{(CT only)}}
& \textbf{Overall} & \textbf{CT} & \textbf{MRI} \\
\midrule
CTFM~\protect\cite{pai2025vision} & \makecell[c]{3.35\\{\fontsize{3.0}{3.3}\selectfont (3.24--3.76)}\\{\fontsize{2.8}{3.0}\selectfont SD 1.72}} & \makecell[c]{3.57\\{\fontsize{3.0}{3.3}\selectfont (3.47--4.20)}\\{\fontsize{2.8}{3.0}\selectfont SD 1.92}} & \makecell[c]{3.18\\{\fontsize{3.0}{3.3}\selectfont (2.89--3.66)}\\{\fontsize{2.8}{3.0}\selectfont SD 1.58}} & \makecell[c]{3.19\\{\fontsize{3.0}{3.3}\selectfont (2.62--3.69)}\\{\fontsize{2.8}{3.0}\selectfont SD 1.36}} & \makecell[c]{2.75\\{\fontsize{3.0}{3.3}\selectfont (2.17--3.50)}\\{\fontsize{2.8}{3.0}\selectfont SD 1.25}} & \makecell[c]{4.50\\{\fontsize{3.0}{3.3}\selectfont (3.25--5.00)}\\{\fontsize{2.8}{3.0}\selectfont SD 0.71}} & \makecell[c]{3.70\\{\fontsize{3.0}{3.3}\selectfont (3.20--4.60)}\\{\fontsize{2.8}{3.0}\selectfont SD 1.64}} & \makecell[c]{4.12\\{\fontsize{3.0}{3.3}\selectfont (3.44--5.12)}\\{\fontsize{2.8}{3.0}\selectfont SD 1.55}} & \makecell[c]{2.00\\{\fontsize{3.0}{3.3}\selectfont (1.00--3.50)}\\{\fontsize{2.8}{3.0}\selectfont SD 0.00}} & \makecell[c]{3.25\\{\fontsize{3.0}{3.3}\selectfont (2.25--3.25)}\\{\fontsize{2.8}{3.0}\selectfont SD 0.35}} & \makecell[c]{3.37\\{\fontsize{3.0}{3.3}\selectfont (3.05--3.66)}\\{\fontsize{2.8}{3.0}\selectfont SD 0.23}} & \makecell[c]{3.42\\{\fontsize{3.0}{3.3}\selectfont (3.14--3.79)}\\{\fontsize{2.8}{3.0}\selectfont SD 0.58}} & \makecell[c]{3.23\\{\fontsize{3.0}{3.3}\selectfont (2.74--3.80)}\\{\fontsize{2.8}{3.0}\selectfont SD 1.25}} \\
CURIA~\protect\cite{dancette2025curia} & \makecell[c]{6.53\\{\fontsize{3.0}{3.3}\selectfont (6.24--6.75)}\\{\fontsize{2.8}{3.0}\selectfont SD 2.03}} & \makecell[c]{6.73\\{\fontsize{3.0}{3.3}\selectfont (6.27--7.03)}\\{\fontsize{2.8}{3.0}\selectfont SD 1.32}} & \makecell[c]{6.37\\{\fontsize{3.0}{3.3}\selectfont (6.05--6.66)}\\{\fontsize{2.8}{3.0}\selectfont SD 2.48}} & \makecell[c]{3.38\\{\fontsize{3.0}{3.3}\selectfont (2.75--4.00)}\\{\fontsize{2.8}{3.0}\selectfont SD 1.19}} & \makecell[c]{2.83\\{\fontsize{3.0}{3.3}\selectfont (2.17--3.58)}\\{\fontsize{2.8}{3.0}\selectfont SD 0.75}} & \makecell[c]{5.00\\{\fontsize{3.0}{3.3}\selectfont (4.00--6.25)}\\{\fontsize{2.8}{3.0}\selectfont SD 0.00}} & \makecell[c]{6.40\\{\fontsize{3.0}{3.3}\selectfont (5.50--6.70)}\\{\fontsize{2.8}{3.0}\selectfont SD 1.07}} & \makecell[c]{6.25\\{\fontsize{3.0}{3.3}\selectfont (5.12--6.75)}\\{\fontsize{2.8}{3.0}\selectfont SD 1.16}} & \makecell[c]{7.00\\{\fontsize{3.0}{3.3}\selectfont (7.00--7.50)}\\{\fontsize{2.8}{3.0}\selectfont SD 0.00}} & \makecell[c]{2.00\\{\fontsize{3.0}{3.3}\selectfont (1.50--3.00)}\\{\fontsize{2.8}{3.0}\selectfont SD 0.00}} & \makecell[c]{4.58\\{\fontsize{3.0}{3.3}\selectfont (4.27--4.93)}\\{\fontsize{2.8}{3.0}\selectfont SD 2.25}} & \makecell[c]{4.45\\{\fontsize{3.0}{3.3}\selectfont (4.04--4.83)}\\{\fontsize{2.8}{3.0}\selectfont SD 2.39}} & \makecell[c]{6.12\\{\fontsize{3.0}{3.3}\selectfont (5.73--6.57)}\\{\fontsize{2.8}{3.0}\selectfont SD 1.02}} \\
MERLIN~\protect\cite{Blankemeier2026} & \makecell[c]{4.41\\{\fontsize{3.0}{3.3}\selectfont (4.15--4.69)}\\{\fontsize{2.8}{3.0}\selectfont SD 1.79}} & \makecell[c]{4.90\\{\fontsize{3.0}{3.3}\selectfont (4.27--5.10)}\\{\fontsize{2.8}{3.0}\selectfont SD 1.72}} & \makecell[c]{4.03\\{\fontsize{3.0}{3.3}\selectfont (3.84--4.58)}\\{\fontsize{2.8}{3.0}\selectfont SD 1.80}} & \makecell[c]{5.12\\{\fontsize{3.0}{3.3}\selectfont (4.50--5.38)}\\{\fontsize{2.8}{3.0}\selectfont SD 2.64}} & \makecell[c]{6.50\\{\fontsize{3.0}{3.3}\selectfont (5.58--6.67)}\\{\fontsize{2.8}{3.0}\selectfont SD 0.84}} & \makecell[c]{1.00\\{\fontsize{3.0}{3.3}\selectfont (1.00--2.25)}\\{\fontsize{2.8}{3.0}\selectfont SD 0.00}} & \makecell[c]{2.25\\{\fontsize{3.0}{3.3}\selectfont (1.75--3.10)}\\{\fontsize{2.8}{3.0}\selectfont SD 1.51}} & \makecell[c]{1.81\\{\fontsize{3.0}{3.3}\selectfont (1.44--3.00)}\\{\fontsize{2.8}{3.0}\selectfont SD 1.36}} & \makecell[c]{4.00\\{\fontsize{3.0}{3.3}\selectfont (2.00--4.50)}\\{\fontsize{2.8}{3.0}\selectfont SD 0.00}} & \makecell[c]{3.75\\{\fontsize{3.0}{3.3}\selectfont (2.75--3.75)}\\{\fontsize{2.8}{3.0}\selectfont SD 0.35}} & \makecell[c]{3.88\\{\fontsize{3.0}{3.3}\selectfont (3.55--4.08)}\\{\fontsize{2.8}{3.0}\selectfont SD 1.23}} & \makecell[c]{4.24\\{\fontsize{3.0}{3.3}\selectfont (3.84--4.44)}\\{\fontsize{2.8}{3.0}\selectfont SD 1.97}} & \makecell[c]{3.01\\{\fontsize{3.0}{3.3}\selectfont (2.45--3.42)}\\{\fontsize{2.8}{3.0}\selectfont SD 1.74}} \\
MIS-FM~\protect\cite{wang2023mis} & \makecell[c]{4.13\\{\fontsize{3.0}{3.3}\selectfont (3.91--4.44)}\\{\fontsize{2.8}{3.0}\selectfont SD 2.16}} & \makecell[c]{3.83\\{\fontsize{3.0}{3.3}\selectfont (3.47--4.20)}\\{\fontsize{2.8}{3.0}\selectfont SD 2.50}} & \makecell[c]{4.37\\{\fontsize{3.0}{3.3}\selectfont (4.03--4.79)}\\{\fontsize{2.8}{3.0}\selectfont SD 1.88}} & \makecell[c]{8.00\\{\fontsize{3.0}{3.3}\selectfont (7.88--8.00)}\\{\fontsize{2.8}{3.0}\selectfont SD 0.00}} & \makecell[c]{8.00\\{\fontsize{3.0}{3.3}\selectfont (7.83--8.00)}\\{\fontsize{2.8}{3.0}\selectfont SD 0.00}} & \makecell[c]{8.00\\{\fontsize{3.0}{3.3}\selectfont (8.00--8.00)}\\{\fontsize{2.8}{3.0}\selectfont SD 0.00}} & \makecell[c]{8.00\\{\fontsize{3.0}{3.3}\selectfont (7.20--8.00)}\\{\fontsize{2.8}{3.0}\selectfont SD 0.00}} & \makecell[c]{8.00\\{\fontsize{3.0}{3.3}\selectfont (7.00--8.00)}\\{\fontsize{2.8}{3.0}\selectfont SD 0.00}} & \makecell[c]{8.00\\{\fontsize{3.0}{3.3}\selectfont (7.50--8.00)}\\{\fontsize{2.8}{3.0}\selectfont SD 0.00}} & \makecell[c]{8.00\\{\fontsize{3.0}{3.3}\selectfont (8.00--8.00)}\\{\fontsize{2.8}{3.0}\selectfont SD 0.00}} & \makecell[c]{7.03\\{\fontsize{3.0}{3.3}\selectfont (6.82--7.09)}\\{\fontsize{2.8}{3.0}\selectfont SD 1.93}} & \makecell[c]{6.96\\{\fontsize{3.0}{3.3}\selectfont (6.69--7.02)}\\{\fontsize{2.8}{3.0}\selectfont SD 2.08}} & \makecell[c]{6.79\\{\fontsize{3.0}{3.3}\selectfont (6.59--6.92)}\\{\fontsize{2.8}{3.0}\selectfont SD 2.10}} \\
SwinUNETR~\protect\cite{tang2022self} & \makecell[c]{6.81\\{\fontsize{3.0}{3.3}\selectfont (6.46--6.82)}\\{\fontsize{2.8}{3.0}\selectfont SD 1.39}} & \makecell[c]{6.57\\{\fontsize{3.0}{3.3}\selectfont (6.10--6.60)}\\{\fontsize{2.8}{3.0}\selectfont SD 1.91}} & \makecell[c]{7.00\\{\fontsize{3.0}{3.3}\selectfont (6.58--7.11)}\\{\fontsize{2.8}{3.0}\selectfont SD 0.80}} & \makecell[c]{5.75\\{\fontsize{3.0}{3.3}\selectfont (5.12--6.00)}\\{\fontsize{2.8}{3.0}\selectfont SD 1.07}} & \makecell[c]{5.42\\{\fontsize{3.0}{3.3}\selectfont (4.58--5.83)}\\{\fontsize{2.8}{3.0}\selectfont SD 1.02}} & \makecell[c]{6.75\\{\fontsize{3.0}{3.3}\selectfont (5.75--7.00)}\\{\fontsize{2.8}{3.0}\selectfont SD 0.35}} & \makecell[c]{3.50\\{\fontsize{3.0}{3.3}\selectfont (2.90--4.10)}\\{\fontsize{2.8}{3.0}\selectfont SD 1.58}} & \makecell[c]{3.00\\{\fontsize{3.0}{3.3}\selectfont (2.25--3.75)}\\{\fontsize{2.8}{3.0}\selectfont SD 1.31}} & \makecell[c]{5.50\\{\fontsize{3.0}{3.3}\selectfont (5.00--6.00)}\\{\fontsize{2.8}{3.0}\selectfont SD 0.71}} & \makecell[c]{6.00\\{\fontsize{3.0}{3.3}\selectfont (6.00--6.00)}\\{\fontsize{2.8}{3.0}\selectfont SD 1.41}} & \makecell[c]{5.51\\{\fontsize{3.0}{3.3}\selectfont (5.23--5.63)}\\{\fontsize{2.8}{3.0}\selectfont SD 1.42}} & \makecell[c]{5.25\\{\fontsize{3.0}{3.3}\selectfont (4.90--5.42)}\\{\fontsize{2.8}{3.0}\selectfont SD 1.57}} & \makecell[c]{6.42\\{\fontsize{3.0}{3.3}\selectfont (5.93--6.57)}\\{\fontsize{2.8}{3.0}\selectfont SD 0.80}} \\
VISTA3D~\protect\cite{he2025vista3d} & \makecell[c]{3.88\\{\fontsize{3.0}{3.3}\selectfont (3.50--4.03)}\\{\fontsize{2.8}{3.0}\selectfont SD 1.58}} & \makecell[c]{3.73\\{\fontsize{3.0}{3.3}\selectfont (3.53--4.20)}\\{\fontsize{2.8}{3.0}\selectfont SD 1.88}} & \makecell[c]{4.00\\{\fontsize{3.0}{3.3}\selectfont (3.29--4.13)}\\{\fontsize{2.8}{3.0}\selectfont SD 1.33}} & \makecell[c]{3.12\\{\fontsize{3.0}{3.3}\selectfont (2.44--3.50)}\\{\fontsize{2.8}{3.0}\selectfont SD 1.96}} & \makecell[c]{3.50\\{\fontsize{3.0}{3.3}\selectfont (2.75--4.00)}\\{\fontsize{2.8}{3.0}\selectfont SD 2.17}} & \makecell[c]{2.00\\{\fontsize{3.0}{3.3}\selectfont (1.25--2.75)}\\{\fontsize{2.8}{3.0}\selectfont SD 0.00}} & \makecell[c]{4.60\\{\fontsize{3.0}{3.3}\selectfont (3.70--5.00)}\\{\fontsize{2.8}{3.0}\selectfont SD 1.71}} & \makecell[c]{5.00\\{\fontsize{3.0}{3.3}\selectfont (3.88--5.38)}\\{\fontsize{2.8}{3.0}\selectfont SD 1.69}} & \makecell[c]{3.00\\{\fontsize{3.0}{3.3}\selectfont (2.00--4.00)}\\{\fontsize{2.8}{3.0}\selectfont SD 0.00}} & \makecell[c]{6.00\\{\fontsize{3.0}{3.3}\selectfont (5.50--6.50)}\\{\fontsize{2.8}{3.0}\selectfont SD 1.41}} & \makecell[c]{4.40\\{\fontsize{3.0}{3.3}\selectfont (4.03--4.53)}\\{\fontsize{2.8}{3.0}\selectfont SD 1.22}} & \makecell[c]{4.56\\{\fontsize{3.0}{3.3}\selectfont (4.20--4.78)}\\{\fontsize{2.8}{3.0}\selectfont SD 1.17}} & \makecell[c]{3.00\\{\fontsize{3.0}{3.3}\selectfont (2.37--3.33)}\\{\fontsize{2.8}{3.0}\selectfont SD 1.00}} \\
VOCO~\protect\cite{wu2024voco} & \makecell[c]{4.82\\{\fontsize{3.0}{3.3}\selectfont (4.57--5.09)}\\{\fontsize{2.8}{3.0}\selectfont SD 1.77}} & \makecell[c]{4.23\\{\fontsize{3.0}{3.3}\selectfont (3.80--4.63)}\\{\fontsize{2.8}{3.0}\selectfont SD 1.55}} & \makecell[c]{5.29\\{\fontsize{3.0}{3.3}\selectfont (4.97--5.63)}\\{\fontsize{2.8}{3.0}\selectfont SD 1.84}} & \makecell[c]{5.06\\{\fontsize{3.0}{3.3}\selectfont (4.62--5.75)}\\{\fontsize{2.8}{3.0}\selectfont SD 1.90}} & \makecell[c]{4.83\\{\fontsize{3.0}{3.3}\selectfont (4.33--5.67)}\\{\fontsize{2.8}{3.0}\selectfont SD 2.14}} & \makecell[c]{5.75\\{\fontsize{3.0}{3.3}\selectfont (4.50--6.75)}\\{\fontsize{2.8}{3.0}\selectfont SD 1.06}} & \makecell[c]{5.50\\{\fontsize{3.0}{3.3}\selectfont (4.80--6.30)}\\{\fontsize{2.8}{3.0}\selectfont SD 1.18}} & \makecell[c]{5.50\\{\fontsize{3.0}{3.3}\selectfont (4.75--6.38)}\\{\fontsize{2.8}{3.0}\selectfont SD 1.31}} & \makecell[c]{5.50\\{\fontsize{3.0}{3.3}\selectfont (4.50--6.00)}\\{\fontsize{2.8}{3.0}\selectfont SD 0.71}} & \makecell[c]{6.00\\{\fontsize{3.0}{3.3}\selectfont (5.50--6.50)}\\{\fontsize{2.8}{3.0}\selectfont SD 0.00}} & \makecell[c]{5.35\\{\fontsize{3.0}{3.3}\selectfont (5.11--5.65)}\\{\fontsize{2.8}{3.0}\selectfont SD 0.52}} & \makecell[c]{5.14\\{\fontsize{3.0}{3.3}\selectfont (4.88--5.49)}\\{\fontsize{2.8}{3.0}\selectfont SD 0.77}} & \makecell[c]{5.51\\{\fontsize{3.0}{3.3}\selectfont (4.98--5.93)}\\{\fontsize{2.8}{3.0}\selectfont SD 0.23}} \\
\midrule
nnFoundation & \makecell[c]{2.06\\{\fontsize{3.0}{3.3}\selectfont (1.93--2.46)}\\{\fontsize{2.8}{3.0}\selectfont SD 1.56}} & \makecell[c]{2.43\\{\fontsize{3.0}{3.3}\selectfont (2.10--3.00)}\\{\fontsize{2.8}{3.0}\selectfont SD 1.91}} & \makecell[c]{1.76\\{\fontsize{3.0}{3.3}\selectfont (1.61--2.26)}\\{\fontsize{2.8}{3.0}\selectfont SD 1.18}} & \makecell[c]{2.38\\{\fontsize{3.0}{3.3}\selectfont (2.19--3.31)}\\{\fontsize{2.8}{3.0}\selectfont SD 1.06}} & \makecell[c]{2.17\\{\fontsize{3.0}{3.3}\selectfont (1.92--3.25)}\\{\fontsize{2.8}{3.0}\selectfont SD 1.17}} & \makecell[c]{3.00\\{\fontsize{3.0}{3.3}\selectfont (2.00--4.25)}\\{\fontsize{2.8}{3.0}\selectfont SD 0.00}} & \makecell[c]{2.05\\{\fontsize{3.0}{3.3}\selectfont (1.85--3.20)}\\{\fontsize{2.8}{3.0}\selectfont SD 0.90}} & \makecell[c]{2.31\\{\fontsize{3.0}{3.3}\selectfont (2.06--3.62)}\\{\fontsize{2.8}{3.0}\selectfont SD 0.80}} & \makecell[c]{1.00\\{\fontsize{3.0}{3.3}\selectfont (1.00--3.00)}\\{\fontsize{2.8}{3.0}\selectfont SD 0.00}} & \makecell[c]{1.00\\{\fontsize{3.0}{3.3}\selectfont (1.00--2.50)}\\{\fontsize{2.8}{3.0}\selectfont SD 0.00}} & \makecell[c]{1.87\\{\fontsize{3.0}{3.3}\selectfont (1.89--2.46)}\\{\fontsize{2.8}{3.0}\selectfont SD 0.60}} & \makecell[c]{1.98\\{\fontsize{3.0}{3.3}\selectfont (1.93--2.61)}\\{\fontsize{2.8}{3.0}\selectfont SD 0.66}} & \makecell[c]{1.92\\{\fontsize{3.0}{3.3}\selectfont (1.75--2.69)}\\{\fontsize{2.8}{3.0}\selectfont SD 1.01}} \\
\bottomrule
\end{tabular*}
}
\end{table}

\begin{table}[ht]
\caption{Pairwise comparisons corresponding to the overall rank endpoints in Table~\ref{tab:supp_task_modality_rankings}. Cells report the comparator-minus-nnFoundation rank difference with paired 95\% CI in parentheses; positive values favor nnFoundation because lower ranks are better. A dagger marks comparisons for which the paired 95\% CI does not lie entirely above zero; these comparisons therefore do not support nnFoundation as top-ranked versus that comparator. For the global endpoint, task-specific paired rank differences are averaged with equal task-type weighting within each bootstrap replicate.}
\label{tab:supp_overall_pairwise_rank_support}
\centering
{\fontsize{5.8}{6.5}\selectfont
\setlength{\tabcolsep}{0.9pt}
\renewcommand{\arraystretch}{0.98}
\begin{tabular*}{\textwidth}{@{\extracolsep{\fill}}lccccc@{}}
\toprule
Comparator & \makecell[c]{Seg.\\(\Name{cnn})} & \makecell[c]{Class.\\(\Name{vit})} & \makecell[c]{Retrieval\\(\Name{vit})} & \makecell[c]{Report\\Gen\\(\Name{vit})} & Global \\
\midrule
CTFM~\protect\cite{pai2025vision} & 1.29 (0.93--1.75) & 0.81 (-0.44--1.31)$^\dagger$ & 1.65 (0.35--2.45) & 2.25 (0.25--2.25) & 1.50 (0.74--1.64) \\
CURIA~\protect\cite{dancette2025curia} & 4.47 (3.94--4.68) & 1.00 (-0.31--1.62)$^\dagger$ & 4.35 (2.65--4.55) & 1.00 (0.00--2.00)$^\dagger$ & 2.71 (1.97--2.93) \\
MERLIN~\protect\cite{Blankemeier2026} & 2.35 (1.82--2.63) & 2.75 (1.44--3.00) & 0.20 (-1.10--0.90)$^\dagger$ & 2.75 (0.75--2.75) & 2.01 (1.22--2.08) \\
MIS-FM~\protect\cite{wang2023mis} & 2.07 (1.57--2.40) & 5.62 (4.69--5.81) & 5.95 (4.25--5.95) & 7.00 (5.50--7.00) & 5.16 (4.49--5.13) \\
SwinUNETR~\protect\cite{tang2022self} & 4.75 (4.09--4.78) & 3.38 (2.12--3.62) & 1.45 (-0.10--2.00)$^\dagger$ & 5.00 (3.50--5.00) & 3.64 (2.91--3.65) \\
VISTA3D~\protect\cite{he2025vista3d} & 1.82 (1.15--1.97) & 0.75 (-0.50--1.06)$^\dagger$ & 2.55 (0.70--2.80) & 5.00 (3.50--5.50) & 2.53 (1.74--2.51) \\
VOCO~\protect\cite{wu2024voco} & 2.76 (2.25--3.01) & 2.69 (1.69--3.31) & 3.45 (1.95--3.95) & 5.00 (3.50--5.50) & 3.48 (2.83--3.63) \\
\bottomrule
\end{tabular*}
}
\end{table}

\clearpage
\subsection{Figure 2 Extended Results}
\label{app:figure2-metric-tables}

\subsubsection{Summary Tables}
\label{app:figure2-summary-tables}

\begin{table}[!htbp]
\caption{Figure 2 ranking overview. Ranks are lower-is-better mean panel ranks with 95\% confidence intervals. Cells report mean rank, bootstrap 95\% CI in parentheses, and SD across contributing datasets. Each column uses the exact model pool shown in the corresponding Figure 2 panel, so ranking pools differ by task.}
\label{tab:fig2-panel-rank-overview}
\centering
{\fontsize{5.8}{6.6}\selectfont
\setlength{\tabcolsep}{1.6pt}
\renewcommand{\arraystretch}{0.95}
\begin{tabular*}{\textwidth}{@{\extracolsep{\fill}}l cccc@{}}
\toprule
\textbf{Name} & \textbf{Segmentation Rank} & \textbf{Classification Rank} & \textbf{Report Gen Rank} & \textbf{Retrieval Rank} \\
\midrule
CTFM~\protect\cite{pai2025vision} & \makecell[c]{5.09\\{\fontsize{3.6}{4.0}\selectfont (4.76--5.56)}\\{\fontsize{3.3}{3.7}\selectfont SD 2.71}} & \makecell[c]{3.62\\{\fontsize{3.6}{4.0}\selectfont (2.94--4.50)}\\{\fontsize{3.3}{3.7}\selectfont SD 1.58}} & \makecell[c]{3.25\\{\fontsize{3.6}{4.0}\selectfont (2.25--3.25)}\\{\fontsize{3.3}{3.7}\selectfont SD 0.35}} & \makecell[c]{3.70\\{\fontsize{3.6}{4.0}\selectfont (3.20--4.80)}\\{\fontsize{3.3}{3.7}\selectfont SD 1.64}} \\
CURIA~\protect\cite{dancette2025curia} & \makecell[c]{9.59\\{\fontsize{3.6}{4.0}\selectfont (9.16--9.96)}\\{\fontsize{3.3}{3.7}\selectfont SD 2.87}} & \makecell[c]{3.81\\{\fontsize{3.6}{4.0}\selectfont (2.94--4.94)}\\{\fontsize{3.3}{3.7}\selectfont SD 1.56}} & \makecell[c]{2.00\\{\fontsize{3.6}{4.0}\selectfont (1.50--3.00)}\\{\fontsize{3.3}{3.7}\selectfont SD 0.00}} & \makecell[c]{6.90\\{\fontsize{3.6}{4.0}\selectfont (5.90--7.40)}\\{\fontsize{3.3}{3.7}\selectfont SD 1.45}} \\
MERLIN~\protect\cite{Blankemeier2026} & \makecell[c]{6.44\\{\fontsize{3.6}{4.0}\selectfont (6.07--6.88)}\\{\fontsize{3.3}{3.7}\selectfont SD 2.44}} & \makecell[c]{6.62\\{\fontsize{3.6}{4.0}\selectfont (5.75--7.12)}\\{\fontsize{3.3}{3.7}\selectfont SD 3.67}} & \makecell[c]{3.75\\{\fontsize{3.6}{4.0}\selectfont (2.75--3.75)}\\{\fontsize{3.3}{3.7}\selectfont SD 0.35}} & \makecell[c]{2.25\\{\fontsize{3.6}{4.0}\selectfont (1.75--3.20)}\\{\fontsize{3.3}{3.7}\selectfont SD 1.51}} \\
MIS-FM~\protect\cite{wang2023mis} & \makecell[c]{6.09\\{\fontsize{3.6}{4.0}\selectfont (5.76--6.51)}\\{\fontsize{3.3}{3.7}\selectfont SD 3.20}} & \makecell[c]{11.00\\{\fontsize{3.6}{4.0}\selectfont (10.87--11.00)}\\{\fontsize{3.3}{3.7}\selectfont SD 0.00}} & \makecell[c]{9.00\\{\fontsize{3.6}{4.0}\selectfont (9.00--9.00)}\\{\fontsize{3.3}{3.7}\selectfont SD 0.00}} & \makecell[c]{9.00\\{\fontsize{3.6}{4.0}\selectfont (8.00--9.00)}\\{\fontsize{3.3}{3.7}\selectfont SD 0.00}} \\
SwinUNETR~\protect\cite{tang2022self} & \makecell[c]{9.79\\{\fontsize{3.6}{4.0}\selectfont (9.28--9.85)}\\{\fontsize{3.3}{3.7}\selectfont SD 2.14}} & \makecell[c]{7.44\\{\fontsize{3.6}{4.0}\selectfont (6.31--7.88)}\\{\fontsize{3.3}{3.7}\selectfont SD 1.50}} & \makecell[c]{6.50\\{\fontsize{3.6}{4.0}\selectfont (6.50--6.50)}\\{\fontsize{3.3}{3.7}\selectfont SD 2.12}} & \makecell[c]{3.50\\{\fontsize{3.6}{4.0}\selectfont (2.90--4.20)}\\{\fontsize{3.3}{3.7}\selectfont SD 1.58}} \\
VISTA3D~\protect\cite{he2025vista3d} & \makecell[c]{5.65\\{\fontsize{3.6}{4.0}\selectfont (5.16--5.90)}\\{\fontsize{3.3}{3.7}\selectfont SD 2.57}} & \makecell[c]{4.06\\{\fontsize{3.6}{4.0}\selectfont (3.06--4.56)}\\{\fontsize{3.3}{3.7}\selectfont SD 3.10}} & \makecell[c]{6.50\\{\fontsize{3.6}{4.0}\selectfont (6.00--7.50)}\\{\fontsize{3.3}{3.7}\selectfont SD 2.12}} & \makecell[c]{4.80\\{\fontsize{3.6}{4.0}\selectfont (3.70--5.20)}\\{\fontsize{3.3}{3.7}\selectfont SD 1.99}} \\
VOCO~\protect\cite{wu2024voco} & \makecell[c]{6.76\\{\fontsize{3.6}{4.0}\selectfont (6.47--7.22)}\\{\fontsize{3.3}{3.7}\selectfont SD 2.67}} & \makecell[c]{6.69\\{\fontsize{3.6}{4.0}\selectfont (5.94--7.50)}\\{\fontsize{3.3}{3.7}\selectfont SD 2.78}} & \makecell[c]{7.00\\{\fontsize{3.6}{4.0}\selectfont (6.50--7.50)}\\{\fontsize{3.3}{3.7}\selectfont SD 0.00}} & \makecell[c]{5.70\\{\fontsize{3.6}{4.0}\selectfont (4.90--6.70)}\\{\fontsize{3.3}{3.7}\selectfont SD 1.49}} \\
nnU-Net Default~\protect\cite{isensee2021nnu} & \makecell[c]{4.43\\{\fontsize{3.6}{4.0}\selectfont (4.15--4.94)}\\{\fontsize{3.3}{3.7}\selectfont SD 3.15}} & NA & NA & NA \\
\midrule
\Name{cnn} (pretrained) & \makecell[c]{2.82\\{\fontsize{3.6}{4.0}\selectfont (2.57--3.34)}\\{\fontsize{3.3}{3.7}\selectfont SD 2.32}} & \makecell[c]{7.12\\{\fontsize{3.6}{4.0}\selectfont (6.37--8.00)}\\{\fontsize{3.3}{3.7}\selectfont SD 3.00}} & \makecell[c]{6.00\\{\fontsize{3.6}{4.0}\selectfont (5.50--6.00)}\\{\fontsize{3.3}{3.7}\selectfont SD 0.00}} & \makecell[c]{7.10\\{\fontsize{3.6}{4.0}\selectfont (6.10--7.60)}\\{\fontsize{3.3}{3.7}\selectfont SD 0.57}} \\
\Name{cnn} (scratch) & \makecell[c]{4.18\\{\fontsize{3.6}{4.0}\selectfont (3.94--4.66)}\\{\fontsize{3.3}{3.7}\selectfont SD 2.37}} & \makecell[c]{4.62\\{\fontsize{3.6}{4.0}\selectfont (3.88--5.50)}\\{\fontsize{3.3}{3.7}\selectfont SD 1.73}} & NA & NA \\
\Name{vit} (pretrained) & \makecell[c]{7.19\\{\fontsize{3.6}{4.0}\selectfont (6.60--7.41)}\\{\fontsize{3.3}{3.7}\selectfont SD 2.84}} & \makecell[c]{2.62\\{\fontsize{3.6}{4.0}\selectfont (2.38--3.81)}\\{\fontsize{3.3}{3.7}\selectfont SD 1.19}} & \makecell[c]{1.00\\{\fontsize{3.6}{4.0}\selectfont (1.00--2.50)}\\{\fontsize{3.3}{3.7}\selectfont SD 0.00}} & \makecell[c]{2.05\\{\fontsize{3.6}{4.0}\selectfont (1.85--3.35)}\\{\fontsize{3.3}{3.7}\selectfont SD 0.90}} \\
\Name{vit} (scratch) & \makecell[c]{9.97\\{\fontsize{3.6}{4.0}\selectfont (9.62--10.16)}\\{\fontsize{3.3}{3.7}\selectfont SD 2.56}} & \makecell[c]{8.38\\{\fontsize{3.6}{4.0}\selectfont (7.56--9.00)}\\{\fontsize{3.3}{3.7}\selectfont SD 1.30}} & NA & NA \\
\bottomrule
\end{tabular*}
}
\end{table}

\begin{table}[!htbp]
\caption{Pairwise rank-difference comparisons corresponding to \cref{tab:fig2-panel-rank-overview}. Cells report comparator-minus-nnFoundation mean rank differences with paired 95\% CIs in parentheses; positive values favor nnFoundation because lower ranks are better. A dagger marks comparisons for which the paired 95\% CI does not lie entirely above zero.}
\label{tab:fig2-panel-rank-pairwise-support}
\centering
{\fontsize{6.1}{6.9}\selectfont
\setlength{\tabcolsep}{1.2pt}
\renewcommand{\arraystretch}{0.96}
\begin{tabular*}{\textwidth}{@{\extracolsep{\fill}}lcccc@{}}
\toprule
Comparator & \makecell[c]{Seg.\\(\Name{cnn})} & \makecell[c]{Class.\\(\Name{vit})} & \makecell[c]{Report Gen\\(\Name{vit})} & \makecell[c]{Retrieval\\(\Name{vit})} \\
\midrule
CTFM~\protect\cite{pai2025vision} & 2.26 (1.62--2.79) & 1.00 (-0.56--1.81)$^\dagger$ & 2.25 (0.25--2.25) & 1.65 (0.35--2.55) \\
CURIA~\protect\cite{dancette2025curia} & 6.76 (6.01--7.13) & 1.19 (-0.44--2.19)$^\dagger$ & 1.00 (0.00--2.00)$^\dagger$ & 4.85 (3.00--5.20) \\
MERLIN~\protect\cite{Blankemeier2026} & 3.62 (2.93--4.06) & 4.00 (2.31--4.44) & 2.75 (0.75--2.75) & 0.20 (-1.20--1.00)$^\dagger$ \\
MIS-FM~\protect\cite{wang2023mis} & 3.26 (2.63--3.72) & 8.38 (7.12--8.62) & 8.00 (6.50--8.00) & 6.95 (5.05--6.95) \\
SwinUNETR~\protect\cite{tang2022self} & 6.97 (6.07--7.07) & 4.81 (2.94--5.06) & 5.50 (4.00--5.50) & 1.45 (-0.15--2.05)$^\dagger$ \\
VISTA3D~\protect\cite{he2025vista3d} & 2.82 (1.97--3.15) & 1.44 (-0.31--1.81)$^\dagger$ & 5.50 (4.50--6.50) & 2.75 (0.80--3.00) \\
VOCO~\protect\cite{wu2024voco} & 3.94 (3.34--4.41) & 4.06 (2.50--4.75) & 6.00 (4.50--6.50) & 3.65 (2.15--4.35) \\
nnU-Net Default~\protect\cite{isensee2021nnu} & 1.60 (1.06--2.18) & NA & NA & NA \\
\midrule
\Name{cnn} (pretrained) & Reference & 4.50 (2.94--5.25) & 5.00 (3.50--5.00) & 5.05 (3.15--5.30) \\
\Name{cnn} (scratch) & 1.35 (0.82--1.85) & 2.00 (0.44--2.81) & NA & NA \\
\Name{vit} (pretrained) & 4.37 (3.50--4.60) & Reference & Reference & Reference \\
\Name{vit} (scratch) & 7.15 (6.49--7.35) & 5.75 (4.12--6.19) & NA & NA \\
\bottomrule
\end{tabular*}
}
\end{table}

\begin{table}[!htbp]
\caption{Figure 2 summary metric scores for semantic segmentation across 34 datasets. Values are dataset-equal means with fixed-dataset within-dataset bootstrap 95\% CIs; SD reports variability across dataset-level metric means.}
\label{tab:fig2-summary-segmentation}
\centering
\tablebodyfont
\setlength{\tabcolsep}{2pt}
\renewcommand{\arraystretch}{0.90}
\begin{tabular}{@{}l@{\hspace{1.20em}}c@{\hspace{1.20em}}c@{}}
\toprule
\textbf{Model} & \multicolumn{1}{c}{\textbf{DSC}} & \multicolumn{1}{c}{\textbf{NSD}} \\
\cmidrule(lr){2-2}\cmidrule(lr){3-3}
 & \multicolumn{1}{c}{\makecell[c]{\textbf{Mean (95\% CI)}\\\textbf{SD}}} & \multicolumn{1}{c}{\makecell[c]{\textbf{Mean (95\% CI)}\\\textbf{SD}}} \\
\midrule
\Name{cnn} & \makecell[c]{0.715\\{\fontsize{5.7}{6.2}\selectfont (0.705--0.724)}\\{\fontsize{5.4}{5.9}\selectfont SD 0.165}} & \makecell[c]{0.640\\{\fontsize{5.7}{6.2}\selectfont (0.631--0.648)}\\{\fontsize{5.4}{5.9}\selectfont SD 0.194}} \\
\Name{cnn} (scratch) & \makecell[c]{0.704\\{\fontsize{5.7}{6.2}\selectfont (0.695--0.713)}\\{\fontsize{5.4}{5.9}\selectfont SD 0.175}} & \makecell[c]{0.628\\{\fontsize{5.7}{6.2}\selectfont (0.620--0.636)}\\{\fontsize{5.4}{5.9}\selectfont SD 0.203}} \\
nnU-Net Default~\protect\cite{isensee2021nnu} & \makecell[c]{0.696\\{\fontsize{5.7}{6.2}\selectfont (0.687--0.705)}\\{\fontsize{5.4}{5.9}\selectfont SD 0.184}} & \makecell[c]{0.621\\{\fontsize{5.7}{6.2}\selectfont (0.614--0.629)}\\{\fontsize{5.4}{5.9}\selectfont SD 0.209}} \\
CTFM~\protect\cite{pai2025vision} & \makecell[c]{0.685\\{\fontsize{5.7}{6.2}\selectfont (0.677--0.695)}\\{\fontsize{5.4}{5.9}\selectfont SD 0.196}} & \makecell[c]{0.611\\{\fontsize{5.7}{6.2}\selectfont (0.603--0.619)}\\{\fontsize{5.4}{5.9}\selectfont SD 0.228}} \\
VISTA3D~\protect\cite{he2025vista3d} & \makecell[c]{0.665\\{\fontsize{5.7}{6.2}\selectfont (0.653--0.678)}\\{\fontsize{5.4}{5.9}\selectfont SD 0.187}} & \makecell[c]{0.594\\{\fontsize{5.7}{6.2}\selectfont (0.584--0.604)}\\{\fontsize{5.4}{5.9}\selectfont SD 0.217}} \\
MISFM~\protect\cite{wang2023mis} & \makecell[c]{0.665\\{\fontsize{5.7}{6.2}\selectfont (0.655--0.675)}\\{\fontsize{5.4}{5.9}\selectfont SD 0.207}} & \makecell[c]{0.594\\{\fontsize{5.7}{6.2}\selectfont (0.585--0.602)}\\{\fontsize{5.4}{5.9}\selectfont SD 0.234}} \\
MERLIN~\protect\cite{Blankemeier2026} & \makecell[c]{0.680\\{\fontsize{5.7}{6.2}\selectfont (0.671--0.688)}\\{\fontsize{5.4}{5.9}\selectfont SD 0.191}} & \makecell[c]{0.606\\{\fontsize{5.7}{6.2}\selectfont (0.599--0.614)}\\{\fontsize{5.4}{5.9}\selectfont SD 0.219}} \\
VOCO~\protect\cite{wu2024voco} & \makecell[c]{0.661\\{\fontsize{5.7}{6.2}\selectfont (0.651--0.671)}\\{\fontsize{5.4}{5.9}\selectfont SD 0.216}} & \makecell[c]{0.592\\{\fontsize{5.7}{6.2}\selectfont (0.584--0.600)}\\{\fontsize{5.4}{5.9}\selectfont SD 0.237}} \\
\Name{vit} & \makecell[c]{0.674\\{\fontsize{5.7}{6.2}\selectfont (0.665--0.684)}\\{\fontsize{5.4}{5.9}\selectfont SD 0.195}} & \makecell[c]{0.601\\{\fontsize{5.7}{6.2}\selectfont (0.593--0.609)}\\{\fontsize{5.4}{5.9}\selectfont SD 0.223}} \\
CURIA~\protect\cite{dancette2025curia} & \makecell[c]{0.637\\{\fontsize{5.7}{6.2}\selectfont (0.629--0.645)}\\{\fontsize{5.4}{5.9}\selectfont SD 0.198}} & \makecell[c]{0.545\\{\fontsize{5.7}{6.2}\selectfont (0.538--0.553)}\\{\fontsize{5.4}{5.9}\selectfont SD 0.217}} \\
SwinUNETR~\protect\cite{tang2022self} & \makecell[c]{0.618\\{\fontsize{5.7}{6.2}\selectfont (0.609--0.628)}\\{\fontsize{5.4}{5.9}\selectfont SD 0.238}} & \makecell[c]{0.549\\{\fontsize{5.7}{6.2}\selectfont (0.541--0.558)}\\{\fontsize{5.4}{5.9}\selectfont SD 0.260}} \\
\Name{vit} (scratch) & \makecell[c]{0.585\\{\fontsize{5.7}{6.2}\selectfont (0.576--0.594)}\\{\fontsize{5.4}{5.9}\selectfont SD 0.249}} & \makecell[c]{0.511\\{\fontsize{5.7}{6.2}\selectfont (0.502--0.519)}\\{\fontsize{5.4}{5.9}\selectfont SD 0.266}} \\
\bottomrule
\end{tabular}
\end{table}
\normalsize

\begin{table}[!htbp]
\caption{Pairwise metric-difference comparisons corresponding to \cref{tab:fig2-summary-segmentation}. Cells report nnFoundationCNN-minus-comparator dataset-equal mean differences with paired 95\% CIs in parentheses; positive values favor nnFoundationCNN because higher DSC and NSD are better. A dagger marks comparisons for which the paired 95\% CI does not lie entirely above zero.}
\label{tab:fig2-summary-segmentation-pairwise-support}
\centering
\tablebodyfont
\setlength{\tabcolsep}{2pt}
\renewcommand{\arraystretch}{0.90}
\begin{tabular}{@{}l@{\hspace{1.20em}}c@{\hspace{1.20em}}c@{}}
\toprule
\textbf{Comparator} & \multicolumn{1}{c}{\textbf{DSC}} & \multicolumn{1}{c}{\textbf{NSD}} \\
\cmidrule(lr){2-2}\cmidrule(lr){3-3}
 & \multicolumn{1}{c}{\textbf{\(\Delta\) (95\% CI)}} & \multicolumn{1}{c}{\textbf{\(\Delta\) (95\% CI)}} \\
\midrule
\Name{cnn} & Reference & Reference \\
\Name{cnn} (scratch) & 0.012 (0.006--0.017) & 0.012 (0.008--0.016) \\
nnU-Net Default~\protect\cite{isensee2021nnu} & 0.020 (0.014--0.026) & 0.019 (0.014--0.023) \\
CTFM~\protect\cite{pai2025vision} & 0.030 (0.023--0.038) & 0.028 (0.023--0.034) \\
VISTA3D~\protect\cite{he2025vista3d} & 0.050 (0.041--0.060) & 0.046 (0.038--0.054) \\
MISFM~\protect\cite{wang2023mis} & 0.051 (0.043--0.058) & 0.046 (0.040--0.052) \\
MERLIN~\protect\cite{Blankemeier2026} & 0.036 (0.030--0.043) & 0.034 (0.029--0.039) \\
VOCO~\protect\cite{wu2024voco} & 0.055 (0.047--0.063) & 0.048 (0.042--0.053) \\
\Name{vit} & 0.041 (0.034--0.048) & 0.039 (0.033--0.045) \\
CURIA~\protect\cite{dancette2025curia} & 0.078 (0.071--0.085) & 0.094 (0.088--0.100) \\
SwinUNETR~\protect\cite{tang2022self} & 0.097 (0.089--0.105) & 0.090 (0.083--0.097) \\
\Name{vit} (scratch) & 0.130 (0.122--0.139) & 0.129 (0.123--0.136) \\
\bottomrule
\end{tabular}
\end{table}
\normalsize

\begin{table}[!htbp]
\caption{Figure 2 summary metric scores for classification across 8 datasets. Values are dataset-equal means with fixed-dataset within-dataset bootstrap 95\% CIs; SD reports variability across dataset-level metric means.}
\label{tab:fig2-summary-classification}
\centering
\tablebodyfont
\setlength{\tabcolsep}{2pt}
\renewcommand{\arraystretch}{0.90}
\begin{tabular}{@{}l@{\hspace{1.20em}}c@{\hspace{1.20em}}c@{}}
\toprule
\textbf{Model} & \multicolumn{1}{c}{\textbf{AUROC}} & \multicolumn{1}{c}{\textbf{AUPRC}} \\
\cmidrule(lr){2-2}\cmidrule(lr){3-3}
 & \multicolumn{1}{c}{\makecell[c]{\textbf{Mean (95\% CI)}\\\textbf{SD}}} & \multicolumn{1}{c}{\makecell[c]{\textbf{Mean (95\% CI)}\\\textbf{SD}}} \\
\midrule
\Name{vit} & \makecell[c]{0.748\\{\fontsize{5.7}{6.2}\selectfont (0.731--0.765)}\\{\fontsize{5.4}{5.9}\selectfont SD 0.064}} & \makecell[c]{0.496\\{\fontsize{5.7}{6.2}\selectfont (0.479--0.526)}\\{\fontsize{5.4}{5.9}\selectfont SD 0.115}} \\
CTFM~\protect\cite{pai2025vision} & \makecell[c]{0.749\\{\fontsize{5.7}{6.2}\selectfont (0.732--0.765)}\\{\fontsize{5.4}{5.9}\selectfont SD 0.088}} & \makecell[c]{0.499\\{\fontsize{5.7}{6.2}\selectfont (0.482--0.531)}\\{\fontsize{5.4}{5.9}\selectfont SD 0.124}} \\
CURIA~\protect\cite{dancette2025curia} & \makecell[c]{0.739\\{\fontsize{5.7}{6.2}\selectfont (0.723--0.756)}\\{\fontsize{5.4}{5.9}\selectfont SD 0.075}} & \makecell[c]{0.478\\{\fontsize{5.7}{6.2}\selectfont (0.462--0.511)}\\{\fontsize{5.4}{5.9}\selectfont SD 0.118}} \\
VISTA3D~\protect\cite{he2025vista3d} & \makecell[c]{0.744\\{\fontsize{5.7}{6.2}\selectfont (0.727--0.761)}\\{\fontsize{5.4}{5.9}\selectfont SD 0.110}} & \makecell[c]{0.498\\{\fontsize{5.7}{6.2}\selectfont (0.484--0.524)}\\{\fontsize{5.4}{5.9}\selectfont SD 0.148}} \\
\Name{cnn} (scratch) & \makecell[c]{0.729\\{\fontsize{5.7}{6.2}\selectfont (0.712--0.746)}\\{\fontsize{5.4}{5.9}\selectfont SD 0.061}} & \makecell[c]{0.469\\{\fontsize{5.7}{6.2}\selectfont (0.452--0.499)}\\{\fontsize{5.4}{5.9}\selectfont SD 0.121}} \\
MERLIN~\protect\cite{Blankemeier2026} & \makecell[c]{0.696\\{\fontsize{5.7}{6.2}\selectfont (0.680--0.713)}\\{\fontsize{5.4}{5.9}\selectfont SD 0.092}} & \makecell[c]{0.429\\{\fontsize{5.7}{6.2}\selectfont (0.416--0.456)}\\{\fontsize{5.4}{5.9}\selectfont SD 0.103}} \\
VOCO~\protect\cite{wu2024voco} & \makecell[c]{0.678\\{\fontsize{5.7}{6.2}\selectfont (0.662--0.695)}\\{\fontsize{5.4}{5.9}\selectfont SD 0.088}} & \makecell[c]{0.407\\{\fontsize{5.7}{6.2}\selectfont (0.391--0.438)}\\{\fontsize{5.4}{5.9}\selectfont SD 0.127}} \\
\Name{cnn} & \makecell[c]{0.681\\{\fontsize{5.7}{6.2}\selectfont (0.665--0.699)}\\{\fontsize{5.4}{5.9}\selectfont SD 0.081}} & \makecell[c]{0.409\\{\fontsize{5.7}{6.2}\selectfont (0.392--0.438)}\\{\fontsize{5.4}{5.9}\selectfont SD 0.118}} \\
SwinUNETR~\protect\cite{tang2022self} & \makecell[c]{0.686\\{\fontsize{5.7}{6.2}\selectfont (0.667--0.704)}\\{\fontsize{5.4}{5.9}\selectfont SD 0.089}} & \makecell[c]{0.426\\{\fontsize{5.7}{6.2}\selectfont (0.410--0.457)}\\{\fontsize{5.4}{5.9}\selectfont SD 0.122}} \\
\Name{vit} (scratch) & \makecell[c]{0.665\\{\fontsize{5.7}{6.2}\selectfont (0.648--0.683)}\\{\fontsize{5.4}{5.9}\selectfont SD 0.073}} & \makecell[c]{0.394\\{\fontsize{5.7}{6.2}\selectfont (0.379--0.422)}\\{\fontsize{5.4}{5.9}\selectfont SD 0.117}} \\
MISFM~\protect\cite{wang2023mis} & \makecell[c]{0.500\\{\fontsize{5.7}{6.2}\selectfont (0.500--0.500)}\\{\fontsize{5.4}{5.9}\selectfont SD 0.000}} & \makecell[c]{0.258\\{\fontsize{5.7}{6.2}\selectfont (0.249--0.268)}\\{\fontsize{5.4}{5.9}\selectfont SD 0.108}} \\
\bottomrule
\end{tabular}
\end{table}
\normalsize

\begin{table}[!htbp]
\caption{Pairwise metric-difference comparisons corresponding to \cref{tab:fig2-summary-classification}. Cells report nnFoundationViT-minus-comparator dataset-equal mean differences with paired 95\% CIs in parentheses; positive values favor nnFoundationViT because higher AUROC and AUPRC are better. A dagger marks comparisons for which the paired 95\% CI does not lie entirely above zero.}
\label{tab:fig2-summary-classification-pairwise-support}
\centering
\tablebodyfont
\setlength{\tabcolsep}{2pt}
\renewcommand{\arraystretch}{0.90}
\begin{tabular}{@{}l@{\hspace{1.20em}}c@{\hspace{1.20em}}c@{}}
\toprule
\textbf{Comparator} & \multicolumn{1}{c}{\textbf{AUROC}} & \multicolumn{1}{c}{\textbf{AUPRC}} \\
\cmidrule(lr){2-2}\cmidrule(lr){3-3}
 & \multicolumn{1}{c}{\textbf{\(\Delta\) (95\% CI)}} & \multicolumn{1}{c}{\textbf{\(\Delta\) (95\% CI)}} \\
\midrule
\Name{vit} & Reference & Reference \\
CTFM~\protect\cite{pai2025vision} & -0.001 (-0.017--0.016)$^\dagger$ & -0.003 (-0.027--0.020)$^\dagger$ \\
CURIA~\protect\cite{dancette2025curia} & 0.009 (-0.009--0.028)$^\dagger$ & 0.019 (-0.010--0.043)$^\dagger$ \\
VISTA3D~\protect\cite{he2025vista3d} & 0.004 (-0.016--0.023)$^\dagger$ & -0.002 (-0.026--0.022)$^\dagger$ \\
\Name{cnn} (scratch) & 0.019 (0.001--0.036) & 0.027 (0.003--0.049) \\
MERLIN~\protect\cite{Blankemeier2026} & 0.051 (0.032--0.072) & 0.068 (0.041--0.093) \\
VOCO~\protect\cite{wu2024voco} & 0.070 (0.051--0.089) & 0.089 (0.068--0.109) \\
\Name{cnn} & 0.066 (0.049--0.081) & 0.087 (0.062--0.110) \\
SwinUNETR~\protect\cite{tang2022self} & 0.062 (0.041--0.083) & 0.070 (0.043--0.095) \\
\Name{vit} (scratch) & 0.082 (0.064--0.102) & 0.102 (0.079--0.125) \\
MISFM~\protect\cite{wang2023mis} & 0.248 (0.229--0.265) & 0.238 (0.222--0.267) \\
\bottomrule
\end{tabular}
\end{table}
\normalsize

\begin{table}[!htbp]
\caption{Figure 2 summary metric scores for report generation across 2 datasets. Values are dataset-equal means with fixed-dataset within-dataset bootstrap 95\% CIs; SD reports variability across dataset-level metric means where at least two datasets contribute.}
\label{tab:fig2-summary-report-generation}
\centering
\tablebodyfont
\setlength{\tabcolsep}{2pt}
\renewcommand{\arraystretch}{0.90}
\begin{tabular}{@{}l@{\hspace{0.90em}}c@{\hspace{0.90em}}c@{\hspace{0.90em}}c@{}}
\toprule
\textbf{Model} & \multicolumn{1}{c}{\textbf{BLEU}} & \multicolumn{1}{c}{\textbf{F1}} & \multicolumn{1}{c}{\textbf{CRG}} \\
\cmidrule(lr){2-2}\cmidrule(lr){3-3}\cmidrule(lr){4-4}
 & \multicolumn{1}{c}{\makecell[c]{\textbf{Mean (95\% CI)}\\\textbf{SD}}} & \multicolumn{1}{c}{\makecell[c]{\textbf{Mean (95\% CI)}\\\textbf{SD}}} & \multicolumn{1}{c}{\makecell[c]{\textbf{Mean (95\% CI)}\\\textbf{SD}}} \\
\midrule
\Name{vit} & \makecell[c]{0.244\\{\fontsize{5.7}{6.2}\selectfont (0.241--0.248)}\\{\fontsize{5.4}{5.9}\selectfont SD 0.103}} & \makecell[c]{0.437\\{\fontsize{5.7}{6.2}\selectfont (0.427--0.448)}\\{\fontsize{5.4}{5.9}\selectfont SD --}} & \makecell[c]{0.444\\{\fontsize{5.7}{6.2}\selectfont (0.440--0.448)}\\{\fontsize{5.4}{5.9}\selectfont SD --}} \\
CURIA~\protect\cite{dancette2025curia} & \makecell[c]{0.243\\{\fontsize{5.7}{6.2}\selectfont (0.240--0.247)}\\{\fontsize{5.4}{5.9}\selectfont SD 0.103}} & \makecell[c]{0.428\\{\fontsize{5.7}{6.2}\selectfont (0.415--0.439)}\\{\fontsize{5.4}{5.9}\selectfont SD --}} & \makecell[c]{0.426\\{\fontsize{5.7}{6.2}\selectfont (0.422--0.429)}\\{\fontsize{5.4}{5.9}\selectfont SD --}} \\
CTFM~\protect\cite{pai2025vision} & \makecell[c]{0.228\\{\fontsize{5.7}{6.2}\selectfont (0.224--0.232)}\\{\fontsize{5.4}{5.9}\selectfont SD 0.081}} & \makecell[c]{0.276\\{\fontsize{5.7}{6.2}\selectfont (0.265--0.287)}\\{\fontsize{5.4}{5.9}\selectfont SD --}} & \makecell[c]{0.384\\{\fontsize{5.7}{6.2}\selectfont (0.381--0.387)}\\{\fontsize{5.4}{5.9}\selectfont SD --}} \\
MERLIN~\protect\cite{Blankemeier2026} & \makecell[c]{0.223\\{\fontsize{5.7}{6.2}\selectfont (0.220--0.227)}\\{\fontsize{5.4}{5.9}\selectfont SD 0.075}} & \makecell[c]{0.227\\{\fontsize{5.7}{6.2}\selectfont (0.216--0.237)}\\{\fontsize{5.4}{5.9}\selectfont SD --}} & \makecell[c]{0.372\\{\fontsize{5.7}{6.2}\selectfont (0.369--0.374)}\\{\fontsize{5.4}{5.9}\selectfont SD --}} \\
\Name{cnn} & \makecell[c]{0.180\\{\fontsize{5.7}{6.2}\selectfont (0.176--0.184)}\\{\fontsize{5.4}{5.9}\selectfont SD 0.123}} & \makecell[c]{0.146\\{\fontsize{5.7}{6.2}\selectfont (0.138--0.154)}\\{\fontsize{5.4}{5.9}\selectfont SD --}} & \makecell[c]{0.354\\{\fontsize{5.7}{6.2}\selectfont (0.352--0.356)}\\{\fontsize{5.4}{5.9}\selectfont SD --}} \\
SwinUNETR~\protect\cite{tang2022self} & \makecell[c]{0.173\\{\fontsize{5.7}{6.2}\selectfont (0.169--0.177)}\\{\fontsize{5.4}{5.9}\selectfont SD 0.104}} & \makecell[c]{0.022\\{\fontsize{5.7}{6.2}\selectfont (0.018--0.026)}\\{\fontsize{5.4}{5.9}\selectfont SD --}} & \makecell[c]{0.336\\{\fontsize{5.7}{6.2}\selectfont (0.335--0.336)}\\{\fontsize{5.4}{5.9}\selectfont SD --}} \\
VISTA3D~\protect\cite{he2025vista3d} & \makecell[c]{0.175\\{\fontsize{5.7}{6.2}\selectfont (0.171--0.179)}\\{\fontsize{5.4}{5.9}\selectfont SD 0.136}} & \makecell[c]{0.135\\{\fontsize{5.7}{6.2}\selectfont (0.126--0.144)}\\{\fontsize{5.4}{5.9}\selectfont SD --}} & \makecell[c]{0.355\\{\fontsize{5.7}{6.2}\selectfont (0.353--0.357)}\\{\fontsize{5.4}{5.9}\selectfont SD --}} \\
VOCO~\protect\cite{wu2024voco} & \makecell[c]{0.172\\{\fontsize{5.7}{6.2}\selectfont (0.168--0.176)}\\{\fontsize{5.4}{5.9}\selectfont SD 0.129}} & \makecell[c]{0.132\\{\fontsize{5.7}{6.2}\selectfont (0.122--0.141)}\\{\fontsize{5.4}{5.9}\selectfont SD --}} & \makecell[c]{0.354\\{\fontsize{5.7}{6.2}\selectfont (0.352--0.355)}\\{\fontsize{5.4}{5.9}\selectfont SD --}} \\
MISFM~\protect\cite{wang2023mis} & \makecell[c]{0.130\\{\fontsize{5.7}{6.2}\selectfont (0.126--0.134)}\\{\fontsize{5.4}{5.9}\selectfont SD 0.157}} & \makecell[c]{0.000\\{\fontsize{5.7}{6.2}\selectfont (0.000--0.000)}\\{\fontsize{5.4}{5.9}\selectfont SD --}} & \makecell[c]{0.333\\{\fontsize{5.7}{6.2}\selectfont (0.333--0.333)}\\{\fontsize{5.4}{5.9}\selectfont SD --}} \\
\bottomrule
\end{tabular}
\end{table}
\normalsize

\begin{table}[!htbp]
\caption{Pairwise metric-difference comparisons corresponding to \cref{tab:fig2-summary-report-generation}. Cells report nnFoundationViT-minus-comparator dataset-equal mean differences with paired 95\% CIs in parentheses; positive values favor nnFoundationViT because higher BLEU, F1, and CRG are better. A dagger marks comparisons for which the paired 95\% CI does not lie entirely above zero.}
\label{tab:fig2-summary-report-generation-pairwise-support}
\centering
\tablebodyfont
\setlength{\tabcolsep}{2pt}
\renewcommand{\arraystretch}{0.90}
\begin{tabular}{@{}l@{\hspace{0.90em}}c@{\hspace{0.90em}}c@{\hspace{0.90em}}c@{}}
\toprule
\textbf{Comparator} & \multicolumn{1}{c}{\textbf{BLEU}} & \multicolumn{1}{c}{\textbf{F1}} & \multicolumn{1}{c}{\textbf{CRG}} \\
\cmidrule(lr){2-2}\cmidrule(lr){3-3}\cmidrule(lr){4-4}
 & \multicolumn{1}{c}{\textbf{\(\Delta\) (95\% CI)}} & \multicolumn{1}{c}{\textbf{\(\Delta\) (95\% CI)}} & \multicolumn{1}{c}{\textbf{\(\Delta\) (95\% CI)}} \\
\midrule
\Name{vit} & Reference & Reference & Reference \\
CURIA~\protect\cite{dancette2025curia} & 0.001 (-0.001--0.004)$^\dagger$ & 0.009 (-0.005--0.022)$^\dagger$ & 0.018 (0.014--0.023) \\
CTFM~\protect\cite{pai2025vision} & 0.017 (0.014--0.019) & 0.161 (0.147--0.173) & 0.060 (0.056--0.064) \\
MERLIN~\protect\cite{Blankemeier2026} & 0.021 (0.018--0.024) & 0.210 (0.197--0.223) & 0.072 (0.068--0.076) \\
\Name{cnn} & 0.064 (0.062--0.067) & 0.291 (0.279--0.303) & 0.090 (0.086--0.094) \\
SwinUNETR~\protect\cite{tang2022self} & 0.071 (0.069--0.074) & 0.415 (0.404--0.427) & 0.108 (0.104--0.112) \\
VISTA3D~\protect\cite{he2025vista3d} & 0.070 (0.067--0.072) & 0.302 (0.289--0.315) & 0.089 (0.085--0.093) \\
VOCO~\protect\cite{wu2024voco} & 0.072 (0.069--0.075) & 0.306 (0.293--0.317) & 0.090 (0.086--0.094) \\
MISFM~\protect\cite{wang2023mis} & 0.114 (0.112--0.117) & 0.437 (0.427--0.448) & 0.111 (0.107--0.115) \\
\bottomrule
\end{tabular}
\end{table}
\normalsize

\begin{table}[!htbp]
\caption{Figure 2 summary metric scores for image retrieval across 5 datasets. Values are dataset-equal means with fixed-dataset within-query bootstrap 95\% CIs; SD reports variability across dataset-level metric means.}
\label{tab:fig2-summary-image-retrieval}
\centering
\tablebodyfont
\setlength{\tabcolsep}{2pt}
\renewcommand{\arraystretch}{0.90}
\begin{tabular}{@{}l@{\hspace{1.20em}}c@{\hspace{1.20em}}c@{}}
\toprule
\textbf{Model} & \multicolumn{1}{c}{\textbf{AP (tumor flag)}} & \multicolumn{1}{c}{\textbf{AP (tumor stage)}} \\
\cmidrule(lr){2-2}\cmidrule(lr){3-3}
 & \multicolumn{1}{c}{\makecell[c]{\textbf{Mean (95\% CI)}\\\textbf{SD}}} & \multicolumn{1}{c}{\makecell[c]{\textbf{Mean (95\% CI)}\\\textbf{SD}}} \\
\midrule
\Name{vit} & \makecell[c]{0.921\\{\fontsize{5.7}{6.2}\selectfont (0.902--0.942)}\\{\fontsize{5.4}{5.9}\selectfont SD 0.071}} & \makecell[c]{0.651\\{\fontsize{5.7}{6.2}\selectfont (0.608--0.694)}\\{\fontsize{5.4}{5.9}\selectfont SD 0.062}} \\
MERLIN~\protect\cite{Blankemeier2026} & \makecell[c]{0.934\\{\fontsize{5.7}{6.2}\selectfont (0.905--0.958)}\\{\fontsize{5.4}{5.9}\selectfont SD 0.092}} & \makecell[c]{0.657\\{\fontsize{5.7}{6.2}\selectfont (0.611--0.702)}\\{\fontsize{5.4}{5.9}\selectfont SD 0.084}} \\
SwinUNETR~\protect\cite{tang2022self} & \makecell[c]{0.917\\{\fontsize{5.7}{6.2}\selectfont (0.887--0.942)}\\{\fontsize{5.4}{5.9}\selectfont SD 0.090}} & \makecell[c]{0.639\\{\fontsize{5.7}{6.2}\selectfont (0.592--0.684)}\\{\fontsize{5.4}{5.9}\selectfont SD 0.069}} \\
CTFM~\protect\cite{pai2025vision} & \makecell[c]{0.876\\{\fontsize{5.7}{6.2}\selectfont (0.838--0.911)}\\{\fontsize{5.4}{5.9}\selectfont SD 0.075}} & \makecell[c]{0.615\\{\fontsize{5.7}{6.2}\selectfont (0.569--0.660)}\\{\fontsize{5.4}{5.9}\selectfont SD 0.058}} \\
VISTA3D~\protect\cite{he2025vista3d} & \makecell[c]{0.867\\{\fontsize{5.7}{6.2}\selectfont (0.828--0.903)}\\{\fontsize{5.4}{5.9}\selectfont SD 0.100}} & \makecell[c]{0.605\\{\fontsize{5.7}{6.2}\selectfont (0.557--0.653)}\\{\fontsize{5.4}{5.9}\selectfont SD 0.084}} \\
VOCO~\protect\cite{wu2024voco} & \makecell[c]{0.841\\{\fontsize{5.7}{6.2}\selectfont (0.809--0.874)}\\{\fontsize{5.4}{5.9}\selectfont SD 0.075}} & \makecell[c]{0.586\\{\fontsize{5.7}{6.2}\selectfont (0.543--0.626)}\\{\fontsize{5.4}{5.9}\selectfont SD 0.056}} \\
CURIA~\protect\cite{dancette2025curia} & \makecell[c]{0.811\\{\fontsize{5.7}{6.2}\selectfont (0.771--0.852)}\\{\fontsize{5.4}{5.9}\selectfont SD 0.097}} & \makecell[c]{0.558\\{\fontsize{5.7}{6.2}\selectfont (0.511--0.601)}\\{\fontsize{5.4}{5.9}\selectfont SD 0.083}} \\
\Name{cnn} & \makecell[c]{0.802\\{\fontsize{5.7}{6.2}\selectfont (0.759--0.843)}\\{\fontsize{5.4}{5.9}\selectfont SD 0.094}} & \makecell[c]{0.563\\{\fontsize{5.7}{6.2}\selectfont (0.520--0.608)}\\{\fontsize{5.4}{5.9}\selectfont SD 0.076}} \\
MISFM~\protect\cite{wang2023mis} & \makecell[c]{0.678\\{\fontsize{5.7}{6.2}\selectfont (0.642--0.710)}\\{\fontsize{5.4}{5.9}\selectfont SD 0.120}} & \makecell[c]{0.481\\{\fontsize{5.7}{6.2}\selectfont (0.441--0.527)}\\{\fontsize{5.4}{5.9}\selectfont SD 0.072}} \\
\bottomrule
\end{tabular}
\end{table}
\normalsize

\begin{table}[!htbp]
\caption{Pairwise metric-difference comparisons corresponding to \cref{tab:fig2-summary-image-retrieval}. Cells report nnFoundationViT-minus-comparator dataset-equal mean AP differences with paired 95\% CIs in parentheses; positive values favor nnFoundationViT because higher AP is better. A dagger marks comparisons for which the paired 95\% CI does not lie entirely above zero.}
\label{tab:fig2-summary-image-retrieval-pairwise-support}
\centering
\tablebodyfont
\setlength{\tabcolsep}{2pt}
\renewcommand{\arraystretch}{0.90}
\begin{tabular}{@{}l@{\hspace{1.20em}}c@{\hspace{1.20em}}c@{}}
\toprule
\textbf{Comparator} & \multicolumn{1}{c}{\textbf{AP (tumor flag)}} & \multicolumn{1}{c}{\textbf{AP (tumor stage)}} \\
\cmidrule(lr){2-2}\cmidrule(lr){3-3}
 & \multicolumn{1}{c}{\textbf{\(\Delta\) (95\% CI)}} & \multicolumn{1}{c}{\textbf{\(\Delta\) (95\% CI)}} \\
\midrule
\Name{vit} & Reference & Reference \\
MERLIN~\protect\cite{Blankemeier2026} & -0.013 (-0.038--0.013)$^\dagger$ & -0.006 (-0.048--0.034)$^\dagger$ \\
SwinUNETR~\protect\cite{tang2022self} & 0.005 (-0.020--0.033)$^\dagger$ & 0.012 (-0.028--0.051)$^\dagger$ \\
CTFM~\protect\cite{pai2025vision} & 0.045 (0.013--0.079) & 0.036 (-0.008--0.082)$^\dagger$ \\
VISTA3D~\protect\cite{he2025vista3d} & 0.054 (0.027--0.088) & 0.046 (0.002--0.092) \\
VOCO~\protect\cite{wu2024voco} & 0.080 (0.050--0.113) & 0.065 (0.025--0.106) \\
CURIA~\protect\cite{dancette2025curia} & 0.110 (0.076--0.146) & 0.093 (0.055--0.138) \\
\Name{cnn} & 0.120 (0.083--0.159) & 0.088 (0.045--0.135) \\
MISFM~\protect\cite{wang2023mis} & 0.243 (0.206--0.281) & 0.170 (0.108--0.224) \\
\bottomrule
\end{tabular}
\end{table}
\normalsize

\clearpage

\subsubsection{Dataset-Level Results}
\label{app:figure2-dataset-level-results}

\begingroup
\singlespacing
\tablebodyfont
\setlength{\tabcolsep}{2pt}
\renewcommand{\arraystretch}{0.90}
\setlength{\LTcapwidth}{\linewidth}
\begin{longtable}{@{}l@{\hspace{1.20em}}c@{\hspace{1.20em}}c@{}}
\caption{Dataset-level Figure 2 metric scores for semantic segmentation. Values are shown as mean and 95\% CI.}\label{tab:fig2-dataset-segmentation}\\
\toprule
\textbf{Model} & \multicolumn{1}{c}{\textbf{DSC}} & \multicolumn{1}{c}{\textbf{NSD}} \\
\cmidrule(lr){2-2}\cmidrule(lr){3-3}
 & \multicolumn{1}{c}{\textbf{Mean (95\% CI)}} & \multicolumn{1}{c}{\textbf{Mean (95\% CI)}} \\
\midrule
\endfirsthead
\caption[]{Dataset-level Figure 2 metric scores for semantic segmentation. Values are shown as mean and 95\% CI. (continued)}\\
\toprule
\textbf{Model} & \multicolumn{1}{c}{\textbf{DSC}} & \multicolumn{1}{c}{\textbf{NSD}} \\
\cmidrule(lr){2-2}\cmidrule(lr){3-3}
 & \multicolumn{1}{c}{\textbf{Mean (95\% CI)}} & \multicolumn{1}{c}{\textbf{Mean (95\% CI)}} \\
\midrule
\endhead
\midrule
\multicolumn{3}{r}{\footnotesize Continued on next page} \\
\endfoot
\bottomrule
\endlastfoot
\addlinespace[0.18em]
\multicolumn{3}{@{}l}{\textbf{MSD Brain}~\protect\cite{antonelli2022medical}}\\
\addlinespace[0.02em]
\Name{cnn} & 0.735 (0.717--0.754) & 0.781 (0.761--0.798) \\
\Name{cnn} (scratch) & 0.733 (0.714--0.751) & 0.777 (0.758--0.796) \\
nnU-Net Default~\protect\cite{isensee2021nnu} & 0.739 (0.721--0.755) & 0.782 (0.762--0.801) \\
CTFM~\protect\cite{pai2025vision} & 0.731 (0.712--0.750) & 0.774 (0.753--0.794) \\
VISTA3D~\protect\cite{he2025vista3d} & 0.736 (0.715--0.753) & 0.778 (0.757--0.797) \\
MISFM~\protect\cite{wang2023mis} & 0.737 (0.718--0.753) & 0.781 (0.763--0.799) \\
MERLIN~\protect\cite{Blankemeier2026} & 0.727 (0.707--0.744) & 0.769 (0.748--0.788) \\
\end{longtable}
\endgroup

\begingroup
\singlespacing
\tablebodyfont
\setlength{\tabcolsep}{2pt}
\renewcommand{\arraystretch}{0.90}
\setlength{\LTcapwidth}{\linewidth}
\begin{longtable}{@{}l@{\hspace{1.20em}}c@{\hspace{1.20em}}c@{}}
\caption{Dataset-level Figure 2 metric scores for classification. Values are shown as mean and 95\% CI.}\label{tab:fig2-dataset-classification}\\
\toprule
\textbf{Model} & \multicolumn{1}{c}{\textbf{AUROC}} & \multicolumn{1}{c}{\textbf{AUPRC}} \\
\cmidrule(lr){2-2}\cmidrule(lr){3-3}
 & \multicolumn{1}{c}{\textbf{Mean (95\% CI)}} & \multicolumn{1}{c}{\textbf{Mean (95\% CI)}} \\
\midrule
\endfirsthead
\caption[]{Dataset-level Figure 2 metric scores for classification. Values are shown as mean and 95\% CI. (continued)}\\
\toprule
\textbf{Model} & \multicolumn{1}{c}{\textbf{AUROC}} & \multicolumn{1}{c}{\textbf{AUPRC}} \\
\cmidrule(lr){2-2}\cmidrule(lr){3-3}
 & \multicolumn{1}{c}{\textbf{Mean (95\% CI)}} & \multicolumn{1}{c}{\textbf{Mean (95\% CI)}} \\
\midrule
\endhead
\midrule
\multicolumn{3}{r}{\footnotesize Continued on next page} \\
\endfoot
\bottomrule
\endlastfoot
\addlinespace[0.18em]
\multicolumn{3}{@{}l}{\textbf{CT-RATE}~\protect\cite{hamamci2026generalist}}\\
\addlinespace[0.02em]
\Name{vit} & 0.826 (0.817--0.836) & 0.537 (0.519--0.561) \\
CTFM~\protect\cite{pai2025vision} & 0.826 (0.816--0.834) & 0.529 (0.513--0.552) \\
CURIA~\protect\cite{dancette2025curia} & 0.832 (0.823--0.841) & 0.553 (0.534--0.582) \\
VISTA3D~\protect\cite{he2025vista3d} & 0.851 (0.842--0.859) & 0.589 (0.572--0.612) \\
\Name{cnn} (scratch) & 0.815 (0.804--0.825) & 0.505 (0.486--0.531) \\
MERLIN~\protect\cite{Blankemeier2026} & 0.766 (0.754--0.776) & 0.442 (0.427--0.464) \\
VOCO~\protect\cite{wu2024voco} & 0.794 (0.783--0.804) & 0.483 (0.465--0.508) \\
\end{longtable}
\endgroup

\begingroup
\singlespacing
\tablebodyfont
\setlength{\tabcolsep}{2pt}
\renewcommand{\arraystretch}{0.90}
\setlength{\LTcapwidth}{\linewidth}
\begin{longtable}{@{}l@{\hspace{0.90em}}c@{\hspace{0.90em}}c@{\hspace{0.90em}}c@{}}
\caption{Dataset-level Figure 2 metric scores for report generation. Values are shown as mean and 95\% CI.}\label{tab:fig2-dataset-report-generation}\\
\toprule
\textbf{Model} & \multicolumn{1}{c}{\textbf{BLEU}} & \multicolumn{1}{c}{\textbf{F1}} & \multicolumn{1}{c}{\textbf{CRG}} \\
\cmidrule(lr){2-2}\cmidrule(lr){3-3}\cmidrule(lr){4-4}
 & \multicolumn{1}{c}{\textbf{Mean (95\% CI)}} & \multicolumn{1}{c}{\textbf{Mean (95\% CI)}} & \multicolumn{1}{c}{\textbf{Mean (95\% CI)}} \\
\midrule
\endfirsthead
\caption[]{Dataset-level Figure 2 metric scores for report generation. Values are shown as mean and 95\% CI. (continued)}\\
\toprule
\textbf{Model} & \multicolumn{1}{c}{\textbf{BLEU}} & \multicolumn{1}{c}{\textbf{F1}} & \multicolumn{1}{c}{\textbf{CRG}} \\
\cmidrule(lr){2-2}\cmidrule(lr){3-3}\cmidrule(lr){4-4}
 & \multicolumn{1}{c}{\textbf{Mean (95\% CI)}} & \multicolumn{1}{c}{\textbf{Mean (95\% CI)}} & \multicolumn{1}{c}{\textbf{Mean (95\% CI)}} \\
\midrule
\endhead
\midrule
\multicolumn{4}{r}{\footnotesize Continued on next page} \\
\endfoot
\bottomrule
\endlastfoot
\addlinespace[0.18em]
\multicolumn{4}{@{}l}{\textbf{CT-Rate}~\protect\cite{hamamci2026generalist}}\\
\addlinespace[0.02em]
\Name{vit} & 0.317 (0.311--0.323) & 0.437 (0.427--0.448) & 0.444 (0.440--0.448) \\
CURIA~\protect\cite{dancette2025curia} & 0.316 (0.310--0.323) & 0.428 (0.415--0.439) & 0.426 (0.422--0.429) \\
CTFM~\protect\cite{pai2025vision} & 0.285 (0.279--0.293) & 0.276 (0.265--0.287) & 0.384 (0.381--0.387) \\
MERLIN~\protect\cite{Blankemeier2026} & 0.276 (0.269--0.283) & 0.227 (0.216--0.237) & 0.372 (0.369--0.374) \\
\Name{cnn} & 0.267 (0.260--0.275) & 0.146 (0.138--0.154) & 0.354 (0.352--0.356) \\
SwinUNETR~\protect\cite{tang2022self} & 0.247 (0.239--0.255) & 0.022 (0.018--0.026) & 0.336 (0.335--0.336) \\
VISTA3D~\protect\cite{he2025vista3d} & 0.271 (0.263--0.278) & 0.135 (0.126--0.144) & 0.355 (0.353--0.357) \\
\end{longtable}
\endgroup

\begingroup
\singlespacing
\tablebodyfont
\setlength{\tabcolsep}{2pt}
\renewcommand{\arraystretch}{0.90}
\setlength{\LTcapwidth}{\linewidth}
\begin{longtable}{@{}l@{\hspace{1.20em}}c@{\hspace{1.20em}}c@{}}
\caption{Dataset-level Figure 2 metric scores for image retrieval. Values are shown as mean and 95\% CI.}\label{tab:fig2-dataset-image-retrieval}\\
\toprule
\textbf{Model} & \multicolumn{1}{c}{\textbf{AP (tumor flag)}} & \multicolumn{1}{c}{\textbf{AP (tumor stage)}} \\
\cmidrule(lr){2-2}\cmidrule(lr){3-3}
 & \multicolumn{1}{c}{\textbf{Mean (95\% CI)}} & \multicolumn{1}{c}{\textbf{Mean (95\% CI)}} \\
\midrule
\endfirsthead
\caption[]{Dataset-level Figure 2 metric scores for image retrieval. Values are shown as mean and 95\% CI. (continued)}\\
\toprule
\textbf{Model} & \multicolumn{1}{c}{\textbf{AP (tumor flag)}} & \multicolumn{1}{c}{\textbf{AP (tumor stage)}} \\
\cmidrule(lr){2-2}\cmidrule(lr){3-3}
 & \multicolumn{1}{c}{\textbf{Mean (95\% CI)}} & \multicolumn{1}{c}{\textbf{Mean (95\% CI)}} \\
\midrule
\endhead
\midrule
\multicolumn{3}{r}{\footnotesize Continued on next page} \\
\endfoot
\bottomrule
\endlastfoot
\addlinespace[0.18em]
\multicolumn{3}{@{}l}{\textbf{MSD Colon}~\protect\cite{antonelli2022medical}}\\
\addlinespace[0.02em]
\Name{vit} & 0.925 (0.859--0.983) & 0.669 (0.562--0.782) \\
MERLIN~\protect\cite{Blankemeier2026} & 1.000 (1.000--1.000) & 0.678 (0.552--0.801) \\
SwinUNETR~\protect\cite{tang2022self} & 0.972 (0.939--0.999) & 0.665 (0.546--0.774) \\
CTFM~\protect\cite{pai2025vision} & 0.794 (0.649--0.918) & 0.535 (0.421--0.652) \\
VISTA3D~\protect\cite{he2025vista3d} & 0.791 (0.657--0.909) & 0.528 (0.404--0.640) \\
VOCO~\protect\cite{wu2024voco} & 0.830 (0.715--0.939) & 0.536 (0.434--0.640) \\
CURIA~\protect\cite{dancette2025curia} & 0.714 (0.592--0.839) & 0.468 (0.338--0.585) \\
\end{longtable}
\endgroup

\subsection{Figure 3 Extended Results}
\label{app:figure3-metric-tables}
\noindent These tables report the dynamic-adaptation results underlying Figure 3.

\subsubsection{Summary Tables}
\label{app:figure3-summary-tables}

\begin{table}[!htbp]
\caption{Figure 3 summary metric scores for semantic segmentation across 34 datasets. Values are dataset-equal means with dataset-level bootstrap 95\% CIs; SD reports variability across dataset-level metric means.}
\label{tab:fig3-summary-segmentation}
\centering
\tablebodyfont
\setlength{\tabcolsep}{2pt}
\renewcommand{\arraystretch}{0.90}
\begin{tabular}{@{}l@{\hspace{1.20em}}c@{\hspace{1.20em}}c@{}}
\toprule
\textbf{Model variant} & \multicolumn{1}{c}{\textbf{DSC}} & \multicolumn{1}{c}{\textbf{NSD}} \\
\cmidrule(lr){2-2}\cmidrule(lr){3-3}
 & \multicolumn{1}{c}{\makecell[c]{\textbf{Mean (95\% CI)}\\\textbf{SD}}} & \multicolumn{1}{c}{\makecell[c]{\textbf{Mean (95\% CI)}\\\textbf{SD}}} \\
\midrule
\Name{cnn} dynamic & \makecell[c]{0.715\\{\fontsize{5.7}{6.2}\selectfont (0.661--0.770)}\\{\fontsize{5.4}{5.9}\selectfont SD 0.165}} & \makecell[c]{0.640\\{\fontsize{5.7}{6.2}\selectfont (0.575--0.702)}\\{\fontsize{5.4}{5.9}\selectfont SD 0.194}} \\
\Name{cnn} non-dynamic & \makecell[c]{0.703\\{\fontsize{5.7}{6.2}\selectfont (0.645--0.759)}\\{\fontsize{5.4}{5.9}\selectfont SD 0.174}} & \makecell[c]{0.625\\{\fontsize{5.7}{6.2}\selectfont (0.559--0.695)}\\{\fontsize{5.4}{5.9}\selectfont SD 0.205}} \\
\Name{cnn} (scratch) & \makecell[c]{0.704\\{\fontsize{5.7}{6.2}\selectfont (0.646--0.762)}\\{\fontsize{5.4}{5.9}\selectfont SD 0.175}} & \makecell[c]{0.628\\{\fontsize{5.7}{6.2}\selectfont (0.562--0.694)}\\{\fontsize{5.4}{5.9}\selectfont SD 0.203}} \\
\Name{cnn} non-dynamic (scratch) & \makecell[c]{0.683\\{\fontsize{5.7}{6.2}\selectfont (0.621--0.742)}\\{\fontsize{5.4}{5.9}\selectfont SD 0.188}} & \makecell[c]{0.606\\{\fontsize{5.7}{6.2}\selectfont (0.535--0.681)}\\{\fontsize{5.4}{5.9}\selectfont SD 0.220}} \\
\bottomrule
\end{tabular}
\end{table}
\normalsize

\begin{table}[!htbp]
\caption{Pairwise metric-difference comparisons corresponding to \cref{tab:fig3-summary-segmentation}. Cells report \Name{cnn} dynamic-minus-comparator dataset-equal mean differences with paired dataset-level bootstrap 95\% CIs in parentheses; positive values favor \Name{cnn} dynamic because higher DSC and NSD are better. A dagger marks comparisons for which the paired 95\% CI does not lie entirely above zero.}
\label{tab:fig3-summary-segmentation-pairwise-support}
\centering
\tablebodyfont
\setlength{\tabcolsep}{2pt}
\renewcommand{\arraystretch}{0.90}
\begin{tabular}{@{}l@{\hspace{1.20em}}c@{\hspace{1.20em}}c@{}}
\toprule
\textbf{Comparator} & \multicolumn{1}{c}{\textbf{DSC}} & \multicolumn{1}{c}{\textbf{NSD}} \\
\cmidrule(lr){2-2}\cmidrule(lr){3-3}
 & \multicolumn{1}{c}{\textbf{\(\Delta\) (95\% CI)}} & \multicolumn{1}{c}{\textbf{\(\Delta\) (95\% CI)}} \\
\midrule
\Name{cnn} dynamic & Reference & Reference \\
\Name{cnn} non-dynamic & 0.013 (0.004--0.022) & 0.014 (0.005--0.024) \\
\Name{cnn} (scratch) & 0.012 (0.004--0.020) & 0.012 (0.005--0.019) \\
\Name{cnn} non-dynamic (scratch) & 0.033 (0.018--0.050) & 0.033 (0.018--0.051) \\
\bottomrule
\end{tabular}
\end{table}
\normalsize

\begin{table}[!htbp]
\caption{Figure 3 summary metric scores for lesion detection across 9 datasets. Values are dataset-equal means with dataset-level bootstrap 95\% CIs; SD reports variability across dataset-level metric means.}
\label{tab:fig3-summary-detection}
\centering
\tablebodyfont
\setlength{\tabcolsep}{0pt}
\renewcommand{\arraystretch}{0.90}
\begin{tabular}{@{}l@{\hspace{0.12em}}c@{\hspace{0.12em}}c@{\hspace{0.12em}}c@{}}
\toprule
\textbf{Model variant} & \multicolumn{1}{c}{\textbf{mAP}} & \multicolumn{1}{c}{\textbf{FROC}} & \multicolumn{1}{c}{\textbf{AP@IoU 0.10}} \\
\cmidrule(lr){2-2}\cmidrule(lr){3-3}\cmidrule(lr){4-4}
 & \multicolumn{1}{c}{\makecell[c]{\textbf{Mean}\\\textbf{(95\% CI)}\\\textbf{SD}}} & \multicolumn{1}{c}{\makecell[c]{\textbf{Mean}\\\textbf{(95\% CI)}\\\textbf{SD}}} & \multicolumn{1}{c}{\makecell[c]{\textbf{Mean}\\\textbf{(95\% CI)}\\\textbf{SD}}} \\
\midrule
\Name{cnn} dynamic & \makecell[c]{0.726\\{\fontsize{5.7}{6.2}\selectfont (0.628--0.817)}\\{\fontsize{5.4}{5.9}\selectfont SD 0.155}} & \makecell[c]{0.783\\{\fontsize{5.7}{6.2}\selectfont (0.663--0.892)}\\{\fontsize{5.4}{5.9}\selectfont SD 0.186}} & \makecell[c]{0.803\\{\fontsize{5.7}{6.2}\selectfont (0.704--0.887)}\\{\fontsize{5.4}{5.9}\selectfont SD 0.147}} \\
\Name{cnn} non-dynamic & \makecell[c]{0.714\\{\fontsize{5.7}{6.2}\selectfont (0.620--0.810)}\\{\fontsize{5.4}{5.9}\selectfont SD 0.151}} & \makecell[c]{0.776\\{\fontsize{5.7}{6.2}\selectfont (0.658--0.885)}\\{\fontsize{5.4}{5.9}\selectfont SD 0.187}} & \makecell[c]{0.796\\{\fontsize{5.7}{6.2}\selectfont (0.699--0.880)}\\{\fontsize{5.4}{5.9}\selectfont SD 0.147}} \\
\Name{cnn} dynamic (scratch) & \makecell[c]{0.696\\{\fontsize{5.7}{6.2}\selectfont (0.590--0.796)}\\{\fontsize{5.4}{5.9}\selectfont SD 0.169}} & \makecell[c]{0.767\\{\fontsize{5.7}{6.2}\selectfont (0.647--0.875)}\\{\fontsize{5.4}{5.9}\selectfont SD 0.184}} & \makecell[c]{0.778\\{\fontsize{5.7}{6.2}\selectfont (0.669--0.864)}\\{\fontsize{5.4}{5.9}\selectfont SD 0.163}} \\
\Name{cnn} non-dynamic (scratch) & \makecell[c]{0.686\\{\fontsize{5.7}{6.2}\selectfont (0.579--0.784)}\\{\fontsize{5.4}{5.9}\selectfont SD 0.161}} & \makecell[c]{0.754\\{\fontsize{5.7}{6.2}\selectfont (0.643--0.859)}\\{\fontsize{5.4}{5.9}\selectfont SD 0.178}} & \makecell[c]{0.760\\{\fontsize{5.7}{6.2}\selectfont (0.662--0.849)}\\{\fontsize{5.4}{5.9}\selectfont SD 0.154}} \\
\bottomrule
\end{tabular}
\end{table}
\normalsize

\begin{table}[!htbp]
\caption{Pairwise metric-difference comparisons corresponding to \cref{tab:fig3-summary-detection}. Cells report \Name{cnn} dynamic-minus-comparator dataset-equal mean differences with paired dataset-level bootstrap 95\% CIs in parentheses; positive values favor \Name{cnn} dynamic because higher mAP, FROC, and AP@IoU 0.10 are better. A dagger marks comparisons for which the paired 95\% CI does not lie entirely above zero.}
\label{tab:fig3-summary-detection-pairwise-support}
\centering
\tablebodyfont
\fontsize{7.4bp}{8.8bp}\selectfont
\setlength{\tabcolsep}{0pt}
\renewcommand{\arraystretch}{0.90}
\begin{tabular}{@{}l@{\hspace{0.18em}}c@{\hspace{0.18em}}c@{\hspace{0.18em}}c@{}}
\toprule
\textbf{Comparator} & \multicolumn{1}{c}{\textbf{mAP}} & \multicolumn{1}{c}{\textbf{FROC}} & \multicolumn{1}{c}{\textbf{AP@IoU 0.10}} \\
\cmidrule(lr){2-2}\cmidrule(lr){3-3}\cmidrule(lr){4-4}
 & \multicolumn{1}{c}{\textbf{\(\Delta\) (95\% CI)}} & \multicolumn{1}{c}{\textbf{\(\Delta\) (95\% CI)}} & \multicolumn{1}{c}{\textbf{\(\Delta\) (95\% CI)}} \\
\midrule
\Name{cnn} dynamic & Reference & Reference & Reference \\
\Name{cnn} non-dynamic & 0.011 (0.002--0.020) & 0.006 (0.001--0.012) & 0.007 (0.001--0.014) \\
\Name{cnn} dynamic (scratch) & 0.030 (0.012--0.047) & 0.016 (-0.001--0.032)$^\dagger$ & 0.025 (0.004--0.047) \\
\Name{cnn} non-dynamic (scratch) & 0.039 (0.024--0.056) & 0.029 (0.014--0.045) & 0.043 (0.022--0.063) \\
\bottomrule
\end{tabular}
\end{table}
\normalsize

\clearpage
\subsubsection{Dataset-Level Results}
\label{app:figure3-dataset-level-results}

\begingroup
\singlespacing
\tablebodyfont
\setlength{\tabcolsep}{2pt}
\renewcommand{\arraystretch}{0.90}
\setlength{\LTcapwidth}{\linewidth}
\begin{longtable}{@{}l@{\hspace{0.75em}}l@{\hspace{0.90em}}c@{\hspace{0.90em}}c@{}}
\caption{Dataset-level Figure 3 dynamic-adaptation results for semantic segmentation. Values are shown as mean and 95\% CI.}\label{tab:fig3-dataset-segmentation}\\
\toprule
\textbf{Dataset} & \textbf{Variant} & \multicolumn{1}{c}{\textbf{DSC}} & \multicolumn{1}{c}{\textbf{NSD}} \\
\cmidrule(lr){3-3}\cmidrule(lr){4-4}
 & & \multicolumn{1}{c}{\textbf{Mean (95\% CI)}} & \multicolumn{1}{c}{\textbf{Mean (95\% CI)}} \\
\midrule

\endfirsthead
\caption[]{Dataset-level Figure 3 dynamic-adaptation results for semantic segmentation. Values are shown as mean and 95\% CI. (continued)}\\
\toprule
\textbf{Dataset} & \textbf{Variant} & \multicolumn{1}{c}{\textbf{DSC}} & \multicolumn{1}{c}{\textbf{NSD}} \\
\cmidrule(lr){3-3}\cmidrule(lr){4-4}
 & & \multicolumn{1}{c}{\textbf{Mean (95\% CI)}} & \multicolumn{1}{c}{\textbf{Mean (95\% CI)}} \\
\midrule

\endhead
\midrule
\multicolumn{4}{r}{\footnotesize Continued on next page} \\
\endfoot
\bottomrule
\endlastfoot
MSD Colon~\protect\cite{antonelli2022medical} & Non-dynamic & 0.327 (0.238--0.420) & 0.296 (0.219--0.387) \\*
 & Dynamic & 0.440 (0.340--0.539) & 0.411 (0.315--0.503) \\
\addlinespace[0.08em]
HCC-TACE-Seg~\protect\cite{moawad2023hcctace} & Non-dynamic & 0.752 (0.686--0.813) & 0.516 (0.469--0.562) \\*
 & Dynamic & 0.817 (0.764--0.862) & 0.582 (0.532--0.636) \\
\addlinespace[0.08em]
MSD Pancreas~\protect\cite{antonelli2022medical} & Non-dynamic & 0.618 (0.591--0.647) & 0.473 (0.450--0.501) \\*
 & Dynamic & 0.670 (0.641--0.698) & 0.532 (0.506--0.560) \\
\addlinespace[0.08em]
HCC-TACE-MRI~\protect\cite{wawtace2024} & Non-dynamic & 0.669 (0.523--0.835) & 0.317 (0.240--0.390) \\*
 & Dynamic & 0.709 (0.554--0.863) & 0.354 (0.269--0.453) \\
\addlinespace[0.08em]
AutoPETIV Task 2~\protect\cite{kustner2025longitudinalct} & Non-dynamic & 0.373 (0.339--0.407) & 0.326 (0.297--0.356) \\*
 & Dynamic & 0.412 (0.374--0.451) & 0.365 (0.331--0.395) \\
\addlinespace[0.08em]
M\&Ms challenge~\protect\cite{campello2021multi} & Non-dynamic & 0.821 (0.808--0.832) & 0.631 (0.612--0.648) \\*
 & Dynamic & 0.855 (0.846--0.862) & 0.669 (0.655--0.685) \\
\addlinespace[0.08em]
HaNSeg~\protect\cite{podobnik2023hanseg} & Non-dynamic & 0.521 (0.456--0.581) & 0.487 (0.420--0.559) \\*
 & Dynamic & 0.553 (0.501--0.603) & 0.522 (0.463--0.584) \\
\addlinespace[0.08em]
Covid-19-20~\protect\cite{roth2021covid1920} & Non-dynamic & 0.726 (0.690--0.757) & 0.553 (0.518--0.590) \\*
 & Dynamic & 0.749 (0.716--0.781) & 0.591 (0.556--0.624) \\
\addlinespace[0.08em]
Rider Lung~\protect\cite{zhao2008rider} & Non-dynamic & 0.466 (0.302--0.624) & 0.410 (0.273--0.554) \\*
 & Dynamic & 0.483 (0.331--0.625) & 0.418 (0.282--0.556) \\
\addlinespace[0.08em]
MSD Prostate~\protect\cite{antonelli2022medical} & Non-dynamic & 0.721 (0.647--0.781) & 0.580 (0.508--0.635) \\*
 & Dynamic & 0.738 (0.644--0.810) & 0.620 (0.542--0.689) \\
\addlinespace[0.08em]
DeepPSMA FDG~\protect\cite{deeppsma2025} & Non-dynamic & 0.710 (0.629--0.781) & 0.595 (0.521--0.660) \\*
 & Dynamic & 0.727 (0.664--0.785) & 0.572 (0.512--0.632) \\
\addlinespace[0.08em]
ISLES2022~\protect\cite{hernandezpetsche2022isles} & Non-dynamic & 0.770 (0.732--0.805) & 0.748 (0.711--0.782) \\*
 & Dynamic & 0.784 (0.749--0.816) & 0.764 (0.728--0.797) \\
\addlinespace[0.08em]
BraTS2024 Pediatric~\protect\cite{kazerooni2024bratspeds} & Non-dynamic & 0.430 (0.399--0.467) & 0.441 (0.405--0.472) \\*
 & Dynamic & 0.444 (0.413--0.480) & 0.453 (0.417--0.490) \\
\addlinespace[0.08em]
HNTSMRG24~\protect\cite{wahid2024hntsmrg} & Non-dynamic & 0.542 (0.499--0.584) & 0.468 (0.429--0.505) \\*
 & Dynamic & 0.555 (0.513--0.598) & 0.486 (0.447--0.524) \\
\addlinespace[0.08em]
Mama Mia~\protect\cite{garrucho2025large} & Non-dynamic & 0.789 (0.774--0.804) & 0.733 (0.715--0.750) \\*
 & Dynamic & 0.800 (0.784--0.815) & 0.746 (0.730--0.762) \\
\addlinespace[0.08em]
ACDC~\protect\cite{bernard2018deep} & Non-dynamic & 0.908 (0.900--0.916) & 0.810 (0.797--0.823) \\*
 & Dynamic & 0.917 (0.908--0.924) & 0.826 (0.812--0.839) \\
\addlinespace[0.08em]
MSD Hepatic Vessel~\protect\cite{antonelli2022medical} & Non-dynamic & 0.665 (0.642--0.688) & 0.612 (0.591--0.633) \\*
 & Dynamic & 0.670 (0.649--0.694) & 0.615 (0.595--0.637) \\
\addlinespace[0.08em]
MS Ljubljana~\protect\cite{lesjak2018novel} & Non-dynamic & 0.717 (0.698--0.734) & 0.893 (0.880--0.906) \\*
 & Dynamic & 0.722 (0.704--0.738) & 0.898 (0.885--0.908) \\
\addlinespace[0.08em]
MSD Hippocampus~\protect\cite{antonelli2022medical} & Non-dynamic & 0.883 (0.877--0.890) & 0.967 (0.961--0.973) \\*
 & Dynamic & 0.887 (0.881--0.893) & 0.971 (0.966--0.977) \\
\addlinespace[0.08em]
Stanford Knee~\protect\cite{desai2022skm} & Non-dynamic & 0.866 (0.855--0.875) & 0.865 (0.852--0.877) \\*
 & Dynamic & 0.868 (0.858--0.879) & 0.868 (0.856--0.879) \\
\addlinespace[0.08em]
MSD Spleen~\protect\cite{antonelli2022medical} & Non-dynamic & 0.972 (0.969--0.975) & 0.932 (0.915--0.948) \\*
 & Dynamic & 0.974 (0.970--0.977) & 0.939 (0.923--0.954) \\
\addlinespace[0.08em]
Panther Task 1~\protect\cite{panther2025task1,betancourt_tarifa_2025_panther} & Non-dynamic & 0.622 (0.555--0.683) & 0.383 (0.339--0.425) \\*
 & Dynamic & 0.623 (0.567--0.682) & 0.389 (0.351--0.429) \\
\addlinespace[0.08em]
MSD Brain~\protect\cite{antonelli2022medical} & Non-dynamic & 0.734 (0.716--0.753) & 0.781 (0.761--0.799) \\*
 & Dynamic & 0.735 (0.717--0.753) & 0.781 (0.759--0.800) \\
\addlinespace[0.08em]
WMHSegChallenge~\protect\cite{kuijf2019standardized} & Non-dynamic & 0.418 (0.390--0.449) & 0.474 (0.450--0.504) \\*
 & Dynamic & 0.418 (0.393--0.447) & 0.476 (0.452--0.502) \\
\addlinespace[0.08em]
MSD Liver~\protect\cite{antonelli2022medical} & Non-dynamic & 0.828 (0.795--0.859) & 0.667 (0.634--0.703) \\*
 & Dynamic & 0.828 (0.798--0.857) & 0.667 (0.634--0.701) \\
\addlinespace[0.08em]
Pengwin~\protect\cite{sang2025pengwin} & Non-dynamic & 0.988 (0.987--0.989) & 0.990 (0.987--0.993) \\*
 & Dynamic & 0.988 (0.987--0.989) & 0.989 (0.985--0.992) \\
\addlinespace[0.08em]
MSD Heart~\protect\cite{antonelli2022medical} & Non-dynamic & 0.930 (0.922--0.937) & 0.605 (0.575--0.630) \\*
 & Dynamic & 0.930 (0.922--0.936) & 0.599 (0.565--0.625) \\
\addlinespace[0.08em]
BraTS2024 Glioma~\protect\cite{verdier2024bratsgli} & Non-dynamic & 0.734 (0.718--0.749) & 0.781 (0.767--0.795) \\*
 & Dynamic & 0.732 (0.717--0.747) & 0.782 (0.767--0.797) \\
\addlinespace[0.08em]
DeepPSMA PSMA~\protect\cite{deeppsma2025} & Non-dynamic & 0.866 (0.825--0.901) & 0.744 (0.704--0.780) \\*
 & Dynamic & 0.863 (0.822--0.895) & 0.734 (0.692--0.771) \\
\addlinespace[0.08em]
MU-Glioma-Post~\protect\cite{mahmoud2025mugliomapost} & Non-dynamic & 0.711 (0.687--0.734) & 0.750 (0.725--0.775) \\*
 & Dynamic & 0.707 (0.683--0.728) & 0.745 (0.722--0.768) \\
\addlinespace[0.08em]
WORD~\protect\cite{luo2022word} & Non-dynamic & 0.860 (0.849--0.869) & 0.783 (0.768--0.799) \\*
 & Dynamic & 0.855 (0.843--0.866) & 0.787 (0.771--0.801) \\
\addlinespace[0.08em]
Panther Task 2~\protect\cite{panther2025task2,betancourt_tarifa_2025_panther} & Non-dynamic & 0.469 (0.405--0.529) & 0.220 (0.188--0.258) \\*
 & Dynamic & 0.457 (0.403--0.513) & 0.224 (0.198--0.251) \\
\addlinespace[0.08em]
ToothFairy~\protect\cite{cipriano2022mandibular,bolelli2025toothfairy,bolelli2025cvpr_toothfairy2} & Non-dynamic & 0.778 (0.762--0.794) & 0.828 (0.812--0.843) \\*
 & Dynamic & 0.753 (0.738--0.769) & 0.808 (0.793--0.823) \\
\addlinespace[0.08em]
MSD Lung~\protect\cite{antonelli2022medical} & Non-dynamic & 0.703 (0.609--0.781) & 0.601 (0.501--0.692) \\*
 & Dynamic & 0.662 (0.550--0.757) & 0.562 (0.457--0.658) \\
\addlinespace[0.08em]
\end{longtable}
\normalsize
\endgroup

\begingroup
\singlespacing
\tablebodyfont
\setlength{\tabcolsep}{2pt}
\renewcommand{\arraystretch}{0.90}
\setlength{\LTcapwidth}{\linewidth}
\begin{longtable}{@{}l@{\hspace{0.65em}}l@{\hspace{0.70em}}c@{\hspace{0.70em}}c@{\hspace{0.70em}}c@{}}
\caption{Dataset-level Figure 3 dynamic-adaptation results for lesion detection. Values are shown as mean and 95\% CI.}\label{tab:fig3-dataset-detection}\\
\toprule
\textbf{Dataset} & \textbf{Variant} & \multicolumn{1}{c}{\textbf{mAP}} & \multicolumn{1}{c}{\textbf{FROC}} & \multicolumn{1}{c}{\textbf{AP@IoU 0.10}} \\
\cmidrule(lr){3-3}\cmidrule(lr){4-4}\cmidrule(lr){5-5}
 & & \multicolumn{1}{c}{\textbf{Mean (95\% CI)}} & \multicolumn{1}{c}{\textbf{Mean (95\% CI)}} & \multicolumn{1}{c}{\textbf{Mean (95\% CI)}} \\
\midrule

\endfirsthead
\caption[]{Dataset-level Figure 3 dynamic-adaptation results for lesion detection. Values are shown as mean and 95\% CI. (continued)}\\
\toprule
\textbf{Dataset} & \textbf{Variant} & \multicolumn{1}{c}{\textbf{mAP}} & \multicolumn{1}{c}{\textbf{FROC}} & \multicolumn{1}{c}{\textbf{AP@IoU 0.10}} \\
\cmidrule(lr){3-3}\cmidrule(lr){4-4}\cmidrule(lr){5-5}
 & & \multicolumn{1}{c}{\textbf{Mean (95\% CI)}} & \multicolumn{1}{c}{\textbf{Mean (95\% CI)}} & \multicolumn{1}{c}{\textbf{Mean (95\% CI)}} \\
\midrule

\endhead
\midrule
\multicolumn{5}{r}{\footnotesize Continued on next page} \\
\endfoot
\bottomrule
\endlastfoot
KIPA~\protect\cite{he2021meta,he2020dense,shao2011laparoscopic,shao2012precise} & Non-dynamic & 0.869 (0.769--0.960) & 0.948 (0.880--1.000) & 0.925 (0.856--0.991) \\*
 & Dynamic & 0.898 (0.814--0.972) & 0.956 (0.890--1.000) & 0.945 (0.885--1.000) \\
\addlinespace[0.08em]
BraTSMets~\protect\cite{moawad2023bratsmets} & Non-dynamic & 0.710 (0.588--0.825) & 0.687 (0.553--0.814) & 0.789 (0.675--0.890) \\*
 & Dynamic & 0.738 (0.611--0.853) & 0.687 (0.540--0.818) & 0.785 (0.666--0.889) \\
\addlinespace[0.08em]
MRAAneurysms~\protect\cite{lausanne2021tofmraaneurysm} & Non-dynamic & 0.681 (0.548--0.824) & 0.870 (0.765--0.967) & 0.803 (0.669--0.929) \\*
 & Dynamic & 0.704 (0.594--0.823) & 0.892 (0.796--0.982) & 0.831 (0.711--0.936) \\
\addlinespace[0.08em]
MELA~\protect\cite{mela2022} & Non-dynamic & 0.885 (0.823--0.938) & 0.981 (0.957--0.995) & 0.961 (0.933--0.984) \\*
 & Dynamic & 0.904 (0.851--0.951) & 0.973 (0.943--0.992) & 0.959 (0.929--0.985) \\
\addlinespace[0.08em]
Ribfrac~\protect\cite{yang2024ribfrac} & Non-dynamic & 0.649 (0.623--0.679) & 0.664 (0.631--0.691) & 0.792 (0.772--0.812) \\*
 & Dynamic & 0.663 (0.638--0.692) & 0.670 (0.641--0.697) & 0.802 (0.784--0.819) \\
\addlinespace[0.08em]
VALDO~\protect\cite{sudre2021valdo} & Non-dynamic & 0.442 (0.194--0.610) & 0.429 (0.273--0.576) & 0.511 (0.262--0.679) \\*
 & Dynamic & 0.443 (0.208--0.633) & 0.440 (0.292--0.627) & 0.518 (0.312--0.706) \\
\addlinespace[0.08em]
LIDC~\protect\cite{armato2011lidcidri} & Non-dynamic & 0.605 (0.569--0.645) & 0.612 (0.579--0.644) & 0.626 (0.591--0.666) \\*
 & Dynamic & 0.605 (0.568--0.645) & 0.616 (0.581--0.651) & 0.627 (0.591--0.667) \\
\addlinespace[0.08em]
DUKE~\protect\cite{saha2018radiogenomics} & Non-dynamic & 0.682 (0.636--0.736) & 0.869 (0.831--0.902) & 0.833 (0.794--0.876) \\*
 & Dynamic & 0.678 (0.630--0.730) & 0.881 (0.848--0.916) & 0.837 (0.792--0.884) \\
\addlinespace[0.08em]
CADA~\protect\cite{ivantsits2022cada} & Non-dynamic & 0.905 (0.843--0.963) & 0.927 (0.881--0.974) & 0.923 (0.871--0.972) \\*
 & Dynamic & 0.898 (0.832--0.959) & 0.932 (0.879--0.982) & 0.922 (0.867--0.977) \\
\addlinespace[0.08em]
\end{longtable}
\normalsize
\endgroup

\clearpage
\subsection{Figure 4 Extended Results}
\label{app:figure4-metric-tables}
\noindent These tables report the data- and compute-efficiency results underlying Figure 4.

\subsubsection{Low-Data Summary}
\label{app:figure4-low-data-summary}

\begin{table}[!htbp]
\caption{Figure 4 low-data segmentation summary DSC scores across 6 datasets. Values are dataset-equal means; SD reports variability across dataset-level means after averaging the three random-seed trials within each dataset.}
\label{tab:fig4-low-data-segmentation-summary}
\centering
\tablebodyfont
\setlength{\tabcolsep}{1pt}
\renewcommand{\arraystretch}{0.90}
\begin{tabular}{@{}l@{\hspace{0.55em}}c@{\hspace{0.55em}}c@{\hspace{0.55em}}c@{\hspace{0.55em}}c@{}}
\toprule
\textbf{Model variant} & \multicolumn{1}{c}{\textbf{10 cases}} & \multicolumn{1}{c}{\textbf{20 cases}} & \multicolumn{1}{c}{\textbf{40 cases}} & \multicolumn{1}{c}{\textbf{100 cases}} \\
\cmidrule(lr){2-2}\cmidrule(lr){3-3}\cmidrule(lr){4-4}\cmidrule(lr){5-5}
 & \multicolumn{1}{c}{\makecell[c]{\textbf{Mean}\\\textbf{SD}}} & \multicolumn{1}{c}{\makecell[c]{\textbf{Mean}\\\textbf{SD}}} & \multicolumn{1}{c}{\makecell[c]{\textbf{Mean}\\\textbf{SD}}} & \multicolumn{1}{c}{\makecell[c]{\textbf{Mean}\\\textbf{SD}}} \\
\midrule
\Name{cnn} & \makecell[c]{0.533\\{\fontsize{5.4}{5.9}\selectfont SD 0.181}} & \makecell[c]{0.640\\{\fontsize{5.4}{5.9}\selectfont SD 0.159}} & \makecell[c]{0.671\\{\fontsize{5.4}{5.9}\selectfont SD 0.161}} & \makecell[c]{0.754\\{\fontsize{5.4}{5.9}\selectfont SD 0.116}} \\
\Name{cnn} (scratch) & \makecell[c]{0.503\\{\fontsize{5.4}{5.9}\selectfont SD 0.160}} & \makecell[c]{0.611\\{\fontsize{5.4}{5.9}\selectfont SD 0.150}} & \makecell[c]{0.648\\{\fontsize{5.4}{5.9}\selectfont SD 0.160}} & \makecell[c]{0.733\\{\fontsize{5.4}{5.9}\selectfont SD 0.114}} \\
CTFM~\protect\cite{pai2025vision} & \makecell[c]{0.507\\{\fontsize{5.4}{5.9}\selectfont SD 0.179}} & \makecell[c]{0.608\\{\fontsize{5.4}{5.9}\selectfont SD 0.149}} & \makecell[c]{0.657\\{\fontsize{5.4}{5.9}\selectfont SD 0.156}} & \makecell[c]{0.745\\{\fontsize{5.4}{5.9}\selectfont SD 0.113}} \\
VISTA3D~\protect\cite{he2025vista3d} & \makecell[c]{0.503\\{\fontsize{5.4}{5.9}\selectfont SD 0.175}} & \makecell[c]{0.602\\{\fontsize{5.4}{5.9}\selectfont SD 0.167}} & \makecell[c]{0.648\\{\fontsize{5.4}{5.9}\selectfont SD 0.158}} & \makecell[c]{0.751\\{\fontsize{5.4}{5.9}\selectfont SD 0.120}} \\
MERLIN~\protect\cite{Blankemeier2026} & \makecell[c]{0.500\\{\fontsize{5.4}{5.9}\selectfont SD 0.160}} & \makecell[c]{0.579\\{\fontsize{5.4}{5.9}\selectfont SD 0.145}} & \makecell[c]{0.635\\{\fontsize{5.4}{5.9}\selectfont SD 0.167}} & \makecell[c]{0.700\\{\fontsize{5.4}{5.9}\selectfont SD 0.126}} \\
MISFM~\protect\cite{wang2023mis} & \makecell[c]{0.499\\{\fontsize{5.4}{5.9}\selectfont SD 0.153}} & \makecell[c]{0.612\\{\fontsize{5.4}{5.9}\selectfont SD 0.143}} & \makecell[c]{0.659\\{\fontsize{5.4}{5.9}\selectfont SD 0.152}} & \makecell[c]{0.742\\{\fontsize{5.4}{5.9}\selectfont SD 0.120}} \\
VOCO~\protect\cite{wu2024voco} & \makecell[c]{0.490\\{\fontsize{5.4}{5.9}\selectfont SD 0.170}} & \makecell[c]{0.594\\{\fontsize{5.4}{5.9}\selectfont SD 0.155}} & \makecell[c]{0.632\\{\fontsize{5.4}{5.9}\selectfont SD 0.175}} & \makecell[c]{0.704\\{\fontsize{5.4}{5.9}\selectfont SD 0.131}} \\
SwinUNETR~\protect\cite{tang2022self} & \makecell[c]{0.417\\{\fontsize{5.4}{5.9}\selectfont SD 0.144}} & \makecell[c]{0.544\\{\fontsize{5.4}{5.9}\selectfont SD 0.126}} & \makecell[c]{0.598\\{\fontsize{5.4}{5.9}\selectfont SD 0.144}} & \makecell[c]{0.687\\{\fontsize{5.4}{5.9}\selectfont SD 0.114}} \\
CURIA~\protect\cite{dancette2025curia} & \makecell[c]{0.372\\{\fontsize{5.4}{5.9}\selectfont SD 0.130}} & \makecell[c]{0.431\\{\fontsize{5.4}{5.9}\selectfont SD 0.139}} & \makecell[c]{0.473\\{\fontsize{5.4}{5.9}\selectfont SD 0.187}} & \makecell[c]{0.529\\{\fontsize{5.4}{5.9}\selectfont SD 0.167}} \\
\bottomrule
\end{tabular}
\end{table}
\normalsize

\begin{table}[!htbp]
\caption{Pairwise low-data DSC comparisons corresponding to \cref{tab:fig4-low-data-segmentation-summary}. Cells report \Name{cnn}-minus-comparator dataset-equal mean differences with paired dataset-level bootstrap 95\% CIs in parentheses; dataset-level values average the three random-seed trials before differencing. Positive values favor \Name{cnn}. A dagger marks comparisons for which the paired 95\% CI does not lie entirely above zero.}
\label{tab:fig4-low-data-segmentation-pairwise-support}
\centering
\tablebodyfont
\fontsize{7.0bp}{8.3bp}\selectfont
\setlength{\tabcolsep}{0pt}
\renewcommand{\arraystretch}{0.90}
\begin{tabular}{@{}l@{\hspace{0.20em}}c@{\hspace{0.20em}}c@{\hspace{0.20em}}c@{\hspace{0.20em}}c@{}}
\toprule
\textbf{Comparator} & \multicolumn{1}{c}{\textbf{10 cases}} & \multicolumn{1}{c}{\textbf{20 cases}} & \multicolumn{1}{c}{\textbf{40 cases}} & \multicolumn{1}{c}{\textbf{100 cases}} \\
\cmidrule(lr){2-2}\cmidrule(lr){3-3}\cmidrule(lr){4-4}\cmidrule(lr){5-5}
 & \multicolumn{1}{c}{\makecell[c]{\textbf{\(\Delta\)}\\\textbf{(95\% CI)}}} & \multicolumn{1}{c}{\makecell[c]{\textbf{\(\Delta\)}\\\textbf{(95\% CI)}}} & \multicolumn{1}{c}{\makecell[c]{\textbf{\(\Delta\)}\\\textbf{(95\% CI)}}} & \multicolumn{1}{c}{\makecell[c]{\textbf{\(\Delta\)}\\\textbf{(95\% CI)}}} \\
\midrule
\Name{cnn} & Reference & Reference & Reference & Reference \\
\Name{cnn} (scratch) & \makecell[c]{0.030\\{\fontsize{5.2}{5.7}\selectfont (0.007--0.051)}} & \makecell[c]{0.029\\{\fontsize{5.2}{5.7}\selectfont (0.007--0.049)}} & \makecell[c]{0.023\\{\fontsize{5.2}{5.7}\selectfont (0.008--0.037)}} & \makecell[c]{0.021\\{\fontsize{5.2}{5.7}\selectfont (0.004--0.044)}} \\
CTFM~\protect\cite{pai2025vision} & \makecell[c]{0.025\\{\fontsize{5.2}{5.7}\selectfont (0.001--0.053)}} & \makecell[c]{0.032\\{\fontsize{5.2}{5.7}\selectfont (0.008--0.057)}} & \makecell[c]{0.014\\{\fontsize{5.2}{5.7}\selectfont (0.001--0.027)}} & \makecell[c]{0.010\\{\fontsize{5.2}{5.7}\selectfont (0.002--0.018)}} \\
VISTA3D~\protect\cite{he2025vista3d} & \makecell[c]{0.030\\{\fontsize{5.2}{5.7}\selectfont (0.008--0.052)}} & \makecell[c]{0.038\\{\fontsize{5.2}{5.7}\selectfont (0.020--0.058)}} & \makecell[c]{0.023\\{\fontsize{5.2}{5.7}\selectfont (-0.007--0.051)$^\dagger$}} & \makecell[c]{0.003\\{\fontsize{5.2}{5.7}\selectfont (-0.009--0.017)$^\dagger$}} \\
\bottomrule
\end{tabular}
\end{table}
\normalsize

\clearpage
\subsubsection{Low-Data Dataset-Level Results}
\label{app:figure4-low-data-dataset-level}

\begingroup
\singlespacing
\tablebodyfont
\setlength{\tabcolsep}{2pt}
\renewcommand{\arraystretch}{0.90}
\setlength{\LTcapwidth}{\linewidth}
\begin{longtable}{@{}l@{\hspace{0.70em}}l@{\hspace{0.70em}}c@{\hspace{0.70em}}c@{\hspace{0.70em}}c@{}}
\caption{Dataset-level Figure 4 low-data segmentation DSC scores.}\label{tab:fig4-low-data-segmentation-dataset}\\
\toprule
\textbf{Model variant} & \textbf{Training cases} & \multicolumn{1}{c}{\textbf{Trial 1 DSC}} & \multicolumn{1}{c}{\textbf{Trial 2 DSC}} & \multicolumn{1}{c}{\textbf{Trial 3 DSC}} \\
\cmidrule(lr){3-3}\cmidrule(lr){4-4}\cmidrule(lr){5-5}
 & & \multicolumn{1}{c}{\textbf{Mean (95\% CI)}} & \multicolumn{1}{c}{\textbf{Mean (95\% CI)}} & \multicolumn{1}{c}{\textbf{Mean (95\% CI)}} \\
\midrule

\endfirsthead
\caption[]{Dataset-level Figure 4 low-data segmentation DSC scores. (continued)}\\
\toprule
\textbf{Model variant} & \textbf{Training cases} & \multicolumn{1}{c}{\textbf{Trial 1 DSC}} & \multicolumn{1}{c}{\textbf{Trial 2 DSC}} & \multicolumn{1}{c}{\textbf{Trial 3 DSC}} \\
\cmidrule(lr){3-3}\cmidrule(lr){4-4}\cmidrule(lr){5-5}
 & & \multicolumn{1}{c}{\textbf{Mean (95\% CI)}} & \multicolumn{1}{c}{\textbf{Mean (95\% CI)}} & \multicolumn{1}{c}{\textbf{Mean (95\% CI)}} \\
\midrule

\endhead
\midrule
\multicolumn{5}{r}{\footnotesize Continued on next page} \\
\endfoot
\bottomrule
\endlastfoot
\addlinespace[0.18em]\multicolumn{5}{@{}l}{\textbf{Mama Mia}~\protect\cite{garrucho2025large}}\\*
\multicolumn{5}{@{}l}{\hspace{0.75em}\Name{cnn}}\\*
 & 10 & 0.517 (0.491--0.544) & 0.624 (0.601--0.646) & 0.680 (0.660--0.699) \\*
 & 20 & 0.580 (0.557--0.606) & 0.612 (0.588--0.635) & 0.660 (0.637--0.682) \\*
 & 40 & 0.711 (0.691--0.731) & 0.748 (0.730--0.766) & 0.661 (0.636--0.686) \\*
 & 100 & 0.759 (0.740--0.777) & 0.775 (0.758--0.791) & 0.741 (0.722--0.759) \\
\addlinespace[0.06em]
\multicolumn{5}{@{}l}{\hspace{0.75em}\Name{cnn} (scratch)}\\*
 & 10 & 0.481 (0.455--0.506) & 0.549 (0.524--0.573) & 0.609 (0.589--0.630) \\
\end{longtable}
\normalsize
\endgroup

\clearpage

\subsubsection{Low-Compute Summary}
\label{app:figure4-low-compute-summary}

\begin{table}[!htbp]
\caption{Figure 4 low-compute segmentation summary scores across 34 datasets. Values are dataset-equal means with 95\% CIs; SD reports variability across dataset-level metric means.}
\label{tab:fig4-low-compute-segmentation-summary}
\centering
\tablebodyfont
\fontsize{7.2bp}{8.4bp}\selectfont
\setlength{\tabcolsep}{0pt}
\renewcommand{\arraystretch}{0.90}
\begin{tabular}{@{}l@{\hspace{0.18em}}c@{\hspace{0.18em}}c@{\hspace{0.18em}}c@{\hspace{0.18em}}c@{}}
\toprule
\textbf{Model variant} & \multicolumn{1}{c}{\makecell[c]{\textbf{DSC}\\\textbf{150 ep}}} & \multicolumn{1}{c}{\makecell[c]{\textbf{DSC}\\\textbf{1000 ep}}} & \multicolumn{1}{c}{\makecell[c]{\textbf{NSD}\\\textbf{150 ep}}} & \multicolumn{1}{c}{\makecell[c]{\textbf{NSD}\\\textbf{1000 ep}}} \\
\cmidrule(lr){2-2}\cmidrule(lr){3-3}\cmidrule(lr){4-4}\cmidrule(lr){5-5}
 & \multicolumn{1}{c}{\makecell[c]{\textbf{Mean}\\\textbf{(95\% CI)}\\\textbf{SD}}} & \multicolumn{1}{c}{\makecell[c]{\textbf{Mean}\\\textbf{(95\% CI)}\\\textbf{SD}}} & \multicolumn{1}{c}{\makecell[c]{\textbf{Mean}\\\textbf{(95\% CI)}\\\textbf{SD}}} & \multicolumn{1}{c}{\makecell[c]{\textbf{Mean}\\\textbf{(95\% CI)}\\\textbf{SD}}} \\
\midrule
\Name{cnn} & \makecell[c]{0.709\\{\fontsize{5.4}{5.9}\selectfont (0.700--0.719)}\\{\fontsize{5.2}{5.7}\selectfont SD 0.170}} & \makecell[c]{0.715\\{\fontsize{5.4}{5.9}\selectfont (0.705--0.724)}\\{\fontsize{5.2}{5.7}\selectfont SD 0.165}} & \makecell[c]{0.629\\{\fontsize{5.4}{5.9}\selectfont (0.621--0.638)}\\{\fontsize{5.2}{5.7}\selectfont SD 0.197}} & \makecell[c]{0.640\\{\fontsize{5.4}{5.9}\selectfont (0.631--0.648)}\\{\fontsize{5.2}{5.7}\selectfont SD 0.194}} \\
\Name{cnn} (scratch) & \makecell[c]{0.653\\{\fontsize{5.4}{5.9}\selectfont (0.644--0.662)}\\{\fontsize{5.2}{5.7}\selectfont SD 0.206}} & \makecell[c]{0.683\\{\fontsize{5.4}{5.9}\selectfont (0.673--0.692)}\\{\fontsize{5.2}{5.7}\selectfont SD 0.188}} & \makecell[c]{0.566\\{\fontsize{5.4}{5.9}\selectfont (0.559--0.574)}\\{\fontsize{5.2}{5.7}\selectfont SD 0.231}} & \makecell[c]{0.606\\{\fontsize{5.4}{5.9}\selectfont (0.598--0.614)}\\{\fontsize{5.2}{5.7}\selectfont SD 0.220}} \\
\bottomrule
\end{tabular}
\end{table}
\normalsize

\begin{table}[!htbp]
\caption{Figure 4 low-compute classification summary scores across 8 datasets. Values are dataset-equal means with 95\% CIs; SD reports variability across dataset-level metric means.}
\label{tab:fig4-low-compute-classification-summary}
\centering
\tablebodyfont
\fontsize{7.2bp}{8.4bp}\selectfont
\setlength{\tabcolsep}{0pt}
\renewcommand{\arraystretch}{0.90}
\begin{tabular}{@{}l@{\hspace{0.15em}}c@{\hspace{0.15em}}c@{\hspace{0.15em}}c@{\hspace{0.15em}}c@{}}
\toprule
\textbf{Model variant} & \multicolumn{1}{c}{\makecell[c]{\textbf{AUROC}\\\textbf{20 ep}}} & \multicolumn{1}{c}{\makecell[c]{\textbf{AUROC}\\\textbf{100 ep}}} & \multicolumn{1}{c}{\makecell[c]{\textbf{AUPRC}\\\textbf{20 ep}}} & \multicolumn{1}{c}{\makecell[c]{\textbf{AUPRC}\\\textbf{100 ep}}} \\
\cmidrule(lr){2-2}\cmidrule(lr){3-3}\cmidrule(lr){4-4}\cmidrule(lr){5-5}
 & \multicolumn{1}{c}{\makecell[c]{\textbf{Mean}\\\textbf{(95\% CI)}\\\textbf{SD}}} & \multicolumn{1}{c}{\makecell[c]{\textbf{Mean}\\\textbf{(95\% CI)}\\\textbf{SD}}} & \multicolumn{1}{c}{\makecell[c]{\textbf{Mean}\\\textbf{(95\% CI)}\\\textbf{SD}}} & \multicolumn{1}{c}{\makecell[c]{\textbf{Mean}\\\textbf{(95\% CI)}\\\textbf{SD}}} \\
\midrule
\Name{vit} & \makecell[c]{0.699\\{\fontsize{5.4}{5.9}\selectfont (0.683--0.716)}\\{\fontsize{5.2}{5.7}\selectfont SD 0.039}} & \makecell[c]{0.748\\{\fontsize{5.4}{5.9}\selectfont (0.731--0.765)}\\{\fontsize{5.2}{5.7}\selectfont SD 0.064}} & \makecell[c]{0.439\\{\fontsize{5.4}{5.9}\selectfont (0.420--0.470)}\\{\fontsize{5.2}{5.7}\selectfont SD 0.118}} & \makecell[c]{0.496\\{\fontsize{5.4}{5.9}\selectfont (0.479--0.526)}\\{\fontsize{5.2}{5.7}\selectfont SD 0.115}} \\
\Name{vit} (scratch) & \makecell[c]{0.624\\{\fontsize{5.4}{5.9}\selectfont (0.606--0.642)}\\{\fontsize{5.2}{5.7}\selectfont SD 0.055}} & \makecell[c]{0.665\\{\fontsize{5.4}{5.9}\selectfont (0.648--0.683)}\\{\fontsize{5.2}{5.7}\selectfont SD 0.073}} & \makecell[c]{0.351\\{\fontsize{5.4}{5.9}\selectfont (0.340--0.378)}\\{\fontsize{5.2}{5.7}\selectfont SD 0.130}} & \makecell[c]{0.394\\{\fontsize{5.4}{5.9}\selectfont (0.379--0.422)}\\{\fontsize{5.2}{5.7}\selectfont SD 0.117}} \\
\bottomrule
\end{tabular}
\end{table}
\normalsize

\begin{table}[!htbp]
\caption{Pairwise low-compute segmentation comparisons corresponding to \cref{tab:fig4-low-compute-segmentation-summary}. Cells report dataset-equal mean differences (\Name{cnn} minus comparator) with paired dataset-level bootstrap 95\% CIs in parentheses. Positive values favor \Name{cnn}. A dagger marks comparisons for which the paired 95\% CI does not lie entirely above zero.}
\label{tab:fig4-low-compute-segmentation-pairwise-support}
\centering
\tablebodyfont
\fontsize{7.0bp}{8.3bp}\selectfont
\setlength{\tabcolsep}{0pt}
\renewcommand{\arraystretch}{0.90}
\begin{tabular}{@{}l@{\hspace{0.20em}}c@{\hspace{0.20em}}c@{\hspace{0.20em}}c@{\hspace{0.20em}}c@{}}
\toprule
\textbf{Comparator} & \multicolumn{1}{c}{\makecell[c]{\textbf{DSC}\\\textbf{150 ep}}} & \multicolumn{1}{c}{\makecell[c]{\textbf{DSC}\\\textbf{1000 ep}}} & \multicolumn{1}{c}{\makecell[c]{\textbf{NSD}\\\textbf{150 ep}}} & \multicolumn{1}{c}{\makecell[c]{\textbf{NSD}\\\textbf{1000 ep}}} \\
\cmidrule(lr){2-2}\cmidrule(lr){3-3}\cmidrule(lr){4-4}\cmidrule(lr){5-5}
 & \multicolumn{1}{c}{\makecell[c]{\textbf{\(\Delta\)}\\\textbf{(95\% CI)}}} & \multicolumn{1}{c}{\makecell[c]{\textbf{\(\Delta\)}\\\textbf{(95\% CI)}}} & \multicolumn{1}{c}{\makecell[c]{\textbf{\(\Delta\)}\\\textbf{(95\% CI)}}} & \multicolumn{1}{c}{\makecell[c]{\textbf{\(\Delta\)}\\\textbf{(95\% CI)}}} \\
\midrule
\Name{cnn} & Reference & Reference & Reference & Reference \\
\Name{cnn} (scratch) & \makecell[c]{0.056\\{\fontsize{5.3}{5.8}\selectfont (0.035 to 0.079)}} & \makecell[c]{0.033\\{\fontsize{5.3}{5.8}\selectfont (0.018 to 0.050)}} & \makecell[c]{0.063\\{\fontsize{5.3}{5.8}\selectfont (0.041 to 0.086)}} & \makecell[c]{0.033\\{\fontsize{5.3}{5.8}\selectfont (0.019 to 0.051)}} \\
nnU-Net~\protect\cite{isensee2021nnu} & \makecell[c]{0.053\\{\fontsize{5.3}{5.8}\selectfont (0.032 to 0.076)}} & \makecell[c]{0.020\\{\fontsize{5.3}{5.8}\selectfont (0.007 to 0.036)}} & \makecell[c]{0.062\\{\fontsize{5.3}{5.8}\selectfont (0.041 to 0.084)}} & \makecell[c]{0.019\\{\fontsize{5.3}{5.8}\selectfont (0.008 to 0.033)}} \\
CTFM~\protect\cite{pai2025vision} & \makecell[c]{0.019\\{\fontsize{5.3}{5.8}\selectfont (0.004 to 0.036)}} & \makecell[c]{0.030\\{\fontsize{5.3}{5.8}\selectfont (0.013 to 0.054)}} & \makecell[c]{0.021\\{\fontsize{5.3}{5.8}\selectfont (0.008 to 0.037)}} & \makecell[c]{0.028\\{\fontsize{5.3}{5.8}\selectfont (0.011 to 0.051)}} \\
VISTA3D~\protect\cite{he2025vista3d} & \makecell[c]{0.016\\{\fontsize{5.3}{5.8}\selectfont (0.002 to 0.032)}} & \makecell[c]{0.050\\{\fontsize{5.3}{5.8}\selectfont (0.022 to 0.088)}} & \makecell[c]{0.014\\{\fontsize{5.3}{5.8}\selectfont (0.001 to 0.030)}} & \makecell[c]{0.046\\{\fontsize{5.3}{5.8}\selectfont (0.017 to 0.081)}} \\
MISFM~\protect\cite{wang2023mis} & \makecell[c]{0.021\\{\fontsize{5.3}{5.8}\selectfont (0.005 to 0.041)}} & \makecell[c]{0.051\\{\fontsize{5.3}{5.8}\selectfont (0.024 to 0.082)}} & \makecell[c]{0.019\\{\fontsize{5.3}{5.8}\selectfont (0.003 to 0.035)}} & \makecell[c]{0.046\\{\fontsize{5.3}{5.8}\selectfont (0.022 to 0.076)}} \\
MERLIN~\protect\cite{Blankemeier2026} & \makecell[c]{0.013\\{\fontsize{5.3}{5.8}\selectfont (0.002 to 0.026)}} & \makecell[c]{0.036\\{\fontsize{5.3}{5.8}\selectfont (0.019 to 0.057)}} & \makecell[c]{0.012\\{\fontsize{5.3}{5.8}\selectfont (0.002 to 0.024)}} & \makecell[c]{0.034\\{\fontsize{5.3}{5.8}\selectfont (0.020 to 0.051)}} \\
VOCO~\protect\cite{wu2024voco} & \makecell[c]{0.036\\{\fontsize{5.3}{5.8}\selectfont (0.019 to 0.058)}} & \makecell[c]{0.055\\{\fontsize{5.3}{5.8}\selectfont (0.030 to 0.084)}} & \makecell[c]{0.035\\{\fontsize{5.3}{5.8}\selectfont (0.021 to 0.050)}} & \makecell[c]{0.048\\{\fontsize{5.3}{5.8}\selectfont (0.026 to 0.072)}} \\
\Name{vit} & \makecell[c]{0.026\\{\fontsize{5.3}{5.8}\selectfont (0.008 to 0.048)}} & \makecell[c]{0.041\\{\fontsize{5.3}{5.8}\selectfont (0.024 to 0.061)}} & \makecell[c]{0.027\\{\fontsize{5.3}{5.8}\selectfont (0.010 to 0.048)}} & \makecell[c]{0.039\\{\fontsize{5.3}{5.8}\selectfont (0.021 to 0.059)}} \\
CURIA~\protect\cite{dancette2025curia} & \makecell[c]{0.124\\{\fontsize{5.3}{5.8}\selectfont (0.088 to 0.171)}} & \makecell[c]{0.078\\{\fontsize{5.3}{5.8}\selectfont (0.056 to 0.103)}} & \makecell[c]{0.155\\{\fontsize{5.3}{5.8}\selectfont (0.117 to 0.205)}} & \makecell[c]{0.094\\{\fontsize{5.3}{5.8}\selectfont (0.074 to 0.118)}} \\
SwinUNETR~\protect\cite{tang2022self} & \makecell[c]{0.066\\{\fontsize{5.3}{5.8}\selectfont (0.040 to 0.096)}} & \makecell[c]{0.097\\{\fontsize{5.3}{5.8}\selectfont (0.060 to 0.137)}} & \makecell[c]{0.067\\{\fontsize{5.3}{5.8}\selectfont (0.042 to 0.095)}} & \makecell[c]{0.090\\{\fontsize{5.3}{5.8}\selectfont (0.058 to 0.125)}} \\
\bottomrule
\end{tabular}
\end{table}
\normalsize

\begin{table}[!htbp]
\caption{Pairwise low-compute classification comparisons corresponding to \cref{tab:fig4-low-compute-classification-summary}. Cells report dataset-equal mean differences (\Name{vit} minus comparator) with paired dataset-level bootstrap 95\% CIs in parentheses. Positive values favor \Name{vit}. A dagger marks comparisons for which the paired 95\% CI does not lie entirely above zero.}
\label{tab:fig4-low-compute-classification-pairwise-support}
\centering
\tablebodyfont
\fontsize{7.0bp}{8.3bp}\selectfont
\setlength{\tabcolsep}{0pt}
\renewcommand{\arraystretch}{0.90}
\begin{tabular}{@{}l@{\hspace{0.20em}}c@{\hspace{0.20em}}c@{\hspace{0.20em}}c@{\hspace{0.20em}}c@{}}
\toprule
\textbf{Comparator} & \multicolumn{1}{c}{\makecell[c]{\textbf{AUROC}\\\textbf{20 ep}}} & \multicolumn{1}{c}{\makecell[c]{\textbf{AUROC}\\\textbf{100 ep}}} & \multicolumn{1}{c}{\makecell[c]{\textbf{AUPRC}\\\textbf{20 ep}}} & \multicolumn{1}{c}{\makecell[c]{\textbf{AUPRC}\\\textbf{100 ep}}} \\
\cmidrule(lr){2-2}\cmidrule(lr){3-3}\cmidrule(lr){4-4}\cmidrule(lr){5-5}
 & \multicolumn{1}{c}{\makecell[c]{\textbf{\(\Delta\)}\\\textbf{(95\% CI)}}} & \multicolumn{1}{c}{\makecell[c]{\textbf{\(\Delta\)}\\\textbf{(95\% CI)}}} & \multicolumn{1}{c}{\makecell[c]{\textbf{\(\Delta\)}\\\textbf{(95\% CI)}}} & \multicolumn{1}{c}{\makecell[c]{\textbf{\(\Delta\)}\\\textbf{(95\% CI)}}} \\
\midrule
\Name{vit} & Reference & Reference & Reference & Reference \\
\Name{vit} (scratch) & \makecell[c]{0.075\\{\fontsize{5.3}{5.8}\selectfont (0.041 to 0.111)}} & \makecell[c]{0.082\\{\fontsize{5.3}{5.8}\selectfont (0.061 to 0.104)}} & \makecell[c]{0.088\\{\fontsize{5.3}{5.8}\selectfont (0.040 to 0.140)}} & \makecell[c]{0.102\\{\fontsize{5.3}{5.8}\selectfont (0.075 to 0.128)}} \\
CTFM~\protect\cite{pai2025vision} & \makecell[c]{-0.019\\{\fontsize{5.3}{5.8}\selectfont (-0.069 to 0.028)$^\dagger$}} & \makecell[c]{-0.001\\{\fontsize{5.3}{5.8}\selectfont (-0.035 to 0.027)$^\dagger$}} & \makecell[c]{-0.024\\{\fontsize{5.3}{5.8}\selectfont (-0.097 to 0.048)$^\dagger$}} & \makecell[c]{-0.003\\{\fontsize{5.3}{5.8}\selectfont (-0.069 to 0.044)$^\dagger$}} \\
CURIA~\protect\cite{dancette2025curia} & \makecell[c]{-0.035\\{\fontsize{5.3}{5.8}\selectfont (-0.080 to 0.003)$^\dagger$}} & \makecell[c]{0.009\\{\fontsize{5.3}{5.8}\selectfont (-0.012 to 0.030)$^\dagger$}} & \makecell[c]{-0.031\\{\fontsize{5.3}{5.8}\selectfont (-0.101 to 0.034)$^\dagger$}} & \makecell[c]{0.019\\{\fontsize{5.3}{5.8}\selectfont (-0.015 to 0.048)$^\dagger$}} \\
\bottomrule
\end{tabular}
\end{table}
\normalsize

\clearpage

\subsubsection{Low-Compute Dataset-Level Results}
\label{app:figure4-low-compute-dataset-level}

\begingroup
\singlespacing
\tablebodyfont
\setlength{\tabcolsep}{2pt}
\renewcommand{\arraystretch}{0.90}
\setlength{\LTcapwidth}{\linewidth}
\begin{longtable}{@{}l@{\hspace{0.80em}}l@{\hspace{0.90em}}c@{\hspace{0.90em}}c@{}}
\caption{Dataset-level Figure 4 low-compute segmentation scores. Values are shown as mean and 95\% CI.}\label{tab:fig4-low-compute-segmentation-dataset}\\
\toprule
\textbf{Model variant} & \textbf{Schedule} & \multicolumn{1}{c}{\textbf{DSC}} & \multicolumn{1}{c}{\textbf{NSD}} \\
\cmidrule(lr){3-3}\cmidrule(lr){4-4}
 & & \multicolumn{1}{c}{\textbf{Mean (95\% CI)}} & \multicolumn{1}{c}{\textbf{Mean (95\% CI)}} \\
\midrule

\endfirsthead
\caption[]{Dataset-level Figure 4 low-compute segmentation scores. Values are shown as mean and 95\% CI. (continued)}\\
\toprule
\textbf{Model variant} & \textbf{Schedule} & \multicolumn{1}{c}{\textbf{DSC}} & \multicolumn{1}{c}{\textbf{NSD}} \\
\cmidrule(lr){3-3}\cmidrule(lr){4-4}
 & & \multicolumn{1}{c}{\textbf{Mean (95\% CI)}} & \multicolumn{1}{c}{\textbf{Mean (95\% CI)}} \\
\midrule

\endhead
\midrule
\multicolumn{4}{r}{\footnotesize Continued on next page} \\
\endfoot
\bottomrule
\endlastfoot
\addlinespace[0.18em]\multicolumn{4}{@{}l}{\textbf{MSD Brain~\protect\cite{antonelli2022medical}}}\\*
\multicolumn{4}{@{}l}{\hspace{0.75em}\Name{cnn}}\\*
 & 150 epochs & 0.735 (0.718--0.752) & 0.779 (0.761--0.797) \\*
 & 1000 epochs & 0.735 (0.717--0.751) & 0.781 (0.762--0.800) \\
\addlinespace[0.06em]
\multicolumn{4}{@{}l}{\hspace{0.75em}\Name{cnn} (scratch)}\\*
 & 150 epochs & 0.729 (0.710--0.746) & 0.770 (0.750--0.789) \\*
 & 1000 epochs & 0.729 (0.711--0.746) & 0.777 (0.758--0.795) \\
\addlinespace[0.06em]
\multicolumn{4}{@{}l}{\hspace{0.75em}nnU-Net~\protect\cite{isensee2021nnu}}\\
\end{longtable}
\normalsize
\endgroup

\begingroup
\singlespacing
\tablebodyfont
\setlength{\tabcolsep}{2pt}
\renewcommand{\arraystretch}{0.90}
\setlength{\LTcapwidth}{\linewidth}
\begin{longtable}{@{}l@{\hspace{0.80em}}l@{\hspace{0.90em}}c@{\hspace{0.90em}}c@{}}
\caption{Dataset-level Figure 4 low-compute classification scores. Values are shown as mean and 95\% CI.}\label{tab:fig4-low-compute-classification-dataset}\\
\toprule
\textbf{Model variant} & \textbf{Schedule} & \multicolumn{1}{c}{\textbf{AUROC}} & \multicolumn{1}{c}{\textbf{AUPRC}} \\
\cmidrule(lr){3-3}\cmidrule(lr){4-4}
 & & \multicolumn{1}{c}{\textbf{Mean (95\% CI)}} & \multicolumn{1}{c}{\textbf{Mean (95\% CI)}} \\
\midrule

\endfirsthead
\caption[]{Dataset-level Figure 4 low-compute classification scores. Values are shown as mean and 95\% CI. (continued)}\\
\toprule
\textbf{Model variant} & \textbf{Schedule} & \multicolumn{1}{c}{\textbf{AUROC}} & \multicolumn{1}{c}{\textbf{AUPRC}} \\
\cmidrule(lr){3-3}\cmidrule(lr){4-4}
 & & \multicolumn{1}{c}{\textbf{Mean (95\% CI)}} & \multicolumn{1}{c}{\textbf{Mean (95\% CI)}} \\
\midrule

\endhead
\midrule
\multicolumn{4}{r}{\footnotesize Continued on next page} \\
\endfoot
\bottomrule
\endlastfoot
\addlinespace[0.18em]\multicolumn{4}{@{}l}{\textbf{CT-RATE}~\protect\cite{hamamci2026generalist}}\\*
\multicolumn{4}{@{}l}{\hspace{0.75em}\Name{vit}}\\*
 & 20 epochs & 0.681 (0.667--0.693) & 0.318 (0.297--0.344) \\*
 & 100 epochs & 0.826 (0.817--0.835) & 0.537 (0.519--0.562) \\
\addlinespace[0.06em]
\multicolumn{4}{@{}l}{\hspace{0.75em}\Name{vit} (scratch)}\\*
 & 20 epochs & 0.673 (0.658--0.686) & 0.310 (0.289--0.337) \\*
 & 100 epochs & 0.758 (0.745--0.769) & 0.411 (0.391--0.440) \\
\addlinespace[0.06em]
\multicolumn{4}{@{}l}{\hspace{0.75em}CTFM~\protect\cite{pai2025vision}}\\
\end{longtable}
\normalsize
\endgroup

\clearpage
\subsection{Figure 5 Extended Results}
\label{app:figure5-metric-tables}
\noindent These tables report the OOD robustness and external validation results underlying Figure 5.

\subsubsection{OOD Robustness Summary}
\label{app:figure5-ood-summary}

\begin{table}[!htbp]
\caption{Figure 5 OOD robustness summary across 7 paired ID/OOD transfer settings. Values are paired-transfer-setting means in raw metric units with 95\% CIs; SD reports variability across the 7 paired ID/OOD transfer settings.}\label{tab:fig5-ood-robustness-summary}
\centering
\tablebodyfont
\fontsize{7.0bp}{8.2bp}\selectfont
\setlength{\tabcolsep}{0pt}
\renewcommand{\arraystretch}{0.90}
\begin{tabular}{@{}l@{\hspace{0.60em}}c@{\hspace{0.60em}}c@{\hspace{0.60em}}c@{}}
\toprule
\textbf{Model variant} & \multicolumn{1}{c}{\textbf{ID DSC}} & \multicolumn{1}{c}{\textbf{OOD DSC}} & \multicolumn{1}{c}{\textbf{Retained}} \\
\cmidrule(lr){2-2}\cmidrule(lr){3-3}\cmidrule(lr){4-4}
 & \multicolumn{1}{c}{\makecell[c]{\textbf{Mean}\\\textbf{(95\% CI)}\\\textbf{SD}}} & \multicolumn{1}{c}{\makecell[c]{\textbf{Mean}\\\textbf{(95\% CI)}\\\textbf{SD}}} & \multicolumn{1}{c}{\makecell[c]{\textbf{Mean}\\\textbf{(95\% CI)}\\\textbf{SD}}} \\
\midrule
\Name{cnn} & \makecell[c]{0.849\\{\fontsize{5.3}{5.8}\selectfont (0.799--0.890)}\\{\fontsize{5.1}{5.6}\selectfont SD 0.063}} & \makecell[c]{0.749\\{\fontsize{5.3}{5.8}\selectfont (0.665--0.809)}\\{\fontsize{5.1}{5.6}\selectfont SD 0.103}} & \makecell[c]{0.882\\{\fontsize{5.3}{5.8}\selectfont (0.820--0.932)}\\{\fontsize{5.1}{5.6}\selectfont SD 0.083}} \\
\Name{cnn} (scratch) & \makecell[c]{0.824\\{\fontsize{5.3}{5.8}\selectfont (0.767--0.871)}\\{\fontsize{5.1}{5.6}\selectfont SD 0.078}} & \makecell[c]{0.704\\{\fontsize{5.3}{5.8}\selectfont (0.621--0.770)}\\{\fontsize{5.1}{5.6}\selectfont SD 0.106}} & \makecell[c]{0.854\\{\fontsize{5.3}{5.8}\selectfont (0.791--0.912)}\\{\fontsize{5.1}{5.6}\selectfont SD 0.083}} \\
\Name{vit} & \makecell[c]{0.822\\{\fontsize{5.3}{5.8}\selectfont (0.748--0.871)}\\{\fontsize{5.1}{5.6}\selectfont SD 0.088}} & \makecell[c]{0.727\\{\fontsize{5.3}{5.8}\selectfont (0.615--0.800)}\\{\fontsize{5.1}{5.6}\selectfont SD 0.138}} & \makecell[c]{0.885\\{\fontsize{5.3}{5.8}\selectfont (0.804--0.941)}\\{\fontsize{5.1}{5.6}\selectfont SD 0.106}} \\
\Name{vit} (scratch) & \makecell[c]{0.655\\{\fontsize{5.3}{5.8}\selectfont (0.456--0.798)}\\{\fontsize{5.1}{5.6}\selectfont SD 0.269}} & \makecell[c]{0.512\\{\fontsize{5.3}{5.8}\selectfont (0.343--0.646)}\\{\fontsize{5.1}{5.6}\selectfont SD 0.237}} & \makecell[c]{0.782\\{\fontsize{5.3}{5.8}\selectfont (0.710--0.841)}\\{\fontsize{5.1}{5.6}\selectfont SD 0.222}} \\
MERLIN~\protect\cite{Blankemeier2026} & \makecell[c]{0.829\\{\fontsize{5.3}{5.8}\selectfont (0.774--0.870)}\\{\fontsize{5.1}{5.6}\selectfont SD 0.069}} & \makecell[c]{0.705\\{\fontsize{5.3}{5.8}\selectfont (0.594--0.784)}\\{\fontsize{5.1}{5.6}\selectfont SD 0.136}} & \makecell[c]{0.851\\{\fontsize{5.3}{5.8}\selectfont (0.759--0.918)}\\{\fontsize{5.1}{5.6}\selectfont SD 0.122}} \\
nnU-Net Default~\protect\cite{isensee2021nnu} & \makecell[c]{0.827\\{\fontsize{5.3}{5.8}\selectfont (0.779--0.866)}\\{\fontsize{5.1}{5.6}\selectfont SD 0.065}} & \makecell[c]{0.700\\{\fontsize{5.3}{5.8}\selectfont (0.613--0.781)}\\{\fontsize{5.1}{5.6}\selectfont SD 0.122}} & \makecell[c]{0.847\\{\fontsize{5.3}{5.8}\selectfont (0.772--0.913)}\\{\fontsize{5.1}{5.6}\selectfont SD 0.102}} \\
CTFM~\protect\cite{pai2025vision} & \makecell[c]{0.821\\{\fontsize{5.3}{5.8}\selectfont (0.748--0.868)}\\{\fontsize{5.1}{5.6}\selectfont SD 0.090}} & \makecell[c]{0.682\\{\fontsize{5.3}{5.8}\selectfont (0.553--0.773)}\\{\fontsize{5.1}{5.6}\selectfont SD 0.162}} & \makecell[c]{0.831\\{\fontsize{5.3}{5.8}\selectfont (0.724--0.915)}\\{\fontsize{5.1}{5.6}\selectfont SD 0.150}} \\
MISFM~\protect\cite{wang2023mis} & \makecell[c]{0.814\\{\fontsize{5.3}{5.8}\selectfont (0.734--0.867)}\\{\fontsize{5.1}{5.6}\selectfont SD 0.094}} & \makecell[c]{0.667\\{\fontsize{5.3}{5.8}\selectfont (0.539--0.755)}\\{\fontsize{5.1}{5.6}\selectfont SD 0.152}} & \makecell[c]{0.820\\{\fontsize{5.3}{5.8}\selectfont (0.722--0.900)}\\{\fontsize{5.1}{5.6}\selectfont SD 0.134}} \\
VISTA3D~\protect\cite{he2025vista3d} & \makecell[c]{0.807\\{\fontsize{5.3}{5.8}\selectfont (0.719--0.869)}\\{\fontsize{5.1}{5.6}\selectfont SD 0.111}} & \makecell[c]{0.665\\{\fontsize{5.3}{5.8}\selectfont (0.513--0.758)}\\{\fontsize{5.1}{5.6}\selectfont SD 0.188}} & \makecell[c]{0.824\\{\fontsize{5.3}{5.8}\selectfont (0.695--0.903)}\\{\fontsize{5.1}{5.6}\selectfont SD 0.176}} \\
VOCO~\protect\cite{wu2024voco} & \makecell[c]{0.798\\{\fontsize{5.3}{5.8}\selectfont (0.728--0.846)}\\{\fontsize{5.1}{5.6}\selectfont SD 0.086}} & \makecell[c]{0.637\\{\fontsize{5.3}{5.8}\selectfont (0.518--0.730)}\\{\fontsize{5.1}{5.6}\selectfont SD 0.160}} & \makecell[c]{0.799\\{\fontsize{5.3}{5.8}\selectfont (0.693--0.889)}\\{\fontsize{5.1}{5.6}\selectfont SD 0.158}} \\
CURIA~\protect\cite{dancette2025curia} & \makecell[c]{0.737\\{\fontsize{5.3}{5.8}\selectfont (0.644--0.826)}\\{\fontsize{5.1}{5.6}\selectfont SD 0.132}} & \makecell[c]{0.627\\{\fontsize{5.3}{5.8}\selectfont (0.496--0.760)}\\{\fontsize{5.1}{5.6}\selectfont SD 0.193}} & \makecell[c]{0.851\\{\fontsize{5.3}{5.8}\selectfont (0.744--0.936)}\\{\fontsize{5.1}{5.6}\selectfont SD 0.147}} \\
SwinUNETR~\protect\cite{tang2022self} & \makecell[c]{0.772\\{\fontsize{5.3}{5.8}\selectfont (0.670--0.835)}\\{\fontsize{5.1}{5.6}\selectfont SD 0.121}} & \makecell[c]{0.549\\{\fontsize{5.3}{5.8}\selectfont (0.395--0.653)}\\{\fontsize{5.1}{5.6}\selectfont SD 0.193}} & \makecell[c]{0.711\\{\fontsize{5.3}{5.8}\selectfont (0.579--0.816)}\\{\fontsize{5.1}{5.6}\selectfont SD 0.213}} \\
\bottomrule
\end{tabular}
\end{table}
\normalsize

\begin{table}[!htbp]
\caption{Pairwise OOD robustness comparisons corresponding to \cref{tab:fig5-ood-robustness-summary}. Cells report paired mean differences in raw metric units with hierarchical paired bootstrap 95\% CIs. Positive values favor \Name{cnn}. A dagger marks comparisons for which the paired 95\% CI does not lie entirely above zero. The primary support endpoint is OOD DSC.}
\label{tab:fig5-ood-robustness-pairwise-support}
\centering
\tablebodyfont
\fontsize{7.0bp}{8.2bp}\selectfont
\setlength{\tabcolsep}{0pt}
\renewcommand{\arraystretch}{0.90}
\begin{tabular}{@{}l@{\hspace{0.60em}}c@{\hspace{0.60em}}c@{\hspace{0.60em}}c@{}}
\toprule
\textbf{Comparator} & \multicolumn{1}{c}{\textbf{ID DSC}} & \multicolumn{1}{c}{\textbf{OOD DSC}} & \multicolumn{1}{c}{\textbf{Retained}} \\
\cmidrule(lr){2-2}\cmidrule(lr){3-3}\cmidrule(lr){4-4}
 & \multicolumn{1}{c}{\makecell[c]{\textbf{\(\Delta\)}\\\textbf{(95\% CI)}}} & \multicolumn{1}{c}{\makecell[c]{\textbf{\(\Delta\)}\\\textbf{(95\% CI)}}} & \multicolumn{1}{c}{\makecell[c]{\textbf{\(\Delta\)}\\\textbf{(95\% CI)}}} \\
\midrule
\Name{cnn} & Reference & Reference & Reference \\
\Name{cnn} (scratch) & \makecell[c]{0.025\\{\fontsize{5.3}{5.8}\selectfont (0.006 to 0.045)}} & \makecell[c]{0.045\\{\fontsize{5.3}{5.8}\selectfont (0.021 to 0.074)}} & \makecell[c]{0.028\\{\fontsize{5.3}{5.8}\selectfont (-0.007 to 0.056)$^\dagger$}} \\
\Name{vit} & \makecell[c]{0.028\\{\fontsize{5.3}{5.8}\selectfont (0.006 to 0.055)}} & \makecell[c]{0.022\\{\fontsize{5.3}{5.8}\selectfont (-0.005 to 0.055)$^\dagger$}} & \makecell[c]{0.002\\{\fontsize{5.3}{5.8}\selectfont (-0.025 to 0.034)$^\dagger$}} \\
\Name{vit} (scratch) & \makecell[c]{0.194\\{\fontsize{5.3}{5.8}\selectfont (0.073 to 0.350)}} & \makecell[c]{0.237\\{\fontsize{5.3}{5.8}\selectfont (0.132 to 0.365)}} & \makecell[c]{0.165\\{\fontsize{5.3}{5.8}\selectfont (0.065 to 0.287)}} \\
MERLIN~\protect\cite{Blankemeier2026} & \makecell[c]{0.020\\{\fontsize{5.3}{5.8}\selectfont (0.003 to 0.040)}} & \makecell[c]{0.044\\{\fontsize{5.3}{5.8}\selectfont (0.009 to 0.089)}} & \makecell[c]{0.034\\{\fontsize{5.3}{5.8}\selectfont (0.001 to 0.077)}} \\
nnU-Net Default~\protect\cite{isensee2021nnu} & \makecell[c]{0.022\\{\fontsize{5.3}{5.8}\selectfont (-0.001 to 0.056)$^\dagger$}} & \makecell[c]{0.049\\{\fontsize{5.3}{5.8}\selectfont (0.009 to 0.112)}} & \makecell[c]{0.037\\{\fontsize{5.3}{5.8}\selectfont (-0.000 to 0.084)$^\dagger$}} \\
CTFM~\protect\cite{pai2025vision} & \makecell[c]{0.028\\{\fontsize{5.3}{5.8}\selectfont (0.005 to 0.058)}} & \makecell[c]{0.068\\{\fontsize{5.3}{5.8}\selectfont (0.028 to 0.119)}} & \makecell[c]{0.059\\{\fontsize{5.3}{5.8}\selectfont (0.013 to 0.120)}} \\
MISFM~\protect\cite{wang2023mis} & \makecell[c]{0.035\\{\fontsize{5.3}{5.8}\selectfont (0.013 to 0.070)}} & \makecell[c]{0.082\\{\fontsize{5.3}{5.8}\selectfont (0.038 to 0.131)}} & \makecell[c]{0.069\\{\fontsize{5.3}{5.8}\selectfont (0.026 to 0.119)}} \\
VISTA3D~\protect\cite{he2025vista3d} & \makecell[c]{0.042\\{\fontsize{5.3}{5.8}\selectfont (0.008 to 0.089)}} & \makecell[c]{0.084\\{\fontsize{5.3}{5.8}\selectfont (0.029 to 0.164)}} & \makecell[c]{0.072\\{\fontsize{5.3}{5.8}\selectfont (0.018 to 0.154)}} \\
VOCO~\protect\cite{wu2024voco} & \makecell[c]{0.051\\{\fontsize{5.3}{5.8}\selectfont (0.024 to 0.083)}} & \makecell[c]{0.112\\{\fontsize{5.3}{5.8}\selectfont (0.060 to 0.167)}} & \makecell[c]{0.090\\{\fontsize{5.3}{5.8}\selectfont (0.038 to 0.153)}} \\
CURIA~\protect\cite{dancette2025curia} & \makecell[c]{0.112\\{\fontsize{5.3}{5.8}\selectfont (0.039 to 0.202)}} & \makecell[c]{0.122\\{\fontsize{5.3}{5.8}\selectfont (0.040 to 0.231)}} & \makecell[c]{0.045\\{\fontsize{5.3}{5.8}\selectfont (-0.014 to 0.113)$^\dagger$}} \\
SwinUNETR~\protect\cite{tang2022self} & \makecell[c]{0.077\\{\fontsize{5.3}{5.8}\selectfont (0.041 to 0.137)}} & \makecell[c]{0.200\\{\fontsize{5.3}{5.8}\selectfont (0.122 to 0.282)}} & \makecell[c]{0.190\\{\fontsize{5.3}{5.8}\selectfont (0.098 to 0.294)}} \\
\bottomrule
\end{tabular}
\end{table}
\normalsize

\clearpage
\subsubsection{OOD Robustness Dataset-Level Results}
\label{app:figure5-ood-dataset-level-results}

\begingroup
\singlespacing
\tablebodyfont
\setlength{\tabcolsep}{1pt}
\renewcommand{\arraystretch}{0.90}
\setlength{\LTcapwidth}{\linewidth}
\begin{longtable}{@{}l@{\hspace{0.45em}}c@{\hspace{0.45em}}c@{\hspace{0.45em}}c@{\hspace{0.45em}}c@{}}
\caption{Dataset-level Figure 5 OOD robustness scores across 7 paired ID/OOD transfer settings. Values are shown as mean and 95\% CI in raw metric units.}\label{tab:fig5-ood-robustness-dataset}\\
\toprule
\textbf{Model variant} & \multicolumn{2}{c}{\textbf{DSC}} & \multicolumn{2}{c}{\textbf{NSD}} \\
\cmidrule(lr){2-3}\cmidrule(lr){4-5}
 & \multicolumn{1}{c}{\textbf{ID}} & \multicolumn{1}{c}{\textbf{OOD}} & \multicolumn{1}{c}{\textbf{ID}} & \multicolumn{1}{c}{\textbf{OOD}} \\
\midrule
\endfirsthead
\caption[]{Dataset-level Figure 5 OOD robustness scores (continued).}\\
\toprule
\textbf{Model variant} & \multicolumn{2}{c}{\textbf{DSC}} & \multicolumn{2}{c}{\textbf{NSD}} \\
\cmidrule(lr){2-3}\cmidrule(lr){4-5}
 & \multicolumn{1}{c}{\textbf{ID}} & \multicolumn{1}{c}{\textbf{OOD}} & \multicolumn{1}{c}{\textbf{ID}} & \multicolumn{1}{c}{\textbf{OOD}} \\
\midrule
\endhead
\midrule
\multicolumn{5}{r}{\footnotesize Continued on next page} \\
\endfoot
\bottomrule
\endlastfoot
\multicolumn{5}{@{}l}{\textbf{MMs (Held-out center)}}\\*
\Name{cnn} & 0.882 (0.869--0.894) & 0.833 (0.823--0.842) & 0.764 (0.745--0.781) & 0.642 (0.625--0.658) \\
\Name{cnn} (scratch) & 0.876 (0.862--0.888) & 0.796 (0.783--0.810) & 0.749 (0.728--0.769) & 0.587 (0.568--0.607) \\
\Name{vit} & 0.848 (0.818--0.870) & 0.823 (0.809--0.834) & 0.710 (0.681--0.732) & 0.635 (0.617--0.653) \\
\Name{vit} (scratch) & 0.789 (0.746--0.821) & 0.581 (0.544--0.616) & 0.627 (0.591--0.657) & 0.405 (0.376--0.433) \\
MERLIN~\protect\cite{Blankemeier2026} & 0.867 (0.853--0.880) & 0.827 (0.817--0.837) & 0.738 (0.720--0.756) & 0.638 (0.620--0.658) \\
nnU-Net Default~\protect\cite{isensee2021nnu} & 0.877 (0.865--0.889) & 0.801 (0.789--0.813) & 0.751 (0.733--0.768) & 0.590 (0.574--0.607) \\
CTFM~\protect\cite{pai2025vision} & 0.842 (0.817--0.861) & 0.805 (0.792--0.817) & 0.702 (0.675--0.725) & 0.608 (0.592--0.626) \\
MISFM~\protect\cite{wang2023mis} & 0.846 (0.831--0.860) & 0.784 (0.769--0.799) & 0.705 (0.686--0.723) & 0.592 (0.573--0.610) \\
VISTA3D~\protect\cite{he2025vista3d} & 0.825 (0.796--0.847) & 0.761 (0.746--0.776) & 0.675 (0.649--0.699) & 0.559 (0.542--0.577) \\
VOCO~\protect\cite{wu2024voco} & 0.823 (0.794--0.847) & 0.758 (0.741--0.773) & 0.678 (0.651--0.702) & 0.562 (0.543--0.582) \\
CURIA~\protect\cite{dancette2025curia} & 0.875 (0.855--0.891) & 0.836 (0.826--0.845) & 0.748 (0.725--0.768) & 0.645 (0.629--0.660) \\
SwinUNETR~\protect\cite{tang2022self} & 0.823 (0.796--0.844) & 0.629 (0.598--0.659) & 0.675 (0.649--0.699) & 0.460 (0.432--0.485) \\
\addlinespace[0.08em]
\multicolumn{5}{@{}l}{\textbf{MMs $\to$ ACDC (Cross-dataset)}}\\*
\Name{cnn} & 0.882 (0.869--0.894) & 0.825 (0.810--0.838) & 0.764 (0.745--0.782) & 0.647 (0.630--0.663) \\
\Name{cnn} (scratch) & 0.876 (0.863--0.888) & 0.803 (0.790--0.815) & 0.749 (0.729--0.768) & 0.613 (0.597--0.630) \\
\Name{vit} & 0.848 (0.818--0.870) & 0.779 (0.764--0.793) & 0.710 (0.680--0.734) & 0.591 (0.575--0.607) \\
\Name{vit} (scratch) & 0.789 (0.745--0.825) & 0.599 (0.571--0.626) & 0.627 (0.594--0.657) & 0.422 (0.397--0.446) \\
MERLIN~\protect\cite{Blankemeier2026} & 0.867 (0.853--0.881) & 0.792 (0.774--0.806) & 0.738 (0.720--0.756) & 0.608 (0.589--0.626) \\
nnU-Net Default~\protect\cite{isensee2021nnu} & 0.877 (0.864--0.888) & 0.790 (0.774--0.803) & 0.751 (0.731--0.768) & 0.597 (0.579--0.615) \\
CTFM~\protect\cite{pai2025vision} & 0.842 (0.817--0.861) & 0.748 (0.732--0.763) & 0.702 (0.677--0.725) & 0.566 (0.548--0.581) \\
MISFM~\protect\cite{wang2023mis} & 0.846 (0.831--0.861) & 0.712 (0.692--0.731) & 0.705 (0.688--0.723) & 0.528 (0.510--0.547) \\
VISTA3D~\protect\cite{he2025vista3d} & 0.825 (0.798--0.847) & 0.708 (0.687--0.726) & 0.675 (0.648--0.701) & 0.525 (0.505--0.542) \\
VOCO~\protect\cite{wu2024voco} & 0.823 (0.793--0.849) & 0.698 (0.678--0.715) & 0.678 (0.650--0.703) & 0.519 (0.500--0.536) \\
CURIA~\protect\cite{dancette2025curia} & 0.875 (0.854--0.891) & 0.838 (0.829--0.847) & 0.748 (0.724--0.768) & 0.658 (0.644--0.671) \\
SwinUNETR~\protect\cite{tang2022self} & 0.823 (0.795--0.846) & 0.590 (0.561--0.618) & 0.675 (0.650--0.700) & 0.428 (0.405--0.452) \\
\addlinespace[0.08em]
\multicolumn{5}{@{}l}{\textbf{WHS (Held-out center)}}\\*
\Name{cnn} & 0.894 (0.878--0.910) & 0.772 (0.729--0.813) & 0.712 (0.672--0.750) & 0.581 (0.521--0.649) \\
\Name{cnn} (scratch) & 0.885 (0.865--0.902) & 0.696 (0.652--0.745) & 0.687 (0.644--0.726) & 0.493 (0.433--0.557) \\
\Name{vit} & 0.898 (0.882--0.912) & 0.807 (0.767--0.840) & 0.722 (0.682--0.758) & 0.626 (0.567--0.682) \\
\Name{vit} (scratch) & 0.886 (0.866--0.904) & 0.758 (0.698--0.811) & 0.700 (0.655--0.739) & 0.570 (0.507--0.637) \\
MERLIN~\protect\cite{Blankemeier2026} & 0.889 (0.868--0.909) & 0.738 (0.669--0.798) & 0.713 (0.667--0.758) & 0.577 (0.501--0.648) \\
nnU-Net Default~\protect\cite{isensee2021nnu} & 0.891 (0.875--0.907) & 0.753 (0.701--0.804) & 0.711 (0.672--0.752) & 0.565 (0.492--0.636) \\
CTFM~\protect\cite{pai2025vision} & 0.895 (0.878--0.910) & 0.682 (0.603--0.760) & 0.718 (0.671--0.759) & 0.549 (0.465--0.632) \\
MISFM~\protect\cite{wang2023mis} & 0.869 (0.832--0.903) & 0.626 (0.556--0.696) & 0.695 (0.643--0.742) & 0.474 (0.401--0.544) \\
VISTA3D~\protect\cite{he2025vista3d} & 0.887 (0.861--0.910) & 0.699 (0.624--0.762) & 0.709 (0.664--0.752) & 0.533 (0.459--0.610) \\
VOCO~\protect\cite{wu2024voco} & 0.873 (0.836--0.905) & 0.601 (0.525--0.675) & 0.696 (0.646--0.744) & 0.464 (0.378--0.545) \\
CURIA~\protect\cite{dancette2025curia} & 0.838 (0.796--0.872) & 0.648 (0.597--0.699) & 0.599 (0.552--0.647) & 0.437 (0.376--0.499) \\
SwinUNETR~\protect\cite{tang2022self} & 0.858 (0.819--0.891) & 0.572 (0.504--0.629) & 0.655 (0.605--0.704) & 0.387 (0.330--0.435) \\
\addlinespace[0.08em]
\multicolumn{5}{@{}l}{\textbf{BraTS-GLI $\to$ Africa (Geographic)}}\\*
\Name{cnn} & 0.809 (0.798--0.821) & 0.716 (0.663--0.767) & 0.865 (0.854--0.875) & 0.731 (0.675--0.786) \\
\Name{cnn} (scratch) & 0.777 (0.765--0.790) & 0.690 (0.634--0.746) & 0.825 (0.812--0.837) & 0.690 (0.629--0.746) \\
\Name{vit} & 0.777 (0.767--0.788) & 0.705 (0.652--0.749) & 0.851 (0.840--0.862) & 0.725 (0.672--0.776) \\
\Name{vit} (scratch) & 0.726 (0.714--0.739) & 0.640 (0.591--0.688) & 0.796 (0.783--0.809) & 0.646 (0.593--0.698) \\
MERLIN~\protect\cite{Blankemeier2026} & 0.795 (0.783--0.807) & 0.715 (0.659--0.765) & 0.848 (0.835--0.860) & 0.720 (0.662--0.774) \\
nnU-Net Default~\protect\cite{isensee2021nnu} & 0.804 (0.793--0.815) & 0.714 (0.664--0.759) & 0.858 (0.847--0.869) & 0.727 (0.672--0.775) \\
CTFM~\protect\cite{pai2025vision} & 0.798 (0.785--0.809) & 0.688 (0.632--0.737) & 0.846 (0.835--0.859) & 0.700 (0.641--0.753) \\
MISFM~\protect\cite{wang2023mis} & 0.795 (0.783--0.807) & 0.692 (0.638--0.750) & 0.842 (0.830--0.854) & 0.695 (0.631--0.753) \\
VISTA3D~\protect\cite{he2025vista3d} & 0.799 (0.788--0.812) & 0.702 (0.645--0.756) & 0.849 (0.837--0.861) & 0.710 (0.651--0.769) \\
VOCO~\protect\cite{wu2024voco} & 0.794 (0.782--0.806) & 0.688 (0.638--0.739) & 0.843 (0.831--0.854) & 0.706 (0.651--0.758) \\
CURIA~\protect\cite{dancette2025curia} & 0.660 (0.648--0.674) & 0.603 (0.552--0.648) & 0.700 (0.687--0.713) & 0.570 (0.515--0.616) \\
SwinUNETR~\protect\cite{tang2022self} & 0.785 (0.771--0.797) & 0.674 (0.627--0.716) & 0.833 (0.821--0.845) & 0.691 (0.641--0.741) \\
\addlinespace[0.08em]
\multicolumn{5}{@{}l}{\textbf{Pancreas-T1 (Held-out centers)}}\\*
\Name{cnn} & 0.821 (0.798--0.842) & 0.793 (0.772--0.814) & 0.659 (0.636--0.679) & 0.561 (0.537--0.582) \\
\Name{cnn} (scratch) & 0.822 (0.803--0.838) & 0.783 (0.752--0.807) & 0.649 (0.629--0.667) & 0.544 (0.520--0.568) \\
\Name{vit} & 0.826 (0.812--0.839) & 0.791 (0.772--0.810) & 0.636 (0.619--0.651) & 0.547 (0.527--0.567) \\
\Name{vit} (scratch) & 0.527 (0.499--0.553) & 0.433 (0.387--0.476) & 0.294 (0.277--0.309) & 0.226 (0.204--0.247) \\
MERLIN~\protect\cite{Blankemeier2026} & 0.836 (0.822--0.850) & 0.799 (0.779--0.818) & 0.666 (0.649--0.682) & 0.563 (0.542--0.585) \\
nnU-Net Default~\protect\cite{isensee2021nnu} & 0.834 (0.818--0.848) & 0.790 (0.769--0.810) & 0.665 (0.647--0.682) & 0.551 (0.529--0.574) \\
CTFM~\protect\cite{pai2025vision} & 0.832 (0.819--0.845) & 0.789 (0.767--0.808) & 0.655 (0.637--0.672) & 0.555 (0.532--0.577) \\
MISFM~\protect\cite{wang2023mis} & 0.822 (0.803--0.837) & 0.779 (0.753--0.803) & 0.649 (0.628--0.668) & 0.548 (0.522--0.573) \\
VISTA3D~\protect\cite{he2025vista3d} & 0.818 (0.798--0.834) & 0.773 (0.742--0.800) & 0.641 (0.620--0.659) & 0.538 (0.509--0.564) \\
VOCO~\protect\cite{wu2024voco} & 0.818 (0.799--0.836) & 0.782 (0.761--0.802) & 0.644 (0.624--0.663) & 0.545 (0.522--0.569) \\
CURIA~\protect\cite{dancette2025curia} & 0.742 (0.723--0.760) & 0.712 (0.677--0.740) & 0.525 (0.508--0.544) & 0.449 (0.422--0.475) \\
SwinUNETR~\protect\cite{tang2022self} & 0.776 (0.753--0.797) & 0.713 (0.667--0.751) & 0.593 (0.572--0.615) & 0.479 (0.447--0.511) \\
\addlinespace[0.08em]
\multicolumn{5}{@{}l}{\textbf{CTSpine1K $\to$ Spine-Mets (Cross-dataset)}}\\*
\Name{cnn} & 0.918 (0.903--0.932) & 0.772 (0.714--0.824) & 0.930 (0.916--0.944) & 0.730 (0.674--0.780) \\
\Name{cnn} (scratch) & 0.862 (0.844--0.879) & 0.660 (0.616--0.698) & 0.865 (0.846--0.883) & 0.607 (0.565--0.652) \\
\Name{vit} & 0.907 (0.893--0.919) & 0.758 (0.694--0.818) & 0.928 (0.914--0.941) & 0.725 (0.660--0.784) \\
\Name{vit} (scratch) & 0.769 (0.751--0.785) & 0.549 (0.502--0.591) & 0.724 (0.708--0.741) & 0.466 (0.423--0.504) \\
MERLIN~\protect\cite{Blankemeier2026} & 0.861 (0.845--0.878) & 0.630 (0.578--0.678) & 0.874 (0.855--0.891) & 0.596 (0.541--0.643) \\
nnU-Net Default~\protect\cite{isensee2021nnu} & 0.800 (0.780--0.818) & 0.541 (0.498--0.583) & 0.789 (0.768--0.809) & 0.484 (0.443--0.524) \\
CTFM~\protect\cite{pai2025vision} & 0.902 (0.887--0.914) & 0.729 (0.666--0.778) & 0.915 (0.900--0.929) & 0.688 (0.628--0.737) \\
MISFM~\protect\cite{wang2023mis} & 0.901 (0.887--0.915) & 0.730 (0.680--0.774) & 0.910 (0.895--0.924) & 0.683 (0.634--0.726) \\
VISTA3D~\protect\cite{he2025vista3d} & 0.919 (0.905--0.932) & 0.767 (0.707--0.820) & 0.930 (0.916--0.944) & 0.727 (0.671--0.778) \\
VOCO~\protect\cite{wu2024voco} & 0.843 (0.827--0.859) & 0.630 (0.589--0.672) & 0.840 (0.823--0.857) & 0.568 (0.523--0.608) \\
CURIA~\protect\cite{dancette2025curia} & 0.546 (0.529--0.563) & 0.338 (0.291--0.384) & 0.438 (0.426--0.451) & 0.258 (0.223--0.296) \\
SwinUNETR~\protect\cite{tang2022self} & 0.837 (0.820--0.854) & 0.533 (0.478--0.590) & 0.827 (0.808--0.844) & 0.475 (0.424--0.524) \\
\addlinespace[0.08em]
\multicolumn{5}{@{}l}{\textbf{Bladder cancer MRI (Held-out center)}}\\*
\Name{cnn} & 0.737 (0.682--0.792) & 0.532 (0.447--0.616) & 0.743 (0.690--0.795) & 0.506 (0.422--0.587) \\
\Name{cnn} (scratch) & 0.671 (0.606--0.732) & 0.502 (0.410--0.590) & 0.671 (0.602--0.729) & 0.472 (0.385--0.557) \\
\Name{vit} & 0.647 (0.581--0.708) & 0.426 (0.323--0.519) & 0.641 (0.579--0.705) & 0.408 (0.315--0.501) \\
\Name{vit} (scratch) & 0.099 (0.070--0.131) & 0.023 (0.015--0.032) & 0.054 (0.043--0.068) & 0.019 (0.014--0.024) \\
MERLIN~\protect\cite{Blankemeier2026} & 0.688 (0.629--0.743) & 0.436 (0.350--0.527) & 0.684 (0.621--0.747) & 0.424 (0.341--0.516) \\
nnU-Net Default~\protect\cite{isensee2021nnu} & 0.705 (0.646--0.759) & 0.512 (0.435--0.586) & 0.701 (0.643--0.755) & 0.498 (0.420--0.575) \\
CTFM~\protect\cite{pai2025vision} & 0.633 (0.565--0.697) & 0.330 (0.241--0.423) & 0.620 (0.556--0.679) & 0.298 (0.217--0.379) \\
MISFM~\protect\cite{wang2023mis} & 0.615 (0.541--0.679) & 0.344 (0.267--0.422) & 0.613 (0.544--0.674) & 0.325 (0.257--0.402) \\
VISTA3D~\protect\cite{he2025vista3d} & 0.574 (0.504--0.639) & 0.244 (0.175--0.321) & 0.555 (0.487--0.630) & 0.225 (0.161--0.295) \\
VOCO~\protect\cite{wu2024voco} & 0.611 (0.542--0.681) & 0.306 (0.226--0.393) & 0.590 (0.519--0.654) & 0.289 (0.212--0.366) \\
CURIA~\protect\cite{dancette2025curia} & 0.622 (0.559--0.680) & 0.416 (0.342--0.494) & 0.596 (0.540--0.644) & 0.374 (0.301--0.443) \\
SwinUNETR~\protect\cite{tang2022self} & 0.505 (0.433--0.575) & 0.133 (0.077--0.194) & 0.496 (0.432--0.563) & 0.126 (0.080--0.179) \\
\addlinespace[0.08em]
\end{longtable}
\normalsize
\endgroup

\clearpage
\subsubsection{External Partner Validation Summary}
\label{app:figure5-external-validation-summary}

\begin{table}[!htbp]
\caption{Figure 5 external partner validation summary across grouped fine-tuning fractions. Values are dataset-equal means in raw metric units with 95\% CIs; SD reports variability across external datasets. Delta is \Name{cnn} minus scratch with hierarchical paired external dataset/case bootstrap 95\% CIs. A dagger marks delta comparisons for which the paired 95\% CI does not lie entirely above zero. Number of contributing external datasets by fraction: 0.1: 25, 0.2: 26, 0.5: 28, 1.0: 28.}\label{tab:fig5-external-validation-summary}
\centering
\tablebodyfont
\fontsize{7.0bp}{8.2bp}\selectfont
\setlength{\tabcolsep}{0pt}
\renewcommand{\arraystretch}{0.90}
\begin{tabular}{@{}c@{\hspace{0.72em}}c@{\hspace{0.72em}}c@{\hspace{0.72em}}c@{\hspace{0.72em}}c@{}}
\toprule
\textbf{Fine-tuning fraction} & \textbf{Metric} & \multicolumn{1}{c}{\textbf{\Name{cnn}}} & \multicolumn{1}{c}{\textbf{Scratch}} & \multicolumn{1}{c}{\makecell[c]{\textbf{Delta}\\\textbf{vs scratch}}} \\
\cmidrule(lr){3-3}\cmidrule(lr){4-4}\cmidrule(lr){5-5}
 & & \multicolumn{1}{c}{\makecell[c]{\textbf{Mean}\\\textbf{(95\% CI)}\\\textbf{SD}}} & \multicolumn{1}{c}{\makecell[c]{\textbf{Mean}\\\textbf{(95\% CI)}\\\textbf{SD}}} & \multicolumn{1}{c}{\makecell[c]{\textbf{\(\Delta\)}\\\textbf{(95\% CI)}\\\textbf{SD}}} \\
\midrule
0.1 & DSC & \makecell[c]{0.637\\{\fontsize{5.3}{5.8}\selectfont (0.526--0.736)}\\{\fontsize{5.1}{5.6}\selectfont SD 0.272}} & \makecell[c]{0.595\\{\fontsize{5.3}{5.8}\selectfont (0.483--0.696)}\\{\fontsize{5.1}{5.6}\selectfont SD 0.269}} & \makecell[c]{0.042\\{\fontsize{5.3}{5.8}\selectfont (0.019--0.069)}\\{\fontsize{5.1}{5.6}\selectfont SD 0.065}} \\
 & NSD & \makecell[c]{0.603\\{\fontsize{5.3}{5.8}\selectfont (0.499--0.694)}\\{\fontsize{5.1}{5.6}\selectfont SD 0.261}} & \makecell[c]{0.548\\{\fontsize{5.3}{5.8}\selectfont (0.443--0.646)}\\{\fontsize{5.1}{5.6}\selectfont SD 0.259}} & \makecell[c]{0.055\\{\fontsize{5.3}{5.8}\selectfont (0.019--0.095)}\\{\fontsize{5.1}{5.6}\selectfont SD 0.099}} \\
0.2 & DSC & \makecell[c]{0.725\\{\fontsize{5.3}{5.8}\selectfont (0.629--0.807)}\\{\fontsize{5.1}{5.6}\selectfont SD 0.233}} & \makecell[c]{0.707\\{\fontsize{5.3}{5.8}\selectfont (0.612--0.789)}\\{\fontsize{5.1}{5.6}\selectfont SD 0.232}} & \makecell[c]{0.019\\{\fontsize{5.3}{5.8}\selectfont (0.003--0.038)}\\{\fontsize{5.1}{5.6}\selectfont SD 0.044}} \\
 & NSD & \makecell[c]{0.704\\{\fontsize{5.3}{5.8}\selectfont (0.611--0.780)}\\{\fontsize{5.1}{5.6}\selectfont SD 0.228}} & \makecell[c]{0.674\\{\fontsize{5.3}{5.8}\selectfont (0.579--0.755)}\\{\fontsize{5.1}{5.6}\selectfont SD 0.232}} & \makecell[c]{0.030\\{\fontsize{5.3}{5.8}\selectfont (0.005--0.055)}\\{\fontsize{5.1}{5.6}\selectfont SD 0.065}} \\
0.5 & DSC & \makecell[c]{0.758\\{\fontsize{5.3}{5.8}\selectfont (0.673--0.824)}\\{\fontsize{5.1}{5.6}\selectfont SD 0.212}} & \makecell[c]{0.749\\{\fontsize{5.3}{5.8}\selectfont (0.662--0.818)}\\{\fontsize{5.1}{5.6}\selectfont SD 0.216}} & \makecell[c]{0.009\\{\fontsize{5.3}{5.8}\selectfont (-0.005--0.023)$^\dagger$}\\{\fontsize{5.1}{5.6}\selectfont SD 0.038}} \\
 & NSD & \makecell[c]{0.753\\{\fontsize{5.3}{5.8}\selectfont (0.669--0.821)}\\{\fontsize{5.1}{5.6}\selectfont SD 0.220}} & \makecell[c]{0.749\\{\fontsize{5.3}{5.8}\selectfont (0.664--0.824)}\\{\fontsize{5.1}{5.6}\selectfont SD 0.224}} & \makecell[c]{0.004\\{\fontsize{5.3}{5.8}\selectfont (-0.021--0.023)$^\dagger$}\\{\fontsize{5.1}{5.6}\selectfont SD 0.059}} \\
1.0 & DSC & \makecell[c]{0.777\\{\fontsize{5.3}{5.8}\selectfont (0.700--0.849)}\\{\fontsize{5.1}{5.6}\selectfont SD 0.218}} & \makecell[c]{0.765\\{\fontsize{5.3}{5.8}\selectfont (0.679--0.837)}\\{\fontsize{5.1}{5.6}\selectfont SD 0.211}} & \makecell[c]{0.011\\{\fontsize{5.3}{5.8}\selectfont (0.001--0.025)}\\{\fontsize{5.1}{5.6}\selectfont SD 0.031}} \\
 & NSD & \makecell[c]{0.782\\{\fontsize{5.3}{5.8}\selectfont (0.703--0.859)}\\{\fontsize{5.1}{5.6}\selectfont SD 0.228}} & \makecell[c]{0.765\\{\fontsize{5.3}{5.8}\selectfont (0.680--0.836)}\\{\fontsize{5.1}{5.6}\selectfont SD 0.220}} & \makecell[c]{0.017\\{\fontsize{5.3}{5.8}\selectfont (0.000--0.037)$^\dagger$}\\{\fontsize{5.1}{5.6}\selectfont SD 0.051}} \\
\bottomrule
\end{tabular}
\end{table}
\normalsize

\begingroup
\singlespacing
\tablebodyfont
\setlength{\tabcolsep}{2pt}
\renewcommand{\arraystretch}{0.90}
\setlength{\LTcapwidth}{\linewidth}
\begin{longtable}{@{}c@{\hspace{0.90em}}c@{\hspace{0.90em}}c@{\hspace{0.90em}}c@{}}
\caption{Dataset-level Figure 5 external partner validation scores across 28 external datasets. Values are shown as mean and 95\% CI in raw metric units.}\label{tab:fig5-external-validation-dataset}\\
\toprule
\textbf{Fine-tuning fraction} & \textbf{Metric} & \multicolumn{1}{c}{\textbf{\Name{cnn}}} & \multicolumn{1}{c}{\textbf{Scratch}} \\
\cmidrule(lr){3-3}\cmidrule(lr){4-4}
\midrule
\endfirsthead
\caption[]{Dataset-level Figure 5 external partner validation scores (continued).}\\
\toprule
\textbf{Fine-tuning fraction} & \textbf{Metric} & \multicolumn{1}{c}{\textbf{\Name{cnn}}} & \multicolumn{1}{c}{\textbf{Scratch}} \\
\cmidrule(lr){3-3}\cmidrule(lr){4-4}
\midrule
\endhead
\midrule
\multicolumn{4}{r}{\footnotesize Continued on next page} \\
\endfoot
\bottomrule
\endlastfoot
\multicolumn{4}{@{}l}{\textbf{Siemens}}\\*
\multicolumn{4}{@{}l}{\hspace{0.75em}Siemens Pulmonary}\\*
0.1 & DSC & 0.811 (0.798--0.824) & 0.678 (0.655--0.698) \\*
 & NSD & 0.756 (0.728--0.781) & 0.510 (0.485--0.534) \\
0.2 & DSC & 0.840 (0.826--0.853) & 0.837 (0.825--0.847) \\*
 & NSD & 0.830 (0.806--0.855) & 0.820 (0.796--0.844) \\
0.5 & DSC & 0.854 (0.842--0.865) & 0.857 (0.848--0.867) \\*
 & NSD & 0.858 (0.837--0.880) & 0.859 (0.841--0.876) \\
1.0 & DSC & 0.863 (0.852--0.874) & 0.860 (0.849--0.871) \\*
 & NSD & 0.869 (0.847--0.888) & 0.862 (0.843--0.881) \\
\addlinespace[0.06em]
\multicolumn{4}{@{}l}{\hspace{0.75em}Siemens BP}\\*
0.1 & DSC & 0.606 (0.587--0.625) & 0.511 (0.487--0.533) \\*
 & NSD & 0.618 (0.598--0.638) & 0.514 (0.493--0.536) \\
0.2 & DSC & 0.593 (0.572--0.616) & 0.558 (0.539--0.577) \\*
 & NSD & 0.631 (0.607--0.655) & 0.605 (0.584--0.624) \\
0.5 & DSC & 0.646 (0.630--0.662) & 0.627 (0.607--0.645) \\*
 & NSD & 0.679 (0.659--0.697) & 0.670 (0.652--0.690) \\
1.0 & DSC & 0.689 (0.673--0.704) & 0.632 (0.614--0.651) \\*
 & NSD & 0.736 (0.720--0.752) & 0.684 (0.666--0.703) \\
\addlinespace[0.06em]
\multicolumn{4}{@{}l}{\hspace{0.75em}Siemens Bowel Bag}\\*
0.1 & DSC & 0.915 (0.890--0.937) & 0.913 (0.871--0.941) \\*
 & NSD & 0.711 (0.657--0.760) & 0.715 (0.659--0.770) \\
0.2 & DSC & 0.931 (0.896--0.956) & 0.919 (0.887--0.945) \\*
 & NSD & 0.803 (0.747--0.854) & 0.753 (0.695--0.808) \\
0.5 & DSC & 0.938 (0.903--0.961) & 0.946 (0.927--0.960) \\*
 & NSD & 0.825 (0.759--0.878) & 0.823 (0.775--0.865) \\
1.0 & DSC & 0.954 (0.930--0.970) & 0.936 (0.910--0.958) \\*
 & NSD & 0.862 (0.822--0.898) & 0.795 (0.739--0.851) \\
\addlinespace[0.06em]
\multicolumn{4}{@{}l}{\hspace{0.75em}Siemens Brain}\\*
0.1 & DSC & 0.938 (0.904--0.966) & 0.900 (0.841--0.944) \\*
 & NSD & 0.846 (0.775--0.912) & 0.765 (0.676--0.838) \\
0.2 & DSC & 0.888 (0.851--0.929) & 0.818 (0.764--0.870) \\*
 & NSD & 0.704 (0.626--0.784) & 0.547 (0.466--0.659) \\
0.5 & DSC & 0.887 (0.836--0.934) & 0.969 (0.945--0.984) \\*
 & NSD & 0.675 (0.575--0.784) & 0.911 (0.850--0.960) \\
1.0 & DSC & 0.977 (0.971--0.984) & 0.929 (0.889--0.962) \\*
 & NSD & 0.933 (0.902--0.963) & 0.812 (0.736--0.888) \\
\addlinespace[0.06em]
\multicolumn{4}{@{}l}{\hspace{0.75em}Siemens Chest Wall LR}\\*
0.1 & DSC & 0.746 (0.715--0.778) & 0.666 (0.616--0.720) \\*
 & NSD & 0.607 (0.564--0.651) & 0.516 (0.463--0.575) \\
0.2 & DSC & 0.838 (0.814--0.862) & 0.794 (0.732--0.849) \\*
 & NSD & 0.760 (0.723--0.801) & 0.707 (0.635--0.789) \\
0.5 & DSC & 0.823 (0.786--0.860) & 0.857 (0.814--0.898) \\*
 & NSD & 0.758 (0.711--0.807) & 0.795 (0.743--0.847) \\
1.0 & DSC & 0.900 (0.868--0.928) & 0.870 (0.829--0.910) \\*
 & NSD & 0.880 (0.831--0.924) & 0.838 (0.778--0.896) \\
\addlinespace[0.06em]
\multicolumn{4}{@{}l}{\hspace{0.75em}Siemens Constrictor Musc.}\\*
0.1 & DSC & 0.659 (0.640--0.676) & 0.658 (0.638--0.674) \\*
 & NSD & 0.757 (0.733--0.778) & 0.753 (0.732--0.773) \\
0.2 & DSC & 0.683 (0.665--0.697) & 0.685 (0.661--0.703) \\*
 & NSD & 0.802 (0.780--0.821) & 0.795 (0.769--0.817) \\
0.5 & DSC & 0.718 (0.703--0.733) & 0.707 (0.690--0.722) \\*
 & NSD & 0.840 (0.823--0.855) & 0.824 (0.802--0.841) \\
1.0 & DSC & 0.732 (0.719--0.743) & 0.733 (0.718--0.746) \\*
 & NSD & 0.852 (0.834--0.867) & 0.851 (0.834--0.866) \\
\addlinespace[0.06em]
\multicolumn{4}{@{}l}{\hspace{0.75em}Siemens Esophagus}\\*
0.1 & DSC & 0.627 (0.594--0.661) & 0.628 (0.581--0.667) \\*
 & NSD & 0.639 (0.599--0.683) & 0.688 (0.635--0.737) \\
0.2 & DSC & 0.696 (0.655--0.736) & 0.760 (0.732--0.788) \\*
 & NSD & 0.745 (0.699--0.793) & 0.849 (0.814--0.882) \\
0.5 & DSC & 0.752 (0.726--0.774) & 0.725 (0.690--0.763) \\*
 & NSD & 0.820 (0.794--0.850) & 0.814 (0.774--0.856) \\
1.0 & DSC & 0.746 (0.717--0.773) & 0.777 (0.755--0.801) \\*
 & NSD & 0.793 (0.761--0.828) & 0.872 (0.848--0.895) \\
\addlinespace[0.06em]
\multicolumn{4}{@{}l}{\hspace{0.75em}Siemens Jaws}\\*
0.1 & DSC & 0.839 (0.830--0.846) & 0.764 (0.749--0.779) \\*
 & NSD & 0.738 (0.716--0.758) & 0.655 (0.631--0.681) \\
0.2 & DSC & 0.850 (0.842--0.858) & 0.843 (0.835--0.852) \\*
 & NSD & 0.779 (0.758--0.799) & 0.766 (0.744--0.785) \\
0.5 & DSC & 0.857 (0.847--0.866) & 0.853 (0.841--0.863) \\*
 & NSD & 0.793 (0.771--0.811) & 0.784 (0.765--0.803) \\
1.0 & DSC & 0.871 (0.863--0.878) & 0.864 (0.857--0.872) \\*
 & NSD & 0.809 (0.790--0.826) & 0.801 (0.783--0.820) \\
\addlinespace[0.06em]
\multicolumn{4}{@{}l}{\hspace{0.75em}Siemens Ribs}\\*
0.1 & DSC & 0.554 (0.500--0.601) & 0.421 (0.379--0.462) \\*
 & NSD & 0.625 (0.570--0.671) & 0.501 (0.458--0.538) \\
0.2 & DSC & 0.772 (0.711--0.824) & 0.834 (0.782--0.882) \\*
 & NSD & 0.853 (0.789--0.906) & 0.917 (0.859--0.961) \\
0.5 & DSC & 0.878 (0.839--0.911) & 0.873 (0.807--0.912) \\*
 & NSD & 0.956 (0.915--0.989) & 0.953 (0.893--0.991) \\
1.0 & DSC & 0.893 (0.841--0.923) & 0.903 (0.870--0.922) \\*
 & NSD & 0.971 (0.919--0.997) & 0.982 (0.955--0.997) \\
\addlinespace[0.06em]
\multicolumn{4}{@{}l}{\hspace{0.75em}Siemens Pancreas}\\*
0.1 & DSC & 0.681 (0.649--0.708) & 0.434 (0.404--0.464) \\*
 & NSD & 0.669 (0.640--0.698) & 0.304 (0.284--0.324) \\
0.2 & DSC & 0.715 (0.684--0.744) & 0.666 (0.626--0.698) \\*
 & NSD & 0.725 (0.691--0.754) & 0.657 (0.617--0.691) \\
0.5 & DSC & 0.740 (0.708--0.767) & 0.737 (0.708--0.766) \\*
 & NSD & 0.762 (0.729--0.792) & 0.747 (0.716--0.778) \\
1.0 & DSC & 0.772 (0.749--0.796) & 0.754 (0.722--0.782) \\*
 & NSD & 0.802 (0.774--0.828) & 0.781 (0.748--0.809) \\
\addlinespace[0.06em]
\multicolumn{4}{@{}l}{\hspace{0.75em}Siemens Pelvic}\\*
0.1 & DSC & 0.772 (0.717--0.823) & 0.803 (0.760--0.840) \\*
 & NSD & 0.621 (0.543--0.697) & 0.683 (0.622--0.738) \\
0.2 & DSC & 0.822 (0.776--0.861) & 0.811 (0.762--0.855) \\*
 & NSD & 0.722 (0.651--0.784) & 0.700 (0.632--0.775) \\
0.5 & DSC & 0.848 (0.798--0.886) & 0.722 (0.677--0.770) \\*
 & NSD & 0.767 (0.702--0.826) & 0.636 (0.576--0.701) \\
1.0 & DSC & 0.847 (0.818--0.875) & 0.841 (0.787--0.882) \\*
 & NSD & 0.760 (0.692--0.823) & 0.749 (0.679--0.814) \\
\addlinespace[0.06em]
\multicolumn{4}{@{}l}{\hspace{0.75em}Siemens Penile Bulb}\\*
0.1 & DSC & 0.650 (0.624--0.675) & 0.604 (0.574--0.633) \\*
 & NSD & 0.682 (0.653--0.708) & 0.621 (0.587--0.654) \\
0.2 & DSC & 0.700 (0.677--0.721) & 0.672 (0.650--0.695) \\*
 & NSD & 0.740 (0.712--0.764) & 0.701 (0.673--0.727) \\
0.5 & DSC & 0.702 (0.679--0.723) & 0.682 (0.658--0.704) \\*
 & NSD & 0.751 (0.723--0.777) & 0.722 (0.696--0.749) \\
1.0 & DSC & 0.692 (0.667--0.716) & 0.700 (0.674--0.724) \\*
 & NSD & 0.740 (0.711--0.769) & 0.748 (0.721--0.776) \\
\addlinespace[0.06em]
\multicolumn{4}{@{}l}{\hspace{0.75em}Siemens Spinal Cord}\\*
0.1 & DSC & 0.806 (0.780--0.828) & 0.796 (0.769--0.820) \\*
 & NSD & 0.927 (0.893--0.955) & 0.923 (0.901--0.942) \\
0.2 & DSC & 0.835 (0.816--0.852) & 0.799 (0.769--0.826) \\*
 & NSD & 0.953 (0.936--0.967) & 0.919 (0.889--0.947) \\
0.5 & DSC & 0.837 (0.819--0.852) & 0.781 (0.743--0.815) \\*
 & NSD & 0.953 (0.938--0.969) & 0.907 (0.873--0.938) \\
1.0 & DSC & 0.842 (0.824--0.858) & 0.811 (0.779--0.839) \\*
 & NSD & 0.955 (0.940--0.969) & 0.931 (0.904--0.956) \\
\addlinespace[0.06em]
\multicolumn{4}{@{}l}{\hspace{0.75em}Siemens Sternum}\\*
0.1 & DSC & 0.617 (0.560--0.677) & 0.606 (0.557--0.663) \\*
 & NSD & 0.529 (0.464--0.615) & 0.551 (0.495--0.620) \\
0.2 & DSC & 0.850 (0.799--0.891) & 0.713 (0.648--0.776) \\*
 & NSD & 0.866 (0.819--0.911) & 0.683 (0.608--0.760) \\
0.5 & DSC & 0.884 (0.866--0.902) & 0.883 (0.856--0.905) \\*
 & NSD & 0.873 (0.844--0.905) & 0.901 (0.868--0.935) \\
1.0 & DSC & 0.926 (0.912--0.937) & 0.786 (0.746--0.828) \\*
 & NSD & 0.953 (0.933--0.972) & 0.741 (0.679--0.797) \\
\addlinespace[0.06em]
\multicolumn{4}{@{}l}{\hspace{0.75em}Siemens Stomach}\\*
0.1 & DSC & 0.828 (0.793--0.858) & 0.841 (0.806--0.873) \\*
 & NSD & 0.748 (0.713--0.782) & 0.774 (0.737--0.809) \\
0.2 & DSC & 0.892 (0.861--0.915) & 0.877 (0.845--0.900) \\*
 & NSD & 0.870 (0.842--0.896) & 0.834 (0.800--0.863) \\
0.5 & DSC & 0.908 (0.880--0.929) & 0.904 (0.879--0.923) \\*
 & NSD & 0.899 (0.871--0.921) & 0.884 (0.856--0.907) \\
1.0 & DSC & 0.914 (0.887--0.934) & 0.912 (0.886--0.930) \\*
 & NSD & 0.909 (0.880--0.931) & 0.904 (0.875--0.925) \\
\addlinespace[0.06em]
\multicolumn{4}{@{}l}{\hspace{0.75em}Siemens Thyroid}\\*
0.1 & DSC & 0.767 (0.734--0.796) & 0.698 (0.662--0.731) \\*
 & NSD & 0.736 (0.697--0.772) & 0.606 (0.570--0.640) \\
0.2 & DSC & 0.780 (0.746--0.812) & 0.760 (0.716--0.795) \\*
 & NSD & 0.768 (0.727--0.807) & 0.739 (0.689--0.781) \\
0.5 & DSC & 0.798 (0.765--0.827) & 0.772 (0.721--0.811) \\*
 & NSD & 0.793 (0.752--0.831) & 0.767 (0.716--0.810) \\
1.0 & DSC & 0.810 (0.781--0.837) & 0.805 (0.768--0.836) \\*
 & NSD & 0.804 (0.765--0.843) & 0.804 (0.764--0.840) \\
\addlinespace[0.06em]
\multicolumn{4}{@{}l}{\hspace{0.75em}Siemens Trachea Bronchus}\\*
0.1 & DSC & 0.850 (0.832--0.866) & 0.832 (0.816--0.847) \\*
 & NSD & 0.900 (0.880--0.918) & 0.881 (0.865--0.895) \\
0.2 & DSC & 0.878 (0.867--0.888) & 0.873 (0.862--0.883) \\*
 & NSD & 0.928 (0.915--0.939) & 0.914 (0.902--0.926) \\
0.5 & DSC & 0.900 (0.891--0.909) & 0.895 (0.886--0.904) \\*
 & NSD & 0.943 (0.931--0.954) & 0.937 (0.925--0.946) \\
1.0 & DSC & 0.905 (0.896--0.912) & 0.901 (0.892--0.909) \\*
 & NSD & 0.944 (0.934--0.953) & 0.940 (0.928--0.952) \\
\addlinespace[0.06em]
\multicolumn{4}{@{}l}{\hspace{0.75em}Siemens Uterus}\\*
0.1 & DSC & 0.799 (0.771--0.826) & 0.717 (0.687--0.745) \\*
 & NSD & 0.684 (0.649--0.716) & 0.521 (0.485--0.554) \\
0.2 & DSC & 0.813 (0.778--0.843) & 0.799 (0.767--0.830) \\*
 & NSD & 0.728 (0.693--0.759) & 0.688 (0.653--0.722) \\
0.5 & DSC & 0.817 (0.780--0.848) & 0.812 (0.784--0.836) \\*
 & NSD & 0.739 (0.702--0.774) & 0.715 (0.682--0.747) \\
1.0 & DSC & 0.848 (0.825--0.866) & 0.839 (0.810--0.861) \\*
 & NSD & 0.777 (0.750--0.803) & 0.763 (0.732--0.789) \\
\addlinespace[0.06em]
\multicolumn{4}{@{}l}{\hspace{0.75em}Siemens Inf. Venacava}\\*
0.1 & DSC & 0.000 (0.000--0.000) & 0.000 (0.000--0.000) \\*
 & NSD & 0.000 (0.000--0.000) & 0.000 (0.000--0.000) \\
0.2 & DSC & 0.722 (0.674--0.762) & 0.616 (0.554--0.669) \\*
 & NSD & 0.675 (0.617--0.732) & 0.495 (0.429--0.558) \\
0.5 & DSC & 0.790 (0.747--0.823) & 0.742 (0.695--0.780) \\*
 & NSD & 0.803 (0.741--0.849) & 0.729 (0.670--0.787) \\
1.0 & DSC & 0.799 (0.766--0.831) & 0.797 (0.754--0.829) \\*
 & NSD & 0.813 (0.771--0.852) & 0.820 (0.773--0.868) \\
\addlinespace[0.06em]
\multicolumn{4}{@{}l}{\hspace{0.75em}Siemens Sup. Venacava}\\*
0.1 & DSC & 0.574 (0.516--0.624) & 0.623 (0.599--0.648) \\*
 & NSD & 0.495 (0.444--0.539) & 0.544 (0.513--0.572) \\
0.2 & DSC & 0.713 (0.674--0.747) & 0.722 (0.681--0.757) \\*
 & NSD & 0.719 (0.675--0.758) & 0.743 (0.699--0.783) \\
0.5 & DSC & 0.784 (0.746--0.817) & 0.805 (0.771--0.833) \\*
 & NSD & 0.849 (0.809--0.889) & 0.868 (0.827--0.901) \\
1.0 & DSC & 0.815 (0.778--0.845) & 0.819 (0.785--0.848) \\*
 & NSD & 0.886 (0.843--0.919) & 0.890 (0.849--0.922) \\
\addlinespace[0.06em]
\multicolumn{4}{@{}l}{\hspace{0.75em}Siemens Whole Bowel}\\*
0.1 & DSC & 0.636 (0.611--0.659) & 0.568 (0.542--0.592) \\*
 & NSD & 0.590 (0.561--0.618) & 0.509 (0.483--0.535) \\
0.2 & DSC & 0.688 (0.663--0.712) & 0.684 (0.658--0.708) \\*
 & NSD & 0.662 (0.635--0.691) & 0.654 (0.627--0.680) \\
0.5 & DSC & 0.760 (0.736--0.782) & 0.716 (0.692--0.740) \\*
 & NSD & 0.764 (0.736--0.787) & 0.710 (0.681--0.738) \\
1.0 & DSC & 0.779 (0.758--0.798) & 0.774 (0.752--0.795) \\*
 & NSD & 0.784 (0.761--0.806) & 0.780 (0.755--0.804) \\
\addlinespace[0.18em]
\multicolumn{4}{@{}l}{\textbf{Floy}}\\*
\multicolumn{4}{@{}l}{\hspace{0.75em}Floy Abdomen}\\*
0.1 & DSC & 0.924 (0.885--0.957) & 0.921 (0.885--0.953) \\*
 & NSD & 0.904 (0.857--0.941) & 0.903 (0.866--0.934) \\
0.2 & DSC & 0.950 (0.929--0.967) & 0.953 (0.935--0.967) \\*
 & NSD & 0.934 (0.910--0.953) & 0.938 (0.918--0.952) \\
0.5 & DSC & 0.957 (0.938--0.970) & 0.958 (0.939--0.971) \\*
 & NSD & 0.945 (0.926--0.959) & 0.946 (0.928--0.960) \\
1.0 & DSC & 0.961 (0.947--0.972) & 0.960 (0.946--0.971) \\*
 & NSD & 0.949 (0.933--0.961) & 0.948 (0.933--0.961) \\
\addlinespace[0.06em]
\multicolumn{4}{@{}l}{\hspace{0.75em}Floy Spine lesion (CT)}\\*
0.1 & DSC & 0.051 (0.042--0.060) & 0.065 (0.054--0.076) \\*
 & NSD & 0.047 (0.039--0.056) & 0.058 (0.048--0.068) \\
0.2 & DSC & 0.075 (0.064--0.085) & 0.081 (0.069--0.095) \\*
 & NSD & 0.067 (0.058--0.077) & 0.069 (0.058--0.080) \\
0.5 & DSC & 0.111 (0.098--0.124) & 0.128 (0.115--0.142) \\*
 & NSD & 0.096 (0.086--0.106) & 0.114 (0.103--0.127) \\
1.0 & DSC & 0.113 (0.101--0.125) & 0.120 (0.107--0.133) \\*
 & NSD & 0.103 (0.093--0.113) & 0.106 (0.095--0.118) \\
\addlinespace[0.06em]
\multicolumn{4}{@{}l}{\hspace{0.75em}Floy Spine lesion (MR)}\\*
0.1 & DSC & 0.111 (0.094--0.130) & 0.123 (0.107--0.142) \\*
 & NSD & 0.108 (0.090--0.126) & 0.125 (0.108--0.144) \\
0.2 & DSC & 0.132 (0.116--0.150) & 0.156 (0.135--0.176) \\*
 & NSD & 0.128 (0.111--0.146) & 0.154 (0.134--0.175) \\
0.5 & DSC & 0.163 (0.143--0.183) & 0.186 (0.163--0.210) \\*
 & NSD & 0.156 (0.137--0.176) & 0.176 (0.155--0.199) \\
1.0 & DSC & 0.149 (0.131--0.168) & 0.179 (0.157--0.201) \\*
 & NSD & 0.137 (0.121--0.155) & 0.168 (0.147--0.189) \\
\addlinespace[0.18em]
\multicolumn{4}{@{}l}{\textbf{MDC}}\\*
\multicolumn{4}{@{}l}{\hspace{0.75em}MDC Mouse}\\*
0.2 & DSC & 0.966 (0.963--0.968) & 0.966 (0.964--0.969) \\*
 & NSD & 0.730 (0.711--0.748) & 0.743 (0.727--0.758) \\
0.5 & DSC & 0.968 (0.965--0.970) & 0.968 (0.965--0.970) \\*
 & NSD & 0.751 (0.732--0.767) & 0.757 (0.741--0.772) \\
1.0 & DSC & 0.969 (0.967--0.972) & 0.969 (0.966--0.971) \\*
 & NSD & 0.770 (0.754--0.783) & 0.765 (0.750--0.779) \\
\addlinespace[0.18em]
\multicolumn{4}{@{}l}{\textbf{DZNE}}\\*
\multicolumn{4}{@{}l}{\hspace{0.75em}DZNE Cerebellum}\\*
0.5 & DSC & 0.807 (0.791--0.824) & 0.822 (0.805--0.838) \\*
 & NSD & 0.868 (0.848--0.889) & 0.890 (0.872--0.910) \\
1.0 & DSC & 0.831 (0.813--0.848) & 0.825 (0.807--0.842) \\*
 & NSD & 0.890 (0.867--0.910) & 0.890 (0.871--0.909) \\
\addlinespace[0.06em]
\multicolumn{4}{@{}l}{\hspace{0.75em}DZNE Hypothalamus}\\*
0.5 & DSC & 0.744 (0.719--0.765) & 0.771 (0.742--0.802) \\*
 & NSD & 0.903 (0.893--0.915) & 0.937 (0.927--0.948) \\
1.0 & DSC & 0.780 (0.753--0.805) & 0.775 (0.754--0.796) \\*
 & NSD & 0.943 (0.930--0.957) & 0.940 (0.930--0.954) \\
\addlinespace[0.18em]
\multicolumn{4}{@{}l}{\textbf{Munich}}\\*
\multicolumn{4}{@{}l}{\hspace{0.75em}Munich Sarcoma}\\*
0.1 & DSC & 0.165 (0.140--0.192) & 0.107 (0.090--0.127) \\*
 & NSD & 0.131 (0.110--0.151) & 0.085 (0.068--0.101) \\
0.2 & DSC & 0.240 (0.210--0.271) & 0.177 (0.153--0.204) \\*
 & NSD & 0.191 (0.165--0.218) & 0.134 (0.115--0.155) \\
0.5 & DSC & 0.348 (0.314--0.380) & 0.273 (0.242--0.303) \\*
 & NSD & 0.270 (0.240--0.296) & 0.208 (0.181--0.235) \\
1.0 & DSC & 0.367 (0.333--0.397) & 0.353 (0.323--0.384) \\*
 & NSD & 0.276 (0.249--0.304) & 0.266 (0.239--0.293) \\
\end{longtable}
\normalsize
\endgroup

\begin{table}[!htbp]
\caption{External lung nodule detection validation on an NLST-derived chest CT cohort~\cite{team2013data} with partner-provided reader annotations. Performance is reported for scratch and \Name{cnn}-initialized models across fine-tuning data fractions. Values are shown as mean and 95\% CI; differences are \Name{cnn} minus scratch.}
\label{tab:fig5-siemens-detection-validation}
\centering
\tablebodyfont
\setlength{\tabcolsep}{2pt}
\renewcommand{\arraystretch}{0.90}
\begin{tabular}{@{}l@{\hspace{0.55em}}l@{\hspace{0.55em}}c@{\hspace{0.55em}}c@{\hspace{0.55em}}c@{}}
\toprule
\textbf{Fine-tuning data} & \textbf{Metric} & \multicolumn{1}{c}{\textbf{Scratch}} & \multicolumn{1}{c}{\textbf{\Name{cnn}}} & \multicolumn{1}{c}{\textbf{Difference}} \\
\cmidrule(lr){3-3}\cmidrule(lr){4-4}\cmidrule(lr){5-5}
 &  & \multicolumn{1}{c}{\textbf{Mean (95\% CI)}} & \multicolumn{1}{c}{\textbf{Mean (95\% CI)}} & \multicolumn{1}{c}{\textbf{Mean (95\% CI)}} \\
\midrule
1\% & mAP & 0.407 (0.358--0.454) & 0.439 (0.404--0.478) & +0.032 (-0.047--0.086) \\*
 & FROC & 0.437 (0.397--0.484) & 0.484 (0.457--0.518) & +0.047 (-0.015--0.096) \\*
 & AP@IoU 0.10 & 0.499 (0.448--0.554) & 0.551 (0.511--0.591) & +0.052 (-0.021--0.117) \\
\addlinespace[0.06em]
10\% & mAP & 0.549 (0.506--0.591) & 0.561 (0.521--0.610) & +0.011 (-0.047--0.078) \\*
 & FROC & 0.571 (0.533--0.612) & 0.568 (0.533--0.607) & -0.003 (-0.048--0.050) \\*
 & AP@IoU 0.10 & 0.650 (0.604--0.693) & 0.653 (0.609--0.701) & +0.003 (-0.054--0.068) \\
\addlinespace[0.06em]
100\% & mAP & 0.648 (0.618--0.681) & 0.638 (0.599--0.675) & -0.010 (-0.063--0.050) \\*
 & FROC & 0.648 (0.612--0.683) & 0.628 (0.594--0.665) & -0.020 (-0.070--0.034) \\*
 & AP@IoU 0.10 & 0.734 (0.694--0.765) & 0.723 (0.684--0.762) & -0.011 (-0.059--0.052) \\
\bottomrule
\end{tabular}
\end{table}

\normalsize

\clearpage

\subsection{Figure 6 Extended Results}
\label{app:figure6-metric-tables}
\noindent These tables report the pretraining-development results underlying Figure 6.

\subsubsection{Summary Tables}
\label{app:figure6-summary-tables}

\begin{table}[!htbp]
\caption{Figure 6 segmentation summary scores across 5 datasets. Values are dataset-equal means with 95\% CIs; SD reports variability across dataset-level metric means.}\label{tab:fig6-segmentation-summary}
\centering
\tablebodyfont
\fontsize{7.0bp}{8.2bp}\selectfont
\setlength{\tabcolsep}{0pt}
\renewcommand{\arraystretch}{0.90}
\begin{tabular}{@{}l@{\hspace{0.72em}}l@{\hspace{0.72em}}c@{\hspace{0.72em}}c@{}}
\toprule
\textbf{Model variant} & \textbf{Scale} & \multicolumn{1}{c}{\textbf{DSC}} & \multicolumn{1}{c}{\textbf{NSD}} \\
\cmidrule(lr){3-3}\cmidrule(lr){4-4}
 & & \multicolumn{1}{c}{\makecell[c]{\textbf{Mean}\\\textbf{(95\% CI)}\\\textbf{SD}}} & \multicolumn{1}{c}{\makecell[c]{\textbf{Mean}\\\textbf{(95\% CI)}\\\textbf{SD}}} \\
\midrule
CNN MAE & 630k & \makecell[c]{0.775\\{\fontsize{5.3}{5.8}\selectfont (0.678--0.870)}\\{\fontsize{5.1}{5.6}\selectfont SD 0.124}} & \makecell[c]{0.766\\{\fontsize{5.3}{5.8}\selectfont (0.631--0.869)}\\{\fontsize{5.1}{5.6}\selectfont SD 0.157}} \\
CNN CLR & 630k & \makecell[c]{0.760\\{\fontsize{5.3}{5.8}\selectfont (0.662--0.864)}\\{\fontsize{5.1}{5.6}\selectfont SD 0.130}} & \makecell[c]{0.751\\{\fontsize{5.3}{5.8}\selectfont (0.605--0.853)}\\{\fontsize{5.1}{5.6}\selectfont SD 0.163}} \\
ViT MAE & 630k & \makecell[c]{0.756\\{\fontsize{5.3}{5.8}\selectfont (0.661--0.851)}\\{\fontsize{5.1}{5.6}\selectfont SD 0.124}} & \makecell[c]{0.737\\{\fontsize{5.3}{5.8}\selectfont (0.617--0.843)}\\{\fontsize{5.1}{5.6}\selectfont SD 0.153}} \\
ViT DINO & 630k & \makecell[c]{0.731\\{\fontsize{5.3}{5.8}\selectfont (0.600--0.856)}\\{\fontsize{5.1}{5.6}\selectfont SD 0.160}} & \makecell[c]{0.704\\{\fontsize{5.3}{5.8}\selectfont (0.555--0.845)}\\{\fontsize{5.1}{5.6}\selectfont SD 0.193}} \\
ViT CLR & 630k & \makecell[c]{0.717\\{\fontsize{5.3}{5.8}\selectfont (0.595--0.839)}\\{\fontsize{5.1}{5.6}\selectfont SD 0.158}} & \makecell[c]{0.688\\{\fontsize{5.3}{5.8}\selectfont (0.530--0.827)}\\{\fontsize{5.1}{5.6}\selectfont SD 0.191}} \\
\addlinespace[0.28em]
\Name{cnn} Small MAE & 2.1M & \makecell[c]{0.775\\{\fontsize{5.3}{5.8}\selectfont (0.677--0.873)}\\{\fontsize{5.1}{5.6}\selectfont SD 0.125}} & \makecell[c]{0.766\\{\fontsize{5.3}{5.8}\selectfont (0.627--0.872)}\\{\fontsize{5.1}{5.6}\selectfont SD 0.161}} \\
\Name{cnn} Big MAE & 2.1M & \makecell[c]{0.767\\{\fontsize{5.3}{5.8}\selectfont (0.662--0.872)}\\{\fontsize{5.1}{5.6}\selectfont SD 0.136}} & \makecell[c]{0.758\\{\fontsize{5.3}{5.8}\selectfont (0.611--0.873)}\\{\fontsize{5.1}{5.6}\selectfont SD 0.174}} \\
\Name{vit} Small MAE & 2.1M & \makecell[c]{0.737\\{\fontsize{5.3}{5.8}\selectfont (0.623--0.848)}\\{\fontsize{5.1}{5.6}\selectfont SD 0.151}} & \makecell[c]{0.707\\{\fontsize{5.3}{5.8}\selectfont (0.551--0.853)}\\{\fontsize{5.1}{5.6}\selectfont SD 0.195}} \\
\Name{vit} Big MAE & 2.1M & \makecell[c]{0.771\\{\fontsize{5.3}{5.8}\selectfont (0.667--0.876)}\\{\fontsize{5.1}{5.6}\selectfont SD 0.133}} & \makecell[c]{0.759\\{\fontsize{5.3}{5.8}\selectfont (0.614--0.869)}\\{\fontsize{5.1}{5.6}\selectfont SD 0.163}} \\
\addlinespace[0.28em]
\bottomrule
\end{tabular}
\end{table}
\normalsize

\begin{table}[!htbp]
\caption{Pairwise metric-difference comparisons corresponding to \cref{tab:fig6-segmentation-summary}. Cells report reference-minus-comparator dataset-equal mean differences with paired dataset-level bootstrap 95\% CIs in parentheses. For the 630k screen, the reference is CNN MAE; for the 2.1M scale-up, the reference is \Name{cnn} Small MAE. Positive values favor the reference model because higher metric values are better. A dagger marks comparisons for which the paired 95\% CI does not lie entirely above zero.}
\label{tab:fig6-segmentation-pairwise-support}
\centering
\tablebodyfont
\fontsize{7.0bp}{8.2bp}\selectfont
\setlength{\tabcolsep}{0pt}
\renewcommand{\arraystretch}{0.90}
\begin{tabular}{@{}l@{\hspace{0.72em}}c@{\hspace{0.72em}}c@{}}
\toprule
\textbf{Comparator} & \multicolumn{1}{c}{\textbf{DSC}} & \multicolumn{1}{c}{\textbf{NSD}} \\
\cmidrule(lr){2-2}\cmidrule(lr){3-3}
 & \multicolumn{1}{c}{\makecell[c]{\textbf{\(\Delta\)}\\\textbf{(95\% CI)}}} & \multicolumn{1}{c}{\makecell[c]{\textbf{\(\Delta\)}\\\textbf{(95\% CI)}}} \\
\midrule
\multicolumn{3}{@{}l}{\textbf{630k screen (reference: CNN MAE)}}\\*
CNN MAE & Reference & Reference \\
CNN CLR & \makecell[c]{0.015\\{\fontsize{5.3}{5.8}\selectfont (0.009 to 0.020)}} & \makecell[c]{0.014\\{\fontsize{5.3}{5.8}\selectfont (0.010 to 0.020)}} \\
ViT MAE & \makecell[c]{0.019\\{\fontsize{5.3}{5.8}\selectfont (0.001 to 0.037)}} & \makecell[c]{0.028\\{\fontsize{5.3}{5.8}\selectfont (0.004 to 0.053)}} \\
ViT DINO & \makecell[c]{0.044\\{\fontsize{5.3}{5.8}\selectfont (0.015 to 0.079)}} & \makecell[c]{0.061\\{\fontsize{5.3}{5.8}\selectfont (0.019 to 0.110)}} \\
ViT CLR & \makecell[c]{0.057\\{\fontsize{5.3}{5.8}\selectfont (0.029 to 0.088)}} & \makecell[c]{0.078\\{\fontsize{5.3}{5.8}\selectfont (0.038 to 0.117)}} \\
\addlinespace[0.18em]
\multicolumn{3}{@{}l}{\textbf{2.1M scale-up (reference: \Name{cnn} Small MAE)}}\\*
\Name{cnn} Small MAE & Reference & Reference \\
\Name{cnn} Big MAE & \makecell[c]{0.008\\{\fontsize{5.3}{5.8}\selectfont (-0.002 to 0.019)$^\dagger$}} & \makecell[c]{0.008\\{\fontsize{5.3}{5.8}\selectfont (-0.002 to 0.018)$^\dagger$}} \\
\Name{vit} Small MAE & \makecell[c]{0.039\\{\fontsize{5.3}{5.8}\selectfont (0.011 to 0.066)}} & \makecell[c]{0.059\\{\fontsize{5.3}{5.8}\selectfont (0.018 to 0.101)}} \\
\Name{vit} Big MAE & \makecell[c]{0.004\\{\fontsize{5.3}{5.8}\selectfont (-0.004 to 0.016)$^\dagger$}} & \makecell[c]{0.007\\{\fontsize{5.3}{5.8}\selectfont (-0.004 to 0.023)$^\dagger$}} \\
\addlinespace[0.18em]
\bottomrule
\end{tabular}
\end{table}
\normalsize

\begin{table}[!htbp]
\caption{Figure 6 classification summary scores across 2 datasets. Values are dataset-equal means with 95\% CIs; SD reports variability across dataset-level metric means.}\label{tab:fig6-classification-summary}
\centering
\tablebodyfont
\fontsize{7.0bp}{8.2bp}\selectfont
\setlength{\tabcolsep}{0pt}
\renewcommand{\arraystretch}{0.90}
\begin{tabular}{@{}l@{\hspace{0.72em}}l@{\hspace{0.72em}}c@{\hspace{0.72em}}c@{}}
\toprule
\textbf{Model variant} & \textbf{Scale} & \multicolumn{1}{c}{\textbf{AUROC}} & \multicolumn{1}{c}{\textbf{AUPRC}} \\
\cmidrule(lr){3-3}\cmidrule(lr){4-4}
 & & \multicolumn{1}{c}{\makecell[c]{\textbf{Mean}\\\textbf{(95\% CI)}\\\textbf{SD}}} & \multicolumn{1}{c}{\makecell[c]{\textbf{Mean}\\\textbf{(95\% CI)}\\\textbf{SD}}} \\
\midrule
ViT MAE & 630k & \makecell[c]{0.732\\{\fontsize{5.3}{5.8}\selectfont (0.655--0.809)}\\{\fontsize{5.1}{5.6}\selectfont SD 0.109}} & \makecell[c]{0.680\\{\fontsize{5.3}{5.8}\selectfont (0.658--0.703)}\\{\fontsize{5.1}{5.6}\selectfont SD 0.032}} \\
ViT DINO & 630k & \makecell[c]{0.701\\{\fontsize{5.3}{5.8}\selectfont (0.611--0.790)}\\{\fontsize{5.1}{5.6}\selectfont SD 0.126}} & \makecell[c]{0.644\\{\fontsize{5.3}{5.8}\selectfont (0.589--0.698)}\\{\fontsize{5.1}{5.6}\selectfont SD 0.077}} \\
CNN CLR & 630k & \makecell[c]{0.708\\{\fontsize{5.3}{5.8}\selectfont (0.636--0.780)}\\{\fontsize{5.1}{5.6}\selectfont SD 0.101}} & \makecell[c]{0.634\\{\fontsize{5.3}{5.8}\selectfont (0.621--0.646)}\\{\fontsize{5.1}{5.6}\selectfont SD 0.018}} \\
CNN MAE & 630k & \makecell[c]{0.648\\{\fontsize{5.3}{5.8}\selectfont (0.639--0.656)}\\{\fontsize{5.1}{5.6}\selectfont SD 0.012}} & \makecell[c]{0.577\\{\fontsize{5.3}{5.8}\selectfont (0.545--0.609)}\\{\fontsize{5.1}{5.6}\selectfont SD 0.045}} \\
ViT CLR & 630k & \makecell[c]{0.693\\{\fontsize{5.3}{5.8}\selectfont (0.600--0.787)}\\{\fontsize{5.1}{5.6}\selectfont SD 0.133}} & \makecell[c]{0.615\\{\fontsize{5.3}{5.8}\selectfont (0.551--0.680)}\\{\fontsize{5.1}{5.6}\selectfont SD 0.091}} \\
\addlinespace[0.28em]
\Name{cnn} Small MAE & 2.1M & \makecell[c]{0.641\\{\fontsize{5.3}{5.8}\selectfont (0.640--0.642)}\\{\fontsize{5.1}{5.6}\selectfont SD 0.001}} & \makecell[c]{0.578\\{\fontsize{5.3}{5.8}\selectfont (0.551--0.605)}\\{\fontsize{5.1}{5.6}\selectfont SD 0.038}} \\
\Name{cnn} Big MAE & 2.1M & \makecell[c]{0.677\\{\fontsize{5.3}{5.8}\selectfont (0.597--0.757)}\\{\fontsize{5.1}{5.6}\selectfont SD 0.113}} & \makecell[c]{0.591\\{\fontsize{5.3}{5.8}\selectfont (0.563--0.620)}\\{\fontsize{5.1}{5.6}\selectfont SD 0.041}} \\
\Name{vit} Small MAE & 2.1M & \makecell[c]{0.692\\{\fontsize{5.3}{5.8}\selectfont (0.639--0.744)}\\{\fontsize{5.1}{5.6}\selectfont SD 0.074}} & \makecell[c]{0.633\\{\fontsize{5.3}{5.8}\selectfont (0.623--0.644)}\\{\fontsize{5.1}{5.6}\selectfont SD 0.014}} \\
\Name{vit} Big MAE & 2.1M & \makecell[c]{0.741\\{\fontsize{5.3}{5.8}\selectfont (0.680--0.802)}\\{\fontsize{5.1}{5.6}\selectfont SD 0.086}} & \makecell[c]{0.681\\{\fontsize{5.3}{5.8}\selectfont (0.673--0.689)}\\{\fontsize{5.1}{5.6}\selectfont SD 0.011}} \\
\addlinespace[0.28em]
\bottomrule
\end{tabular}
\end{table}
\normalsize

\begin{table}[!htbp]
\caption{Pairwise metric-difference comparisons corresponding to \cref{tab:fig6-classification-summary}. Cells report reference-minus-comparator dataset-equal mean differences with paired dataset-level bootstrap 95\% CIs in parentheses. For the 630k screen, the reference is ViT MAE; for the 2.1M scale-up, the reference is \Name{vit} Big MAE. Positive values favor the reference model because higher metric values are better. A dagger marks comparisons for which the paired 95\% CI does not lie entirely above zero.}
\label{tab:fig6-classification-pairwise-support}
\centering
\tablebodyfont
\fontsize{7.0bp}{8.2bp}\selectfont
\setlength{\tabcolsep}{0pt}
\renewcommand{\arraystretch}{0.90}
\begin{tabular}{@{}l@{\hspace{0.72em}}c@{\hspace{0.72em}}c@{}}
\toprule
\textbf{Comparator} & \multicolumn{1}{c}{\textbf{AUROC}} & \multicolumn{1}{c}{\textbf{AUPRC}} \\
\cmidrule(lr){2-2}\cmidrule(lr){3-3}
 & \multicolumn{1}{c}{\makecell[c]{\textbf{\(\Delta\)}\\\textbf{(95\% CI)}}} & \multicolumn{1}{c}{\makecell[c]{\textbf{\(\Delta\)}\\\textbf{(95\% CI)}}} \\
\midrule
\multicolumn{3}{@{}l}{\textbf{630k screen (reference: ViT MAE)}}\\*
ViT MAE & Reference & Reference \\
ViT DINO & \makecell[c]{0.031\\{\fontsize{5.3}{5.8}\selectfont (0.019 to 0.043)}} & \makecell[c]{0.037\\{\fontsize{5.3}{5.8}\selectfont (0.004 to 0.069)}} \\
CNN CLR & \makecell[c]{0.024\\{\fontsize{5.3}{5.8}\selectfont (0.018 to 0.030)}} & \makecell[c]{0.046\\{\fontsize{5.3}{5.8}\selectfont (0.037 to 0.056)}} \\
CNN MAE & \makecell[c]{0.084\\{\fontsize{5.3}{5.8}\selectfont (0.016 to 0.153)}} & \makecell[c]{0.103\\{\fontsize{5.3}{5.8}\selectfont (0.049 to 0.157)}} \\
ViT CLR & \makecell[c]{0.038\\{\fontsize{5.3}{5.8}\selectfont (0.022 to 0.055)}} & \makecell[c]{0.065\\{\fontsize{5.3}{5.8}\selectfont (0.023 to 0.107)}} \\
\addlinespace[0.18em]
\multicolumn{3}{@{}l}{\textbf{2.1M scale-up (reference: \Name{vit} Big MAE)}}\\*
\Name{cnn} Small MAE & \makecell[c]{0.100\\{\fontsize{5.3}{5.8}\selectfont (0.039 to 0.161)}} & \makecell[c]{0.102\\{\fontsize{5.3}{5.8}\selectfont (0.083 to 0.122)}} \\
\Name{cnn} Big MAE & \makecell[c]{0.064\\{\fontsize{5.3}{5.8}\selectfont (0.044 to 0.083)}} & \makecell[c]{0.089\\{\fontsize{5.3}{5.8}\selectfont (0.053 to 0.126)}} \\
\Name{vit} Small MAE & \makecell[c]{0.049\\{\fontsize{5.3}{5.8}\selectfont (0.041 to 0.058)}} & \makecell[c]{0.047\\{\fontsize{5.3}{5.8}\selectfont (0.029 to 0.066)}} \\
\Name{vit} Big MAE & Reference & Reference \\
\addlinespace[0.18em]
\bottomrule
\end{tabular}
\end{table}
\normalsize

\begin{table}[!htbp]
\caption{Figure 6 detection summary scores across 2 tasks. Values are dataset-equal means with 95\% CIs; SD reports variability across dataset-level metric means.}\label{tab:fig6-detection-summary}
\centering
\tablebodyfont
\fontsize{7.0bp}{8.2bp}\selectfont
\setlength{\tabcolsep}{0pt}
\renewcommand{\arraystretch}{0.90}
\begin{tabular}{@{}l@{\hspace{0.72em}}l@{\hspace{0.72em}}c@{\hspace{0.72em}}c@{\hspace{0.72em}}c@{}}
\toprule
\textbf{Model variant} & \textbf{Scale} & \multicolumn{1}{c}{\textbf{mAP}} & \multicolumn{1}{c}{\textbf{FROC}} & \multicolumn{1}{c}{\textbf{AP@IoU 0.10}} \\
\cmidrule(lr){3-3}\cmidrule(lr){4-4}\cmidrule(lr){5-5}
 & & \multicolumn{1}{c}{\makecell[c]{\textbf{Mean}\\\textbf{(95\% CI)}\\\textbf{SD}}} & \multicolumn{1}{c}{\makecell[c]{\textbf{Mean}\\\textbf{(95\% CI)}\\\textbf{SD}}} & \multicolumn{1}{c}{\makecell[c]{\textbf{Mean}\\\textbf{(95\% CI)}\\\textbf{SD}}} \\
\midrule
CNN MAE & 630k & \makecell[c]{0.379\\{\fontsize{5.3}{5.8}\selectfont (0.166--0.593)}\\{\fontsize{5.1}{5.6}\selectfont SD 0.302}} & \makecell[c]{0.606\\{\fontsize{5.3}{5.8}\selectfont (0.523--0.689)}\\{\fontsize{5.1}{5.6}\selectfont SD 0.118}} & \makecell[c]{0.462\\{\fontsize{5.3}{5.8}\selectfont (0.214--0.710)}\\{\fontsize{5.1}{5.6}\selectfont SD 0.351}} \\
ViT DINO & 630k & \makecell[c]{0.222\\{\fontsize{5.3}{5.8}\selectfont (0.055--0.389)}\\{\fontsize{5.1}{5.6}\selectfont SD 0.236}} & \makecell[c]{0.468\\{\fontsize{5.3}{5.8}\selectfont (0.432--0.504)}\\{\fontsize{5.1}{5.6}\selectfont SD 0.050}} & \makecell[c]{0.321\\{\fontsize{5.3}{5.8}\selectfont (0.142--0.501)}\\{\fontsize{5.1}{5.6}\selectfont SD 0.254}} \\
ViT MAE & 630k & \makecell[c]{0.367\\{\fontsize{5.3}{5.8}\selectfont (0.180--0.554)}\\{\fontsize{5.1}{5.6}\selectfont SD 0.264}} & \makecell[c]{0.577\\{\fontsize{5.3}{5.8}\selectfont (0.503--0.651)}\\{\fontsize{5.1}{5.6}\selectfont SD 0.104}} & \makecell[c]{0.470\\{\fontsize{5.3}{5.8}\selectfont (0.270--0.671)}\\{\fontsize{5.1}{5.6}\selectfont SD 0.284}} \\
CNN CLR & 630k & \makecell[c]{0.346\\{\fontsize{5.3}{5.8}\selectfont (0.135--0.557)}\\{\fontsize{5.1}{5.6}\selectfont SD 0.298}} & \makecell[c]{0.585\\{\fontsize{5.3}{5.8}\selectfont (0.499--0.671)}\\{\fontsize{5.1}{5.6}\selectfont SD 0.121}} & \makecell[c]{0.444\\{\fontsize{5.3}{5.8}\selectfont (0.198--0.689)}\\{\fontsize{5.1}{5.6}\selectfont SD 0.347}} \\
ViT CLR & 630k & \makecell[c]{0.304\\{\fontsize{5.3}{5.8}\selectfont (0.147--0.462)}\\{\fontsize{5.1}{5.6}\selectfont SD 0.223}} & \makecell[c]{0.539\\{\fontsize{5.3}{5.8}\selectfont (0.505--0.573)}\\{\fontsize{5.1}{5.6}\selectfont SD 0.049}} & \makecell[c]{0.395\\{\fontsize{5.3}{5.8}\selectfont (0.212--0.578)}\\{\fontsize{5.1}{5.6}\selectfont SD 0.259}} \\
\addlinespace[0.28em]
\Name{cnn} Small MAE & 2.1M & \makecell[c]{0.362\\{\fontsize{5.3}{5.8}\selectfont (0.135--0.588)}\\{\fontsize{5.1}{5.6}\selectfont SD 0.320}} & \makecell[c]{0.622\\{\fontsize{5.3}{5.8}\selectfont (0.556--0.687)}\\{\fontsize{5.1}{5.6}\selectfont SD 0.093}} & \makecell[c]{0.456\\{\fontsize{5.3}{5.8}\selectfont (0.203--0.710)}\\{\fontsize{5.1}{5.6}\selectfont SD 0.358}} \\
\Name{cnn} Big MAE & 2.1M & \makecell[c]{0.392\\{\fontsize{5.3}{5.8}\selectfont (0.181--0.602)}\\{\fontsize{5.1}{5.6}\selectfont SD 0.298}} & \makecell[c]{0.653\\{\fontsize{5.3}{5.8}\selectfont (0.609--0.697)}\\{\fontsize{5.1}{5.6}\selectfont SD 0.062}} & \makecell[c]{0.491\\{\fontsize{5.3}{5.8}\selectfont (0.261--0.721)}\\{\fontsize{5.1}{5.6}\selectfont SD 0.325}} \\
\Name{vit} Small MAE & 2.1M & \makecell[c]{0.341\\{\fontsize{5.3}{5.8}\selectfont (0.145--0.537)}\\{\fontsize{5.1}{5.6}\selectfont SD 0.278}} & \makecell[c]{0.566\\{\fontsize{5.3}{5.8}\selectfont (0.497--0.636)}\\{\fontsize{5.1}{5.6}\selectfont SD 0.098}} & \makecell[c]{0.443\\{\fontsize{5.3}{5.8}\selectfont (0.233--0.653)}\\{\fontsize{5.1}{5.6}\selectfont SD 0.297}} \\
\Name{vit} Big MAE & 2.1M & \makecell[c]{0.367\\{\fontsize{5.3}{5.8}\selectfont (0.178--0.556)}\\{\fontsize{5.1}{5.6}\selectfont SD 0.268}} & \makecell[c]{0.622\\{\fontsize{5.3}{5.8}\selectfont (0.593--0.650)}\\{\fontsize{5.1}{5.6}\selectfont SD 0.040}} & \makecell[c]{0.478\\{\fontsize{5.3}{5.8}\selectfont (0.284--0.671)}\\{\fontsize{5.1}{5.6}\selectfont SD 0.274}} \\
\addlinespace[0.28em]
\bottomrule
\end{tabular}
\end{table}
\normalsize

\begin{table}[!htbp]
\caption{Pairwise metric-difference comparisons corresponding to \cref{tab:fig6-detection-summary}. Cells report reference-minus-comparator dataset-equal mean differences with paired dataset-level bootstrap 95\% CIs in parentheses. For the 630k screen, the reference is CNN MAE; for the 2.1M scale-up, the reference is \Name{cnn} Small MAE. Figure 6 ranks are based on mAP and FROC; AP@IoU 0.10 is shown as a supportive endpoint. Positive values favor the reference model because higher metric values are better. A dagger marks comparisons for which the paired 95\% CI does not lie entirely above zero.}
\label{tab:fig6-detection-pairwise-support}
\centering
\tablebodyfont
\fontsize{7.0bp}{8.2bp}\selectfont
\setlength{\tabcolsep}{0pt}
\renewcommand{\arraystretch}{0.90}
\begin{tabular}{@{}l@{\hspace{0.72em}}c@{\hspace{0.72em}}c@{\hspace{0.72em}}c@{}}
\toprule
\textbf{Comparator} & \multicolumn{1}{c}{\textbf{mAP}} & \multicolumn{1}{c}{\textbf{FROC}} & \multicolumn{1}{c}{\textbf{AP@IoU 0.10}} \\
\cmidrule(lr){2-2}\cmidrule(lr){3-3}\cmidrule(lr){4-4}
 & \multicolumn{1}{c}{\makecell[c]{\textbf{\(\Delta\)}\\\textbf{(95\% CI)}}} & \multicolumn{1}{c}{\makecell[c]{\textbf{\(\Delta\)}\\\textbf{(95\% CI)}}} & \multicolumn{1}{c}{\makecell[c]{\textbf{\(\Delta\)}\\\textbf{(95\% CI)}}} \\
\midrule
\multicolumn{4}{@{}l}{\textbf{630k screen (reference: CNN MAE)}}\\*
CNN MAE & Reference & Reference & Reference \\
ViT DINO & \makecell[c]{0.157\\{\fontsize{5.3}{5.8}\selectfont (0.110 to 0.204)}} & \makecell[c]{0.138\\{\fontsize{5.3}{5.8}\selectfont (0.090 to 0.185)}} & \makecell[c]{0.141\\{\fontsize{5.3}{5.8}\selectfont (0.072 to 0.210)}} \\
ViT MAE & \makecell[c]{0.012\\{\fontsize{5.3}{5.8}\selectfont (-0.015 to 0.039)$^\dagger$}} & \makecell[c]{0.029\\{\fontsize{5.3}{5.8}\selectfont (0.019 to 0.038)}} & \makecell[c]{-0.008\\{\fontsize{5.3}{5.8}\selectfont (-0.056 to 0.039)$^\dagger$}} \\
CNN CLR & \makecell[c]{0.033\\{\fontsize{5.3}{5.8}\selectfont (0.031 to 0.036)}} & \makecell[c]{0.021\\{\fontsize{5.3}{5.8}\selectfont (0.018 to 0.023)}} & \makecell[c]{0.019\\{\fontsize{5.3}{5.8}\selectfont (0.016 to 0.021)}} \\
ViT CLR & \makecell[c]{0.075\\{\fontsize{5.3}{5.8}\selectfont (0.019 to 0.131)}} & \makecell[c]{0.067\\{\fontsize{5.3}{5.8}\selectfont (0.018 to 0.116)}} & \makecell[c]{0.067\\{\fontsize{5.3}{5.8}\selectfont (0.002 to 0.132)}} \\
\addlinespace[0.18em]
\multicolumn{4}{@{}l}{\textbf{2.1M scale-up (reference: \Name{cnn} Small MAE)}}\\*
\Name{cnn} Small MAE & Reference & Reference & Reference \\
\Name{cnn} Big MAE & \makecell[c]{-0.030\\{\fontsize{5.3}{5.8}\selectfont (-0.046 to -0.014)$^\dagger$}} & \makecell[c]{-0.031\\{\fontsize{5.3}{5.8}\selectfont (-0.053 to -0.010)$^\dagger$}} & \makecell[c]{-0.034\\{\fontsize{5.3}{5.8}\selectfont (-0.058 to -0.011)$^\dagger$}} \\
\Name{vit} Small MAE & \makecell[c]{0.021\\{\fontsize{5.3}{5.8}\selectfont (-0.010 to 0.051)$^\dagger$}} & \makecell[c]{0.055\\{\fontsize{5.3}{5.8}\selectfont (0.052 to 0.059)}} & \makecell[c]{0.013\\{\fontsize{5.3}{5.8}\selectfont (-0.030 to 0.057)$^\dagger$}} \\
\Name{vit} Big MAE & \makecell[c]{-0.005\\{\fontsize{5.3}{5.8}\selectfont (-0.043 to 0.032)$^\dagger$}} & \makecell[c]{0.000\\{\fontsize{5.3}{5.8}\selectfont (-0.037 to 0.037)$^\dagger$}} & \makecell[c]{-0.021\\{\fontsize{5.3}{5.8}\selectfont (-0.081 to 0.039)$^\dagger$}} \\
\addlinespace[0.18em]
\bottomrule
\end{tabular}
\end{table}
\normalsize

\clearpage
\subsubsection{Dataset-Level Results}
\label{app:figure6-dataset-level-results}

\begingroup
\singlespacing
\tablebodyfont
\setlength{\tabcolsep}{2pt}
\renewcommand{\arraystretch}{0.90}
\setlength{\LTcapwidth}{\linewidth}
\begin{longtable}{@{}l@{\hspace{0.80em}}l@{\hspace{0.80em}}c@{\hspace{0.80em}}c@{}}
\caption{Dataset-level Figure 6 segmentation scores. Values are shown as mean and 95\% CI.}\label{tab:fig6-segmentation-dataset}\\
\toprule
\textbf{Model variant} & \textbf{Scale} & \multicolumn{1}{c}{\textbf{DSC}} & \multicolumn{1}{c}{\textbf{NSD}} \\
\cmidrule(lr){3-3}\cmidrule(lr){4-4}
 & & \multicolumn{1}{c}{\textbf{Mean (95\% CI)}} & \multicolumn{1}{c}{\textbf{Mean (95\% CI)}} \\
\midrule

\endfirsthead
\caption[]{Dataset-level Figure 6 segmentation scores. Values are shown as mean and 95\% CI. (continued)}\\
\toprule
\textbf{Model variant} & \textbf{Scale} & \multicolumn{1}{c}{\textbf{DSC}} & \multicolumn{1}{c}{\textbf{NSD}} \\
\cmidrule(lr){3-3}\cmidrule(lr){4-4}
 & & \multicolumn{1}{c}{\textbf{Mean (95\% CI)}} & \multicolumn{1}{c}{\textbf{Mean (95\% CI)}} \\
\midrule

\endhead
\midrule
\multicolumn{4}{r}{\footnotesize Continued on next page} \\
\endfoot
\bottomrule
\endlastfoot
\addlinespace[0.18em]\multicolumn{4}{@{}l}{\textbf{BrainMetShare~\protect\cite{stanfordaimi2023brainmetshare,grovik2020deep}}}\\*
CNN MAE & 630k & 0.650 (0.604--0.692) & 0.764 (0.709--0.812) \\
CNN CLR & 630k & 0.633 (0.583--0.675) & 0.755 (0.707--0.802) \\
ViT MAE & 630k & 0.608 (0.558--0.652) & 0.709 (0.657--0.756) \\
ViT DINO & 630k & 0.542 (0.494--0.590) & 0.615 (0.560--0.665) \\
ViT CLR & 630k & 0.537 (0.484--0.585) & 0.615 (0.558--0.666) \\
\Name{cnn} Small MAE & 2.1M & 0.656 (0.613--0.700) & 0.768 (0.714--0.813) \\
\Name{cnn} Big MAE & 2.1M & 0.636 (0.584--0.678) & 0.751 (0.699--0.799) \\
\Name{vit} Small MAE & 2.1M & 0.573 (0.519--0.623) & 0.655 (0.599--0.708) \\
\Name{vit} Big MAE & 2.1M & 0.629 (0.585--0.672) & 0.732 (0.680--0.779) \\
\addlinespace[0.06em]
\addlinespace[0.18em]\multicolumn{4}{@{}l}{\textbf{KiTS~\protect\cite{heller2023kits21}}}\\*
CNN MAE & 630k & 0.881 (0.870--0.892) & 0.793 (0.782--0.805) \\
CNN CLR & 630k & 0.875 (0.864--0.886) & 0.780 (0.768--0.791) \\
ViT MAE & 630k & 0.837 (0.823--0.850) & 0.724 (0.711--0.737) \\
ViT DINO & 630k & 0.860 (0.847--0.871) & 0.754 (0.740--0.766) \\
ViT CLR & 630k & 0.831 (0.816--0.845) & 0.710 (0.696--0.724) \\
\Name{cnn} Small MAE & 2.1M & 0.882 (0.871--0.893) & 0.794 (0.782--0.805) \\
\Name{cnn} Big MAE & 2.1M & 0.874 (0.862--0.885) & 0.789 (0.777--0.801) \\
\Name{vit} Small MAE & 2.1M & 0.826 (0.812--0.841) & 0.694 (0.681--0.707) \\
\Name{vit} Big MAE & 2.1M & 0.882 (0.871--0.892) & 0.793 (0.782--0.805) \\
\addlinespace[0.06em]
\addlinespace[0.18em]\multicolumn{4}{@{}l}{\textbf{AutoPET FDG~\protect\cite{gatidis2022fdgpetctlesions}}}\\*
CNN MAE & 630k & 0.637 (0.608--0.664) & 0.506 (0.482--0.529) \\
CNN CLR & 630k & 0.612 (0.584--0.638) & 0.480 (0.456--0.503) \\
ViT MAE & 630k & 0.643 (0.617--0.668) & 0.506 (0.483--0.527) \\
ViT DINO & 630k & 0.576 (0.549--0.604) & 0.418 (0.395--0.440) \\
ViT CLR & 630k & 0.560 (0.532--0.587) & 0.404 (0.380--0.424) \\
\Name{cnn} Small MAE & 2.1M & 0.628 (0.597--0.655) & 0.498 (0.475--0.521) \\
\Name{cnn} Big MAE & 2.1M & 0.606 (0.578--0.633) & 0.472 (0.448--0.494) \\
\Name{vit} Small MAE & 2.1M & 0.576 (0.547--0.603) & 0.419 (0.396--0.440) \\
\Name{vit} Big MAE & 2.1M & 0.628 (0.601--0.654) & 0.495 (0.470--0.518) \\
\addlinespace[0.06em]
\addlinespace[0.18em]\multicolumn{4}{@{}l}{\textbf{AMOS~\protect\cite{ji2022amos}}}\\*
CNN MAE & 630k & 0.891 (0.885--0.897) & 0.846 (0.836--0.855) \\
CNN CLR & 630k & 0.883 (0.876--0.889) & 0.832 (0.822--0.841) \\
ViT MAE & 630k & 0.890 (0.885--0.896) & 0.843 (0.834--0.851) \\
ViT DINO & 630k & 0.881 (0.875--0.887) & 0.828 (0.819--0.837) \\
ViT CLR & 630k & 0.876 (0.870--0.882) & 0.819 (0.809--0.827) \\
\Name{cnn} Small MAE & 2.1M & 0.892 (0.886--0.898) & 0.847 (0.837--0.855) \\
\Name{cnn} Big MAE & 2.1M & 0.894 (0.888--0.900) & 0.851 (0.841--0.860) \\
\Name{vit} Small MAE & 2.1M & 0.892 (0.887--0.897) & 0.845 (0.836--0.854) \\
\Name{vit} Big MAE & 2.1M & 0.898 (0.892--0.903) & 0.856 (0.847--0.864) \\
\addlinespace[0.06em]
\addlinespace[0.18em]\multicolumn{4}{@{}l}{\textbf{TopCoW~\protect\cite{yang2023topcow}}}\\*
CNN MAE & 630k & 0.814 (0.800--0.831) & 0.919 (0.904--0.935) \\
CNN CLR & 630k & 0.798 (0.782--0.812) & 0.909 (0.893--0.925) \\
ViT MAE & 630k & 0.802 (0.789--0.815) & 0.906 (0.893--0.919) \\
ViT DINO & 630k & 0.797 (0.784--0.810) & 0.908 (0.894--0.922) \\
ViT CLR & 630k & 0.783 (0.769--0.796) & 0.893 (0.879--0.908) \\
\Name{cnn} Small MAE & 2.1M & 0.818 (0.802--0.832) & 0.925 (0.906--0.941) \\
\Name{cnn} Big MAE & 2.1M & 0.825 (0.810--0.839) & 0.929 (0.911--0.944) \\
\Name{vit} Small MAE & 2.1M & 0.815 (0.801--0.828) & 0.922 (0.907--0.935) \\
\Name{vit} Big MAE & 2.1M & 0.818 (0.806--0.832) & 0.919 (0.905--0.932) \\
\addlinespace[0.06em]
\end{longtable}
\normalsize
\endgroup

\begingroup
\singlespacing
\tablebodyfont
\setlength{\tabcolsep}{2pt}
\renewcommand{\arraystretch}{0.90}
\setlength{\LTcapwidth}{\linewidth}
\begin{longtable}{@{}l@{\hspace{0.80em}}l@{\hspace{0.80em}}c@{\hspace{0.80em}}c@{}}
\caption{Dataset-level Figure 6 classification scores. Values are shown as mean and 95\% CI.}\label{tab:fig6-classification-dataset}\\
\toprule
\textbf{Model variant} & \textbf{Scale} & \multicolumn{1}{c}{\textbf{AUROC}} & \multicolumn{1}{c}{\textbf{AUPRC}} \\
\cmidrule(lr){3-3}\cmidrule(lr){4-4}
 & & \multicolumn{1}{c}{\textbf{Mean (95\% CI)}} & \multicolumn{1}{c}{\textbf{Mean (95\% CI)}} \\
\midrule

\endfirsthead
\caption[]{Dataset-level Figure 6 classification scores. Values are shown as mean and 95\% CI. (continued)}\\
\toprule
\textbf{Model variant} & \textbf{Scale} & \multicolumn{1}{c}{\textbf{AUROC}} & \multicolumn{1}{c}{\textbf{AUPRC}} \\
\cmidrule(lr){3-3}\cmidrule(lr){4-4}
 & & \multicolumn{1}{c}{\textbf{Mean (95\% CI)}} & \multicolumn{1}{c}{\textbf{Mean (95\% CI)}} \\
\midrule

\endhead
\midrule
\multicolumn{4}{r}{\footnotesize Continued on next page} \\
\endfoot
\bottomrule
\endlastfoot
\addlinespace[0.18em]\multicolumn{4}{@{}l}{\textbf{MRNet~\protect\cite{bien2018skmtea}}}\\*
ViT MAE & 630k & 0.809 (0.786--0.833) & 0.703 (0.668--0.741) \\
ViT DINO & 630k & 0.790 (0.764--0.818) & 0.698 (0.661--0.739) \\
CNN CLR & 630k & 0.780 (0.755--0.803) & 0.646 (0.613--0.688) \\
CNN MAE & 630k & 0.656 (0.619--0.690) & 0.545 (0.518--0.581) \\
ViT CLR & 630k & 0.787 (0.762--0.811) & 0.680 (0.645--0.721) \\
\Name{cnn} Small MAE & 2.1M & 0.640 (0.606--0.674) & 0.551 (0.523--0.587) \\
\Name{cnn} Big MAE & 2.1M & 0.757 (0.732--0.783) & 0.620 (0.589--0.659) \\
\Name{vit} Small MAE & 2.1M & 0.744 (0.712--0.773) & 0.644 (0.605--0.685) \\
\Name{vit} Big MAE & 2.1M & 0.802 (0.775--0.826) & 0.673 (0.634--0.712) \\
\addlinespace[0.06em]
\addlinespace[0.18em]\multicolumn{4}{@{}l}{\textbf{RSNA Spine~\protect\cite{lin2023rsnacervicalspine}}}\\*
ViT MAE & 630k & 0.655 (0.622--0.687) & 0.658 (0.616--0.700) \\
ViT DINO & 630k & 0.611 (0.575--0.645) & 0.589 (0.541--0.638) \\
CNN CLR & 630k & 0.636 (0.601--0.672) & 0.621 (0.577--0.669) \\
CNN MAE & 630k & 0.639 (0.603--0.674) & 0.609 (0.562--0.659) \\
ViT CLR & 630k & 0.600 (0.565--0.637) & 0.551 (0.507--0.599) \\
\Name{cnn} Small MAE & 2.1M & 0.642 (0.604--0.675) & 0.605 (0.557--0.651) \\
\Name{cnn} Big MAE & 2.1M & 0.597 (0.562--0.631) & 0.563 (0.518--0.611) \\
\Name{vit} Small MAE & 2.1M & 0.639 (0.603--0.673) & 0.623 (0.576--0.669) \\
\Name{vit} Big MAE & 2.1M & 0.680 (0.647--0.714) & 0.689 (0.646--0.733) \\
\addlinespace[0.06em]
\end{longtable}
\normalsize
\endgroup

\begingroup
\singlespacing
\tablebodyfont
\setlength{\tabcolsep}{2pt}
\renewcommand{\arraystretch}{0.90}
\setlength{\LTcapwidth}{\linewidth}
\begin{longtable}{@{}l@{\hspace{0.80em}}l@{\hspace{0.80em}}c@{\hspace{0.80em}}c@{\hspace{0.80em}}c@{}}
\caption{Dataset-level Figure 6 detection scores. Values are shown as mean and 95\% CI.}\label{tab:fig6-detection-dataset}\\
\toprule
\textbf{Model variant} & \textbf{Scale} & \multicolumn{1}{c}{\textbf{mAP}} & \multicolumn{1}{c}{\textbf{FROC}} & \multicolumn{1}{c}{\textbf{AP@IoU 0.10}} \\
\cmidrule(lr){3-3}\cmidrule(lr){4-4}\cmidrule(lr){5-5}
 & & \multicolumn{1}{c}{\textbf{Mean (95\% CI)}} & \multicolumn{1}{c}{\textbf{Mean (95\% CI)}} & \multicolumn{1}{c}{\textbf{Mean (95\% CI)}} \\
\midrule

\endfirsthead
\caption[]{Dataset-level Figure 6 detection scores. Values are shown as mean and 95\% CI. (continued)}\\
\toprule
\textbf{Model variant} & \textbf{Scale} & \multicolumn{1}{c}{\textbf{mAP}} & \multicolumn{1}{c}{\textbf{FROC}} & \multicolumn{1}{c}{\textbf{AP@IoU 0.10}} \\
\cmidrule(lr){3-3}\cmidrule(lr){4-4}\cmidrule(lr){5-5}
 & & \multicolumn{1}{c}{\textbf{Mean (95\% CI)}} & \multicolumn{1}{c}{\textbf{Mean (95\% CI)}} & \multicolumn{1}{c}{\textbf{Mean (95\% CI)}} \\
\midrule

\endhead
\midrule
\multicolumn{5}{r}{\footnotesize Continued on next page} \\
\endfoot
\bottomrule
\endlastfoot
\addlinespace[0.18em]\multicolumn{5}{@{}l}{\textbf{PI-CAI~\protect\cite{saha2024picai}}}\\*
CNN MAE & 630k & 0.166 (0.105--0.245) & 0.523 (0.445--0.604) & 0.214 (0.139--0.292) \\
ViT DINO & 630k & 0.055 (0.033--0.098) & 0.432 (0.358--0.511) & 0.142 (0.087--0.225) \\
ViT MAE & 630k & 0.180 (0.114--0.256) & 0.503 (0.416--0.590) & 0.270 (0.193--0.360) \\
CNN CLR & 630k & 0.135 (0.083--0.214) & 0.499 (0.417--0.583) & 0.198 (0.136--0.285) \\
ViT CLR & 630k & 0.147 (0.093--0.219) & 0.505 (0.424--0.578) & 0.212 (0.140--0.301) \\
\Name{cnn} Small MAE & 2.1M & 0.135 (0.083--0.212) & 0.556 (0.486--0.633) & 0.203 (0.144--0.297) \\
\Name{cnn} Big MAE & 2.1M & 0.181 (0.121--0.263) & 0.609 (0.537--0.685) & 0.261 (0.188--0.359) \\
\Name{vit} Small MAE & 2.1M & 0.145 (0.095--0.221) & 0.497 (0.409--0.591) & 0.233 (0.164--0.333) \\
\Name{vit} Big MAE & 2.1M & 0.178 (0.124--0.260) & 0.593 (0.521--0.676) & 0.284 (0.208--0.376) \\
\addlinespace[0.06em]
\addlinespace[0.18em]\multicolumn{5}{@{}l}{\textbf{PN9~\protect\cite{mei2022sanet}}}\\*
CNN MAE & 630k & 0.593 (0.580--0.607) & 0.689 (0.674--0.701) & 0.710 (0.697--0.724) \\
ViT DINO & 630k & 0.389 (0.375--0.404) & 0.504 (0.490--0.519) & 0.501 (0.484--0.517) \\
ViT MAE & 630k & 0.554 (0.540--0.568) & 0.651 (0.638--0.664) & 0.671 (0.656--0.685) \\
CNN CLR & 630k & 0.557 (0.543--0.570) & 0.671 (0.657--0.684) & 0.689 (0.675--0.703) \\
ViT CLR & 630k & 0.462 (0.448--0.477) & 0.573 (0.557--0.588) & 0.578 (0.561--0.595) \\
\Name{cnn} Small MAE & 2.1M & 0.588 (0.574--0.604) & 0.687 (0.672--0.700) & 0.710 (0.696--0.724) \\
\Name{cnn} Big MAE & 2.1M & 0.602 (0.588--0.618) & 0.697 (0.683--0.710) & 0.721 (0.706--0.735) \\
\Name{vit} Small MAE & 2.1M & 0.537 (0.522--0.552) & 0.636 (0.622--0.650) & 0.653 (0.637--0.668) \\
\Name{vit} Big MAE & 2.1M & 0.556 (0.540--0.573) & 0.650 (0.635--0.667) & 0.671 (0.652--0.689) \\
\addlinespace[0.06em]
\end{longtable}
\normalsize
\endgroup

\clearpage



\end{appendices}


\bibliography{sn-bibliography}

\end{document}